\documentclass[10pt,twocolumn,letterpaper]{article}

\usepackage[pagenumbers]{cvpr} 

\usepackage{microtype}
\usepackage{multirow}
\usepackage{tabularx}

\usepackage{xr-hyper}            
\makeatletter
\long\def\XR@test#1#2#3#4\XR@{%
  \let\XR@tempa\@gobblefour
  \ifx#1\newlabel
    \let\XR@tempa\@firstoffour
  \else\ifx#1\XR@bibcite
    \let\XR@tempa\@secondoffour
  \else\ifx#1\@input
     \let\XR@tempa\@thirdoffour
  \else\ifx#1\new@label@record
     \let\XR@tempa\@fourthoffour
  \fi\fi\fi\fi
   \XR@tempa
    {%
     \in@{@cref}{#2}%
     \ifin@
       \expandafter\protected@xdef\csname r@\XR@prefix#2\endcsname{#3}%
     \else
       \expandafter\protected@xdef\csname r@\XR@prefix#2\endcsname{\XR@addURL{#3}}%
     \fi
    }%
    {\expandafter\bibcite\expandafter{\XR@prefix#2}{#3}}%
    {\edef\XR@list{\XR@list\filename@area#2\relax}}%
    {%
      \edef\next{\noexpand\new@label@record{\XR@prefix#2}{\unexpanded{#3}{xr-url}{\XR@URL}}}%
      \next
    }
  \ifeof\@inputcheck\expandafter\XR@aux
  \else\expandafter\XR@read\fi}
\makeatother
\definecolor{cvprblue}{rgb}{0.21,0.49,0.74}
\usepackage[pagebackref,breaklinks,colorlinks,allcolors=cvprblue]{hyperref}
\usepackage[capitalize]{cleveref}
\crefname{appendix}{Appendix}{Appendices}\Crefname{appendix}{Appendix}{Appendices}\crefname{subappendix}{Appendix}{Appendices}\Crefname{subappendix}{Appendix}{Appendices}   

\newcommand{\methodlong}{Draw-and-Tell\xspace}
\newcommand{\method}{Draw\&Tell\xspace}
\newcommand{\sg}{\mathrm{sg}}
\newcommand{\Limg}{\mathcal L_{\mathrm{img}}}
\newcommand{\Ldesc}{\mathcal L_{\mathrm{desc}}}
\newcommand{\Ltxt}{\mathcal L_{\mathrm{txt}}}

\usepackage[normalem]{ulem}
\newcounter{chg}
\AtBeginDocument{  }

\def\paperID{*****}
\def\confName{CVPR}
\def\confYear{2027}

\title{Draw and Tell: Teaching Text-to-Image Diffusion Models\\to Describe What They Draw}

\author{Muhammad Gul Zain Ali Khan \qquad Dar-Yen Chen \qquad Yi-Zhe Song\\
SketchX, University of Surrey
}

\makeatletter
\let\oldmaketitleDD\@maketitle
\def\@maketitle{\oldmaketitleDD
  \begin{center}\vspace{-6pt}
  \includegraphics[width=\textwidth]{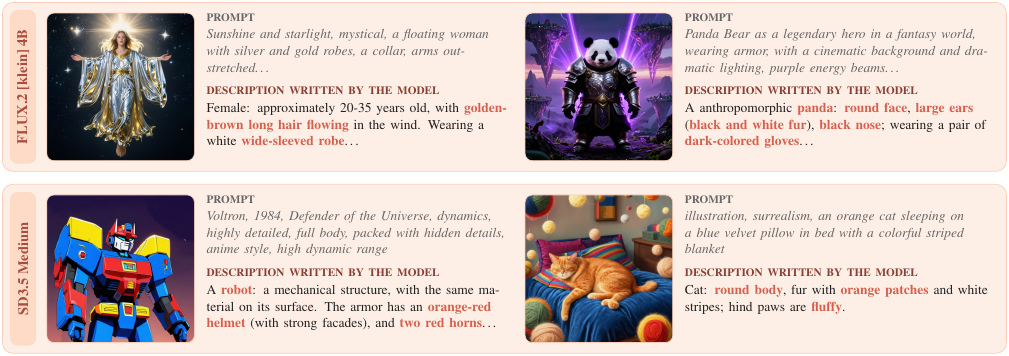}
    \vspace{-2pt}
    \captionof{figure}{\textbf{\methodlong draws and tells}. Alongside the image, the model writes what
    it draws, decoded from its own states by a one-layer read-out. Descriptions have variable length; ``\dots'' marks
    text that continues. Coloured words are details visible in the image.}
    \label{fig:readout}
  \end{center}\vspace{2pt}}
\makeatother

\usepackage{adjustbox}
\usepackage{colortbl}
\usepackage{tikz}
\usepackage[most]{tcolorbox}
\usepackage{float}
\definecolor{accent}{HTML}{D6604D}
\definecolor{accentbar}{HTML}{F4A582}
\definecolor{tint}{HTML}{FDDBC7}
\definecolor{pix}{HTML}{4393C3}
\definecolor{pixtint}{HTML}{D1E5F0}
\definecolor{inkc}{HTML}{222222}
\definecolor{mutedc}{HTML}{666666}
\definecolor{rulec}{HTML}{BDBDBD}
\definecolor{lightc}{HTML}{EFEFEF}
\colorlet{sdcol}{accent}       
\colorlet{fluxcol}{accent}
\colorlet{qmute}{mutedc}       
\colorlet{sdours}{tint!45}     
\colorlet{fluxours}{tint!45}
\colorlet{sdband}{tint}        
\colorlet{fluxband}{tint}
\colorlet{qframe}{accentbar!55}
\colorlet{qlabel}{accent!80!black}
\newtcolorbox{qualband}[1]{enhanced,colback=sdours,colframe=qframe,boxrule=0.3pt,arc=4pt,boxsep=0pt,left=18pt,right=6pt,
  top=5pt,bottom=5pt,before skip=0pt,after skip=3pt,overlay={\fill[sdband,rounded corners=3pt] ([xshift=4pt,yshift=4pt]frame.south west)
  rectangle ([xshift=13pt,yshift=-4pt]frame.north west); \node[rotate=90,text=qlabel,font=\bfseries\fontsize{5.5}{6}\selectfont] at
  ([xshift=8.5pt]frame.west) {#1};}}
\tcbset{judgebox/.style={enhanced,colback=#1,colframe=qframe,boxrule=0.3pt,arc=4pt,boxsep=0pt,left=7pt,right=7pt,top=6pt,
  bottom=6pt,before skip=3pt,after skip=3pt}}
\newcommand{\qchip}[2]{\tikz[baseline=(c.base)]\node[fill=#1,rounded corners=2pt,inner xsep=3.5pt,inner ysep=1.6pt,text=qlabel,
  font=\scriptsize\sffamily\bfseries](c){#2};}
\newcommand{\qhl}[1]{\textcolor{accent}{\bfseries #1}}
\newsavebox{\rimgbox}
\newcommand{\rimg}[2][]{\sbox{\rimgbox}{\includegraphics[#1]{#2}}\tikz[baseline=0pt]{\begin{scope}\clip[rounded corners=2pt]
  (0,0) rectangle (\wd\rimgbox,\ht\rimgbox); \node[anchor=south west,inner sep=0pt] at (0,0) {\usebox{\rimgbox}};\end{scope}
  \draw[qframe,line width=0.3pt,rounded corners=2pt] (0,0) rectangle (\wd\rimgbox,\ht\rimgbox);}}
\newcommand{\qplabel}[1]{{\fontsize{4.6}{5}\selectfont\bfseries\color{mutedc}#1}}
\newcommand{\qdlabel}[1]{{\fontsize{4.6}{5}\selectfont\bfseries\color{accent}#1}}
\newcommand{\dok}[1]{\textcolor{pix!75!black}{#1}}
\newcommand{\dbad}[1]{\textcolor{accent}{#1}}
\begin{document}
\maketitle

\begin{abstract}
Text-to-image (T2I) diffusion models read text and generate images, but they never write. We present \methodlong, which
makes a pre-trained T2I model also write what it draws. Alongside the image, the model co-generates a description
of it as a stream of text-encoder states, denoised by the same rectified flow, and a one-layer language-model read-out
turns the stream into words.  The backbone is adapted with LoRA, and text-only MLPs give the description capacity of its own. The description works like a representation-alignment
target that the backbone must \emph{write} rather than only match: during training it shapes the weights the image is
drawn with, and the image itself never reads it. With minimal resources, 13M text tokens and about 17 GH200 GPU-hours,
the model can be made to write: on SD3.5 Medium and FLUX.2 [klein], its descriptions of held-out prompts are read back
at a perplexity of 46 and 26. And it draws better: GenEval rises from 68.4 to 74.3 on SD3.5 and from 76.3 to 79.0 on
FLUX.2 [klein], above a matched image-only fine-tuning control (73.4 and 77.2), and the gain comes from writing, not from aligning to or training on the descriptions. Blind MLLM judges prefer its images to the control's for fine detail
(56\% and 60\% of decided pairs), and a user study confirms this. The judges also find that, on short, ambiguous prompts, it commits to a meaningful reading
of the prompt. With the description switched off, the network keeps the original architecture and inference cost,
with fine-tuned weights.
\end{abstract}

\section{Introduction}
\label{sec:intro}

A text-to-image diffusion model \cite{lipman2022flow,liu2022flow,albergo2023building,esser2024scaling,sd35,flux2024,flux2,zimage} reads text but does not write it. A frozen encoder turns the prompt into tokens, the network reads them
through attention and denoises an image, and nothing is written back: language flows in, never out. So progress in
prompt following has come from giving the model better text to \emph{read}: denser recaptions \cite{dalle3},
prompts expanded or planned by a language model \cite{llm_grounded_diffusion,gligen,yang2024rpg}, a language model
as the encoder \cite{hu2024ella,qin2025lumina2,wu2025qwenimage}. Writing also sharpens prompt understanding \cite{unraveling2026mmdit,deng2025emerging,mmflow}, but the models that do write are unified models, and
they pay for it: a joint backbone pre-trained on hundreds of billions of tokens \cite{deng2025emerging,mmflow,shi2025muddit}, or,
starting from a text-to-image model, a text decoder trained on tens of millions of image--text pairs
\cite{li2025dual,li2025omniflow,bill2026fullflow}. 

To draw a yellow bird beside
a purple flower, a text-to-image model has to represent the objects, their count, their colours and their layout, which is exactly what a
description of the image says. The knowledge is already there. We therefore ask a simpler question: does an image backbone draw better if it learns to write? Writing here is not a product, as in unified models, but a training signal for the image backbone.

\begin{figure*}[t]
  \centering
  \includegraphics[width=\textwidth]{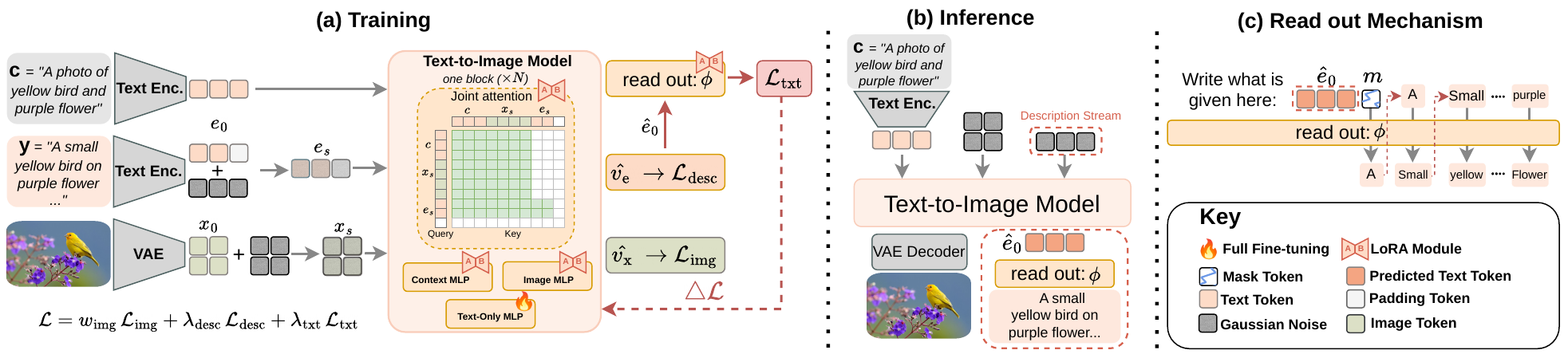}
  \caption{\textbf{Overview of \method.} (a)~\emph{Training.} The description $y$ is encoded by the model's text
encoder into the description stream $e_0$, noised with the image latent by the same rectified flow and denoised
jointly. Joint attention is masked (green: attends) so that the stream reads the prompt and the image but neither
reads the stream. LoRA adapts the blocks and their image and context MLPs; a new text-only MLP is trained in full. The
read-out $\phi$ decodes the predicted stream $\hat e_0$ into words, and its cross-entropy is back-propagated into the backbone. (b)~\emph{Inference.} The stream and read-out (dashed) are optional. (c)~\emph{Read-out.} One LoRA-adapted language-model layer
predicts each word from an instruction, $\hat e_0$ and the previous words.}
  \label{fig:overview}
\end{figure*}

\methodlong enables an image backbone to generate language (\cref{fig:overview}). It makes the text stream of a
multimodal diffusion transformer \cite{esser2024scaling} \emph{read and write}: alongside the image, the model
generates a description of what it is drawing, as states of its own text encoder (T5~\cite{raffel2020exploring} for SD3.5, Qwen3-4B~\cite{yang2025qwen3} for FLUX.2
[klein]) appended to the context sequence of every block and noised and denoised by the same rectified flow as the
image. The description attends to the prompt and the image, but neither attends to it, so it affects the image only
through training. A one-layer read-out initialised from Qwen3-4B \cite{yang2025qwen3} turns the stream into words, and
every wrong word costs: where the states would read as \emph{purple bird}, the loss on the true description says
\emph{yellow}, and its gradient reaches the blocks the image is drawn with. This is a sharper signal than a pixel
loss, under which a rendering can hedge (a half-hidden third cup, a flower of indeterminate colour); a description must
commit to counts, colours and layout. Like representation alignment \cite{yu2024repa}, it is an auxiliary target on
the backbone's states, but one the backbone must \emph{write}, and the read-out shows what those states encode. Two
design choices follow from the premise that the knowledge is already there. The backbone is touched only by LoRA
\cite{LoRA} (rank 32; 47M parameters on SD3.5, 48M on FLUX.2 [klein]), so what the text-to-image model knows is kept; and text-only MLPs, initialised from their block's own MLP and acting on the description tokens alone, hold the new knowledge that writing needs without crowding the image pathway (597M and 566M parameters). Both, with
the read-out, are set aside at inference: with the stream off, the network keeps the original architecture and its
inference cost, with fine-tuned weights.

The read-out feeds the generated states as a prefix to a single pre-trained language-model layer, frozen apart from
LoRA; 
decoding the stream directly instead gives garbled text (perplexity
$3{,}526$ against $46.3$ on SD3.5, \cref{sec:abl_direct}). Training needs minimal resources: 13M text tokens of descriptions
and about 17 GH200 GPU-hours. We validate on two open-weight models, SD3.5 Medium (2.5B)~\cite{sd35} and FLUX.2 [klein] (4B)~\cite{flux2}. On held-out prompts the co-generated description is read back at a perplexity of $46.3$ (SD3.5) and $26.2$ (FLUX.2 [klein]) (\cref{sec:ppl}). A blind MLLM judge (Claude Opus 5.5) finds that most descriptions talk about things in the image (appendix).
Yet some miss the main subject; as our goal is better image generation, we leave improving writing to future work.

It also draws better. GenEval~\cite{ghosh2023geneval} rises from $68.4$ to $74.3$ and DPG-Bench~\cite{hu2024ella}
from $84.2$ to $85.3$ for SD3.5, above a matched fine-tuning control run on the same data, batches and noise without the description
($73.4$, $84.6$); on FLUX.2 [klein] 4B, GenEval rises from $76.3$ to $79.0$ (control $77.2$) (\cref{sec:results}). Aligning the backbone's states to the description with a REPA-style loss \cite{yu2024repa} instead of generating it does not lift GenEval: it is $72.0$ with the prompt tokens aligned and $72.8$ with the image tokens, below the control (\cref{sec:abl_repa}).  Training on the same descriptions without writing them does not help either (GenEval $72.7$, against $74.3$ for \method; \cref{sec:abl_repa}): the gain comes from learning to write. Two properties emerge that benchmarks miss. The images get finer: on prompts every model renders correctly, blind MLLM judges (Claude Opus 5.5) prefer the textures and small parts of \method to the base model's $73.6\%$ to $25.5\%$ ($0.9\%$ draw) on SD3.5 and $63.7\%$ to $34.7\%$ ($1.5\%$ draw) on FLUX.2 [klein], and to the control's $49.5\%$ to $38.4\%$ ($12.0\%$ draw) and $55.2\%$ to $36.3\%$ ($8.5\%$ draw). A user study confirms this preference. And the model commits: given a one-word prompt such as \emph{fairness}, it draws a meaningful scene where the base model renders the word or a malformed object; on 100 PartiPrompts~\cite{yu2022parti} the judges prefer it to the control $43\%$ to $29\%$ ($28\%$ draw) on SD3.5 and $40\%$ to $35\%$ ($25\%$ draw) on FLUX.2 [klein] (\cref{sec:refine}). A text-to-image model that must say what it draws has, in the end, to know what it is drawing.

Our contributions are:
(1)~\methodlong introduces a text stream that a pre-trained diffusion transformer generates with its image, and a
read-out that turns it into words, learned with minimal resources (LoRA, 13M text tokens, 17 GH200 GPU-hours).
(2)~Writing as a training signal: with the words switched off, the same network at the same cost draws better than a
matched image-only control, and the gain comes from writing, not from aligning to the descriptions. (3)~Two emerging properties of a text-to-image model that writes: finer detail,
and commitment to a meaningful reading of ambiguous prompts.

\section{Related Work}
\label{sec:related}
\paragraph{Auxiliary objectives for diffusion transformers.} REPA \cite{yu2024repa} regularises the intermediate
features of a diffusion transformer towards those of a frozen vision encoder and discards the projector at inference;
follow-ups vary the target or regularise without one \cite{leng2025repae,yao2025vavae,zheng2025rae,wang2025dispersive}.
The signal is always a \emph{visual} feature, and the gains are in fidelity and convergence, not prompt following.
Language has reached generator training only indirectly: re-description methods, from MirrorGAN
\cite{qiao2019mirrorgan} to CoMat \cite{jiang2024comat}, score the \emph{generated image} with a frozen captioner, so
the signal arrives through decoded pixels, and UNO \cite{liu2026uno} passes the noised generation states of a unified
model through its frozen understanding expert under a captioning loss. \method keeps the shape of REPA and the lesson
of UNO, in a text-to-image-only transformer that has no understanding expert: the backbone writes the description.

\vspace{-5pt}
\paragraph{Unified text-and-image generation.} One network can generate both modalities, with discrete image tokens
\cite{team2024chameleon,wang2024emu3,xie2025show,yang2026mmada,swerdlow2025unified,wang2026fudoki,nguyen2025oneflow,nie2025large},
continuous images beside discrete text
\cite{zhou2025transfusion,shi2024lmfusion,wu2025janus,chen2025janus,ma2025janusflow,deng2025emerging,xie2026show,li2025dual},
or both continuous \cite{bao2023one,xu2023versatile,tang2023any,li2025omniflow,he2025flowtok,liu2025crossflow,mmflow}.
UniDiffuser \cite{bao2023one} and Multimodal Flow \cite{mmflow} diffuse text embeddings jointly with images, the latter
reaching 0.82 GenEval from scratch with 150B tokens; we add this ingredient to an existing generator. Whether writing
helps drawing inside such models is debated: positive transfer at scale
\cite{tong2024metamorph,deng2025emerging,fan2025unifluid}, interference that leads others to decouple the pathways
\cite{zhou2025transfusion,shi2024lmfusion,wu2025janus}, and a sign that depends on the architecture
\cite{wu2026synergy}. Four start from a pre-trained text-to-image generator, FullFlow \cite{bill2026fullflow}, Muddit
\cite{shi2025muddit}, OmniFlow \cite{li2025omniflow} and, closest to us, D-DiT \cite{li2025dual}, which adds a
masked-diffusion head to the T5 branch of SD3 Medium, trains on 40M recaptioned pairs and reports GenEval rising from
0.62 to 0.65 while counting drops, with no matched image-only control; all four keep the text pathway at inference.
\method is not a unified model: the description is an auxiliary task for the image pathway, trained for a few thousand
steps, measured against a matched control.

\vspace{-5pt}
\paragraph{Reasoning for image generation.} A multimodal language model can write a plan, a layout or a chain of
thought before or between the image tokens
\cite{fang2025got,liao2025imagegencot,jiang2025t2ir1,zhang2025reasongenr1,deng2025emerging,hunyuanimage3,qin2025unicot},
usually trained with reinforcement learning and consumed at inference; some reason in continuous latents
\cite{zhai2026lac,chen2026latentmorph}, and MMaDA-Parallel \cite{tian2025mmadaparallel} denoises text and image in
parallel in a discrete diffusion language model, the mechanism nearest ours. Two findings shaped our design: part of
the benefit of a learnt chain survives when none is produced at inference \cite{qin2025unicot}, and a chain read while
it is still wrong can hurt \cite{tian2025mmadaparallel}. In \method the diffusion trunk itself writes the text, in a
model with no language backbone, and it matters mainly in training; we do not call it reasoning.

\vspace{-5pt}
\paragraph{The MMDiT text stream.} Dense recaptions \cite{dalle3}, prompt expansion and planning
\cite{llm_grounded_diffusion,gligen,yang2024rpg}, language-model encoders \cite{hu2024ella,qin2025lumina2,wu2025qwenimage}
and diffusion heads on multimodal language models
\cite{pan2025transfer,chen2025blip3ofamilyfullyopen,dong2023dreamllm,sun2023generative,tong2024metamorph,xiao2024omnigen}
all give the generator richer text to \emph{read}; in the MMDiT of SD3 \cite{esser2024scaling}, as in CLIP- or
T5-conditioned models \cite{saharia2022photorealistic,ramesh2022hierarchical,chen2024pixartalpha}, the text tokens are
inputs the network never has to produce. The MMDiT text stream drifts during sampling and feeds image errors back into
the conditioning, and training-free interventions on it recover several GenEval points
\cite{unraveling2026mmdit,imagestalkback2026,promptreinjection2026}; soft-token alignment \cite{softrepa2025} gains two
points on SD3 at the cost of counting. \method acts on the same stream, during training and by construction: part of it
becomes an output that must stay decodable, which anchors the states the image reads.
Language enters a generator as an input to read \cite{dalle3,llm_grounded_diffusion,hu2024ella} or as a target to
match \cite{softrepa2025}; unified models also write it, as a separate task or a joint sample \cite{li2025dual,zhou2025transfusion}.
Here a pre-trained text-to-image model generates a description with the image, and the image does not attend to it.

\section{Method}
\label{sec:method}
\method adds a description stream to a pre-trained multimodal diffusion transformer. We describe it for a generic
backbone; backbone-specific choices are in
the appendix.

\subsection{Preliminaries}
\label{sec:prelim}
A multimodal diffusion transformer \cite{esser2024scaling} denoises image latents $x_0$ conditioned on a context
sequence: the prompt $c$ is encoded by frozen text encoders into prompt tokens, and image and context tokens interact
through joint attention in every block. Both backbones are flow models: flow matching \cite{lipman2022flow} with
the rectified-flow path \cite{liu2022flow} moves data $x_0$ to Gaussian noise along a straight line in time $s\in[0,1]$,
\begin{equation}
x_s=(1-s)\,x_0+s\,\epsilon,\qquad \epsilon\sim\mathcal N(0,I),
\label{eq:flow}
\end{equation}
and the network $v_\theta(x_s,s,c)$ is trained to predict the path's velocity $v_x=\mathrm dx_s/\mathrm ds=\epsilon-x_0$,
with $s$ drawn from the backbone's training schedule. Sampling solves the ordinary differential equation
$\mathrm dx_s/\mathrm ds=v_\theta(x_s,s,c)$ from noise ($s{=}1$) to data ($s{=}0$) with Euler steps on a decreasing time
grid $1{=}s_0>s_1>\dots>s_K{=}0$,
\begin{equation}
x_{s_{k+1}}=x_{s_k}+(s_{k+1}-s_k)\,v_\theta(x_{s_k},s_k,c),
\label{eq:euler}
\end{equation}
where the backbone's sampling schedule shifts the grid towards high noise.
In the pre-trained model the context tokens are read by the image but never predicted: the text pathway is read-only.

\subsection{\texorpdfstring{The description stream}{The description stream}}
\label{sec:stream}

\paragraph{The description.} For every training image we have a long, grounded description $y$
(\cref{sec:data}). A frozen text encoder $E_y$ turns it into
states, which we divide by a fixed per-channel scale $\boldsymbol\sigma$ (the per-channel RMS of training
descriptions),
\begin{equation}
e_0=E_y(y)\oslash\boldsymbol\sigma\in\mathbb R^{L_e\times d_e},
\end{equation}
where $\oslash$ is element-wise division. The scale makes the stream commensurate with unit-variance noise, so that it
is noised exactly like the image by \cref{eq:flow}, $e_s=(1-s)\,e_0+s\,\eta$ with $\eta\sim\mathcal N(0,I)$ and velocity target
$v_e=\eta-e_0$.  Training descriptions are a median of 128
and up to 512 tokens long.

\paragraph{The stream.} The context sequence of every block (\cref{fig:overview}a) becomes
\begin{equation}
C=\big[\,E_{ctx}(c)\;\big\|\;E_e(e_s)+\tau+P_{1:L_e}\,\big],
\label{eq:context}
\end{equation}
where $\|$ concatenates along the sequence, $E_{ctx}(c)$ are the prompt tokens, $E_e$ embeds the stream into the
backbone's width $D$, $\tau\in\mathbb R^D$ is a learned, zero-initialised type vector that distinguishes description
tokens from prompt tokens, and $P$ gives the description tokens their positions, which a noisy state no longer
encodes (learnable on SD3.5; learnable plus rotary on FLUX.2 [klein]). Image, prompt and description tokens interact through the native joint attention, with one restriction: the
description attends to the prompt and the image, while neither attends to the description, so the image is drawn
exactly as without the stream and the description reaches the backbone only through the shared weights.
The network thus becomes $v_\theta(x_s,e_s,s,c)=(\hat v_x,\hat v_e)$, in which $\hat v_x$ does not depend on $e_s$. Descriptions
differ in length, so within a batch they are zero-padded to the longest and a
boolean mask removes padded positions as attention keys and from both description losses.

\paragraph{Reading out the description velocity.} The velocity of the stream is read from the description tokens
$C^{(r)}_{\mathrm{desc}}\in\mathbb R^{L_e\times D}$ after a late block $r$ through a fresh AdaLN-continuous norm and a
linear map back to the stream's width,
\begin{equation}
\hat v_e=W_{out}\,\mathrm{AdaLN}\big(C^{(r)}_{\mathrm{desc}},\,t_{emb}\big),
\end{equation}
where $t_{emb}$ is the backbone's conditioning vector, an embedding of the time $s$ (on SD3.5 together with the pooled prompt). $W_{out}$ is zero-initialised, so
$\hat v_e\equiv0$ at initialisation, and with no description ($L_e{=}0$) the network is \emph{exactly} the
original architecture.

\paragraph{Text-only MLPs.} A description carries far more content than a prompt, yet in the pre-trained network
both would share the context MLP (the MLP of the prompt tokens; the image tokens have their own). After every block $i$ with separate text weights we therefore add a residual MLP
that acts only on the description tokens $h_i\in\mathbb R^{L_e\times D}$ output by block $i$,
\begin{equation}
h_i\leftarrow h_i+g_i\odot\mathrm{MLP}_i\big(\mathrm{LN}(h_i)\odot(1{+}a_i)+b_i\big),
\end{equation}
where the scale $a_i$, shift $b_i$ and gate $g_i\in\mathbb R^D$, which weighs the MLP's update per channel, are computed from $t_{emb}$, $(a_i,b_i,g_i)=A_i\,\mathrm{SiLU}(t_{emb})$
(adaLN-Zero \cite{peebles2023scalable}). $A_i$ is zero-initialised, so each MLP is the identity at
insertion; $\mathrm{MLP}_i$ is initialised as a copy of block $i$'s own context MLP, so it starts as the context
MLP and then specialises (its form follows the backbone; see the appendix). 

\subsection{Language-model read-out}
\label{sec:readout}
A flow-matching loss on $\hat v_e$ does not guarantee that a denoised state still says ``three'' rather than ``two'',
so we also supervise the stream at the token level. A read-out $\phi$ decodes the predicted clean stream
$\hat e_0=e_s-s\,\hat v_e$, which follows from \cref{eq:flow} with the predicted velocity (\cref{fig:overview}a): one causal transformer layer initialised from the last layer of Qwen3-4B \cite{yang2025qwen3} and frozen apart from LoRA of rank
16, so it brings a vocabulary and a language prior without becoming a second network to train. Let $y_{1:n}$ be the
Qwen3 tokens of $y$ and $y_{n+1}$ the end-of-text token. The read-out reads
\begin{equation}
\big[\,Q_{-2}(\mathrm{instr.})\;\big\|\;\mathrm{in\_proj}(\hat e_0)\;\big\|\;\tilde y_0,\tilde y_1,\dots,\tilde y_n\,\big],
\end{equation}
where $Q_{-2}(\cdot)$ denotes the states of Qwen3's second-to-last layer, here for a fixed instruction (``Write what is given here:''), the stream mapped to
Qwen3's width by $\mathrm{in\_proj}:\mathbb R^{d_e}\!\to\!\mathbb R^{2560}$, and the teacher-forced tokens
$\tilde y_{1:n}$ (Qwen3's frozen embeddings with a learned projection). A single learned vector $m$ plays two roles.
It opens the token sequence ($\tilde y_0=m$), and the read-out predicts the first word $y_1$ at its position. It also
stands in for dropped tokens: during training each $\tilde y_i$, $i\ge1$, is replaced by $m$ with probability $0.1$,
so that the content must come from the stream rather than from the previous tokens. Position $i-1$ predicts $y_i$, so
the cross-entropy loss is
\begin{equation}
\Ltxt=-\tfrac1{n+1}\sum_{i=1}^{n+1}\log p_\phi\big(y_i\mid Q_{-2},\hat e_0,\tilde y_{0:i-1}\big),
\end{equation}
where predicting $y_{n+1}$ teaches the read-out where the description ends. The read-out never touches image
generation; it turns ``the states carry the description'' into a differentiable statement whose gradient reaches
the backbone through $\hat e_0$.

\subsection{Objective}
\label{sec:objective}
Both streams are trained by velocity matching with the backbone's own flow-matching loss $\ell(\hat v,v)$ (appendix), averaged
over elements (over real description positions only for the stream). Image and description share one time $s$ per sample, drawn from the backbone's training schedule. The full objective models $p(x_0,e_0\mid c)$,
\begin{equation}
\begin{gathered}
\mathcal L=w_{\mathrm{img}}\,\Limg+\lambda_{\mathrm{desc}}\,\Ldesc+\lambda_{\mathrm{txt}}\,\Ltxt,\\
\Limg=\ell(\hat v_x,v_x),\qquad \Ldesc=\ell(\hat v_e,v_e),
\end{gathered}
\label{eq:loss}
\end{equation}

\subsection{Inference: stream off, stream on}
\label{sec:inference}
We sample with 28 Euler steps over the backbone's shifted grid $\{s_k\}_{k=0}^{28}$ ($s_0{=}1$, $s_{28}{=}0$),
integrating the joint ODE $\mathrm d(x_s,e_s)/\mathrm ds=v_\theta(x_s,e_s,s,c)$ with \cref{eq:euler} for each stream $z\in\{x,e\}$, using the velocity $\tilde v_z$ with
classifier-free guidance \cite{ho2022classifier} on both, $\tilde v_z=v_z^{\varnothing}+w_z\,(v_z^{c}-v_z^{\varnothing})$.
The null prompt $\varnothing$ is the backbone's encoding of the empty prompt; it replaces the prompt in $10\%$ of
training samples, so the same network also learns the unconditional velocity. $v_z^{\varnothing}$ is predicted with the
null prompt and the \emph{same} description latents, so guidance contrasts the prompt and nothing else. We use
$w_x{=}w_e{=}4$ on both backbones.  With the stream on, a description of $L_e$ tokens
is initialised from noise and denoised jointly with the image, and $\phi$ decodes it, starting from
$m$, until the end-of-text token. Since each training description is masked to its own length, $L_e$ is free at test time. As in training, the image and prompt
tokens do not attend to the description, so the image is the same with the stream on or off and the description is a
by-product that can be read for the cost of its tokens at each step and one read-out pass. 

\section{\texorpdfstring{Short prompts, long descriptions}{Short prompts, long descriptions}}
\label{sec:data}
Learning to describe needs, for each image, a short prompt $c$ of the kind a user types and a long
description $y$ that states what the prompt leaves out. 

\paragraph{Our corpus.} We use an in-house dataset of
114.7k images, each with a structured description written by a
vision-language model (generation and annotation details in the appendix). The description $y$ names
every object with its count and colours, states where each sits in the frame and how the objects relate to one another. It thus spells out exactly the content that compositional prompts ask for and that the image loss
alone does not name.

\section{Experiments}
\label{sec:experiments}

\begin{table*}[t]
\caption{\textbf{Main results} on FLUX.2 [klein] 4B and SD3.5 Medium (stream off): GenEval, DPG-Bench, the average over the eight T2I-CompBench categories (per category in the appendix) and ImageReward on the HPSv2 benchmark.}
\label{tab:main}
\centering
\footnotesize
\setlength{\tabcolsep}{3pt}
\renewcommand{\arraystretch}{1.0}
\resizebox{\textwidth}{!}{%
\begin{tabular}{lcccccccccccccc}
\toprule
& \multicolumn{7}{c}{GenEval (\%)} & \multicolumn{5}{c}{DPG-Bench} & T2I-Comp. & Image \\
\cmidrule(lr){2-8}\cmidrule(lr){9-13}
Model & Overall & 1 obj. & 2 obj. & Count. & Colours & Pos. & Col.\ attr. & Overall & Global & Entity & Attr. & Relation & Avg.\ (\%) & Reward \\
\midrule
\rowcolor{fluxband}\multicolumn{15}{l}{\textbf{FLUX.2 [klein] 4B}} \\
FLUX.2 [klein] 4B (base) & 76.3 & 98.8 & 88.9 & 72.2 & 87.8 & \textbf{51.5} & 59.0 & 83.15 & 82.07 & 90.13 & 89.48 & 93.50 & 51.20 & 0.961 \\
SFT control (no stream) & 77.2 & \textbf{99.4} & 87.4 & 80.9 & 89.4 & 42.5 & 63.5 & 85.12 & \textbf{83.89} & 91.84 & 89.40 & 94.43 & 51.96 & 1.093 \\
\rowcolor{fluxours}\textbf{\method} & \textbf{79.0} & \textbf{99.4} & \textbf{89.4} & \textbf{81.6} & \textbf{89.9} & 45.3 & \textbf{68.8} & \textbf{85.36} & 83.59 & \textbf{91.97} & \textbf{89.80} & \textbf{94.51} & \textbf{52.29} & \textbf{1.126} \\
\midrule
\rowcolor{sdband}\multicolumn{15}{l}{\textbf{SD3.5 Medium}} \\
SD3.5 Medium (base)     & 68.4 & 99.1 & 84.1 & 64.1 & 82.7 & 26.8 & 53.8 & 84.20 & \textbf{83.89} & 90.34 & 88.40 & 93.27 & 50.01 & 1.076 \\
SFT control (no stream) & 73.4 & 99.1 & \textbf{90.9} & \textbf{72.8} & 84.8 & 30.5 & 62.3 & 84.58 & 82.98 & \textbf{91.11} & \textbf{89.50} & \textbf{93.69} & 52.32 & 1.101 \\
\rowcolor{sdours}\textbf{\method} & \textbf{74.3} & \textbf{99.7} & 89.4 & 71.6 & \textbf{85.6} & \textbf{32.5} & \textbf{67.0} & \textbf{85.25} & 82.37 & \textbf{91.11} & 89.42 & 93.46 & \textbf{52.58} & \textbf{1.139} \\
\bottomrule
\end{tabular}}
\end{table*}

\subsection{Setup}
\label{sec:setup}
\paragraph{Backbones.} We use two open-weight backbones, SD3.5 Medium \cite{esser2024scaling} (2.5B; 24 joint blocks,
CLIP and T5-XXL prompt encoders, T5-XXL description stream) and FLUX.2 [klein] base (4B; 5 double- and 20
single-stream blocks, Qwen3-4B prompt encoder and description stream). The text-only MLPs follow every SD3.5 block that
emits context tokens and the five FLUX.2 double-stream blocks; the remaining backbone-specific choices are in
the appendix.

\paragraph{Training.} We train LoRA on the backbone (rank 32; 47M parameters on
SD3.5, 48M on FLUX.2 [klein]), the stream modules ($\tau$, $P$, the read-out norm and $W_{out}$), the read-out's
projections, mask token $m$ and LoRA (rank 16), and the text-only MLPs (597M and 566M). 8-bit AdamW, constant learning rate $5{\cdot}10^{-5}$, weight decay $10^{-4}$, gradient clipping $1.0$,
batch 128, bf16. Training uses our corpus (114.7k images, \cref{sec:data}, held-out prompts excluded); the corpus holds
13.3M text tokens (Qwen3 tokens of the descriptions). \method
starts from the SFT control at step 800 (the first 800 steps are image-only and shared by both) and adds the
description losses from there, ramped in over 200 steps, with loss weights $w_{\mathrm{img}}{=}1.4$,
$\lambda_{\mathrm{txt}}{=}0.4$ and $\lambda_{\mathrm{desc}}{=}1$. We report step 1{,}800; hardware is in
the appendix. The \emph{SFT control} is plain supervised fine-tuning without descriptions: every
hyperparameter is kept the same (data, steps, batch, LoRA, optimiser and schedule, so both runs see the same
batches and image noise), with $L_e{=}0$ and $\lambda_{\mathrm{desc}}{=}\lambda_{\mathrm{txt}}{=}0$. Everything we
claim about writing is measured against it.

\paragraph{Evaluation.} All benchmark images are generated with the stream off, at $1024^2$, with 28 Euler steps and
guidance 4. We report GenEval \cite{ghosh2023geneval}, DPG-Bench \cite{hu2024ella} for dense prompts, T2I-CompBench \cite{huang2023t2icompbench}, and ImageReward \cite{xu2023imagereward} on the HPSv2 prompts \cite{wu2023hpsv2} as a check that image quality does not regress; blind MLLM judges compare images in \cref{sec:refine}. Each model is trained once, with one training seed; evaluation follows the standard protocol of each benchmark (four sampling seeds per prompt on GenEval and DPG-Bench, ten on T2I-CompBench), and the base models are evaluated under the same protocol.

\subsection{\texorpdfstring{Main results}{Main results}}
\label{sec:results}

\Cref{tab:main} gives the main result: on both backbones, learning to write makes the model draw better. On GenEval, \method raises SD3.5 from $68.4$ to $74.3$ and FLUX.2 [klein]
from $76.3$ to $79.0$, ahead of the SFT control on both ($73.4$ for SD3.5 and $77.2$ for FLUX.2). On SD3.5 fine-tuning on our corpus adds $5.0$ and writing a further $0.9$. On FLUX.2
the control adds $0.9$ and writing a further $1.8$. 
 The largest gain is on what the descriptions state most often: they give each
object's colours in $99\%$ of cases and its position in $83\%$ (appendix), and colour attribution, which asks
which colour belongs to which object, gains $+4.7$ and $+5.3$ over the control.
Counts are stated less often ($66\%$), and counting trails the control on SD3.5 and leads on FLUX.2. Position is where fine-tuning hurts FLUX.2 most: the control falls from $51.5$ to $42.5$, and \method recovers part ($45.3$). These benchmarks check that the content is present; \cref{sec:refine} evaluates how well it is drawn.

DPG-Bench gives the same ordering on dense prompts: \method scores $85.25$ on SD3.5 and $85.36$ on FLUX.2, above the
SFT control ($84.58$, $85.12$) and the base model ($84.20$, $83.15$). On
FLUX.2 simple finetuning provides considerable improvement, and \method adds to it; on
SD3.5 \method's gain in overall performance to the control is larger, and entities, attributes and relations improve over the base model. \method gains over the control where its descriptions are explicit, in entities and attributes, and not in global prompts (style, lighting, mood), which our dataset leaves to other fields.

On T2I-CompBench \method has the highest average on both backbones ($52.58$ and $52.29$; $+0.3$ over the SFT control and
$+2.6$ and $+1.1$ over the base model), with its gains in colour ($+0.8$ and $+0.4$ over the control) and spatial prompts
(2D spatial $+1.0$ and $+1.6$; appendix). GenEval and T2I-CompBench thus agree: \method gains most where a prompt binds colours to objects or places them, which is what its descriptions spell out. ImageReward on the HPSv2 prompts checks that the gains are not bought with image quality: \method scores above the base model and the SFT control on both backbones.
 The appendix places \method next to generation-only and unified models.

\subsection{Finer detail and ambiguous prompts}
\label{sec:refine}

\begin{figure}[t]
  \centering
  \includegraphics[width=\linewidth]{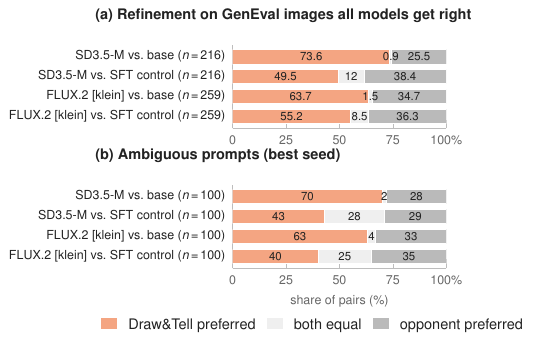}
  \caption{\textbf{Blind judge preferences}. (a)~Refinement on images all models get
  right. (b)~Ambiguous prompts. Share of pairs.}
  \label{fig:judge_sr}
\end{figure}

\begin{figure}[t]
  \centering
  \setlength{\tabcolsep}{1pt}
  \renewcommand{\arraystretch}{0.6}
  \begin{qualband}{SD3.5 Medium}\begin{adjustbox}{max width=\linewidth}%
  \begin{tabular}{@{}ccc@{}}
  \scriptsize SD3.5-M & \scriptsize SFT control & \scriptsize \textcolor{sdcol}{\textbf{\method}} \\
  \rimg[width=0.318\linewidth]{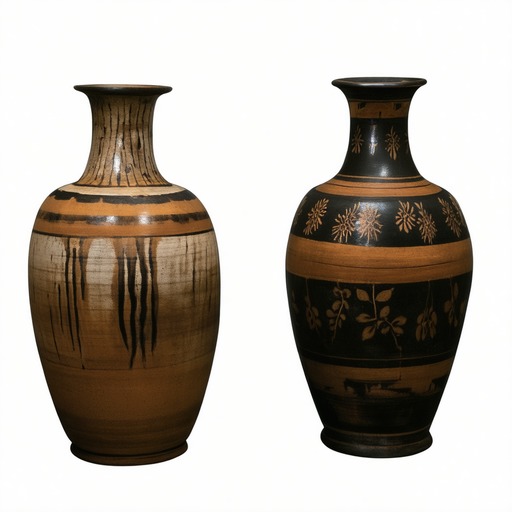} & \rimg[width=0.318\linewidth]{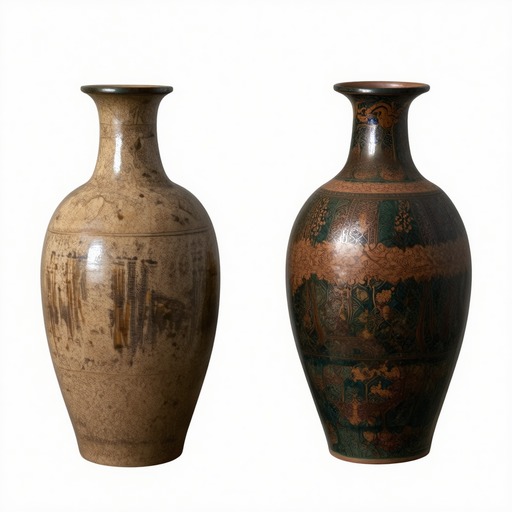} & \rimg[width=0.318\linewidth]{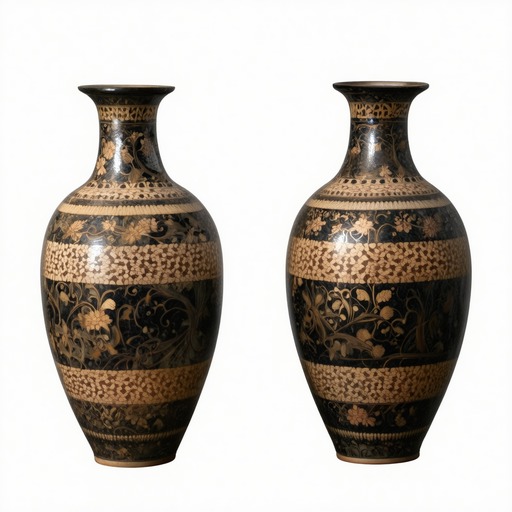} \\[-1pt]
  \multicolumn{3}{c}{\scriptsize\itshape\color{qmute} ``a photo of two vases''} \\
  \end{tabular}\end{adjustbox}\end{qualband}
  \begin{qualband}{FLUX.2 [klein]}\begin{adjustbox}{max width=\linewidth}%
  \begin{tabular}{@{}ccc@{}}
  \scriptsize FLUX.2 [klein] & \scriptsize SFT control & \scriptsize \textcolor{sdcol}{\textbf{\method}} \\
  \rimg[width=0.318\linewidth]{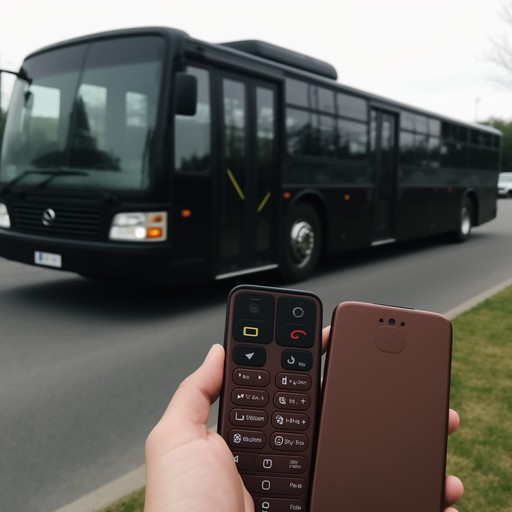} & \rimg[width=0.318\linewidth]{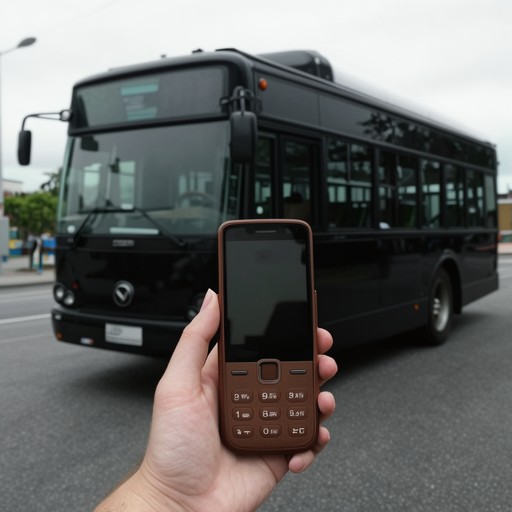} & \rimg[width=0.318\linewidth]{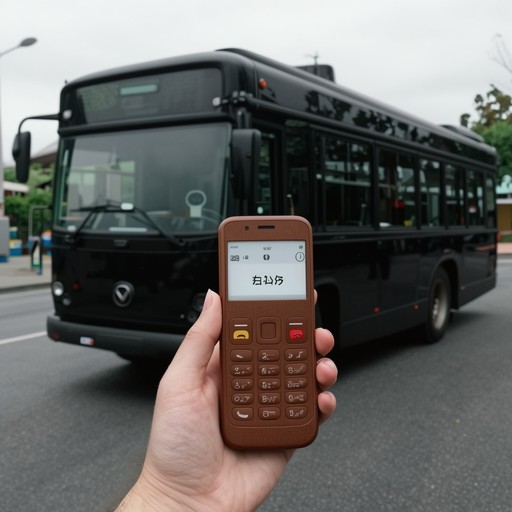} \\
  \multicolumn{3}{c}{\scriptsize\itshape\color{qmute} ``a photo of a black bus and a brown cell phone''} \\
  \end{tabular}\end{adjustbox}\end{qualband}
  \caption{\textbf{Fine details in images.} (GenEval, same prompt and seed). The prompt content is the
  same; what differs is how well it is drawn: the painted
  bands of the vases, the screen and keypad of the phone.}
  \label{fig:qual_refine}
\end{figure}

\method also draws finer detail. GenEval's detector checks whether the requested content is there, not how well it is drawn. We therefore
take, for each backbone, the prompts that the base model, the SFT control and \method all get right, so the judge compares only how the content is drawn. A
blind judge (Claude Opus 5.5) compares each pair on refinement: how crisp, realistic and well resolved the fine textures and small parts are, and may answer that both
are equally good. The detailed prompt is in the appendix.

\method is preferred to the base model on both backbones, and to the SFT control, trained on the same data, batches and noise, by 11 and 19 points (\cref{fig:judge_sr}a); the four-seed averages are in the appendix. \Cref{fig:qual_refine} shows examples; a tie-free variant, the per-category breakdown, single-image ratings,
more examples and the open-weight Qwen3-VL-32B judge, which also prefers \method to the control, are in the appendix. 

\begin{figure}[t]
  \centering
  \includegraphics[width=\linewidth]{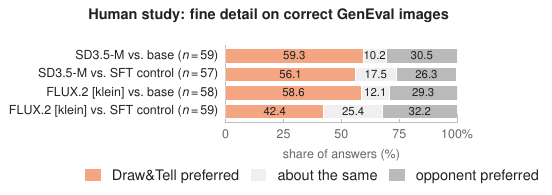}
  \caption{\textbf{Human preferences} on refinement pairs from \cref{fig:judge_sr}a.}
  \label{fig:human}
\end{figure}

\paragraph{Human study.}The user study gives the same result. Asked the judge's question on pairs it compared, participants prefer \method in all four comparisons: to the base model by 29 points on both backbones, and to the SFT control by 30 and 10 points (\cref{fig:human}). A participant agrees with the judge as often as with another participant (appendix).

\vspace{-9pt}
\paragraph{Ambiguous prompts.} GenEval prompts state what to draw. A one-word prompt does not: for
\emph{inertia}, \emph{fairness} or \emph{water} the model has to decide what the image is about. We take the 100 shortest PartiPrompts (appendix) and ask a blind judge, as above, which of two images better answers the open-ended request (criteria in the appendix). \method is preferred to the base model and to the SFT control on both backbones (\cref{fig:judge_sr}b); examples are in the appendix. 

\subsection{\texorpdfstring{Can the description be read back?}{Can the description be read back?}}
\label{sec:ppl}

\begin{table}[t]
\caption{\textbf{Read-out perplexity} of the reference description. Random is a uniform guess over the $151{,}936$-token vocabulary. } 
\label{tab:ppl}
\centering
\footnotesize
\setlength{\tabcolsep}{5pt}
\begin{tabular}{l>{\columncolor{sdours}}ccc}
\toprule
\method on & generated & random & true (ceiling) \\
\midrule
SD3.5 Medium (2.5B) & \textbf{46.3} & 151{,}936 & 13.5 \\
FLUX.2 [klein] (4B) & \textbf{26.2} & 151{,}936 & 17.5 \\
\bottomrule
\end{tabular}
\end{table}

 For 400 held-out prompts we sample the description
jointly with the image from noise and score the reference description by the read-out's teacher-forced perplexity
(\cref{tab:ppl}). As a ceiling, the read-out reads the true encoding of the
description. The generated stream is read back at a perplexity of $46.3$, against
$13.5$ for the true encoding. On FLUX.2 [klein] the same reference descriptions are read back at $26.2$, against $17.5$ for its true encoding. Decoded descriptions (\cref{fig:readout} and the appendix) 
 misstate details: a blind judge finds that most of them talk about
things in the image, but some miss its main subject (appendix), so the words are a readable account of the image, not yet an accurate one.

\begin{table}[t]
\caption{\textbf{Read-out ablation}: read-back perplexity (random: uniform guess over the $151{,}936$-token vocabulary).  }
\label{tab:abl_direct}
\centering
\footnotesize
\setlength{\tabcolsep}{4pt}
\begin{tabular}{lccc}
\toprule
Read-out & generated & true & random \\
\midrule
\rowcolor{sdours}Qwen3-4B layer (\textbf{\method}) & \textbf{46.3} & \textbf{13.5} & 151{,}936 \\
Direct mapping & 3{,}526 & 17.0 & 151{,}936 \\
\bottomrule
\end{tabular}
\end{table}

\paragraph{A light read-out.}
\label{sec:abl_direct} Our read-out uses the generated stream as a prefix for a pre-trained language-model layer. We ablate this
by decoding the stream directly: a separate run trains the stream towards Qwen3 states and maps position $i$ to token
$i{+}1$, without the instruction prefix. The output is degenerate (\cref{tab:abl_direct} and the appendix): the
generated stream is read back at $3{,}526$ against $46.3$ with our read-out,  while the true stream decodes almost as well as with ours ($17.0$ against $13.5$). Sampled states need a language model to be read.  Larger read-outs (the last Qwen3-8B layer, or the T5-XXL decoder) read the stream better but lower GenEval (appendix). 

\subsection{\texorpdfstring{Using the description without writing it}{Using the description without writing it}}
\label{sec:abl_repa}

\begin{table}[t]
\caption{\textbf{Using the description without writing it} (SD3.5). }
\label{tab:abl_repa}
\centering
\footnotesize
\setlength{\tabcolsep}{3pt}
\resizebox{\linewidth}{!}{%
\begin{tabular}{rlcccccccc}
\toprule
& & \multicolumn{7}{c}{GenEval (\%)} & DPG \\
\cmidrule(lr){3-9}
\# & Training & Overall & 1 obj. & 2 obj. & Count. & Colours & Pos. & Col.\ attr. & Overall \\
\midrule
(a) & SFT control & 73.4 & 99.1 & \textbf{90.9} & \textbf{72.8} & 84.8 & 30.5 & 62.3 & 84.58 \\
(b) & Description appended to the prompt & 72.7 & \textbf{100.0} & 88.4 & 66.9 & \textbf{86.2} & 31.5 & 63.0 & 85.13 \\
(c) & REPA: prompt tokens $\rightarrow$ description & 72.0 & 99.7 & 88.4 & 70.9 & 83.8 & 28.0 & 61.3 & 85.07 \\
(d) & REPA: image tokens $\rightarrow$ description & 72.8 & 99.7 & 87.6 & 70.6 & 84.6 & 30.8 & 63.8 & 84.90 \\
\midrule
\rowcolor{sdours}(e) & \textbf{\method} & \textbf{74.3} & 99.7 & 89.4 & 71.6 & 85.6 & \textbf{32.5} & \textbf{67.0} & \textbf{85.25} \\
\bottomrule
\end{tabular}}
\end{table}

To test whether the gain comes from writing the descriptions or from their content, we give the SFT control the same descriptions to align to or train on. Following REPA \cite{yu2024repa}, a head predicts
the frozen T5-XXL features of the description from the prompt or the image tokens after block 8 (cosine loss, weight
$0.5$; discarded after training). As a second control, the description is appended to the prompt text
without a stream. All runs continue the SFT control from step 800 to $1{,}800$ with the data,
noise and schedule of \method. None beats the control on GenEval (\cref{tab:abl_repa}), and colour attribution, \method's largest gain, stays near the control. On DPG-Bench all three lift the control but stay below \method. 

\subsection{Ablating the components}
\label{sec:abl_comp}

\begin{table}[t]
\caption{\textbf{Component ablation} (SD3.5, GenEval; per category in the appendix). }
\label{tab:abl_comp}
\centering
\footnotesize
\setlength{\tabcolsep}{3pt}
\begin{tabular}{rlc}
\toprule
\# & Training & GenEval \\
\midrule
(a) & SFT control (no stream) & 73.4 \\
\midrule
\rowcolor{sdours}(b) & \textbf{\method} & 74.3 \\
\midrule
(c) & no description flow loss (read-out loss only) & 74.1 \\
(d) & no text-only MLPs & 72.0 \\
(e) & text-only MLPs as LoRA (rank 32) & 73.7 \\
(f) & full attention (image and prompt read the stream) & 73.1 \\
\bottomrule
\end{tabular}
\end{table}

\begin{table}[t]
\caption{\textbf{Reading the description at inference} (SD3.5, GenEval): image and prompt tokens read the co-denoised description on the given steps; columns give how each model was trained.}
\label{tab:attn}
\centering
\footnotesize
\setlength{\tabcolsep}{4pt}
\begin{tabular}{lcc}
\toprule
Read on steps & \textbf{\method} (one-way) & Full attention \\
\midrule
all 28 steps & 54.8 & 72.1 \\
0--7 and 27 & 63.7 & 72.9 \\
\rowcolor{sdours}none (stream off) & \textbf{74.3} & 73.1 \\
\bottomrule
\end{tabular}
\end{table}

\Cref{tab:abl_comp} removes one component at a time.
Token-level supervision carries most of the gain: without the description flow loss, the read-out loss alone stays on par with \method. Writing needs its own MLPs: when the description shares the context MLPs with the prompt,
GenEval falls to $72.0$, below the SFT control, with the largest losses on position and two-object prompts
(appendix). LoRA text-only MLPs recover only part of it. This points to interference between reading a prompt and writing a description when they share
weights.  Information should flow one way: full
attention, where the image and prompt also attend to the description during training, lowers GenEval to $73.1$, below the SFT control, and reading the description at inference does not recover the gain (\cref{tab:attn}): the description helps the image as a training signal, not as an input (\cref{sec:why}).
More capacity does not help: full fine-tuning \method falls below control, and text-only MLPs in all 25 FLUX.2 blocks lower GenEval (appendix).

\subsection{Ablating the loss weights}
\label{sec:abl_loss}

\begin{table}[t]
\caption{\textbf{Loss weights} ($\lambda_{\mathrm{txt}}$, $\lambda_{\mathrm{desc}}$ rows change the $w_{\mathrm{img}}{=}1$ run). }
\label{tab:abl_loss}
\centering
\footnotesize
\setlength{\tabcolsep}{3pt}
\resizebox{\linewidth}{!}{%
\begin{tabular}{rlccccccc}
\toprule
\# & Weights & Overall & 1 obj & 2 obj & Count & Col. & Pos. & Attr. \\
\midrule
\rowcolor{sdours}(a) & \textbf{\method}\ ($w_{\mathrm{img}}{=}1.4$) & \textbf{74.3} & 99.7 & 89.4 & 71.6 & 85.6 & \textbf{32.5} & 67.0 \\
(b) & $w_{\mathrm{img}}{=}2.0$ & 73.6 & 99.7 & 90.9 & 71.2 & 85.1 & 29.5 & 65.2 \\
\midrule
(c) & $w_{\mathrm{img}}{=}1$ & 74.1 & 99.7 & \textbf{92.2} & 71.9 & 85.6 & 31.0 & 64.5 \\
(d) & \quad $\lambda_{\mathrm{txt}}{=}0.2$ & 74.1 & 99.7 & \textbf{92.2} & \textbf{72.8} & 85.6 & 30.0 & 64.0 \\
(e) & \quad $\lambda_{\mathrm{txt}}{=}0.8$ & 73.5 & 99.7 & 90.4 & 71.9 & 85.1 & 27.2 & 66.5 \\
(f) & \quad $\lambda_{\mathrm{desc}}{=}0.5$ & 73.6 & 99.7 & 89.9 & 72.5 & \textbf{86.2} & 26.2 & \textbf{67.2} \\
(g) & \quad $\lambda_{\mathrm{desc}}{=}2.0$ & 73.1 & 99.7 & 88.6 & \textbf{72.8} & 84.3 & 28.5 & 64.5 \\
\bottomrule
\end{tabular}}
\end{table}

\Cref{tab:abl_loss} varies one loss weight at a time from the switch point at step 800. It also shows where the gain
comes from: with the image-loss weight left at $1$, the description losses alone lift the SFT control from $73.4$ to
$74.1$ GenEval, so the gain is not a re-weighting of the image loss. The
image-loss weight peaks at $1.4$ (\method); every other setting stays within $1.2$ points of it. Colour attribution, the largest gain, stays at $64.0$--$67.2$ in every setting, above every run that does not write ($61.3$--$63.8$, \cref{tab:abl_repa}).

\subsection{Why does writing help?}
\label{sec:why}
\begin{table}[t]
\caption{\textbf{Blind judge} classifies cause of GenEval colour-attribution image winning. We report two reasons below. \emph{Objects and their colours}: objects present and distinct, each in its own colour; \emph{colour coverage}: how much of an object is coloured.}
\label{tab:why}
\centering
\footnotesize
\setlength{\tabcolsep}{4pt}
\begin{tabular}{l>{\columncolor{sdours}}cc>{\columncolor{sdours}}cc}
\toprule
 & \multicolumn{2}{c}{SD3.5} & \multicolumn{2}{c}{FLUX.2 [klein]} \\
Main reason & \textbf{\method} & SFT & \textbf{\method} & SFT \\
\midrule
Objects and their colours & \textbf{71} & 62 & \textbf{86} & 49 \\
Colour coverage & 35 & \textbf{49} & 28 & \textbf{60} \\
\bottomrule
\end{tabular}
\end{table}
A description is a list of discrete claims about separate things: in a yellow bird beside a purple flower,
each object is a noun with its own colour word. For the language loss to be low, each noun and its properties must be in
the image: the description is written from the states the image is drawn from, so a missing object, a hybrid or a
smeared colour sends a gradient back into the backbone. The image loss, a pixel regression, does not
require this: a blend of both costs little. Nor does aligning to the descriptions (\cref{tab:abl_repa}).
Accordingly, colour attribution is \method's largest GenEval gain ($+4.7$,
$+5.3$). In a blind judge (Claude Opus 5.5) study,
\method wins colour-attribution pairs mainly on the presence of the objects and their
colours, while the control wins on colour coverage, which language does not enforce (\cref{tab:why}). Writing therefore pushes the backbone to bind each colour to its object.  Finer detail and committed readings of
ambiguous prompts (\cref{sec:refine}) reflect the same. The
gradient flows even for a wrong word, so the description need not be accurate; but the image must not read it, or the
description, would make the decisions (\cref{tab:attn}).

\subsection{Limitations}
\label{sec:limitations}
 Descriptions are readable but not yet faithful (\cref{sec:ppl} and the appendix),
and they inherit the biases of our corpus.  We train on one corpus and two backbones of up to 4B parameters; scaling model and data to improve both writing and drawing is the natural next step. With faithful descriptions, the stream could also probe the model's choices, as the image is the same with the stream on or off (\cref{sec:inference}). We also see this as an exciting new avenue of research, in which text data is used to improve text-to-image models.
\vspace{-4pt}
\section{Conclusion}
\label{sec:conclusion}
Show and tell asks a child to put words to what they made; \method asks the same of a text-to-image model, and
finds that it can. The read-only text stream of a pre-trained diffusion transformer becomes read--write. The model draws
better than image-only fine-tuning.
Furthermore, learning to write outperforms aligning with language. Writing has been treated as the other half of a unified model, bought with hundreds of billions of
tokens. Here it is a training signal that a text-to-image model learns in an afternoon, and then sets aside.
\newpage


\clearpage

{
\footnotesize
\setlength{\bibsep}{1pt plus 0.5pt}
\bibliographystyle{ieeenat_fullname}
\bibliography{main}
}
\newpage

\maketitlesupplementary
\appendix
\makeatletter   
\renewcommand\thesection{\ifnum\c@section>26 A\@Alph{\numexpr\c@section-26\relax}\else\@Alph\c@section\fi}
\makeatother

{\makeatletter
\let\supp@contentsline\contentsline
\def\supptocstart{\let\contentsline\supp@contentsline}%
\def\contentsline#1#2#3#4{}%
\c@tocdepth=2
\def\l@section#1#2{\addvspace{3pt}\@dottedtocline{1}{0em}{1.3em}{\bfseries #1}{#2}}%
\def\l@subsection#1#2{\@dottedtocline{2}{1.3em}{2.1em}{#1}{#2}}%
\def\@tocrmarg{2.55em plus 1fil}
\def\numberline#1#2.#3{\hb@xt@\@tempdima{#1.\hfil}}
\tableofcontents
\makeatother}
\addtocontents{toc}{\protect\supptocstart}

\paragraph{Summary.} \Cref{sec:supp_g_train} gives the training details: hardware, backbone-specific choices, hyperparameters, cost and
the course of training; it extends the method (\cref{sec:method} of the main paper) and the setup (\cref{sec:setup} of
the main paper). \Cref{sec:supp_g_data} describes our corpus and checks that its descriptions refer to their images and
that the benchmark prompts are not in it; it extends \cref{sec:data} of the main paper. \Cref{sec:supp_g_results} adds
per-category T2I-CompBench, a comparison with generation-only and unified models, qualitative comparisons, training seeds
and the significance of the GenEval gains; it extends the main results (\cref{sec:results} of the main paper).
\Cref{sec:supp_g_eval} gives the judge prompts and the full fine-detail, human and ambiguous-prompt studies behind
\cref{sec:refine} of the main paper. \Cref{sec:supp_g_readout} shows decoded descriptions in and out of domain and ablates
the read-out decoder, extending \cref{sec:ppl} of the main paper. \Cref{sec:supp_g_abl} reports further ablations:
components by GenEval category, where to add capacity, attention to the description at inference and the image-loss
weight, extending \cref{sec:abl_comp,sec:abl_loss} of the main paper. \Cref{sec:supp_theory} provides theoretical explanation on why writing helps. \Cref{sec:supp_g_why} details the judge study
behind \cref{sec:why} of the main paper. \Cref{sec:supp_failures} shows failure cases of the generated descriptions, to
which the limitations (\cref{sec:limitations} of the main paper) refer. \Cref{sec:supp_open} trains \method on an
open-source dataset, FLUX-Reason-6M.
\makeatletter
\setlength{\@fptop}{0pt}\setlength{\@fpsep}{8pt plus 2pt}\setlength{\@fpbot}{0pt plus 1fil}
\setlength{\@dblfptop}{0pt}\setlength{\@dblfpsep}{8pt plus 2pt}\setlength{\@dblfpbot}{0pt plus 1fil}
\makeatother
\renewcommand{\topfraction}{0.9}\renewcommand{\textfraction}{0.07}\renewcommand{\floatpagefraction}{0.5}
\renewcommand{\dbltopfraction}{0.95}\renewcommand{\dblfloatpagefraction}{0.5}
\setcounter{topnumber}{3}\setcounter{dbltopnumber}{3}
\section{Training details}
\label{sec:supp_g_train}
Hardware, backbone-specific choices, hyperparameters, cost and the course of training.

\subsection{Training hardware}
\label{sec:supp_train}
\method was trained on NVIDIA GH200 GPUs and the SFT control on NVIDIA H200 GPUs, with the same code, seeds and data order,
so both see the same batches and image noise. Learning to describe (from the shared image-only checkpoint to the reported model) takes about 17 GH200 GPU-hours,
under 6 hours on 4 GPUs. Of the loss-weight ablations, the run with image-loss weight $1$ were also trained on GH200. On FLUX.2 [klein] both \method and the SFT control were trained on H200. Within each method, the control SFT and \method was kept on the same partition of GPUs and same data order for fair comparison.

\subsection{Backbone-specific details}
\label{sec:supp_backbone}
\Cref{sec:method} of the main paper describes \method for a generic multimodal diffusion transformer. Where a choice
depends on the backbone, we follow the backbone's own design; this section gives those choices for both backbones.

\paragraph{SD3.5 Medium.} SD3.5 \cite{esser2024scaling} is an MMDiT with $N{=}24$ joint blocks of width $D{=}1536$
over 16-channel VAE latents. A prompt is encoded by CLIP-L and CLIP-G \cite{radford2021learning} (77 tokens) and
T5-XXL \cite{raffel2020exploring} (256 tokens), giving a context sequence of 333 tokens and a pooled CLIP vector. The
noise level follows SD3: $u\sim\mathcal U(0,1)$ is quantised to a 1000-step grid and shifted by
$s\leftarrow\alpha s/(1+(\alpha{-}1)s)$ with $\alpha{=}3$.
\begin{itemize}
\item \emph{Description encoder.} $E_y$ is the backbone's own frozen T5-XXL, with an attention mask so that padding
never enters; we take the last hidden states and divide by $\boldsymbol\sigma_{T5}\in\mathbb R^{4096}$, so
$e_0\in\mathbb R^{L_e\times 4096}$ with $L_e\le 512$.
\item \emph{Stream embedding.} $E_e(e_s)=E_{ctx}(e_s\odot\boldsymbol\sigma_{T5})$: SD3's own context embedder, shared
with the prompt, applied after rescaling the stream to raw T5 scale so that the embedder sees the distribution it was
trained on. $P\in\mathbb R^{512\times D}$ is a learned position table.
\item \emph{Velocity read-out.} After block $r{=}N{-}2$, the last block that emits context tokens; $W_{out}$ maps back
to 4096 dimensions, and $t_{emb}$ is SD3's conditioning vector (timestep embedding plus pooled CLIP vector). The norm's
modulation is zero-initialised as well.
\item \emph{Text-only MLPs.} After blocks $i\in\{0,\dots,N{-}2\}$ (23 MLPs, 597M parameters), each a copy of the
block's context MLP: $\mathrm{MLP}_i(u)=\mathrm{FC}_2(\mathrm{GELU}(\mathrm{FC}_1u))$ with hidden width $4D$.
\item \emph{Read-out.} $\mathrm{in\_proj}:\mathbb R^{4096}\!\to\!\mathbb R^{2560}$ maps T5 space to Qwen3's width.
\item \emph{Loss.} An adaptively weighted $\ell_2$,
$\ell(\hat v,v)=\overline{(\hat v-v)^2}\,/\,\big(\sg[\overline{(\hat v-v)^2}]+c\big)^{1-\gamma}$, where the overline
averages over elements and $\sg[\cdot]$ is stop-gradient, with $\gamma{=}0.5$ and $c{=}10^{-4}$.
\item \emph{LoRA.} Rank 32 on all block linears and the context embedder (47M parameters).
\end{itemize}

\paragraph{FLUX.2 [klein] 4B.} FLUX.2 [klein] has 5 double-stream blocks, with separate text and image weights, and
20 single-stream blocks of width 3072. It encodes the prompt with Qwen3-4B (layers 9, 18 and 27 stacked; 512 tokens)
and keeps its own exponential time shift, which depends on the image resolution.
\begin{itemize}
\item \emph{Description encoder.} $E_y$ is Qwen3-4B, the language model of the read-out; we take its
second-to-last-layer states and divide by a fixed per-channel scale, so $e_0\in\mathbb R^{L_e\times 2560}$ with
$L_e\le 384$.
\item \emph{Stream embedding.} A fresh linear embedder; the stream is appended to the text sequence, with rotary
positions continuing after the prompt, and gets a learned position table $P\in\mathbb R^{L_{\max}\times D}$ as on SD3.5.
\item \emph{Velocity read-out.} From the stream tokens after the last single-stream block; $W_{out}$ maps back to
2560 dimensions.
\item \emph{Text-only MLPs.} After the five double-stream blocks, the only blocks with separate text weights (566M
parameters), each a copy of the block's context MLP, which in FLUX.2 is a SwiGLU MLP with ratio 3:
$\mathrm{MLP}_i(u)=W_2\big(\mathrm{SiLU}(W_1u)\odot W_1'u\big)$, with $W_1$ and $W_1'$ fused in one linear layer.
\item \emph{Read-out.} The stream is already in Qwen3's space; $\mathrm{in\_proj}:\mathbb R^{2560}\!\to\!\mathbb R^{2560}$.
\item \emph{Loss.} A plain mean squared error.
\item \emph{LoRA.} Rank 32 on every linear layer of both block types (48M parameters).
\end{itemize}

\subsection{Training hyperparameters}
\label{sec:supp_hparams}
\Cref{tab:supp_hparams} lists the training hyperparameters of \method on both backbones. The SFT control uses the same
values with the description stream, its losses and the read-out removed: it trains the backbone LoRA on the image loss
alone, on the same data, batches and noise, for the same number of steps. Both start from the same image-only
checkpoint at step $800$, which the SFT control's own first $800$ steps produce.

\begin{table}[t]
\caption{\textbf{Training hyperparameters} of \method. The SFT control shares every value that does not concern the
description stream or the read-out.}
\label{tab:supp_hparams}
\centering
\footnotesize
\setlength{\tabcolsep}{3pt}
\begin{adjustbox}{max width=\linewidth}
\begin{tabular}{lll}
\toprule
 & SD3.5 Medium & FLUX.2 [klein] base 4B \\
\midrule
\multicolumn{3}{l}{\emph{Trainable parameters}} \\
Backbone & LoRA rank $32$ ($47$M) & LoRA rank $32$ ($48$M) \\
Text-only MLPs & every block emitting context & the $5$ double-stream blocks \\
 & tokens ($597$M) & ($566$M) \\
Text-only MLP init. & the block's context MLP & the block's context MLP \\
Read-out & last Qwen3-4B layer, LoRA $16$ & last Qwen3-4B layer, LoRA $16$ \\
\midrule
\multicolumn{3}{l}{\emph{Optimisation}} \\
Optimiser & \multicolumn{2}{l}{8-bit AdamW, weight decay $10^{-4}$, gradient clipping $1.0$} \\
Learning rate & \multicolumn{2}{l}{$5{\cdot}10^{-5}$ for all trainable weights, constant, no warm-up} \\
Batch size & \multicolumn{2}{l}{$128$ (gradient accumulation)} \\
Precision & \multicolumn{2}{l}{bf16, gradient checkpointing} \\
Steps & \multicolumn{2}{l}{$800$ image-only (shared with the SFT control), then $1{,}000$ with descriptions} \\
Seed & \multicolumn{2}{l}{$2026011015$ (LoRA init., data order, noise)} \\
\midrule
\multicolumn{3}{l}{\emph{Losses}} \\
Weights & \multicolumn{2}{l}{$w_{\mathrm{img}}{=}1.4$, $\lambda_{\mathrm{txt}}{=}0.4$, $\lambda_{\mathrm{desc}}{=}1$} \\
Description-loss ramp & linear, $200$ steps & none \\
Read-out input dropout & \multicolumn{2}{l}{$0.1$ (teacher-forced token replaced by $m$)} \\
End of description & \multicolumn{2}{l}{end-of-text token supervised by the read-out} \\
Timestep & \multicolumn{2}{l}{one per micro-batch, uniform on the $1{,}000$-step grid} \\
Timestep shift & $3.0$ (SD3.5 scheduler) & FLUX.2's resolution-dependent shift \\
Prompt dropout & \multicolumn{2}{l}{$0.1$ (null prompt, for classifier-free guidance)} \\
\midrule
\multicolumn{3}{l}{\emph{Data}} \\
Images & \multicolumn{2}{l}{our corpus (\cref{sec:supp_data}), about $1$MP, aspect-ratio buckets} \\
Max.\ description length & $384$ T5 tokens & $384$ Qwen3 tokens \\
Shuffle buffer & $200$ samples per worker & $200$ samples per worker \\
\bottomrule
\end{tabular}
\end{adjustbox}
\end{table}

\subsection{Training and inference cost}
\label{sec:supp_cost}
\Cref{tab:supp_cost} lists what \method adds to the SFT control. Both train the same backbone LoRA ($47$M parameters on
SD3.5, $48$M on FLUX.2 [klein]). \method adds the text-only MLPs, the stream modules (embedder, positions, output
norm and projection) and the read-out's projections and LoRA. In training, the extra description tokens and the
read-out make a step $33\%$ slower: on the same node ($4{\times}$ H200), the same data and the same batch of $128$, a
step of \method takes $8.68$\,s and a step of the SFT control $6.53$\,s of wall-clock time for the whole node, so the
$1{,}000$ description steps take $2.4$ against $1.8$ hours on the node, that is $9.6$ against $7.3$ GPU-hours. At
inference with the stream off, the setting of all our benchmarks, \method costs exactly as much as the SFT control: the
text-only MLPs act only on description tokens, so without a stream they, the stream modules and the read-out are never
run, and the network evaluated is the backbone with its LoRA, which can be merged into the weights. Turning the stream off
also loses none of the gains: the image and prompt tokens never attend to the description, in training or at inference
(\cref{sec:stream,sec:inference}), so the image drawn with the stream off is the same as with it on, and every result we
report with the stream off holds for the model that writes the description.

\begin{table}[t]
\caption{\textbf{Cost of \method} relative to the SFT control. Parameters counted from the checkpoints; training time is
the wall-clock time per step of one node of $4{\times}$ H200 (batch $128$), measured on the same node, data and batch
(SD3.5); GPU-hours are for the $1{,}000$ steps after step $800$ on the four GPUs.}
\label{tab:supp_cost}
\centering
\footnotesize
\setlength{\tabcolsep}{4pt}
\begin{adjustbox}{max width=\linewidth}
\begin{tabular}{lcc}
\toprule
 & SD3.5 & FLUX.2 [klein] \\
\midrule
\multicolumn{3}{l}{\emph{Trainable parameters}} \\
Backbone LoRA (SFT control and \method) & $47.2$M & $48.2$M \\
Text-only MLPs (\method) & $597.2$M & $566.3$M \\
Stream modules (\method) & $11.8$M & $35.8$M \\
Read-out projections and LoRA (\method) & $24.5$M & $20.6$M \\
\midrule
\multicolumn{3}{l}{\emph{Training, per step (batch $128$)}} \\
SFT control & $6.53$\,s & -- \\
\method & $8.68$\,s ($+33\%$) & -- \\
\multicolumn{3}{l}{\emph{Training, $1{,}000$ steps on $4{\times}$ H200}} \\
SFT control & $1.8$\,h ($7.3$ GPU-h) & -- \\
\method & $2.4$\,h ($9.6$ GPU-h) & -- \\
\midrule
\multicolumn{3}{l}{\emph{Inference, stream off}} \\
Extra computation over the SFT control & none & none \\
\bottomrule
\end{tabular}
\end{adjustbox}
\end{table}

\subsection{Training curve}
\label{sec:supp_curve}
\begin{figure}[H]
  \centering
  \includegraphics[width=0.9\linewidth]{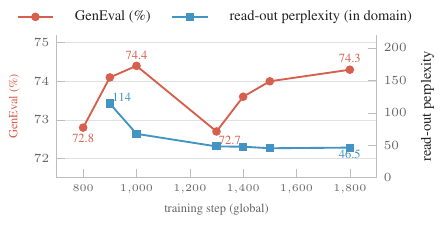}
  \caption{\textbf{GenEval and read-out perplexity of \method over training.}}
  \label{fig:supp_curve}
\end{figure}

\Cref{fig:supp_curve} follows \method over training. GenEval rises from $72.8$ to $74.1$ within 100 steps of adding the
description losses and then stays within about a point of $74$ ($72.7$--$74.4$). The
in-domain read-out perplexity of the generated description falls from $114$ at step 900 to $48$ at step 1{,}300 and levels
off at $45$--$47$. We report step 1{,}800, where both have settled.

\subsection{Losses during training}
\label{sec:supp_losses}
\begin{figure}[H]
  \centering
  \includegraphics[width=\linewidth]{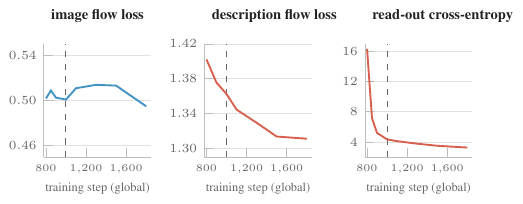}
  \caption{\textbf{Training losses of \method on SD3.5 Medium} (exponential moving average, factor 0.99 for the image loss and
  0.95 otherwise). The dashed line marks the end of the 200-step ramp of the language losses.}
  \label{fig:supp_losses}
\end{figure}

\Cref{fig:supp_losses} shows the training losses from the switch at step 800 to step 1{,}800: the image flow loss, the flow
loss of the description stream and the read-out cross-entropy on the target description. The image flow loss stays flat at
about $0.50$, while the two description losses fall steadily, so the model learns to write the description it is supervised
on rather than reaching a fixed level at the switch. The cross-entropy falls fastest in the first 200 steps, over which the
language losses are ramped in, and keeps decreasing to the end.

\subsection{Training losses on FLUX.2 [klein]}
\label{sec:supp_flux_losses}
\begin{figure}[H]
  \centering
  \includegraphics[width=\linewidth]{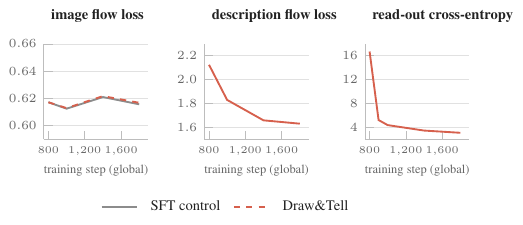}
  \caption{\textbf{Training losses on FLUX.2 [klein] 4B} (exponential moving average, factor 0.99 for the image loss and
  0.95 otherwise), from step 800, where \method starts from the SFT control's checkpoint.}
  \label{fig:supp_flux_losses}
\end{figure}

\Cref{fig:supp_flux_losses} shows the training losses on FLUX.2 [klein] 4B. The SFT control trains from step 0 to
1{,}800; \method starts from its step-800 checkpoint and sees the same samples and noise levels from there on. The image
flow loss of the two runs is the same within about $0.001$ throughout (at step 1{,}800, $0.616$ and $0.617$), so writing the
description costs the image nothing. The description flow loss falls from $2.12$ to $1.63$ and the read-out
cross-entropy from $16.7$ to $3.1$, most of it in the first 100 steps after the switch, and both keep decreasing to the end.

\section{Data}
\label{sec:supp_g_data}
How the descriptions relate to the images, how the corpus was built, and a check that it holds no benchmark prompts.

\subsection{Do the descriptions refer to the image?}
\label{sec:supp_reference}
We use all 400 held-out prompts on both backbones, with greedy
descriptions. The stream has no fixed length: every training description is masked to its own length, so \method can write a
description of any length (\cref{sec:method}). We therefore read it out in two ways: at the length of the prompt's
reference description (the length of its encoding), and at the fixed length used for the other results of this paper
($L_e{=}200$). The numbers cited in the main paper are those at the reference length. A blind judge (Claude Opus 5.5; a
fresh session for every item, which sees one image and one description and nothing else) scores whether the description
talks about the overall subject and the main objects of the image: 0, about something
else entirely; 1, only a generic overlap such as ``a person'' or ``food''; 2, some things in the image but not its main
subject; 3, the main subject, but much that is not in the image; 4, the main subject and most of the main things, a few
that are not there; 5, everything it talks about is in the image.

Each image is also scored with two baseline descriptions of other prompts of the same backbone and length: a random one,
and the nearest one, the description among the 400 held-out ones closest to the image's own description by cosine
similarity of Qwen3-4B embeddings (mean of the last-layer states). The nearest description is a hard baseline: it is
written about the same kind of subject, often a portrait or a dish, but for a different image. The 4{,}800 items (two
backbones, two lengths, 400 images, three descriptions per image) are shuffled together and the judge is not told where a
description comes from.

The descriptions refer to their own image (\cref{tab:supp_reference}). At the reference length they score $3.1$ on SD3.5
and $3.2$ on FLUX.2 [klein]; $90\%$ and $94\%$ of them talk about things in the image (score ${\ge}2$), against $17\%$ and
$22\%$ for a random description, and $72\%$ and $74\%$ name its main subject (${\ge}3$), against $9\%$ and $13\%$. The nearest
description reaches the main subject far more often ($56\%$ and $47\%$), since it is about the same kind of thing, but the
own description still scores higher ($+0.6$ and $+0.9$ points). The description is therefore about the
image it accompanies.

\paragraph{Variable length.} Since the stream has no fixed length, the description can be given the length its content
needs. At the fixed length $L_e{=}200$ the descriptions refer to their image as often ($91\%$ and $94\%$ score ${\ge}2$; $+0.5$
and $+0.9$ points over the nearest description), but they are less exact: on SD3.5 the share that mentions nothing that is
not in the image halves ($6\%$ against $13\%$), the share that names most of the main things falls from $39\%$ to $30\%$, and
the mean score from $3.13$ to $2.96$; on FLUX.2 [klein] the share scoring 5 falls from $11\%$ to $8\%$. A length that fits the
content leaves less room for things that are not in the image, which a stream of fixed length cannot offer.

At the reference length $28\%$ of the descriptions on SD3.5 and $26\%$ on FLUX.2 [klein] miss the main subject
(score ${\le}2$). Making the description faithful in its details is left to future work.

\begin{table*}[!t]
\caption{\textbf{Do the descriptions refer to the image?} Blind judge score (0--5) for the image's own description
and two baseline descriptions of other prompts, and the share of descriptions that talk about things in the image (score ${\ge}2$), name its main subject
(${\ge}3$), most of its main things (${\ge}4$) or nothing that is not in it ($5$). All 400 held-out prompts, greedy decoding,
read out at the length of the reference description (top; the numbers cited in the main paper) and at the fixed length
$L_e{=}200$ (bottom); one fresh judge per item.}
\label{tab:supp_reference}
\centering\footnotesize
\setlength{\tabcolsep}{3.5pt}
\begin{tabular}{@{}lcccccccccc@{}}
\toprule
& \multicolumn{5}{c}{SD3.5 Medium} & \multicolumn{5}{c}{FLUX.2 [klein]} \\
\cmidrule(lr){2-6}\cmidrule(lr){7-11}
Description & Score & ${\ge}2$ (\%) & ${\ge}3$ (\%) & ${\ge}4$ (\%) & $5$ (\%) & Score & ${\ge}2$ (\%) & ${\ge}3$ (\%) & ${\ge}4$ (\%) & $5$ (\%) \\
\midrule
\multicolumn{11}{@{}l}{\emph{Reference length}} \\
\rowcolor{sdours}Own (\method) & \textbf{3.13} & \textbf{90} & \textbf{72} & \textbf{39} & \textbf{13} & \textbf{3.20} & \textbf{94} & \textbf{74} & \textbf{41} & \textbf{11} \\
Nearest of another prompt & 2.49 & 76 & 56 & 19 & 5 & 2.27 & 72 & 47 & 15 & 3 \\
Random of another prompt & 0.73 & 17 & 9 & 3 & 1 & 0.89 & 22 & 13 & 3 & 0 \\
Own vs.\ nearest: wins / ties / losses & \multicolumn{5}{c}{203 / 145 / 52} & \multicolumn{5}{c}{251 / 119 / 30} \\
\midrule
\multicolumn{11}{@{}l}{\emph{$L_e{=}200$}} \\
\rowcolor{sdours}Own (\method) & \textbf{2.96} & \textbf{91} & \textbf{70} & \textbf{30} & \textbf{6} & \textbf{3.19} & \textbf{94} & \textbf{79} & \textbf{38} & \textbf{8} \\
Nearest of another prompt & 2.43 & 74 & 54 & 18 & 4 & 2.27 & 73 & 46 & 12 & 2 \\
Random of another prompt & 0.79 & 21 & 12 & 1 & 0 & 0.83 & 22 & 10 & 1 & 0 \\
Own vs.\ nearest: wins / ties / losses & \multicolumn{5}{c}{179 / 176 / 45} & \multicolumn{5}{c}{252 / 119 / 29} \\
\bottomrule
\end{tabular}
\end{table*}

\subsection{Dataset}
\label{sec:supp_data}
\paragraph{Sources.} Our
dataset holds 114.7k images. Prompts written by users of
jimeng, lexica and prompthero are expanded by Qwen3.5-35B-A3B \cite{qwen35} into a structured prompt, which
Z-Image-Turbo \cite{zimage} renders. The same model then checks each image against its prompt and discards those
that do not match; where the image differs only in details, it rewrites the prompt to match the image. The resulting
prompt, translated to English, is the short prompt $c$. We hold out
6{,}040 prompts (spread over the three sources) for evaluation and exclude them from training; the read-out perplexities
and description qualitatives use 400 of them.

\paragraph{Structured description.} A vision-language model, Qwen3.5-35B-A3B \cite{qwen35}, which sees only the image and not the prompt, describes each image in nine fields
(\cref{tab:supp_fields}). The description $y$ that \method learns to write is the subject field. It is organised by
entity: one labelled block per object or person with its count, appearance and colours, its position in the frame (left,
right, centre, foreground, background, on top of, in front of) and its relations to the other objects (holding, standing
on, next to, wearing). Setting, lighting, camera and style are left to the other fields, so the subject field states the
composition of the image and nothing else.

\begin{table}[H]
\caption{\textbf{Fields of the structured description.} \method learns to write the subject field.}
\label{tab:supp_fields}
\centering\footnotesize
\begin{adjustbox}{max width=\linewidth}
\begin{tabular}{@{}lll@{}}
\toprule
Field & Present & Content \\
\midrule
\textbf{Subject} ($y$) & always & objects and people, count, appearance, colours, position, relations \\
Action & optional & actions, expression, gaze, interaction \\
Typography & if text is visible & position, style, literal string, colour of the text \\
Context & optional & environment, background objects, depth \\
Colour & optional & palette, per-object colours, light source \\
Layout & optional & per-subject position, frame proportion, composition \\
Medium & always & medium category (2--4 words) \\
Medium detail & optional & depth of field, light, brushwork, substrate \\
Style & if distinctive & aesthetic style \\
\bottomrule
\end{tabular}
\end{adjustbox}
\end{table}

\paragraph{Statistics.} The descriptions hold 13.3M text tokens (Qwen3 tokens), against 451M image tokens; a description
has a median of 86 words (128 tokens, at most 512 are used in training). On 300 held-out descriptions, colours are stated
in $99\%$, positions in $83\%$, relations between objects in $69\%$ and counts in $66\%$; $65\%$ begin with an entity
label (``Young man:'').

\begin{figure*}[!tp]
\centering
\begin{tcbraster}[raster columns=2,raster equal height,raster column skip=5pt,raster left skip=0pt,raster right skip=0pt]
\begin{tcolorbox}[judgebox=sdours]
\qchip{sdband}{Image}\\[2pt]
\centerline{\rimg[width=0.62\linewidth]{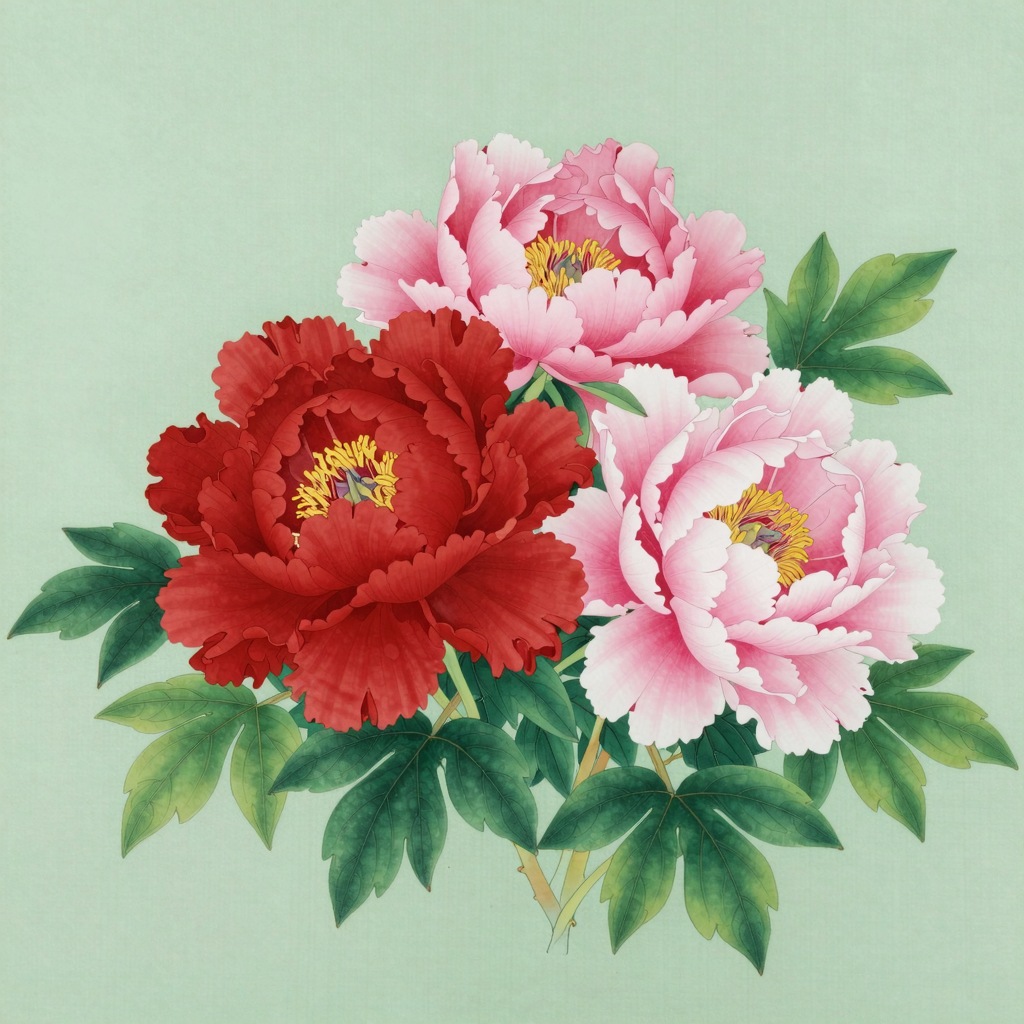}}\\[3pt]
\qchip{sdband}{Prompt $c$}\\[2pt]
{\footnotesize\itshape Beautiful Chinese peonies decorate the soft light green canvas}\\[4pt]
\qchip{sdband}{Description $y$}\\[2pt]
{\footnotesize Three peonies clustered together: Left peony: deep red, petals layered tightly in a spherical shape, with golden
yellow stamens; Top peony: pink, petals wide and spread out, edges wavy, with golden yellow stamens; Right peony: light pink,
petals layered multiple times, shape spread out, with golden yellow stamens; Green leaves: multiple green leaves, palmately
lobed, with clear veins, surrounding the flowers; Stem: light yellow-green, slender, partially visible.}
\end{tcolorbox}
\begin{tcolorbox}[judgebox=sdours]
\qchip{sdband}{Image}\\[2pt]
\centerline{\rimg[width=0.62\linewidth]{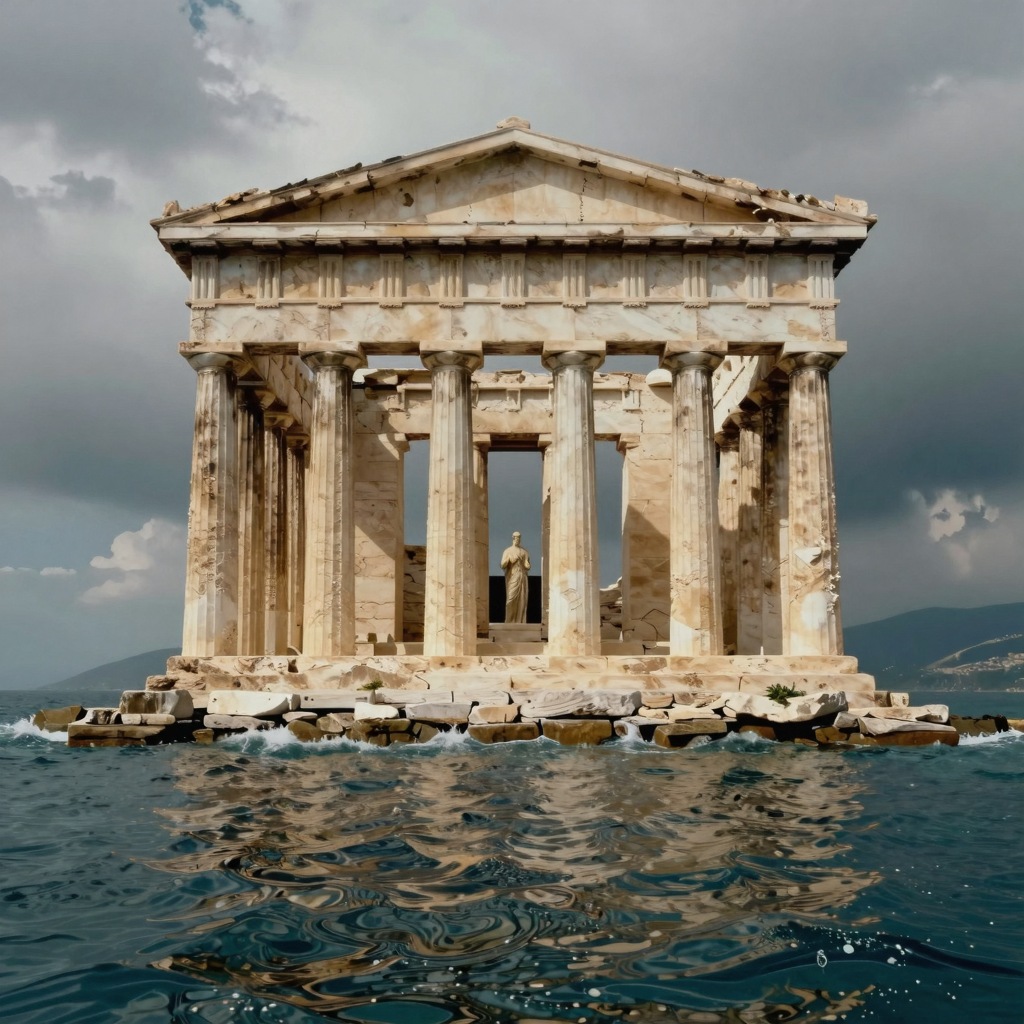}}\\[3pt]
\qchip{sdband}{Prompt $c$}\\[2pt]
{\footnotesize\itshape A Greek temple floating in the middle of the sea, digital rendering, cinematic composition.}\\[4pt]
\qchip{sdband}{Description $y$}\\[2pt]
{\footnotesize A ruined site of a Greek temple-style building: the main structure is a rectangular hall, constructed of beige
marble with weathered and weathered traces; on the front, six Doric columns are visible, with several columns visible at the
rear and sides, the column shafts having vertical grooves; the top features a triangular gable and cornice, with edges damaged;
the building's base is made of rough rock, partially submerged in water; within the niche stands a white classical-style human
statue, draped in a long robe, in a static pose.}
\end{tcolorbox}
\end{tcbraster}
\caption{\textbf{Two samples of our corpus.} The prompt is what the user wrote; the description states the objects, their
counts, colours and positions, which the prompt leaves out. Images are decoded from the stored training latents.}
\label{fig:supp_data_ex}
\end{figure*}

\paragraph{Examples.} \Cref{fig:supp_data_ex} shows two samples of the corpus: the short prompt $c$ that a user wrote and the
description $y$ that \method learns to write. Texts are unedited, except that the trailing Chinese full stop of our description
format is printed as a period.

\subsection{Benchmark prompts in the corpus}
\label{sec:supp_contam}
Our corpus is built from prompts written by users of public image generators (\cref{sec:supp_data}). To make sure that
our GenEval and DPG-Bench results do not come from training on the benchmarks' own prompts, we search the corpus for
every prompt of both benchmarks. The search covers all text a training sample can carry, a superset of what training
reads: the short prompt and the description of all $114{,}724$ samples. Neither benchmark has reference images (both score images
generated from their prompts), so overlap can only enter through the prompts. All text is lower-cased and stripped of
punctuation before matching, and we run five tests (\cref{tab:supp_contam}):
\begin{itemize}
\item \emph{exact match} of a benchmark prompt with a corpus text;
\item \emph{containment}: the benchmark prompt occurs, word for word, inside a longer corpus text;
\item \emph{long word sequences}: any 13 consecutive words of a DPG-Bench prompt found in the corpus, and the share of each prompt's 8-word sequences found in it;
\item \emph{compositions}: for GenEval's colour-attribution prompts, a corpus text that contains both attribute--object
phrases (\eg \emph{red umbrella} and \emph{green cow}); for its position prompts, a corpus text that names both
objects with the requested relation between them; for its two-object prompts, the object phrase (\emph{a fork and a
knife});
\item \emph{nearest neighbour}: the most similar corpus text by TF-IDF cosine similarity.
\end{itemize}

\paragraph{Result.} No prompt of either benchmark occurs in the corpus. No GenEval or DPG-Bench prompt matches a corpus
text exactly, no DPG-Bench prompt is contained in one, and none shares a 13-word sequence with it; on average
$0.2\%$ of a DPG-Bench prompt's 8-word sequences occur somewhere in the corpus (at most $7\%$).
GenEval's prompts are template sentences over the 80 COCO object classes, so the shortest of them are
ordinary English: \emph{a photo of a dog} or \emph{a photo of a person} occurs inside a longer prompt for $14$ of the
$80$ single-object prompts, and \emph{a photo of a black backpack} for one colour prompt. The categories that test
composition do not occur at all: no corpus text binds the two colours of a colour-attribution prompt to its two
objects ($0$ of $100$), none places the two objects of a position prompt in the requested relation ($0$ of $100$), and
the object phrase of one two-object prompt (\emph{a fork and a knife}) occurs in two corpus texts. The nearest corpus text
is a different scene (\cref{tab:supp_contam}, bottom): \emph{a photo of a fire hydrant} is closest to \emph{cartoon
fire hydrant in bold red}, and \emph{a photo of a stop sign} to \emph{red stop sign with white ``STOP'' text}.

Therefore, the gains we observe are the outcome of writing, not of benchmark prompts seen in training. Colour
attribution and position, the two categories where \method gains most (\cref{sec:why}), are exactly the ones whose
prompts are absent from the corpus, and the SFT control, which trains on the same corpus, data order and noise without
writing, sees the same prompts as \method.

\begin{table}[t]
\caption{\textbf{Benchmark prompts in our corpus}. Counts of benchmark prompts that pass each test against the
$549{,}448$ corpus texts (short prompts, descriptions and prompt files; \cref{sec:supp_contam}). Bottom: cosine
similarity of each prompt's nearest corpus text (TF-IDF).}
\label{tab:supp_contam}
\centering
\footnotesize
\setlength{\tabcolsep}{4pt}
\begin{adjustbox}{max width=\linewidth}
\begin{tabular}{lcc}
\toprule
Test & GenEval ($553$) & DPG-Bench ($1{,}065$) \\
\midrule
Exact match with a corpus text & $0$ & $0$ \\
Prompt contained in a corpus text & $15$\textsuperscript{a} & $0$ \\
Shares a 13-word sequence & -- & $0$ \\
8-word sequences found (mean / max share) & -- & $0.2\%$ / $7\%$ \\
\midrule
Colour attribution: both attribute--object phrases & $0$ / $100$ & -- \\
Position: both objects in the requested relation & $0$ / $100$ & -- \\
Two objects: object phrase & $1$ / $99$ & -- \\
\midrule
Nearest corpus text, cosine: median & $0.59$ & $0.31$ \\
\hspace{1em}99th percentile / max & $0.82$ / $0.85$ & $0.57$ / $0.64$ \\
\bottomrule
\end{tabular}
\end{adjustbox}
\\[2pt]{\scriptsize \textsuperscript{a}$14$ single-object prompts (\eg \emph{a photo of a dog}) and one colour prompt
(\emph{a photo of a black backpack}), each inside a longer prompt.}
\end{table}

\section{Additional results}
\label{sec:supp_g_results}
Per-category T2I-CompBench, unified models, more qualitative comparisons, training seeds and the significance of the GenEval gains.

\subsection{T2I-CompBench}
\label{sec:supp_t2icomp}
\Cref{tab:supp_t2icomp} gives T2I-CompBench \cite{huang2023t2icompbench} per category. We use the official validation
prompts (300 per category), ten images per prompt from fixed seeds, the sampling settings of the main paper and the
official scorers: BLIP-VQA for colour, shape and texture, UniDet for 2D spatial, 3D spatial and numeracy, CLIPScore for
non-spatial and the 3-in-1 score for complex prompts. The base SD3.5 Medium and FLUX.2 [klein] rows are not quoted: we
reproduce both base models under this same protocol, so that every row in our blocks shares prompts, seeds, sampler and scorers. The average in \cref{tab:main} is the unweighted mean of the eight
categories.

On SD3.5 \method has the highest average ($52.58$, against $52.32$ for the SFT control and $50.01$ for the base model) and is
highest on colour, texture, 2D and 3D spatial and complex prompts. Over the SFT control, \method gains most on 2D spatial
($+1.0$) and colour ($+0.8$), followed by texture ($+0.3$), and is level with it (within $0.1$ points) on shape, non-spatial
and numeracy. On FLUX.2 [klein] \method also has the highest average ($52.29$ against
$51.96$ and $51.20$) and is highest on colour, 2D and 3D spatial and complex prompts; the control is ahead on texture and
numeracy, and the base model on shape. The largest change on both backbones is in spatial relations: on SD3.5 2D spatial
rises from $28.06$ to $35.02$ and 3D spatial from $37.98$ to $42.66$.

Prior results are quoted from the papers and serve as context only. The first block comes from T2I-CompBench++
\cite{huang2025t2icompbenchpp}, which uses the same protocol and reports all eight categories. The second block collects
compositional methods that report a subset of the categories; their baselines and protocols are not always stated and
differ between papers (for example, SD3.5 Medium as scored in \cite{lv2025taca} is $0.6$ points above our measurement on
colour and $2.4$ on texture, and $7.2$ below on 2D spatial), so only rows within one block, or within one paper, compare
directly.

\begin{table*}[!tp]
\caption{\textbf{T2I-CompBench per category} ($\times 100$). Our rows, including the base SD3.5 Medium and FLUX.2 [klein],
are reproduced under our protocol: official scorers, 300 validation prompts per category, ten images per prompt, stream off. Prior rows are quoted from the cited papers.}
\label{tab:supp_t2icomp}
\centering
\footnotesize
\setlength{\tabcolsep}{4pt}
\renewcommand{\arraystretch}{1.06}
\begin{adjustbox}{max width=\textwidth}
\begin{tabular}{llccccccccc}
\toprule
& & \multicolumn{3}{c}{Attribute binding} & \multicolumn{3}{c}{Object relationship} & & & \\
\cmidrule(lr){3-5}\cmidrule(lr){6-8}
Model & Built on & Colour & Shape & Texture & 2D spatial & 3D spatial & Non-spatial & Numeracy & Complex & Avg.$\uparrow$ \\
\midrule
\multicolumn{11}{@{}l}{\textit{Quoted from T2I-CompBench++~\cite{huang2025t2icompbenchpp}, all eight categories}} \\
SD v1.4~\cite{rombach2022high} & -- & 37.65 & 35.76 & 41.56 & 12.46 & 30.30 & 30.79 & 44.56 & 30.80 & 32.98 \\
SD v2~\cite{rombach2022high} & -- & 50.65 & 42.21 & 49.22 & 13.42 & 32.30 & 31.27 & 45.82 & 33.86 & 37.34 \\
Composable Diffusion~\cite{liu2022composable} & SD v2 & 40.63 & 32.99 & 36.45 & 8.00 & 28.47 & 29.80 & 42.72 & 28.98 & 31.00 \\
Structured Diffusion~\cite{feng2023structured} & SD v2 & 49.90 & 42.18 & 49.00 & 13.86 & 32.24 & 31.11 & 45.57 & 33.55 & 37.18 \\
Attend-and-Excite~\cite{chefer2023attend} & SD v2 & 64.00 & 45.17 & 59.63 & 14.55 & 32.22 & 31.09 & 47.73 & 34.01 & 41.05 \\
GORS~\cite{huang2023t2icompbench} & SD v2 & 66.03 & 47.85 & 62.87 & 18.15 & 35.72 & 31.93 & 48.30 & 33.28 & 43.02 \\
SDXL~\cite{podell2024sdxl} & -- & 58.79 & 46.87 & 52.99 & 21.33 & 35.66 & 31.19 & 49.91 & 32.37 & 41.14 \\
PixArt-$\alpha$~\cite{chen2024pixartalpha} & -- & 66.90 & 49.27 & 64.77 & 20.64 & 39.01 & 31.97 & 50.32 & 34.33 & 44.65 \\
DALL-E 3$^{\ddagger}$~\cite{dalle3} & -- & 77.85 & 62.05 & 70.36 & 28.65 & 37.44 & 30.03 & 59.26 & 37.73 & 50.42 \\
SD3$^{\S}$~\cite{esser2024scaling} & -- & 81.32 & 58.85 & 73.34 & 32.00 & 40.84 & 31.40 & 61.74 & 37.71 & 52.15 \\
FLUX.1 [schnell]~\cite{flux2024} & -- & 74.07 & 57.18 & 69.22 & 28.63 & 38.66 & 31.27 & 61.85 & 37.03 & 49.74 \\
\midrule
\multicolumn{11}{@{}l}{\textit{Compositional methods, quoted from their papers (categories they report)}} \\
CoMat-SDXL~\cite{jiang2024comat} & SDXL & 78.27 & 53.29 & 64.68 & 24.28 & -- & 31.87 & -- & 36.80 & -- \\
ELLA-SDXL~\cite{hu2024ella} & SDXL & 72.60 & 56.34 & 66.86 & 22.14 & -- & 30.69 & -- & -- & -- \\
IterComp~\cite{zhang2025itercomp} & SDXL & 79.82 & 62.17 & 76.83 & 31.96 & -- & 33.71 & -- & 48.73 & -- \\
RPG$^{\P}$~\cite{yang2024rpg} & SDXL & 83.35 & 68.01 & 81.29 & 45.47 & -- & 34.62 & -- & 54.08 & -- \\
Janus-Pro-7B (from \cite{jiang2025t2ir1}) & -- & 63.59 & 35.28 & 49.36 & 20.61 & -- & 30.85 & -- & 35.59 & -- \\
T2I-R1~\cite{jiang2025t2ir1} & Janus-Pro-7B & 81.30 & 58.52 & 72.43 & 33.78 & -- & 30.90 & -- & 39.93 & -- \\
Emu3-Gen~\cite{wang2024emu3} & -- & 61.07 & 47.34 & 61.78 & -- & -- & -- & -- & -- & -- \\
SD3 Medium (from \cite{li2025dual}) & -- & 79.93 & 58.16 & 73.89 & -- & -- & -- & -- & -- & -- \\
D-DiT~\cite{li2025dual} & SD3 Medium & 80.01 & 57.03 & 68.56 & -- & -- & -- & -- & -- & -- \\
SD3.5 Medium (from \cite{lv2025taca}) & -- & 78.90 & 57.70 & 73.28 & 20.87 & -- & 31.04 & -- & 44.41 & -- \\
TACA~\cite{lv2025taca} & SD3.5 Medium & 80.74 & 59.38 & 75.22 & 26.78 & -- & 31.06 & -- & 44.70 & -- \\
\midrule
\rowcolor{sdband}\multicolumn{11}{l}{\textbf{Ours (SD3.5 Medium)}} \\
Base (our measurement) & -- & 78.26 & 57.36 & 70.86 & 28.06 & 37.98 & 31.32 & 59.84 & 36.41 & 50.01 \\
SFT control (no stream) & -- & 81.33 & \textbf{57.84} & 71.28 & 34.05 & 42.51 & \textbf{31.35} & \textbf{63.05} & 37.11 & 52.32 \\
\rowcolor{sdours}\textbf{\method} & -- & \textbf{82.12} & 57.76 & \textbf{71.59} & \textbf{35.02} & \textbf{42.66} & 31.32 & 63.02 & \textbf{37.19} & \textbf{52.58} \\
\midrule
\rowcolor{fluxband}\multicolumn{11}{l}{\textbf{Ours (FLUX.2 [klein] 4B)}} \\
Base (our measurement) & -- & 80.47 & \textbf{56.31} & 70.71 & 38.24 & 36.94 & 30.78 & 59.37 & 36.83 & 51.20 \\
SFT control (no stream) & -- & 81.80 & 54.76 & \textbf{72.18} & 37.16 & 39.61 & \textbf{31.00} & \textbf{60.87} & 38.30 & 51.96 \\
\rowcolor{fluxours}\textbf{\method} & -- & \textbf{82.23} & 55.43 & 71.44 & \textbf{38.77} & \textbf{40.37} & 30.99 & 60.53 & \textbf{38.55} & \textbf{52.29} \\
\bottomrule
\end{tabular}
\end{adjustbox}
\\[2pt]
{\footnotesize $^{\ddagger}$Three images per prompt without fixed seeds \cite{huang2025t2icompbenchpp}.
$^{\S}$The checkpoint is not stated in \cite{huang2025t2icompbenchpp}.
$^{\P}$Training-free; plans the layout with GPT-4.}
\end{table*}

\subsection{Comparison with unified models}
\label{sec:supp_unified}
The comparison is indicative only: the baselines differ in backbone, scale and budget. Its point is the cost of the
language ability. Unified models trained from scratch reach their scores with tens to hundreds of billions of tokens;
the two that, like us, start from an SD3 Medium checkpoint use 30M pairs (OmniFlow \cite{li2025omniflow}, which
matches its base) and 40M pairs (D-DiT \cite{li2025dual}, three GenEval points above its base while losing nine on
counting). \method gains $5.9$ GenEval points with minimal resources, 13M text and 0.45B image tokens and about 17 GH200 GPU-hours
(under 6 hours on 4 GPUs) of description training, keeps the base model's input pathway and inference cost, and on DPG-Bench is above every
generation-only and unified baseline in this comparison.

\Cref{tab:unified,tab:dpg} place \method next to generation-only and unified models on GenEval and DPG-Bench. Baseline budgets are fine-tuning data for models initialised from a text-to-image model and pre-training tokens otherwise (Show-o reports 0.68 at $512^2$ in its own paper); our rows use the official scorers under the protocol of the main paper. The baselines differ from ours in backbone, scale and budget, so the controlled comparisons are the last two blocks of each table.

\subsection{Additional qualitative comparisons}
\label{sec:supp_qual}
\begin{figure*}[t]
  \centering
  \setlength{\tabcolsep}{0.6pt}
  \renewcommand{\arraystretch}{0.6}
  \begin{qualband}{SD3.5 Medium}\begin{adjustbox}{max width=\linewidth}%
  \begin{tabular}{@{}ccc@{\hspace{5pt}}ccc@{\hspace{5pt}}ccc@{}}
  \scriptsize SD3.5-M & \scriptsize SFT control & \scriptsize \textcolor{sdcol}{\textbf{\method}} & \scriptsize SD3.5-M & \scriptsize SFT control & \scriptsize \textcolor{sdcol}{\textbf{\method}} & \scriptsize SD3.5-M & \scriptsize SFT control & \scriptsize \textcolor{sdcol}{\textbf{\method}} \\
  \rimg[width=0.104\textwidth]{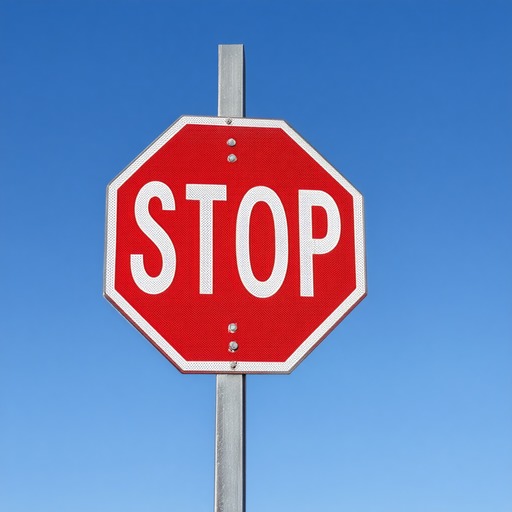} & \rimg[width=0.104\textwidth]{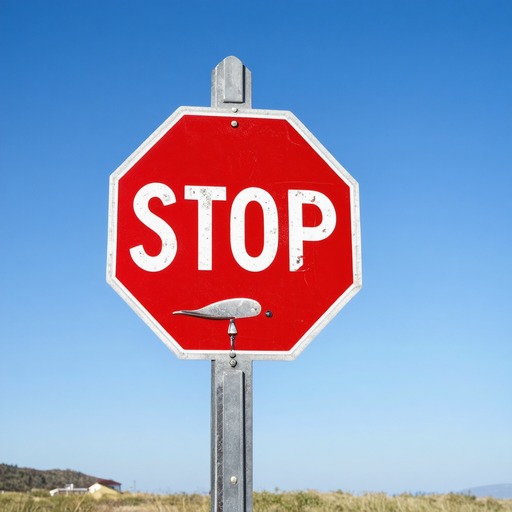} & \rimg[width=0.104\textwidth]{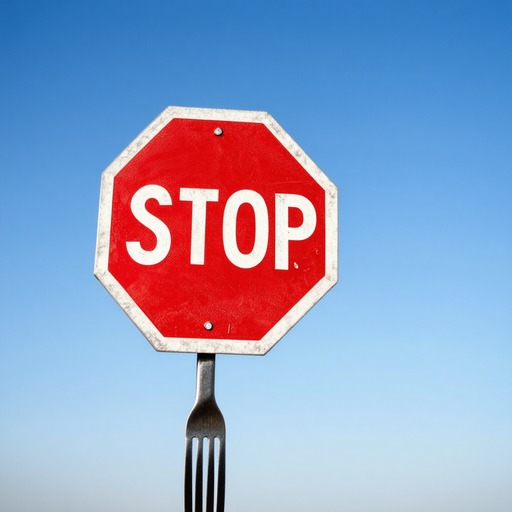} & \rimg[width=0.104\textwidth]{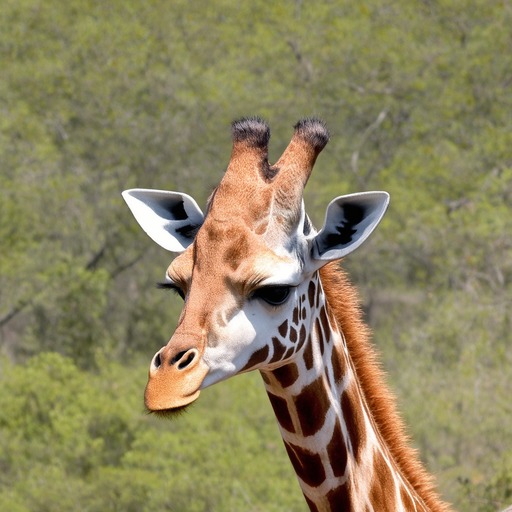} & \rimg[width=0.104\textwidth]{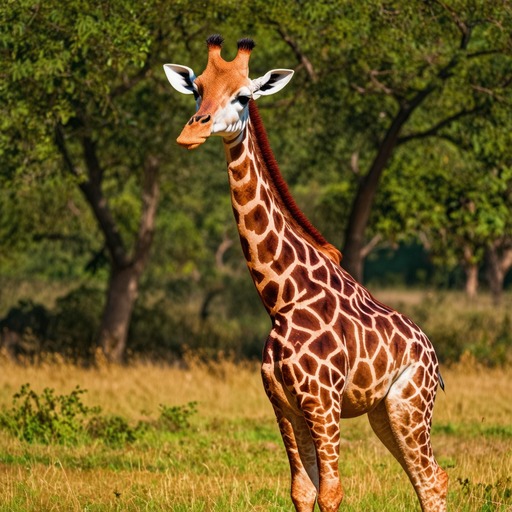} & \rimg[width=0.104\textwidth]{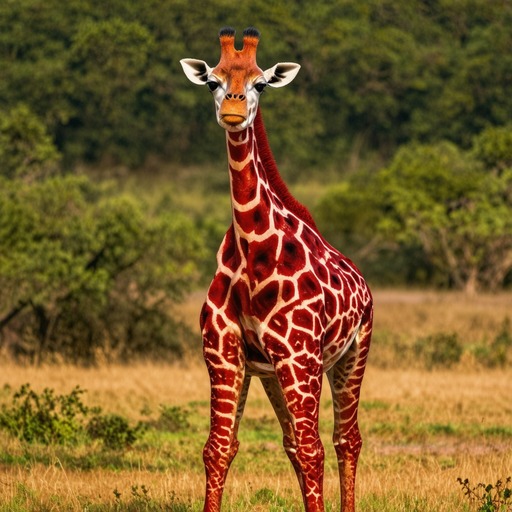} & \rimg[width=0.104\textwidth]{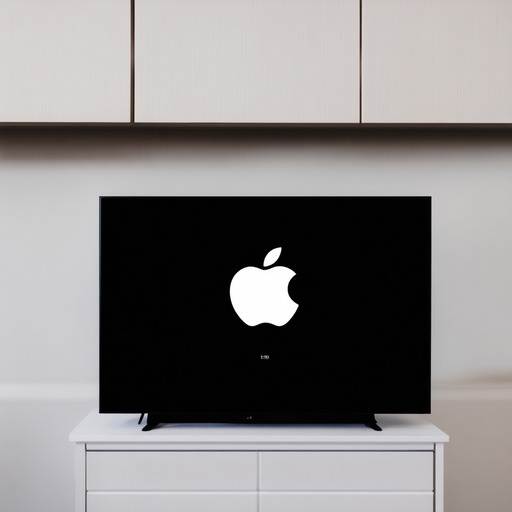} & \rimg[width=0.104\textwidth]{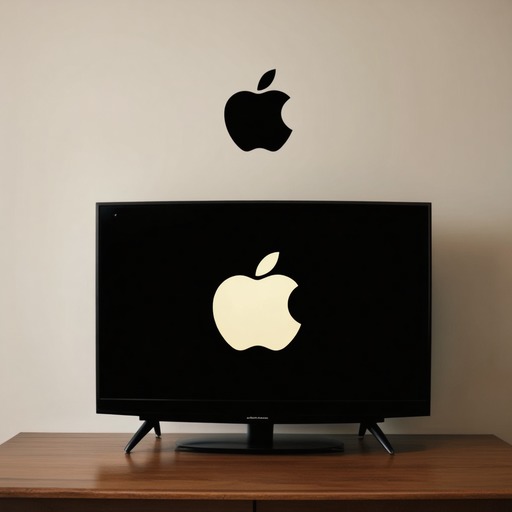} & \rimg[width=0.104\textwidth]{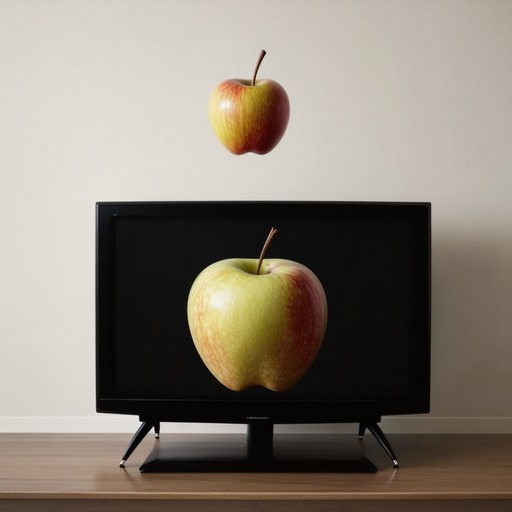} \\[-1pt]
  \multicolumn{3}{c}{\scriptsize\itshape\color{qmute} ``a photo of a stop sign and a fork''} & \multicolumn{3}{c}{\scriptsize\itshape\color{qmute} ``a photo of a red giraffe''} & \multicolumn{3}{c}{\scriptsize\itshape\color{qmute} ``a photo of an apple above a tv''} \\[5pt]
  \scriptsize SD3.5-M & \scriptsize SFT control & \scriptsize \textcolor{sdcol}{\textbf{\method}} & \scriptsize SD3.5-M & \scriptsize SFT control & \scriptsize \textcolor{sdcol}{\textbf{\method}} & \scriptsize SD3.5-M & \scriptsize SFT control & \scriptsize \textcolor{sdcol}{\textbf{\method}} \\
  \rimg[width=0.104\textwidth]{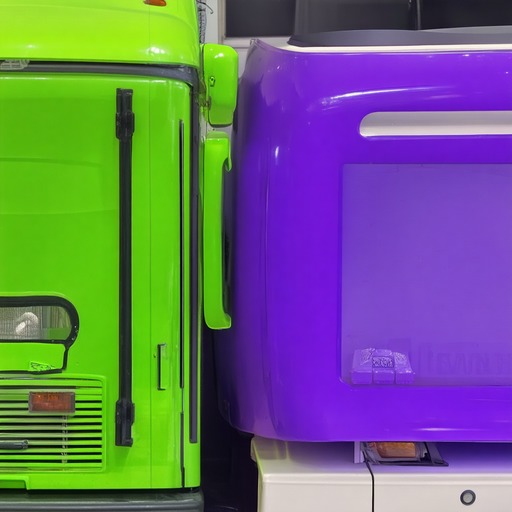} & \rimg[width=0.104\textwidth]{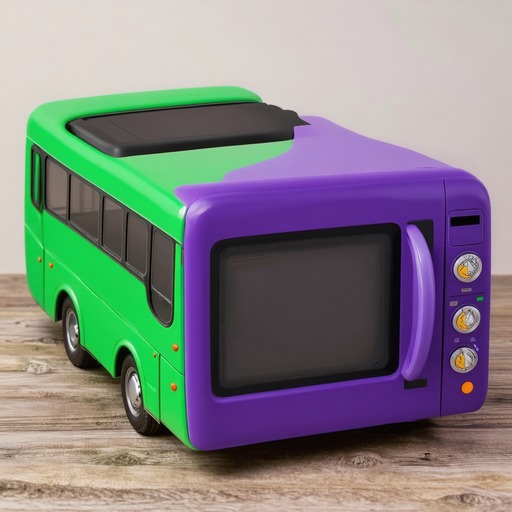} & \rimg[width=0.104\textwidth]{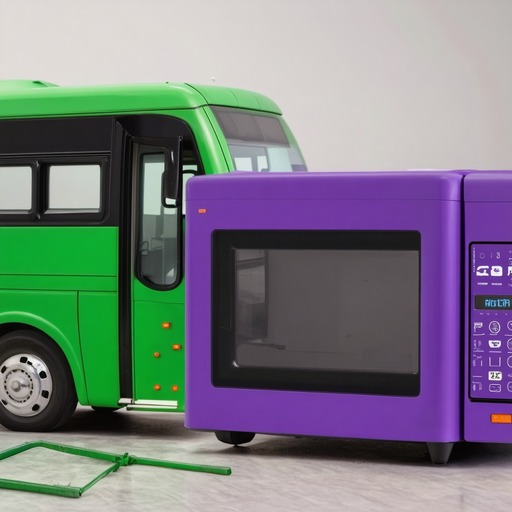} & \rimg[width=0.104\textwidth]{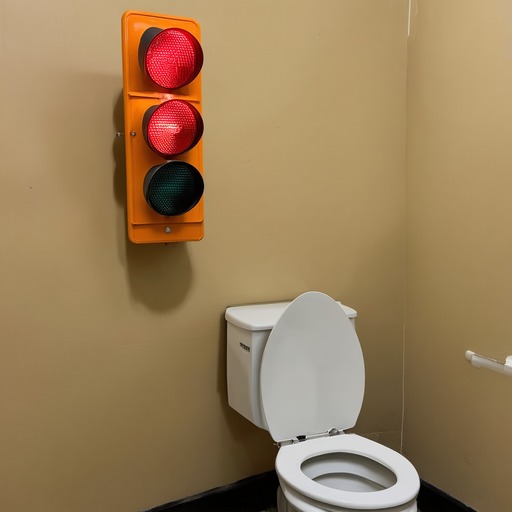} & \rimg[width=0.104\textwidth]{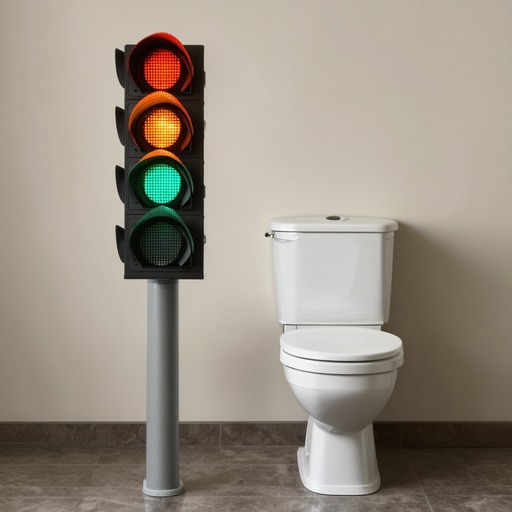} & \rimg[width=0.104\textwidth]{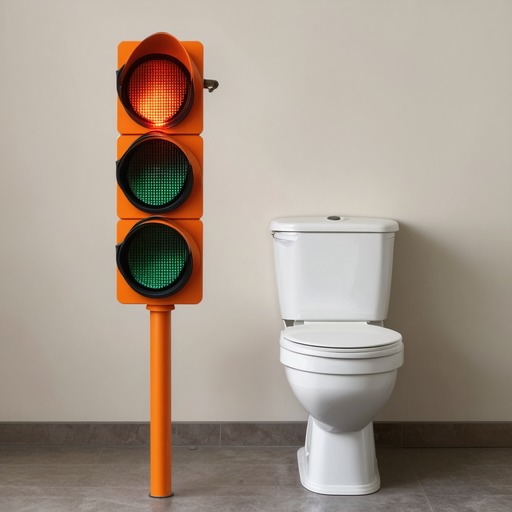} & \rimg[width=0.104\textwidth]{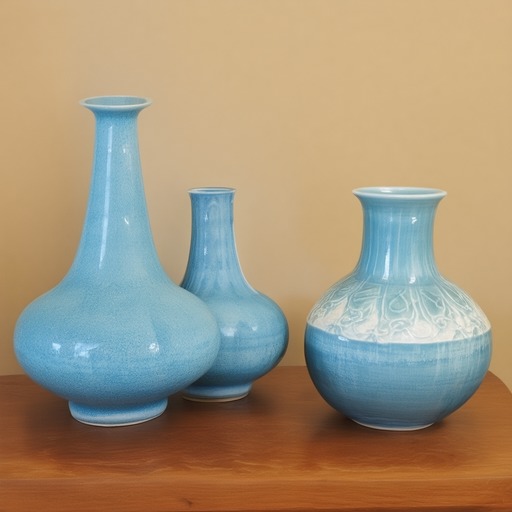} & \rimg[width=0.104\textwidth]{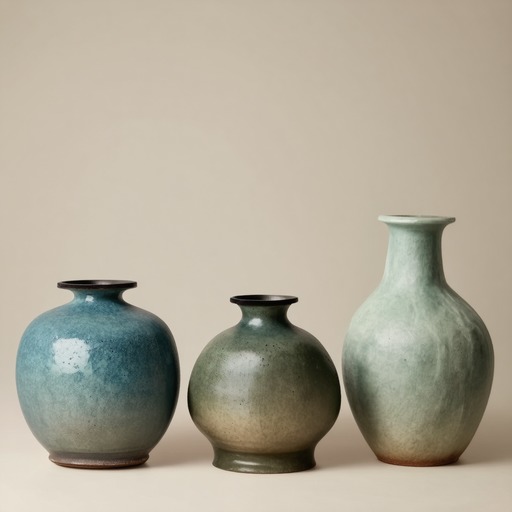} & \rimg[width=0.104\textwidth]{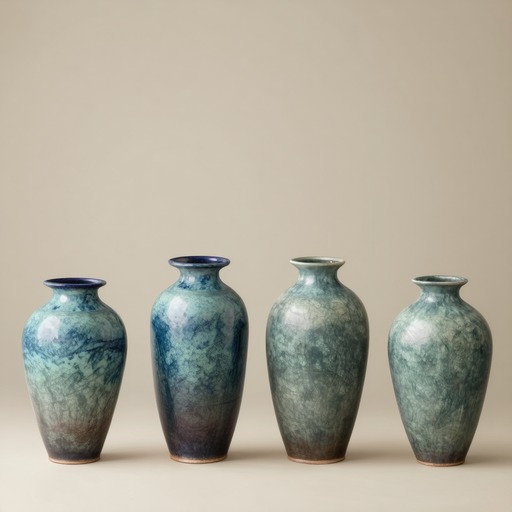} \\[-1pt]
  \multicolumn{3}{c}{\scriptsize\itshape\color{qmute} ``a photo of a green bus and a purple microwave''} & \multicolumn{3}{c}{\scriptsize\itshape\color{qmute} ``a photo of an orange traffic light and a white toilet''} & \multicolumn{3}{c}{\scriptsize\itshape\color{qmute} ``a photo of four vases''} \\[5pt]
  \end{tabular}\end{adjustbox}\end{qualband}
  \caption{\textbf{GenEval prompts}: SD3.5 Medium, the SFT control and \method (same seed). It draws the object the others drop (the fork), keeps two objects apart (bus and microwave), binds colours to the right object (red giraffe, orange traffic light), places them as asked (apple above the TV) and counts (four vases).}
  \label{fig:qual}
\end{figure*}

\begin{figure*}[!tp]
  \centering
  \setlength{\tabcolsep}{0.5pt}
  \renewcommand{\arraystretch}{0.6}
  \begin{qualband}{SD3.5 Medium}\begin{adjustbox}{max width=\linewidth}%
  \begin{tabular}{@{}ccc@{\hspace{5pt}}ccc@{\hspace{5pt}}ccc@{\hspace{5pt}}ccc@{}}
  \scriptsize SD3.5-M & \scriptsize SFT control & \scriptsize \textcolor{sdcol}{\textbf{\method}} & \scriptsize SD3.5-M & \scriptsize SFT control & \scriptsize \textcolor{sdcol}{\textbf{\method}} & \scriptsize SD3.5-M & \scriptsize SFT control & \scriptsize \textcolor{sdcol}{\textbf{\method}} & \scriptsize SD3.5-M & \scriptsize SFT control & \scriptsize \textcolor{sdcol}{\textbf{\method}} \\
  \rimg[width=0.077\textwidth]{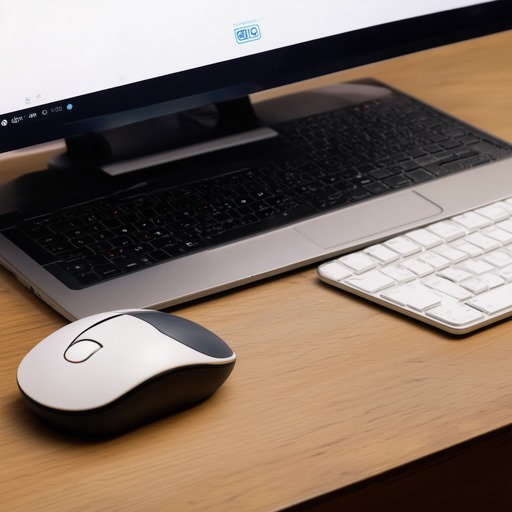} & \rimg[width=0.077\textwidth]{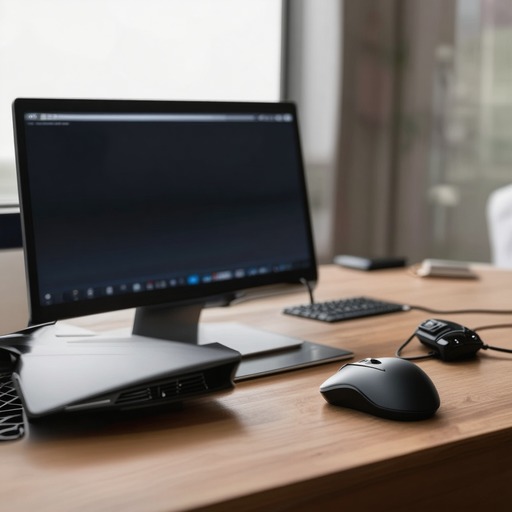} & \rimg[width=0.077\textwidth]{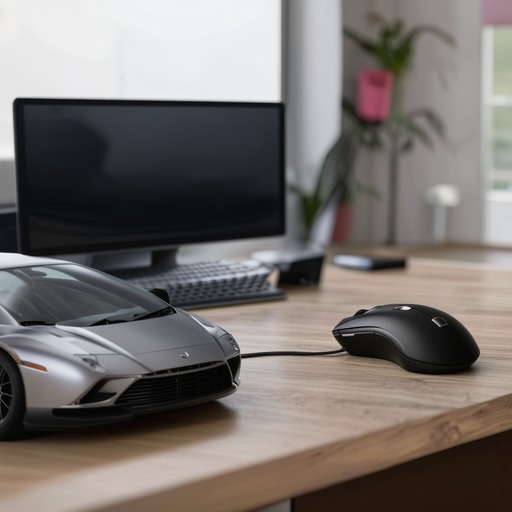} & \rimg[width=0.077\textwidth]{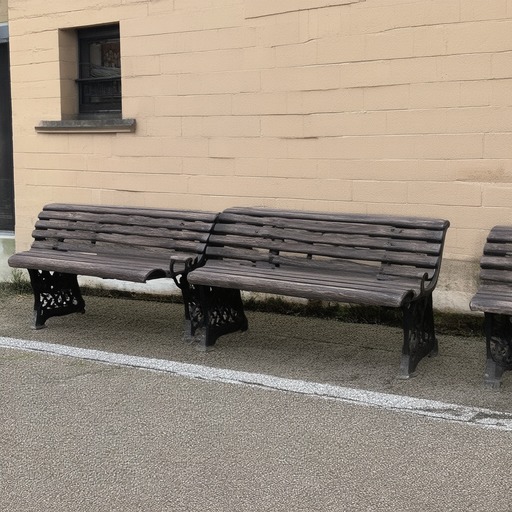} & \rimg[width=0.077\textwidth]{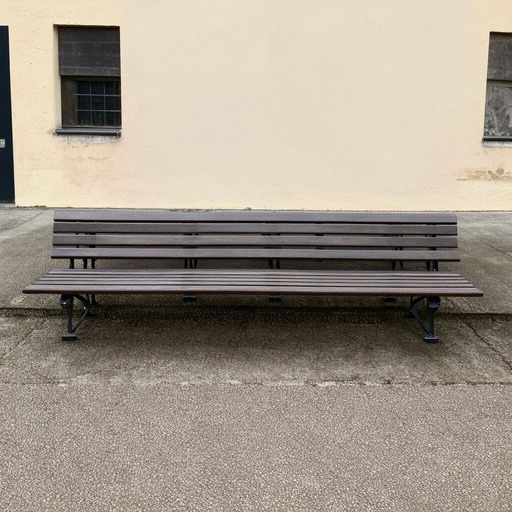} & \rimg[width=0.077\textwidth]{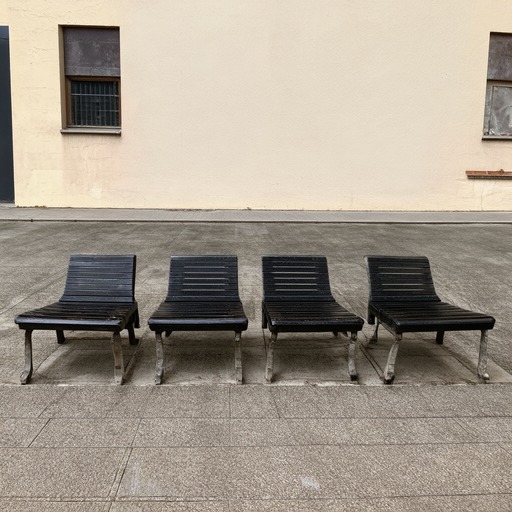} & \rimg[width=0.077\textwidth]{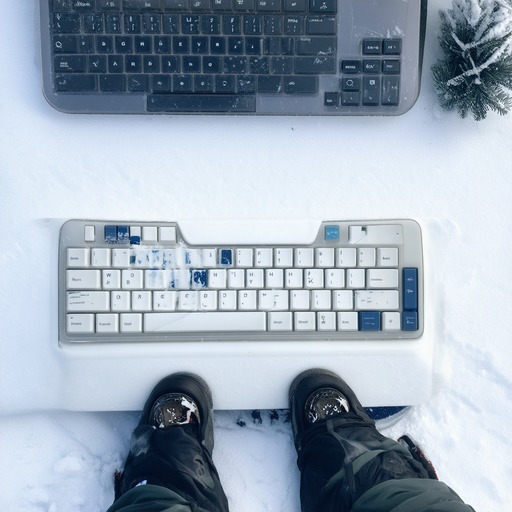} & \rimg[width=0.077\textwidth]{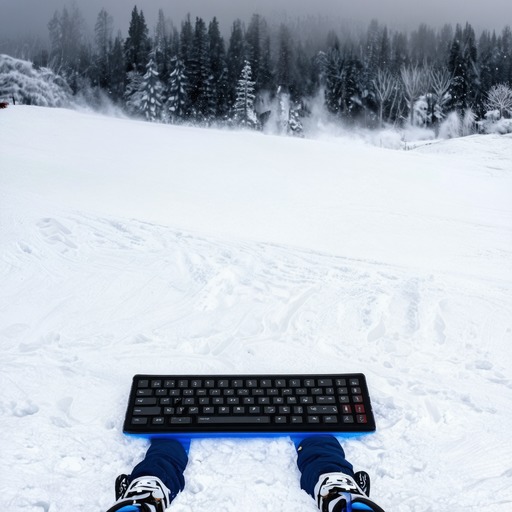} & \rimg[width=0.077\textwidth]{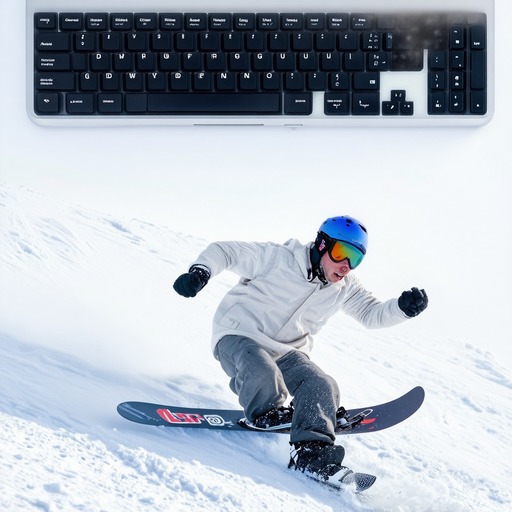} & \rimg[width=0.077\textwidth]{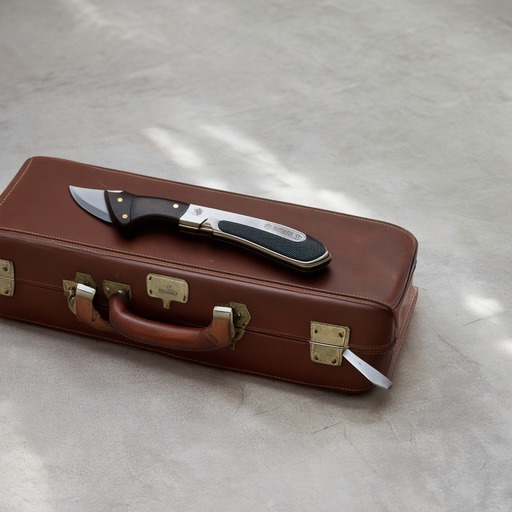} & \rimg[width=0.077\textwidth]{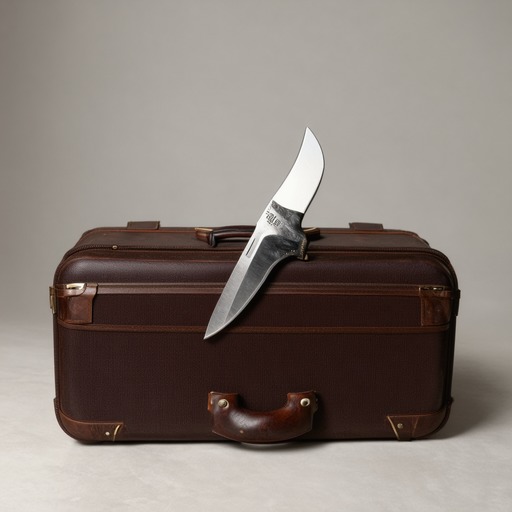} & \rimg[width=0.077\textwidth]{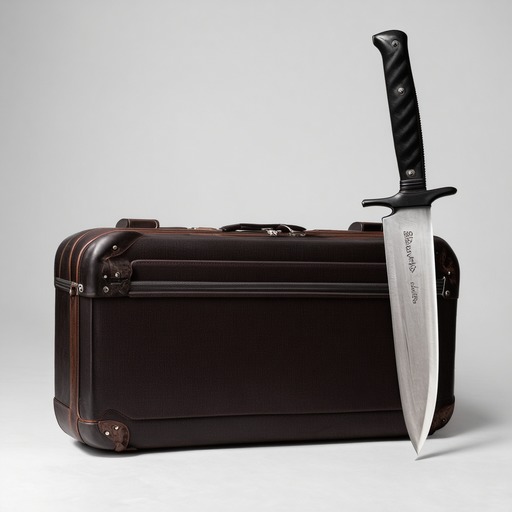} \\
  \multicolumn{3}{c}{\parbox[t]{0.23\textwidth}{\centering\scriptsize\textcolor{qmute}{\textit{``a photo of a car and a computer mouse''}}\\\tiny A car beside the mouse; the baselines draw only the desk.}} & \multicolumn{3}{c}{\parbox[t]{0.23\textwidth}{\centering\scriptsize\textcolor{qmute}{\textit{``a photo of four benchs''}}\\\tiny Four seats; base two to three, SFT one long bench.}} & \multicolumn{3}{c}{\parbox[t]{0.23\textwidth}{\centering\scriptsize\textcolor{qmute}{\textit{``a photo of a computer keyboard above a snowboard''}}\\\tiny Keyboard above a snowboarder; the baselines put keyboards on the snow.}} & \multicolumn{3}{c}{\parbox[t]{0.23\textwidth}{\centering\scriptsize\textcolor{qmute}{\textit{``a photo of a knife right of a suitcase''}}\\\tiny Knife standing right of the suitcase; the baselines lay it on top.}} \\[3pt]
  \rimg[width=0.077\textwidth]{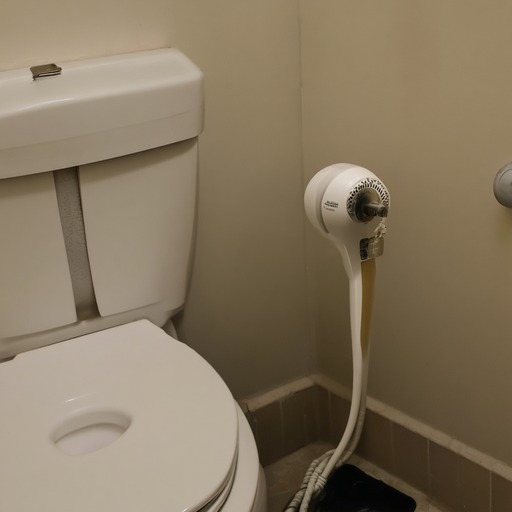} & \rimg[width=0.077\textwidth]{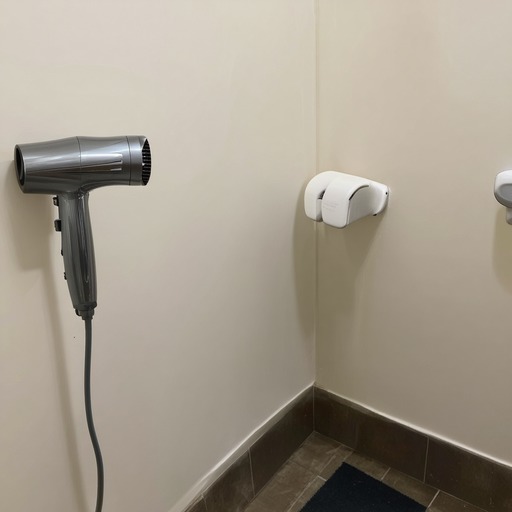} & \rimg[width=0.077\textwidth]{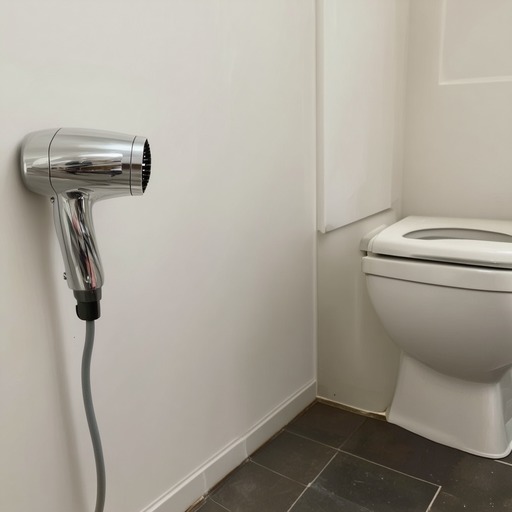} & \rimg[width=0.077\textwidth]{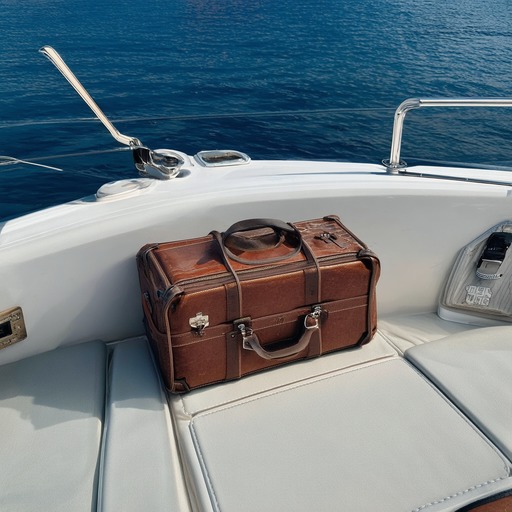} & \rimg[width=0.077\textwidth]{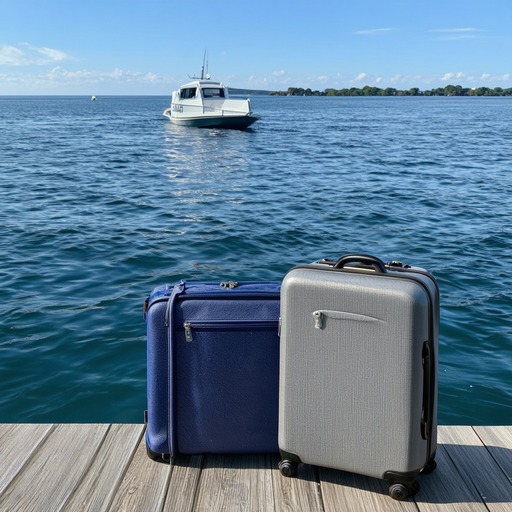} & \rimg[width=0.077\textwidth]{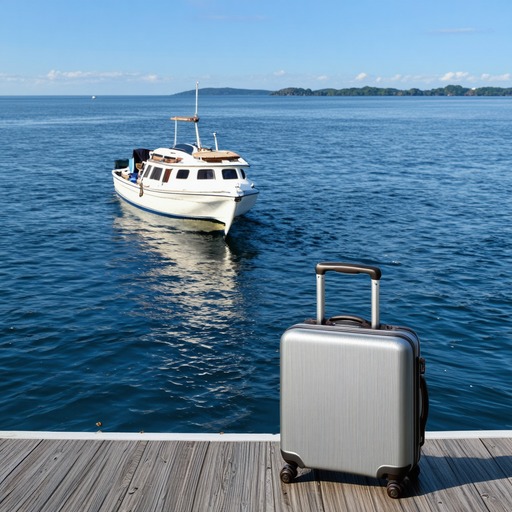} & \rimg[width=0.077\textwidth]{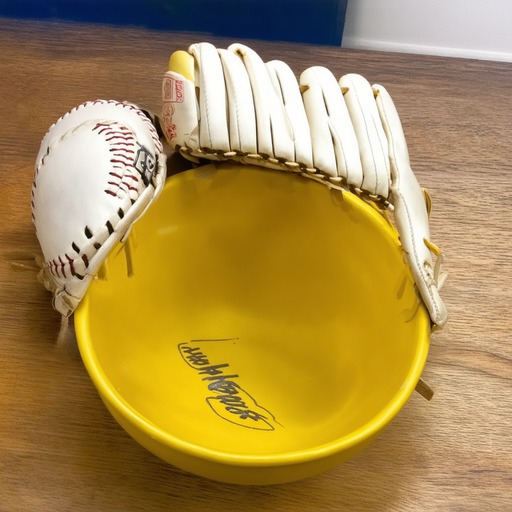} & \rimg[width=0.077\textwidth]{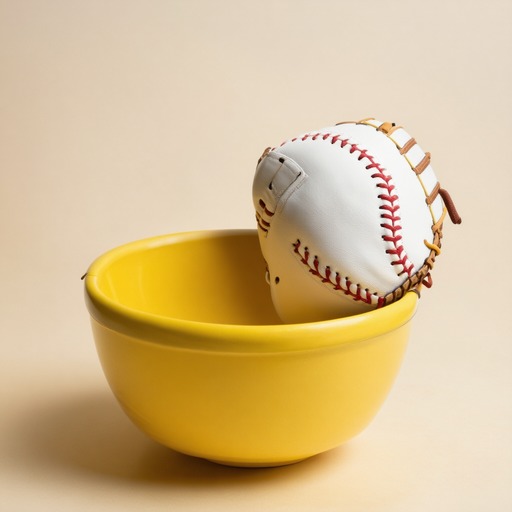} & \rimg[width=0.077\textwidth]{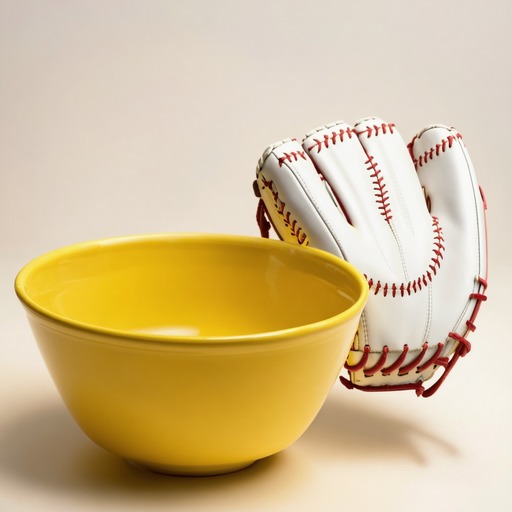} & \rimg[width=0.077\textwidth]{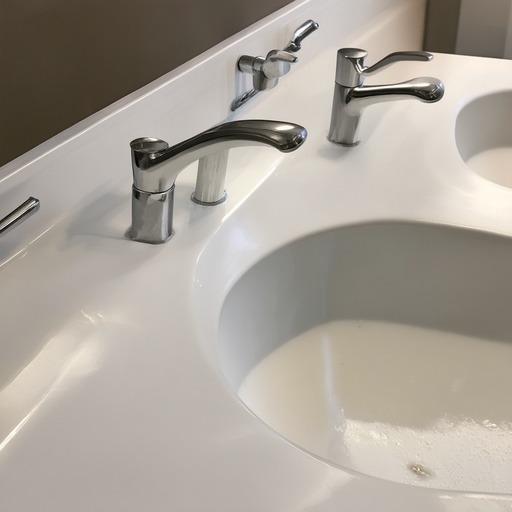} & \rimg[width=0.077\textwidth]{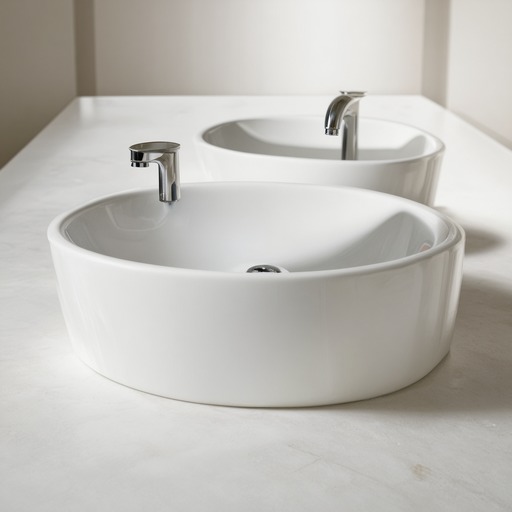} & \rimg[width=0.077\textwidth]{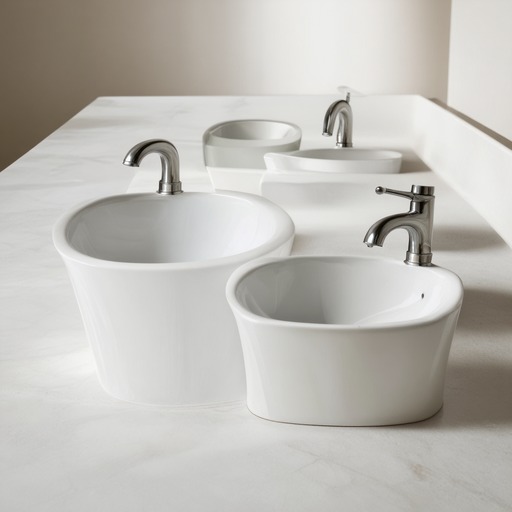} \\
  \multicolumn{3}{c}{\parbox[t]{0.23\textwidth}{\centering\scriptsize\textcolor{qmute}{\textit{``a photo of a hair drier left of a toilet''}}\\\tiny Hair drier left of the toilet; base puts it on the right, SFT draws no toilet.}} & \multicolumn{3}{c}{\parbox[t]{0.23\textwidth}{\centering\scriptsize\textcolor{qmute}{\textit{``a photo of a suitcase right of a boat''}}\\\tiny One suitcase right of the boat; base puts it in the boat, SFT draws two.}} & \multicolumn{3}{c}{\parbox[t]{0.23\textwidth}{\centering\scriptsize\textcolor{qmute}{\textit{``a photo of a yellow bowl and a white baseball glove''}}\\\tiny A glove beside the bowl; SFT draws a ball in it.}} & \multicolumn{3}{c}{\parbox[t]{0.23\textwidth}{\centering\scriptsize\textcolor{qmute}{\textit{``a photo of three sinks''}}\\\tiny Three basins; base one, SFT two.}} \\[3pt]
  \rimg[width=0.077\textwidth]{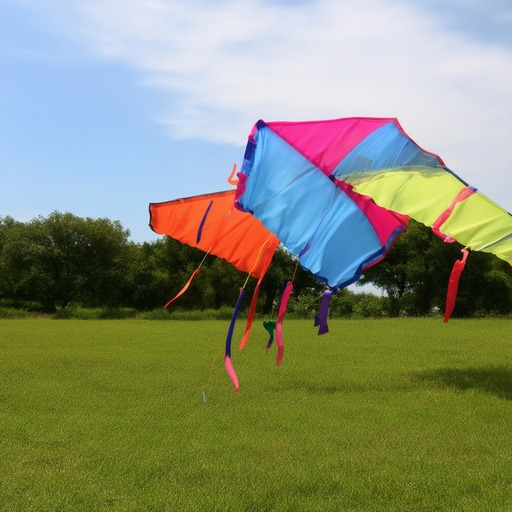} & \rimg[width=0.077\textwidth]{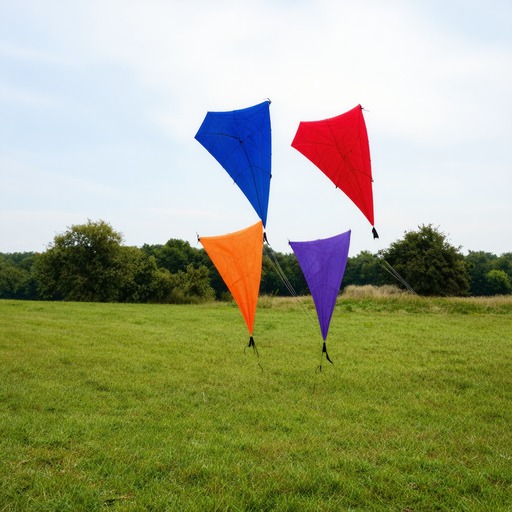} & \rimg[width=0.077\textwidth]{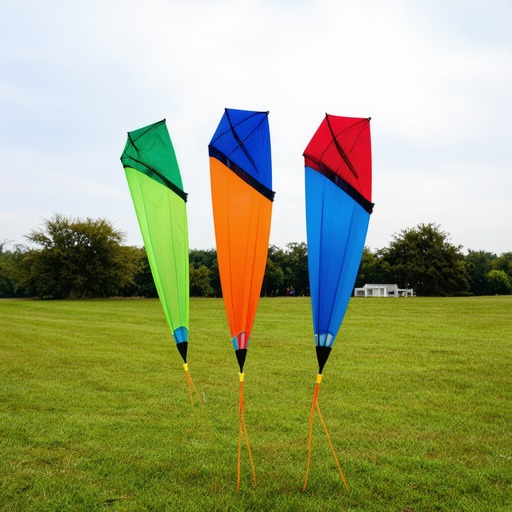} & \rimg[width=0.077\textwidth]{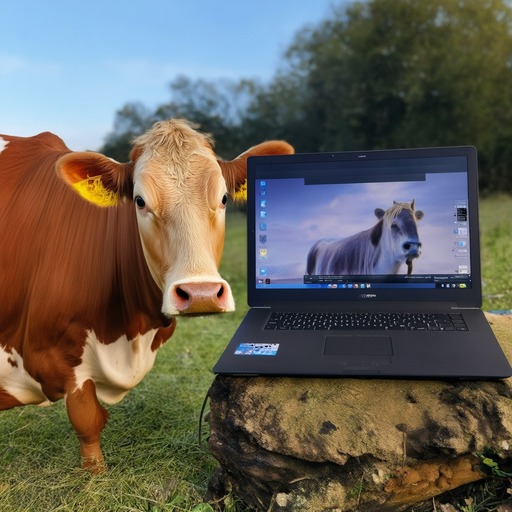} & \rimg[width=0.077\textwidth]{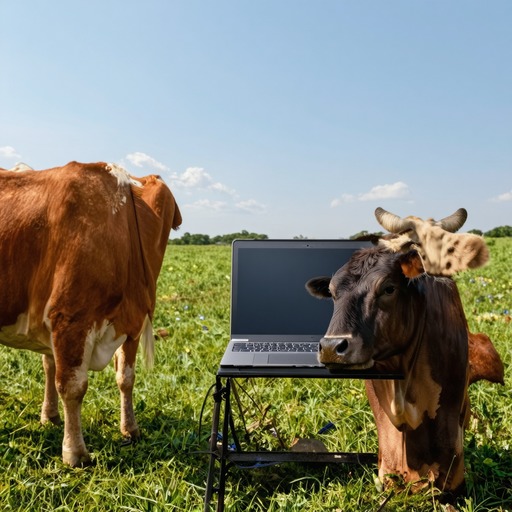} & \rimg[width=0.077\textwidth]{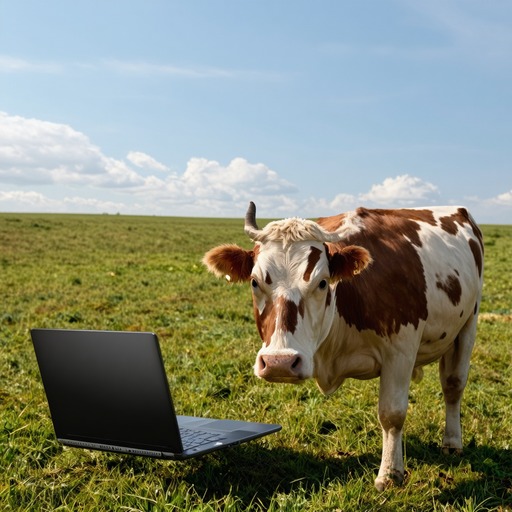} & \rimg[width=0.077\textwidth]{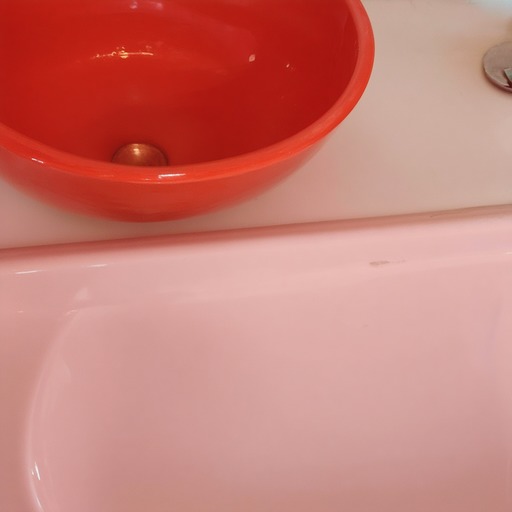} & \rimg[width=0.077\textwidth]{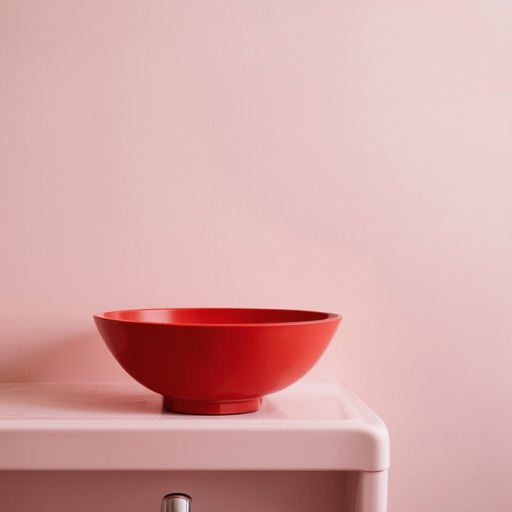} & \rimg[width=0.077\textwidth]{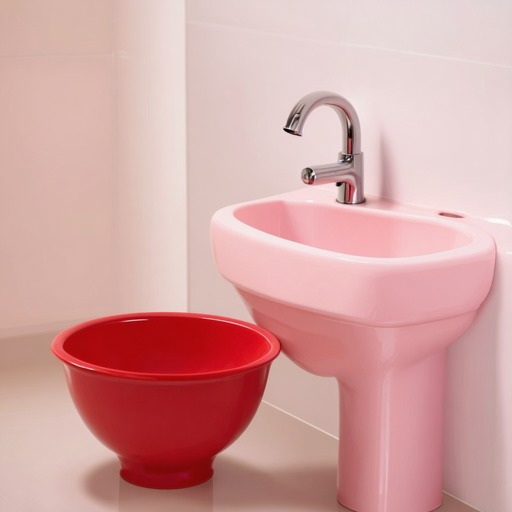} & \rimg[width=0.077\textwidth]{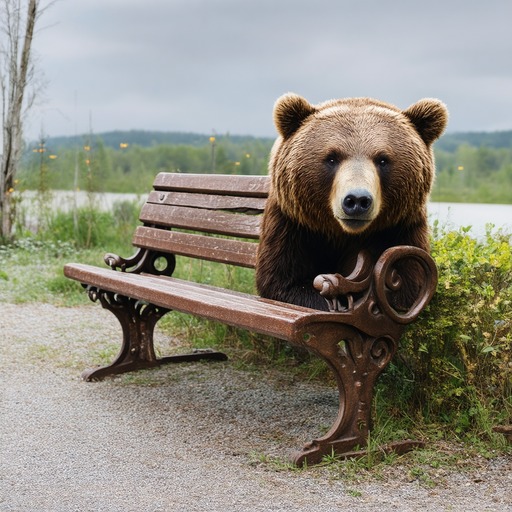} & \rimg[width=0.077\textwidth]{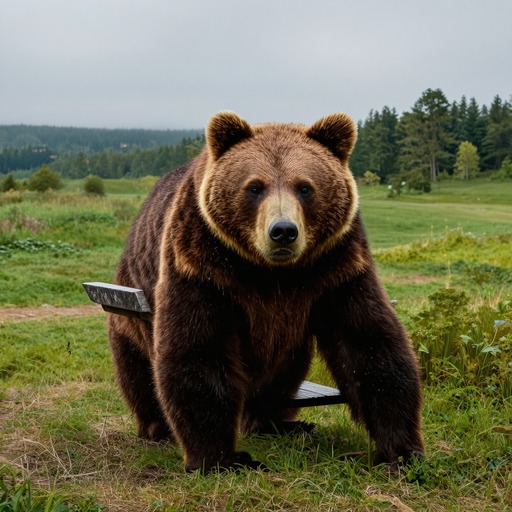} & \rimg[width=0.077\textwidth]{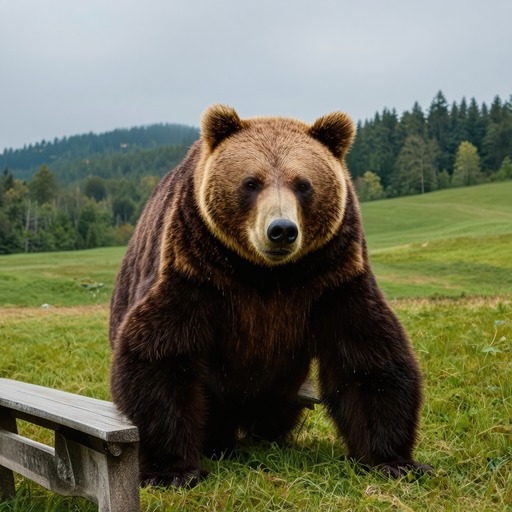} \\
  \multicolumn{3}{c}{\parbox[t]{0.23\textwidth}{\centering\scriptsize\textcolor{qmute}{\textit{``a photo of three kites''}}\\\tiny Three separate kites; base fuses them, SFT draws four.}} & \multicolumn{3}{c}{\parbox[t]{0.23\textwidth}{\centering\scriptsize\textcolor{qmute}{\textit{``a photo of a laptop left of a cow''}}\\\tiny Laptop left of the cow; base puts it right, SFT adds a second cow.}} & \multicolumn{3}{c}{\parbox[t]{0.23\textwidth}{\centering\scriptsize\textcolor{qmute}{\textit{``a photo of a red bowl and a pink sink''}}\\\tiny A pink sink beside the red bowl; SFT draws no sink.}} & \multicolumn{3}{c}{\parbox[t]{0.23\textwidth}{\centering\scriptsize\textcolor{qmute}{\textit{``a photo of a bench left of a bear''}}\\\tiny Bench at the bear's left; base seats the bear on it.}} \\[3pt]
  \rimg[width=0.077\textwidth]{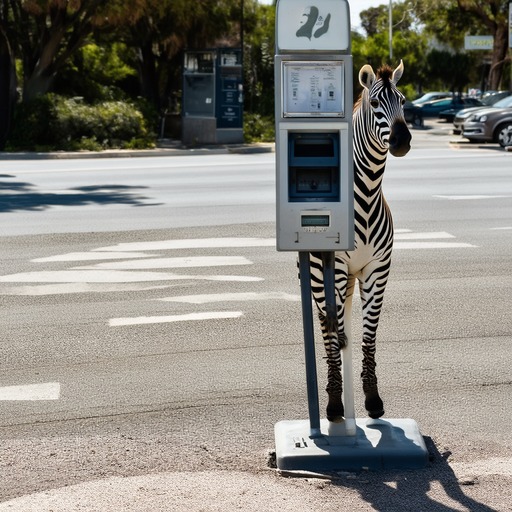} & \rimg[width=0.077\textwidth]{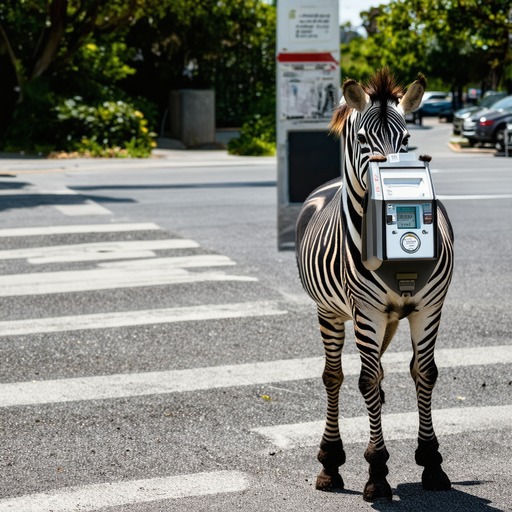} & \rimg[width=0.077\textwidth]{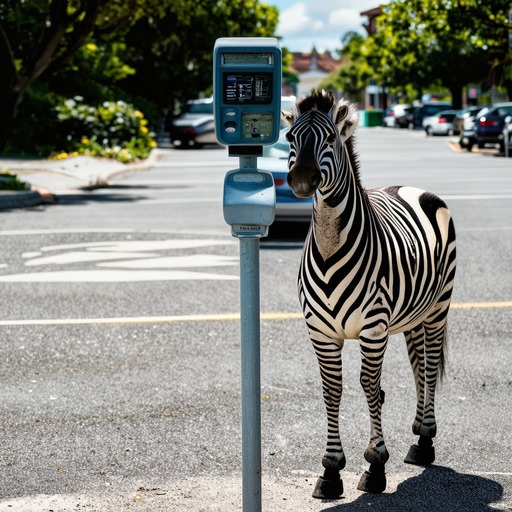} & \rimg[width=0.077\textwidth]{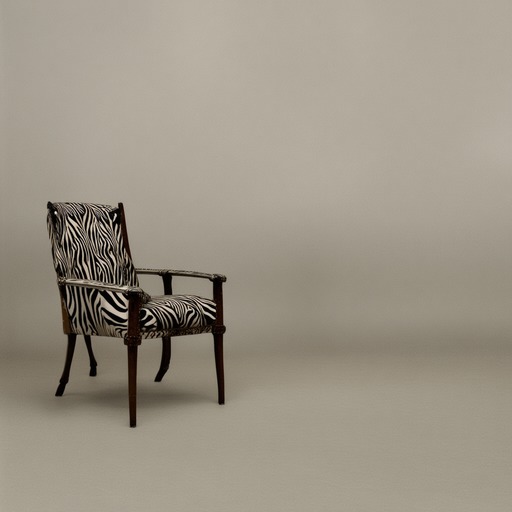} & \rimg[width=0.077\textwidth]{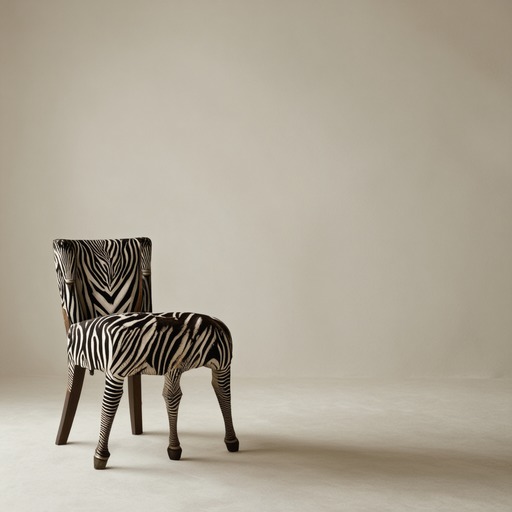} & \rimg[width=0.077\textwidth]{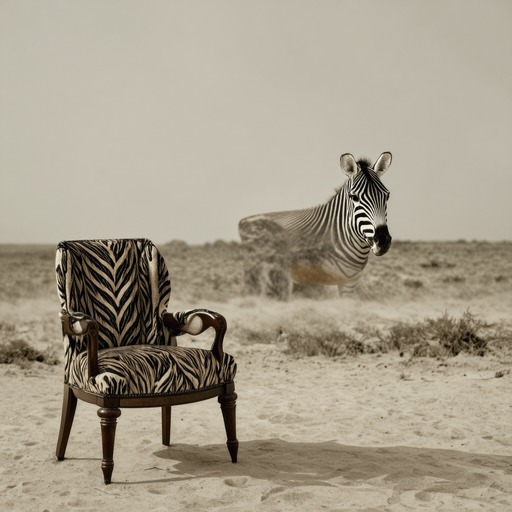} & \rimg[width=0.077\textwidth]{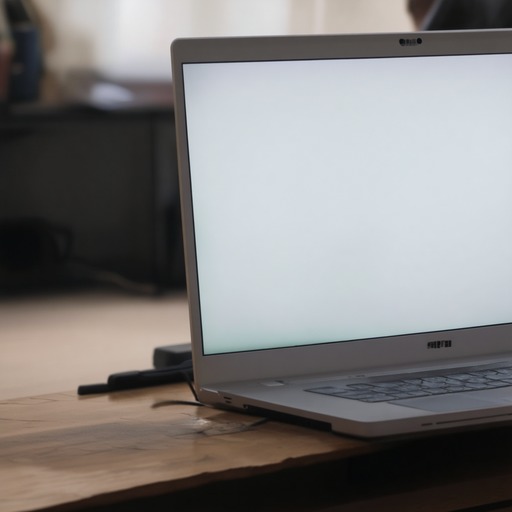} & \rimg[width=0.077\textwidth]{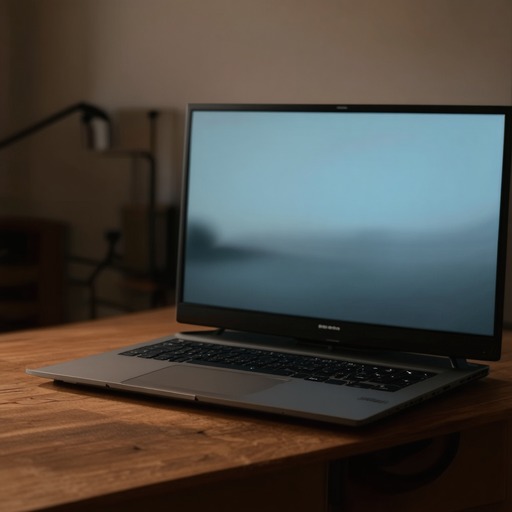} & \rimg[width=0.077\textwidth]{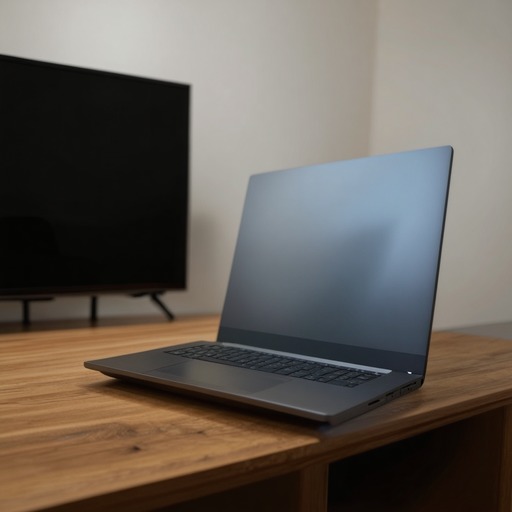} & \rimg[width=0.077\textwidth]{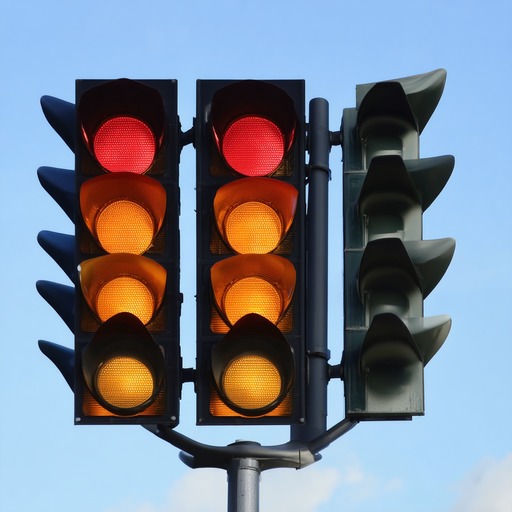} & \rimg[width=0.077\textwidth]{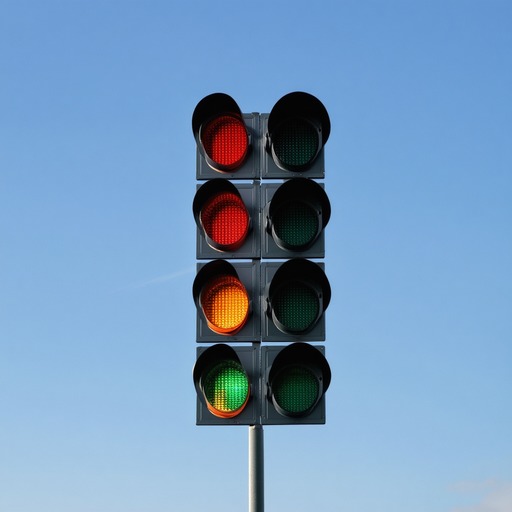} & \rimg[width=0.077\textwidth]{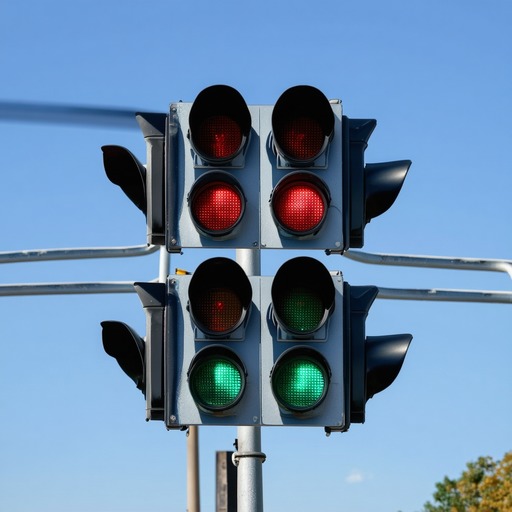} \\
  \multicolumn{3}{c}{\parbox[t]{0.23\textwidth}{\centering\scriptsize\textcolor{qmute}{\textit{``a photo of a zebra right of a parking meter''}}\\\tiny Zebra right of the meter; the baselines merge them.}} & \multicolumn{3}{c}{\parbox[t]{0.23\textwidth}{\centering\scriptsize\textcolor{qmute}{\textit{``a photo of a chair left of a zebra''}}\\\tiny A zebra right of the chair; the baselines draw only a zebra-print chair.}} & \multicolumn{3}{c}{\parbox[t]{0.23\textwidth}{\centering\scriptsize\textcolor{qmute}{\textit{``a photo of a laptop right of a tv''}}\\\tiny A TV with the laptop at its right; the baselines miss the TV.}} & \multicolumn{3}{c}{\parbox[t]{0.23\textwidth}{\centering\scriptsize\textcolor{qmute}{\textit{``a photo of four traffic lights''}}\\\tiny Four traffic lights; base draws three, SFT one.}} \\[3pt]
  \end{tabular}\end{adjustbox}\end{qualband}
  \caption{\textbf{More GenEval prompts} (same seed).}
  \label{fig:supp_gen}
\end{figure*}

\Cref{fig:qual,fig:supp_gen} show qualitative comparisons on GenEval prompts: SD3.5 Medium, the SFT control
and \method at the same seed.

\begin{figure*}[!tp]
  \centering
  \setlength{\tabcolsep}{0.5pt}
  \renewcommand{\arraystretch}{0.6}
  \begin{qualband}{FLUX.2 [klein]}\begin{adjustbox}{max width=\linewidth}%
  \begin{tabular}{@{}ccc@{\hspace{5pt}}ccc@{\hspace{5pt}}ccc@{\hspace{5pt}}ccc@{}}
  \scriptsize FLUX.2 & \scriptsize SFT control & \scriptsize \textcolor{sdcol}{\textbf{\method}} & \scriptsize FLUX.2 & \scriptsize SFT control & \scriptsize \textcolor{sdcol}{\textbf{\method}} & \scriptsize FLUX.2 & \scriptsize SFT control & \scriptsize \textcolor{sdcol}{\textbf{\method}} & \scriptsize FLUX.2 & \scriptsize SFT control & \scriptsize \textcolor{sdcol}{\textbf{\method}} \\
  \rimg[width=0.077\textwidth]{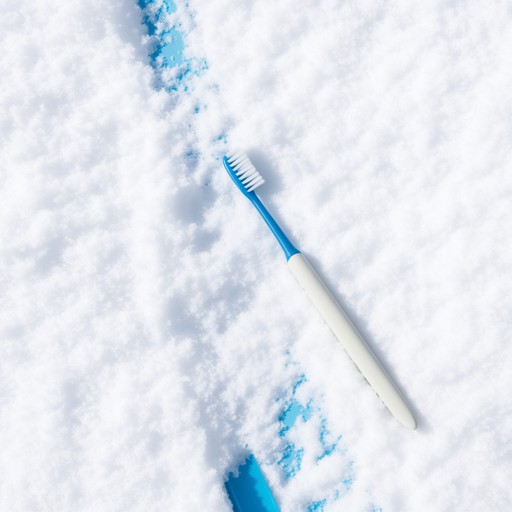} & \rimg[width=0.077\textwidth]{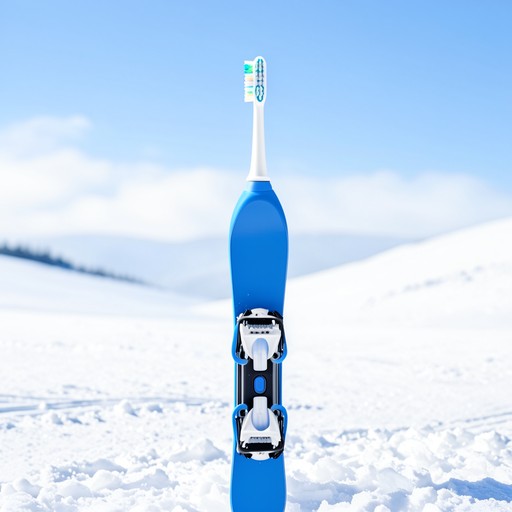} & \rimg[width=0.077\textwidth]{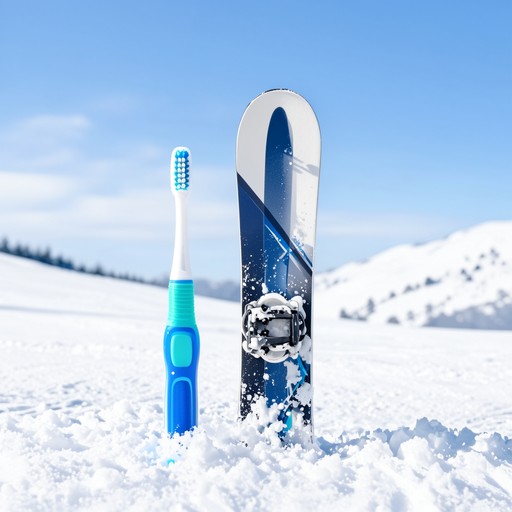} & \rimg[width=0.077\textwidth]{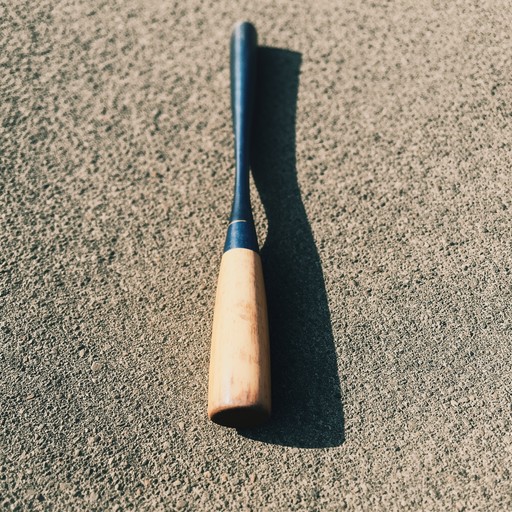} & \rimg[width=0.077\textwidth]{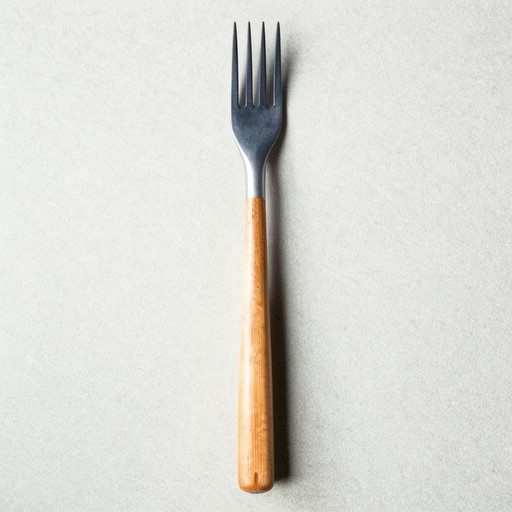} & \rimg[width=0.077\textwidth]{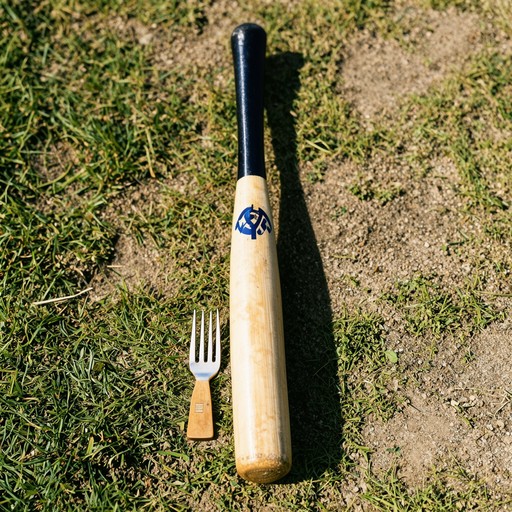} & \rimg[width=0.077\textwidth]{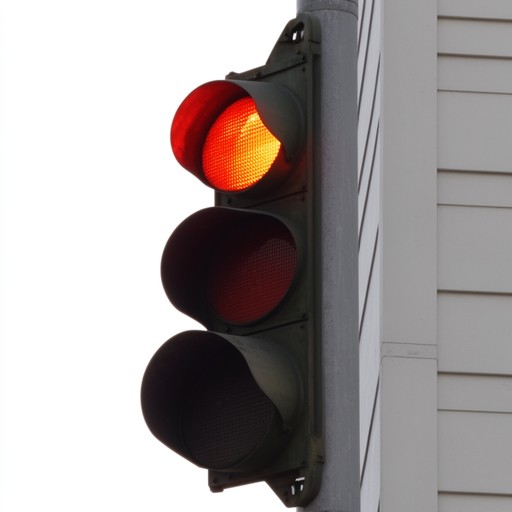} & \rimg[width=0.077\textwidth]{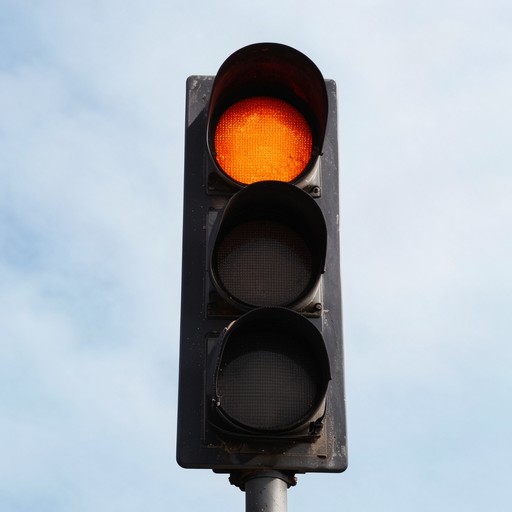} & \rimg[width=0.077\textwidth]{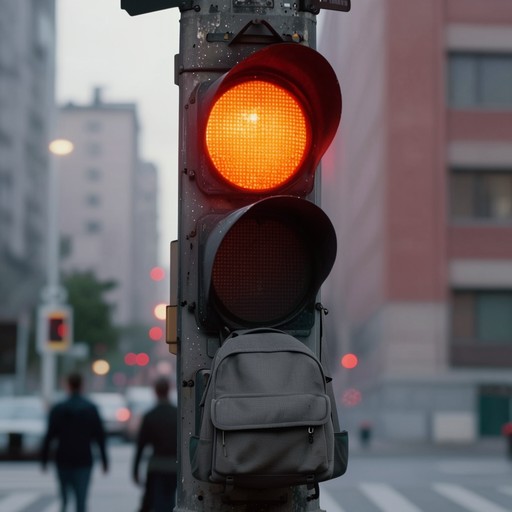} & \rimg[width=0.077\textwidth]{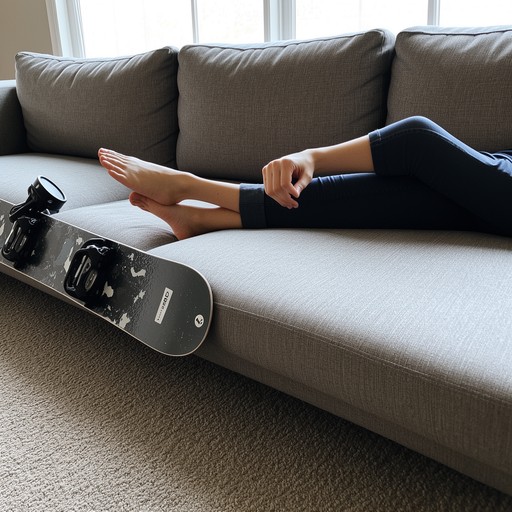} & \rimg[width=0.077\textwidth]{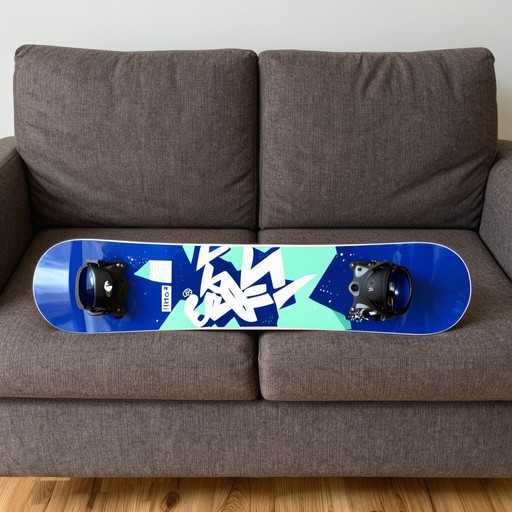} & \rimg[width=0.077\textwidth]{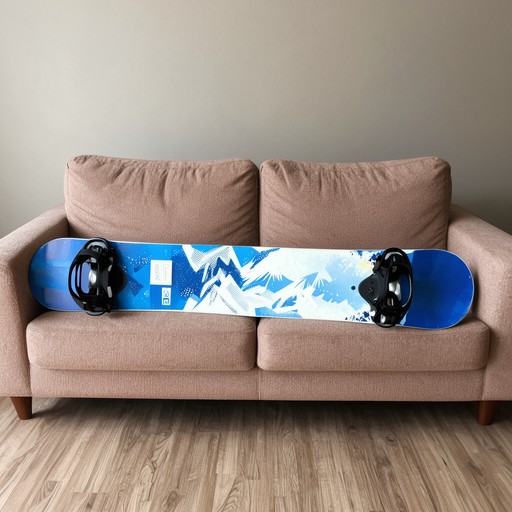} \\
  \multicolumn{3}{c}{\parbox[t]{0.23\textwidth}{\centering\scriptsize\textcolor{qmute}{\textit{``a photo of a toothbrush and a snowboard''}}}} & \multicolumn{3}{c}{\parbox[t]{0.23\textwidth}{\centering\scriptsize\textcolor{qmute}{\textit{``a photo of a baseball bat and a fork''}}}} & \multicolumn{3}{c}{\parbox[t]{0.23\textwidth}{\centering\scriptsize\textcolor{qmute}{\textit{``a photo of a traffic light and a backpack''}}}} & \multicolumn{3}{c}{\parbox[t]{0.23\textwidth}{\centering\scriptsize\textcolor{qmute}{\textit{``a photo of a couch and a snowboard''}}}} \\[3pt]
  \rimg[width=0.077\textwidth]{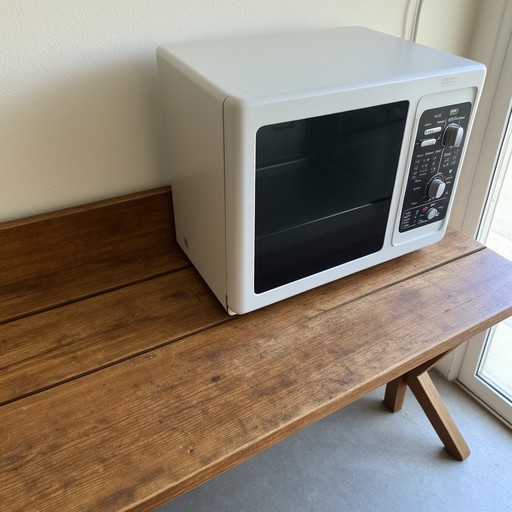} & \rimg[width=0.077\textwidth]{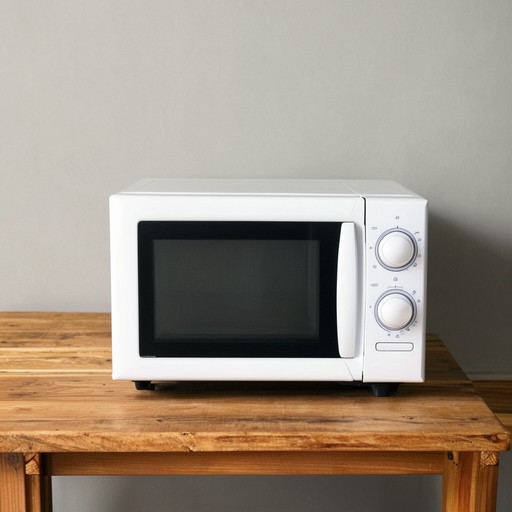} & \rimg[width=0.077\textwidth]{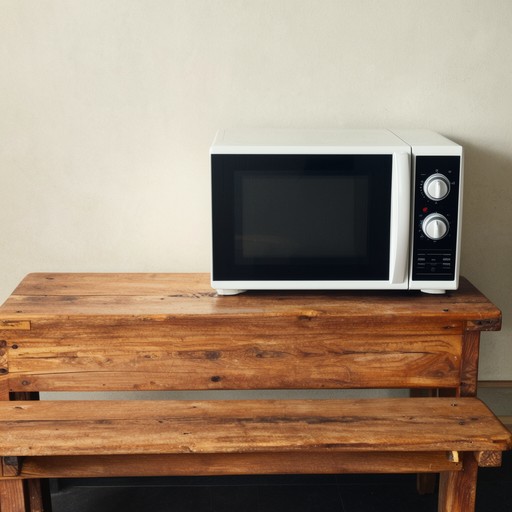} & \rimg[width=0.077\textwidth]{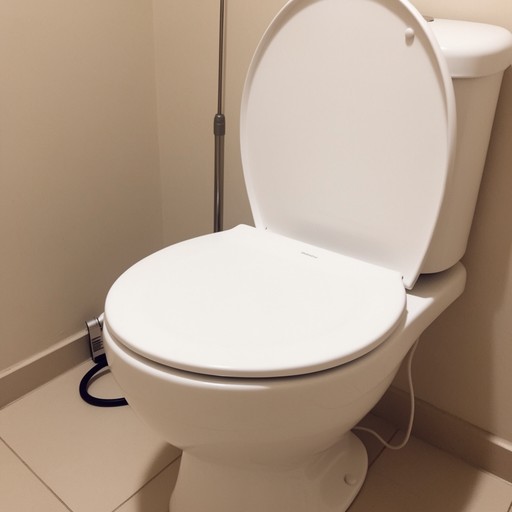} & \rimg[width=0.077\textwidth]{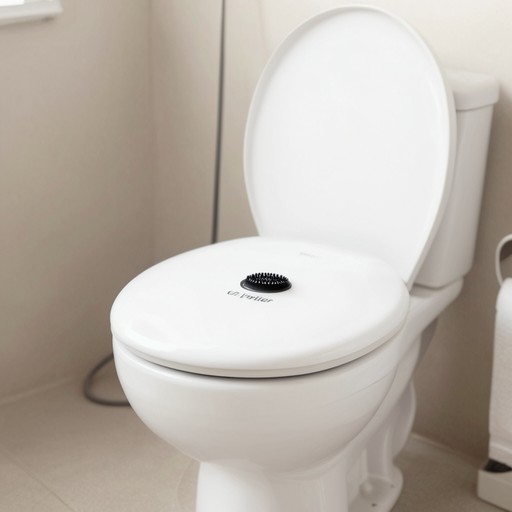} & \rimg[width=0.077\textwidth]{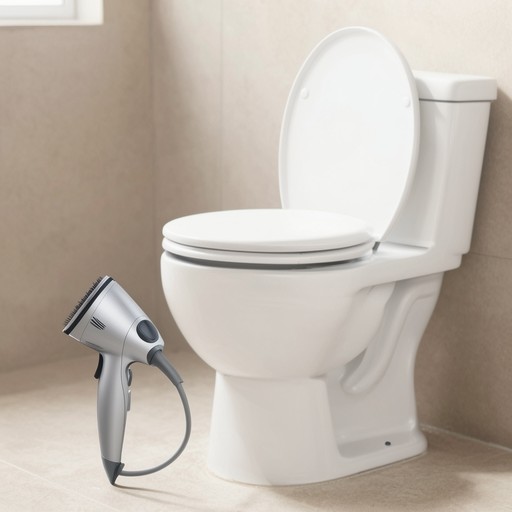} & \rimg[width=0.077\textwidth]{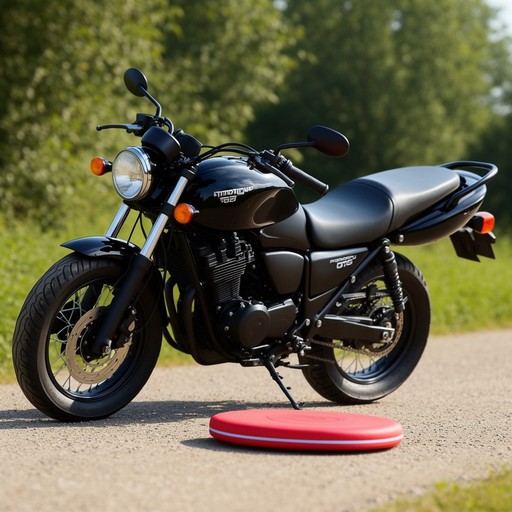} & \rimg[width=0.077\textwidth]{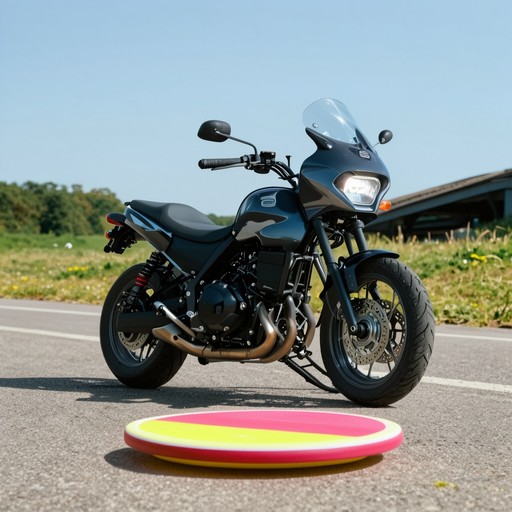} & \rimg[width=0.077\textwidth]{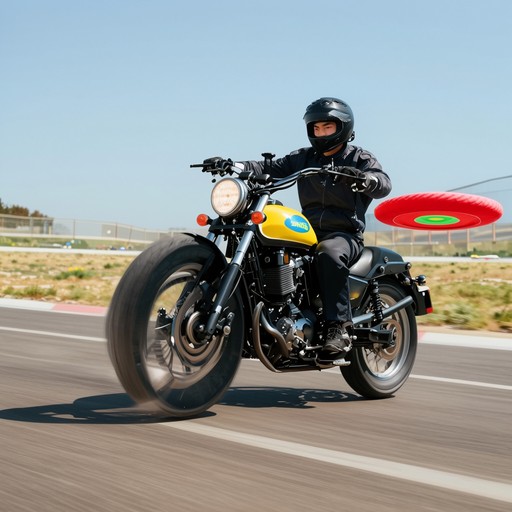} & \rimg[width=0.077\textwidth]{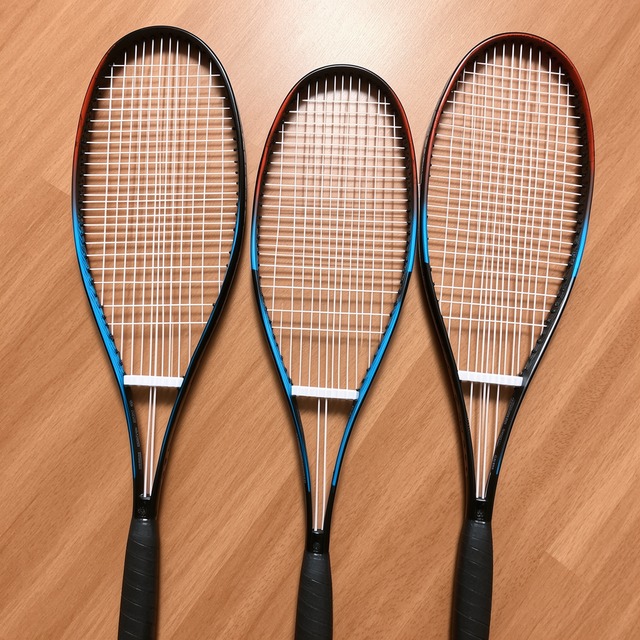} & \rimg[width=0.077\textwidth]{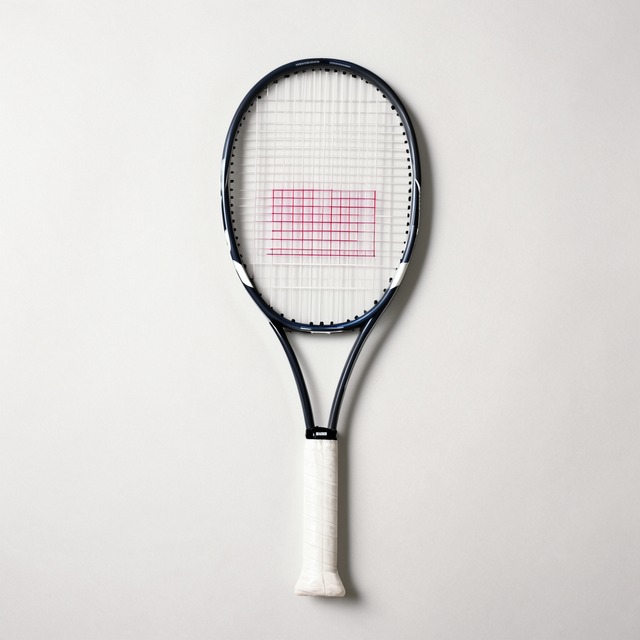} & \rimg[width=0.077\textwidth]{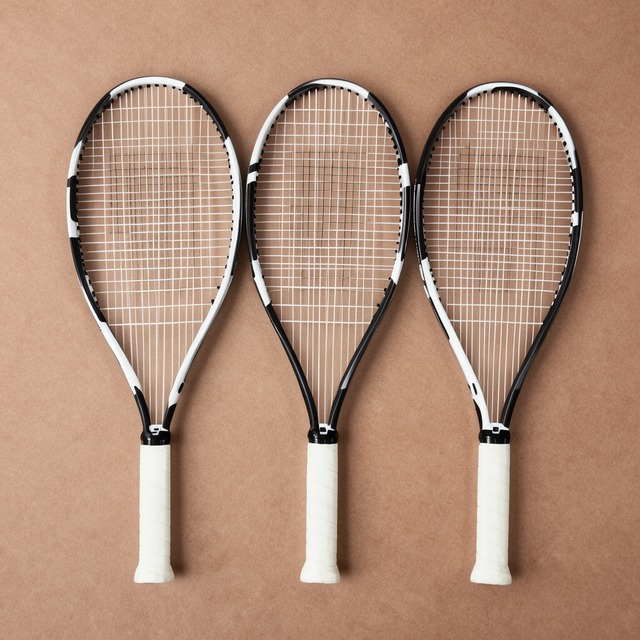} \\
  \multicolumn{3}{c}{\parbox[t]{0.23\textwidth}{\centering\scriptsize\textcolor{qmute}{\textit{``a photo of a microwave and a bench''}}}} & \multicolumn{3}{c}{\parbox[t]{0.23\textwidth}{\centering\scriptsize\textcolor{qmute}{\textit{``a photo of a hair drier left of a toilet''}}}} & \multicolumn{3}{c}{\parbox[t]{0.23\textwidth}{\centering\scriptsize\textcolor{qmute}{\textit{``a photo of a frisbee right of a motorcycle''}}}} & \multicolumn{3}{c}{\parbox[t]{0.23\textwidth}{\centering\scriptsize\textcolor{qmute}{\textit{``a photo of three tennis rackets''}}}} \\[3pt]
  \rimg[width=0.077\textwidth]{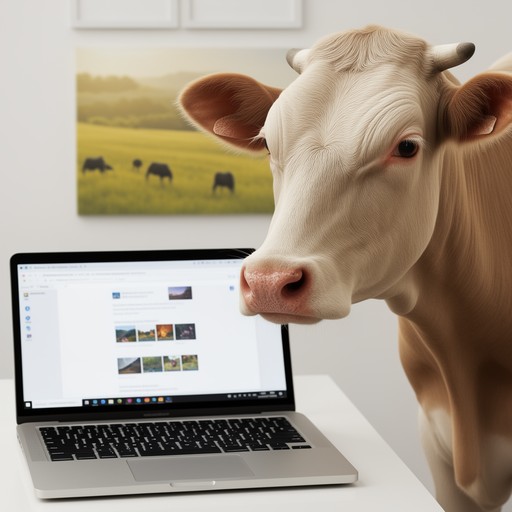} & \rimg[width=0.077\textwidth]{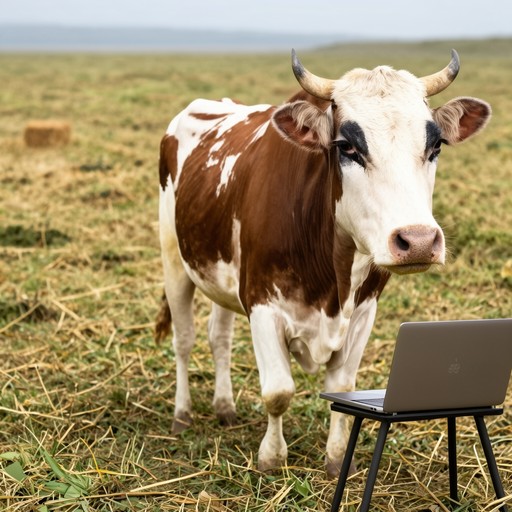} & \rimg[width=0.077\textwidth]{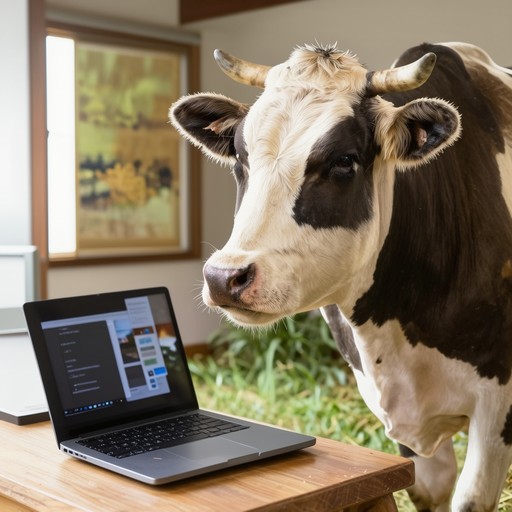} & \rimg[width=0.077\textwidth]{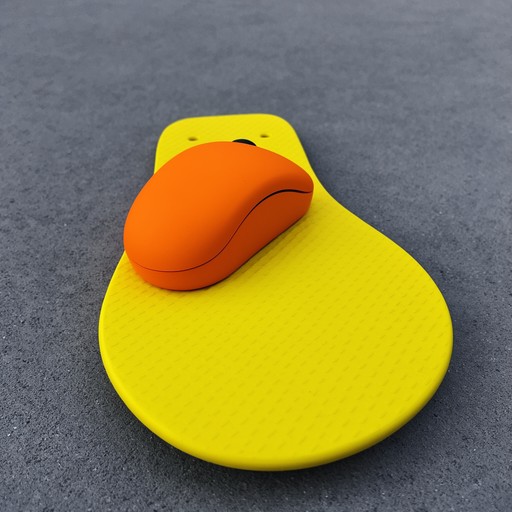} & \rimg[width=0.077\textwidth]{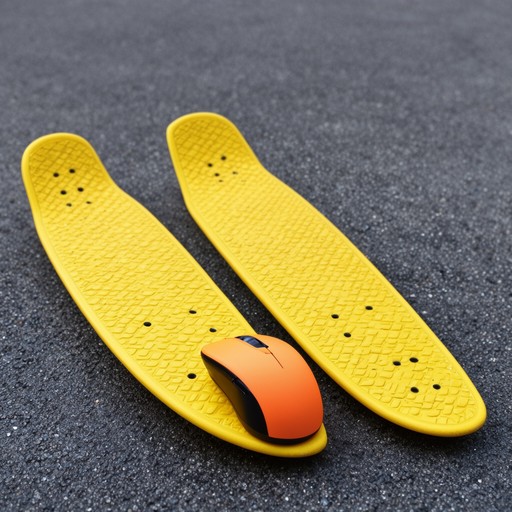} & \rimg[width=0.077\textwidth]{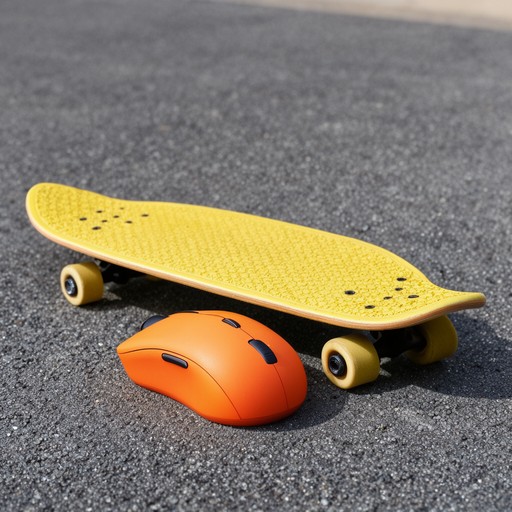} & \rimg[width=0.077\textwidth]{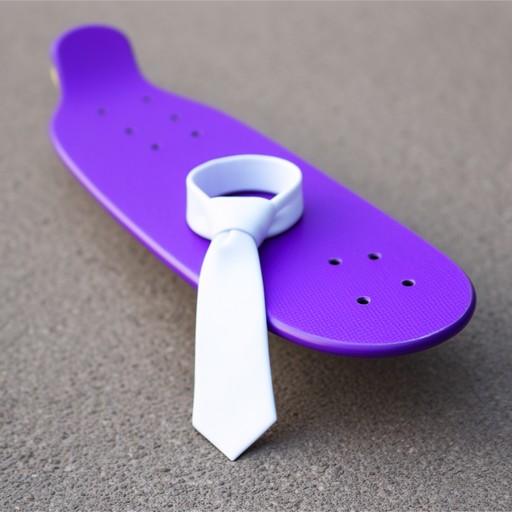} & \rimg[width=0.077\textwidth]{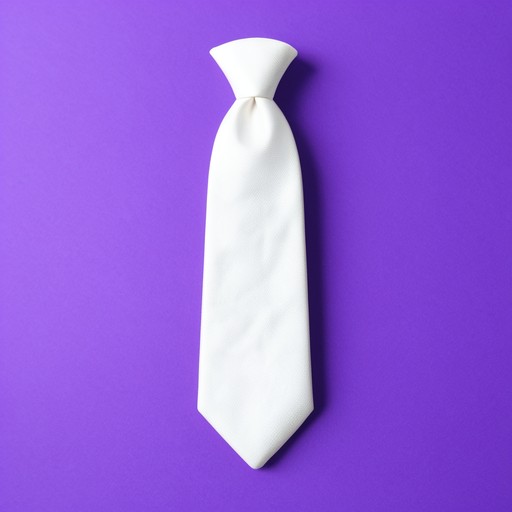} & \rimg[width=0.077\textwidth]{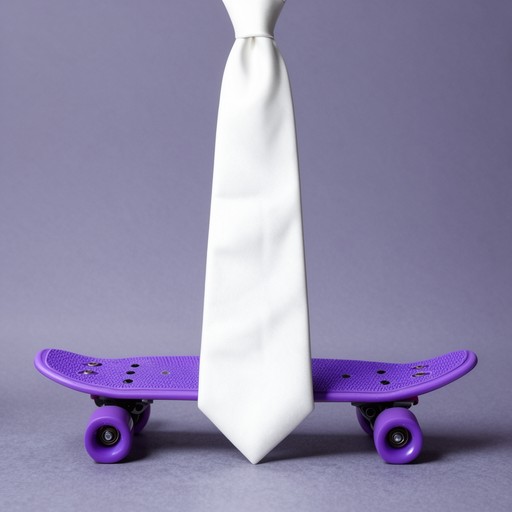} & \rimg[width=0.077\textwidth]{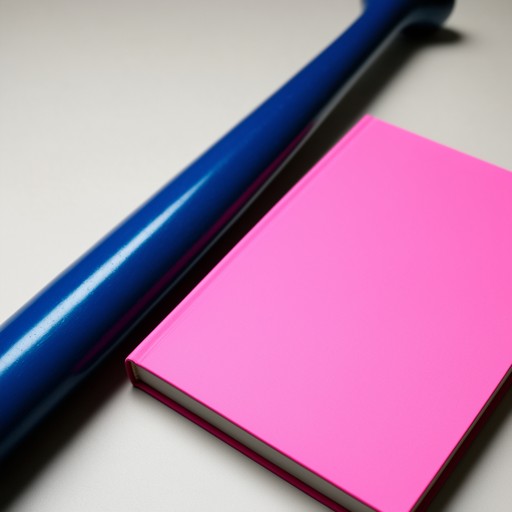} & \rimg[width=0.077\textwidth]{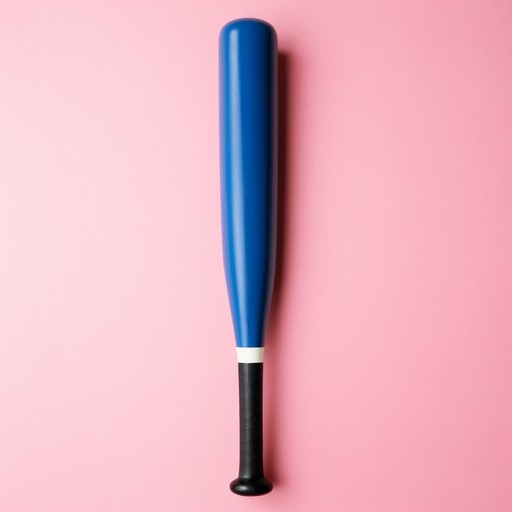} & \rimg[width=0.077\textwidth]{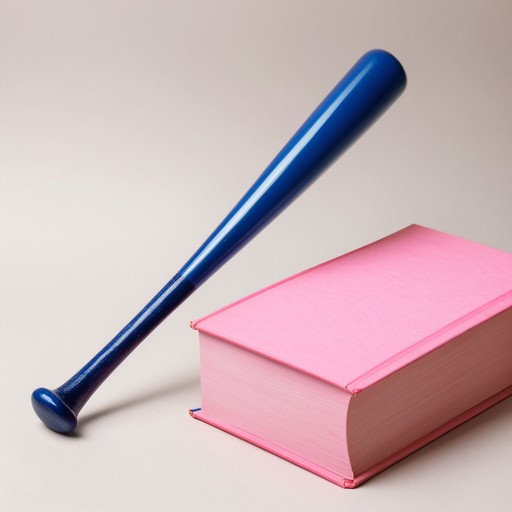} \\
  \multicolumn{3}{c}{\parbox[t]{0.23\textwidth}{\centering\scriptsize\textcolor{qmute}{\textit{``a photo of a cow right of a laptop''}}}} & \multicolumn{3}{c}{\parbox[t]{0.23\textwidth}{\centering\scriptsize\textcolor{qmute}{\textit{``a photo of a yellow skateboard and an orange computer mouse''}}}} & \multicolumn{3}{c}{\parbox[t]{0.23\textwidth}{\centering\scriptsize\textcolor{qmute}{\textit{``a photo of a white tie and a purple skateboard''}}}} & \multicolumn{3}{c}{\parbox[t]{0.23\textwidth}{\centering\scriptsize\textcolor{qmute}{\textit{``a photo of a blue baseball bat and a pink book''}}}} \\[3pt]
  \rimg[width=0.077\textwidth]{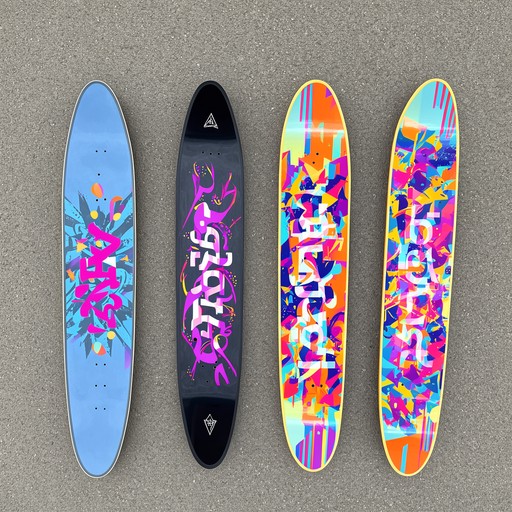} & \rimg[width=0.077\textwidth]{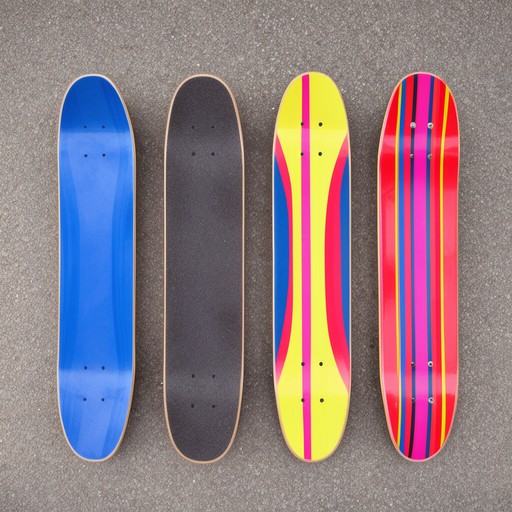} & \rimg[width=0.077\textwidth]{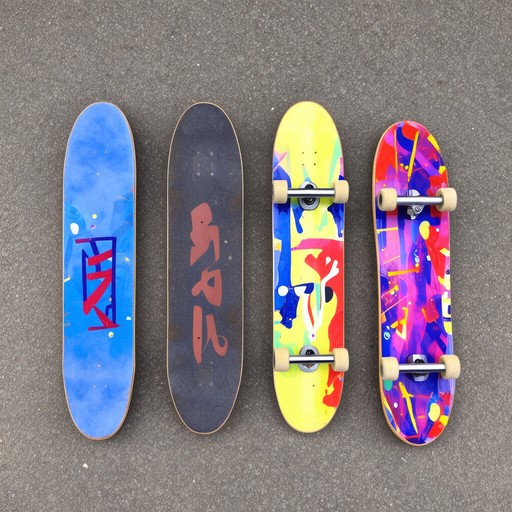} & \rimg[width=0.077\textwidth]{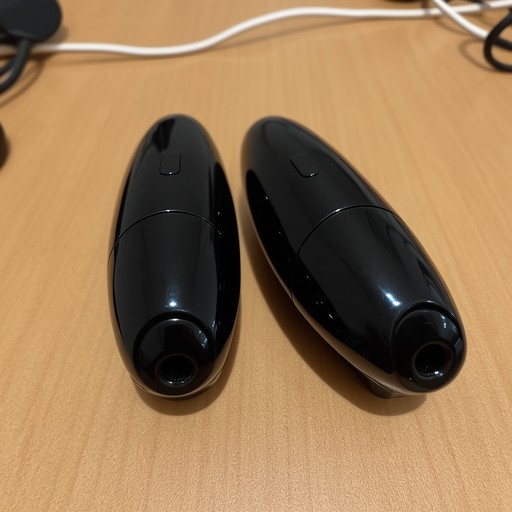} & \rimg[width=0.077\textwidth]{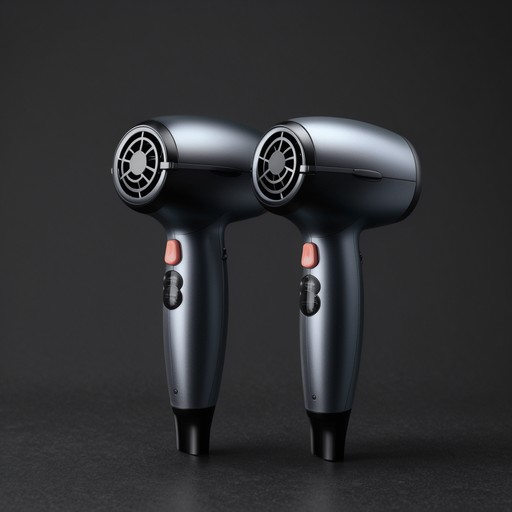} & \rimg[width=0.077\textwidth]{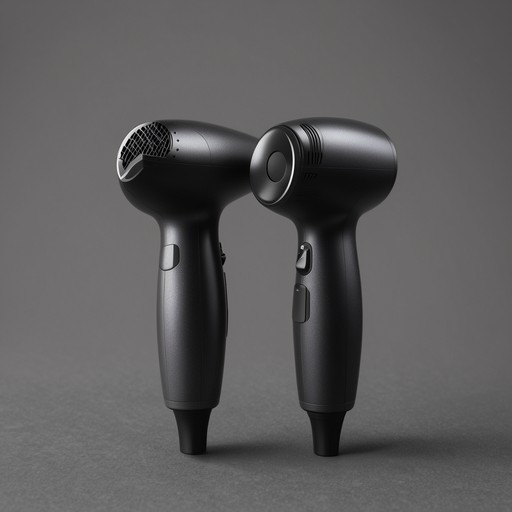} & \rimg[width=0.077\textwidth]{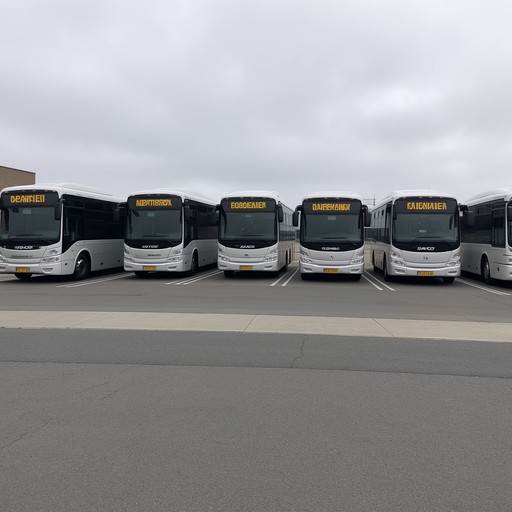} & \rimg[width=0.077\textwidth]{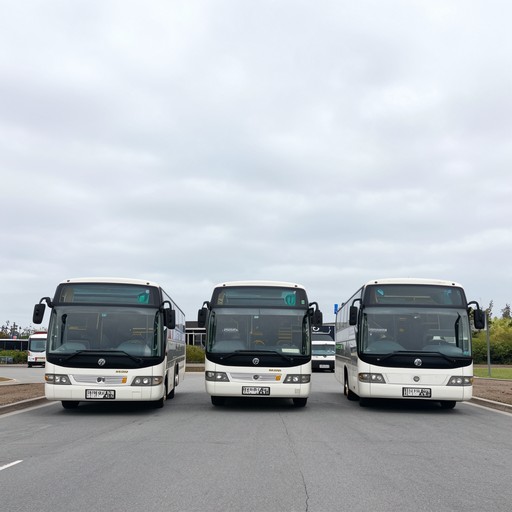} & \rimg[width=0.077\textwidth]{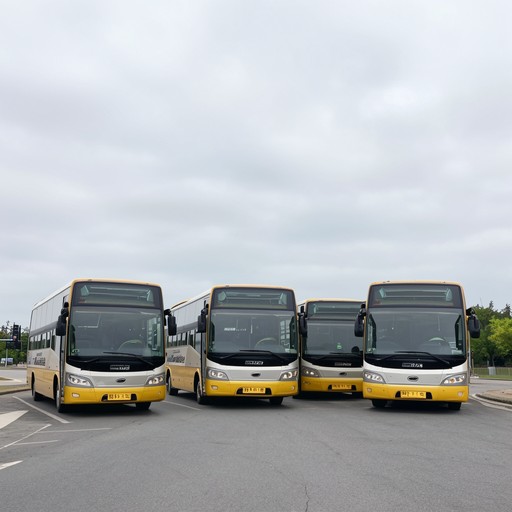} & \rimg[width=0.077\textwidth]{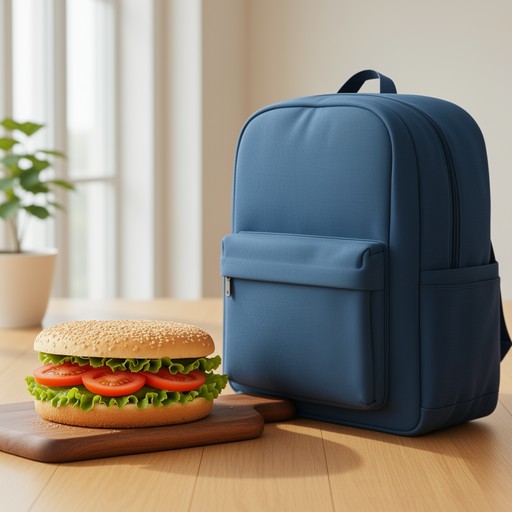} & \rimg[width=0.077\textwidth]{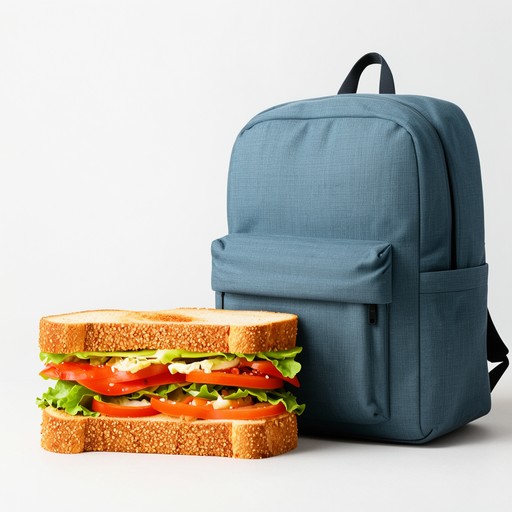} & \rimg[width=0.077\textwidth]{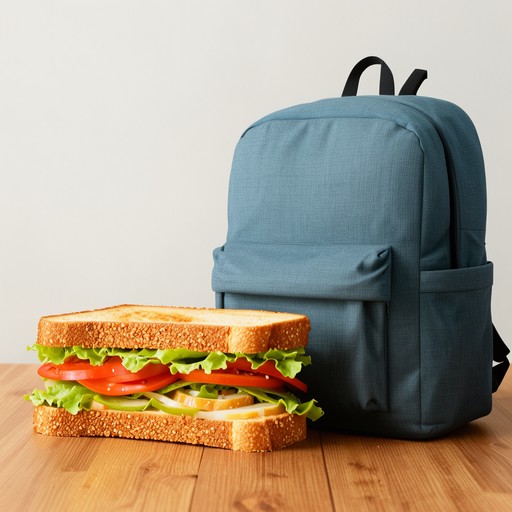} \\
  \multicolumn{3}{c}{\parbox[t]{0.23\textwidth}{\centering\scriptsize\textcolor{qmute}{\textit{``a photo of four skateboards''}}}} & \multicolumn{3}{c}{\parbox[t]{0.23\textwidth}{\centering\scriptsize\textcolor{qmute}{\textit{``a photo of two hair driers''}}}} & \multicolumn{3}{c}{\parbox[t]{0.23\textwidth}{\centering\scriptsize\textcolor{qmute}{\textit{``a photo of four buses''}}}} & \multicolumn{3}{c}{\parbox[t]{0.23\textwidth}{\centering\scriptsize\textcolor{qmute}{\textit{``a photo of a backpack right of a sandwich''}}}} \\[3pt]
  \rimg[width=0.077\textwidth]{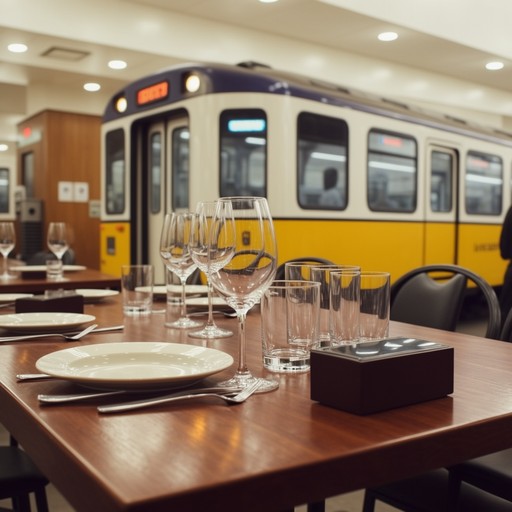} & \rimg[width=0.077\textwidth]{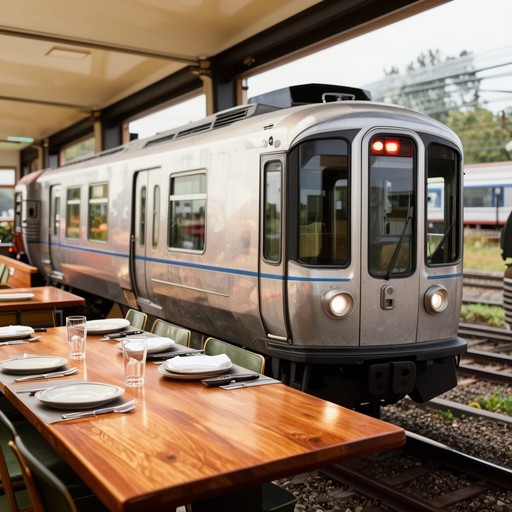} & \rimg[width=0.077\textwidth]{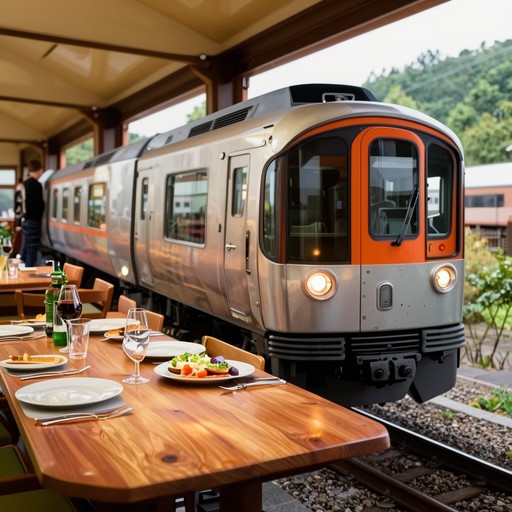} & \rimg[width=0.077\textwidth]{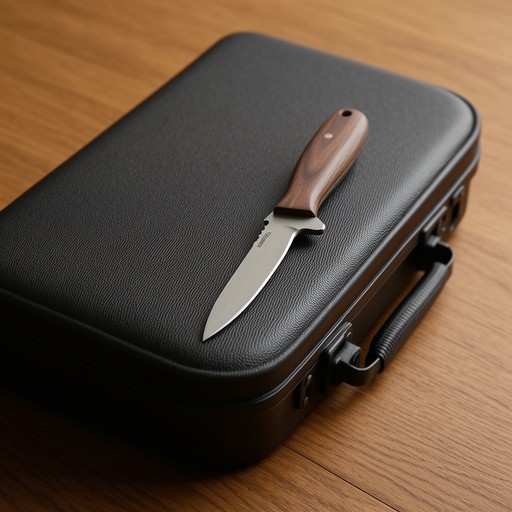} & \rimg[width=0.077\textwidth]{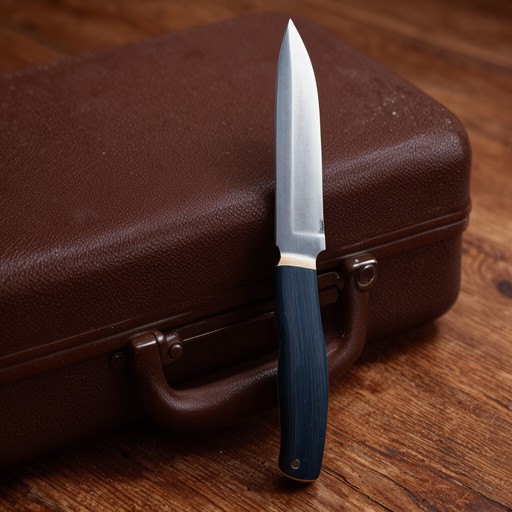} & \rimg[width=0.077\textwidth]{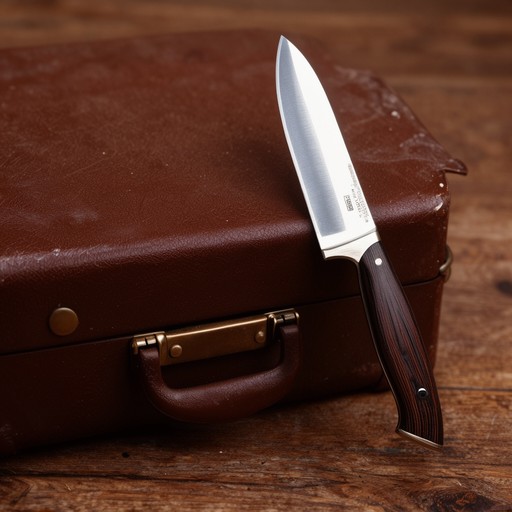} & \rimg[width=0.077\textwidth]{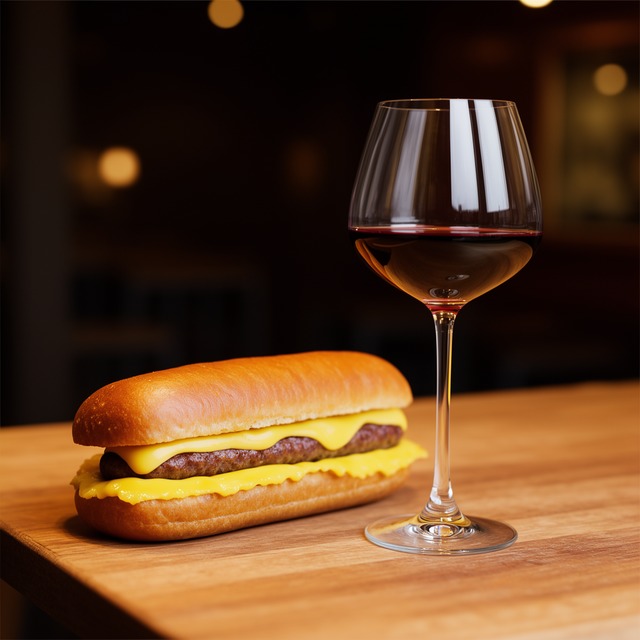} & \rimg[width=0.077\textwidth]{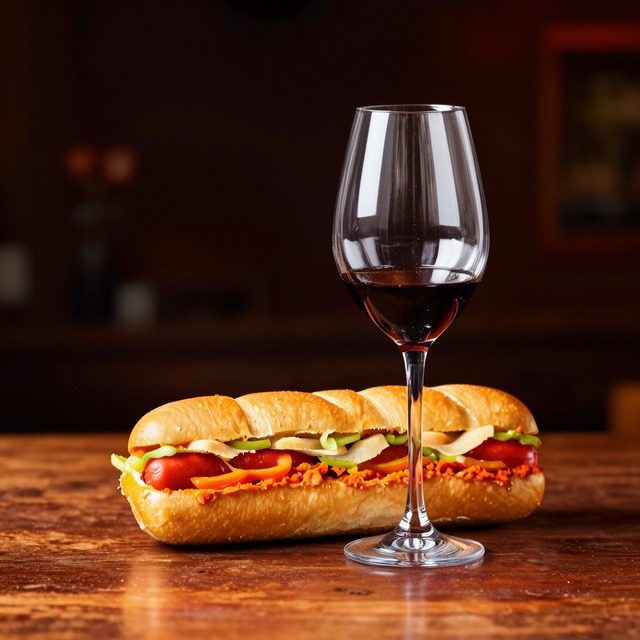} & \rimg[width=0.077\textwidth]{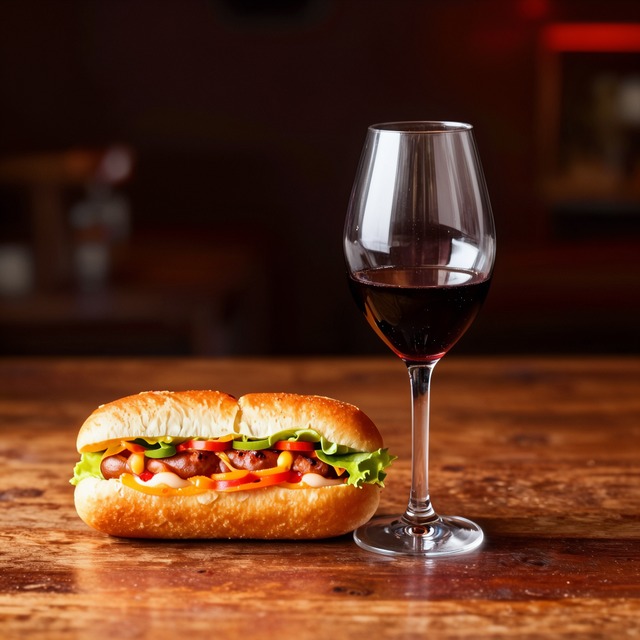} & \rimg[width=0.077\textwidth]{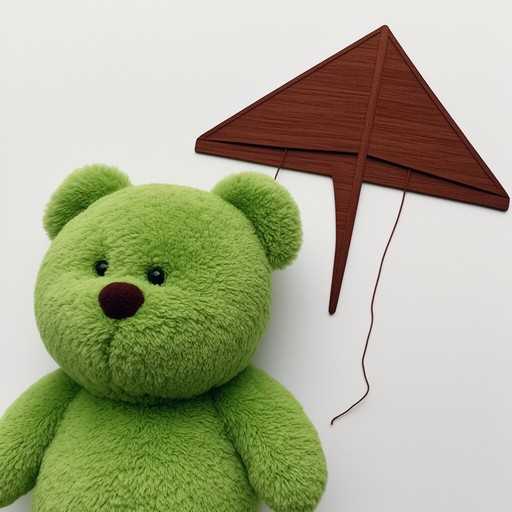} & \rimg[width=0.077\textwidth]{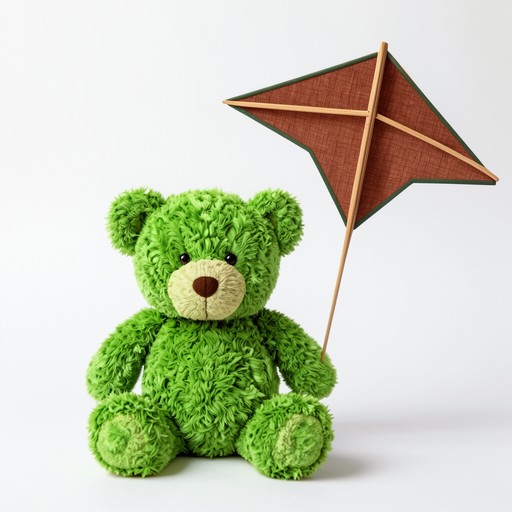} & \rimg[width=0.077\textwidth]{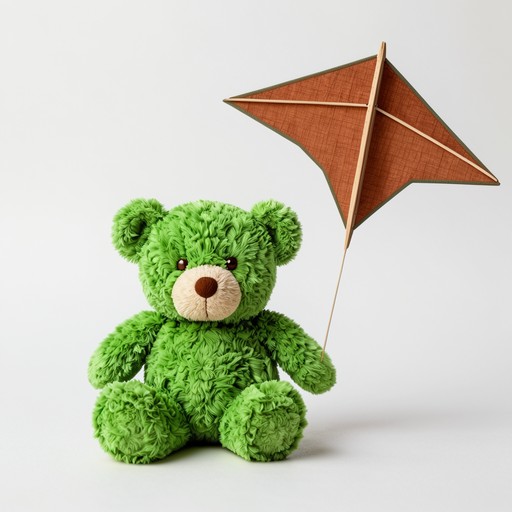} \\
  \multicolumn{3}{c}{\parbox[t]{0.23\textwidth}{\centering\scriptsize\textcolor{qmute}{\textit{``a photo of a train right of a dining table''}}}} & \multicolumn{3}{c}{\parbox[t]{0.23\textwidth}{\centering\scriptsize\textcolor{qmute}{\textit{``a photo of a knife right of a suitcase''}}}} & \multicolumn{3}{c}{\parbox[t]{0.23\textwidth}{\centering\scriptsize\textcolor{qmute}{\textit{``a photo of a wine glass right of a hot dog''}}}} & \multicolumn{3}{c}{\parbox[t]{0.23\textwidth}{\centering\scriptsize\textcolor{qmute}{\textit{``a photo of a green teddy bear and a brown kite''}}}} \\[3pt]
  \rimg[width=0.077\textwidth]{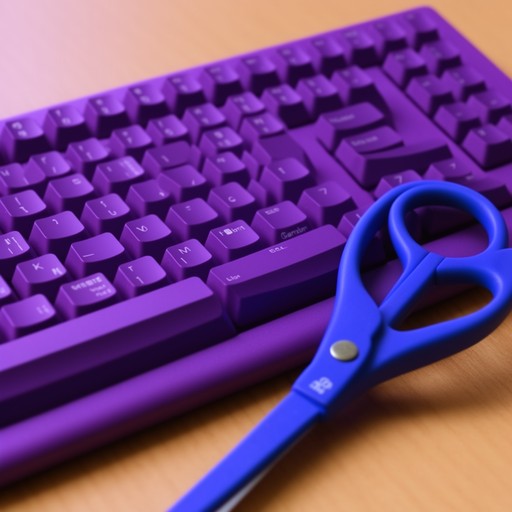} & \rimg[width=0.077\textwidth]{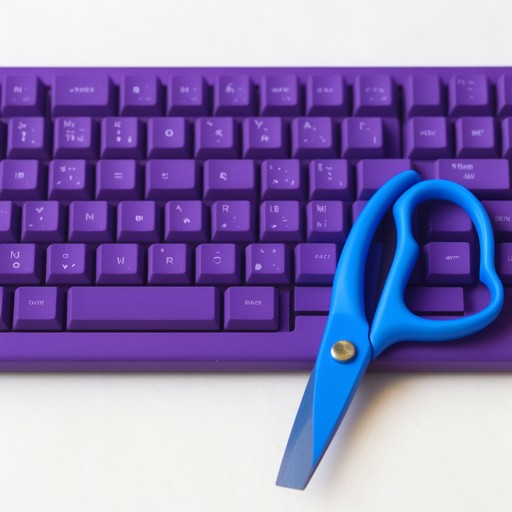} & \rimg[width=0.077\textwidth]{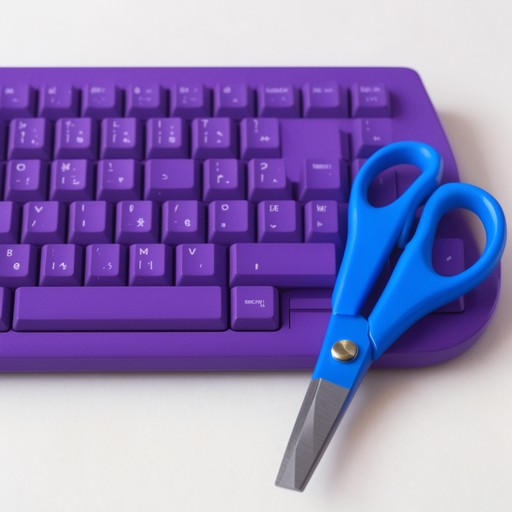} & \rimg[width=0.077\textwidth]{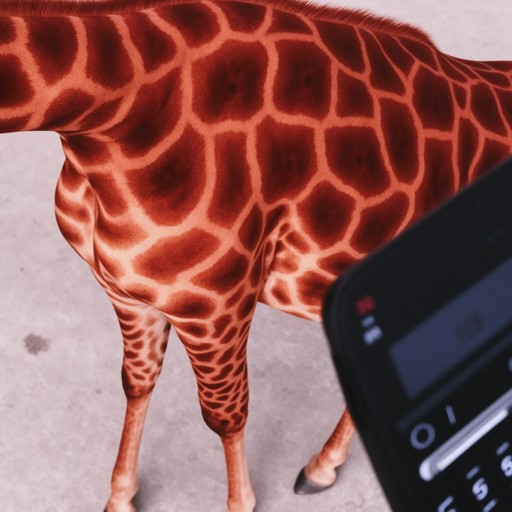} & \rimg[width=0.077\textwidth]{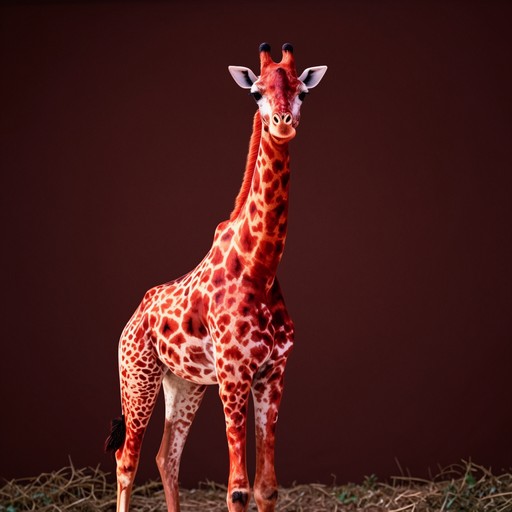} & \rimg[width=0.077\textwidth]{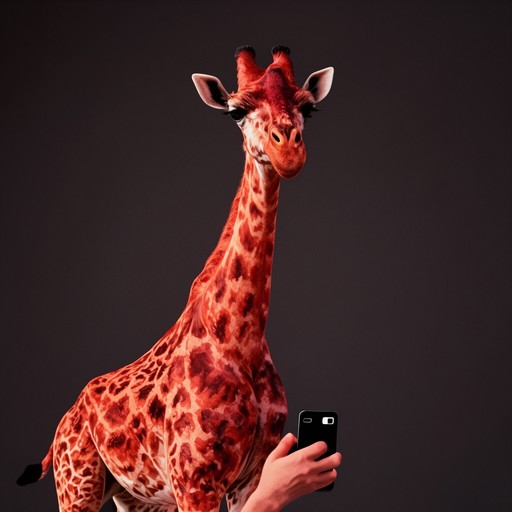} & \rimg[width=0.077\textwidth]{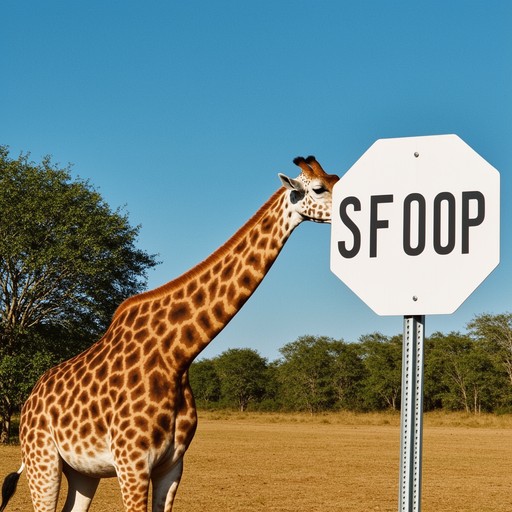} & \rimg[width=0.077\textwidth]{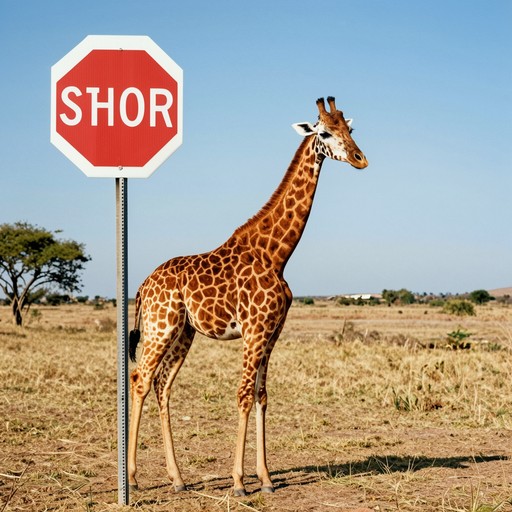} & \rimg[width=0.077\textwidth]{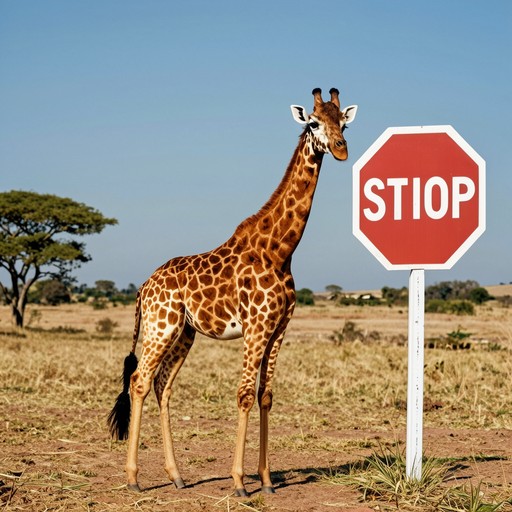} & \rimg[width=0.077\textwidth]{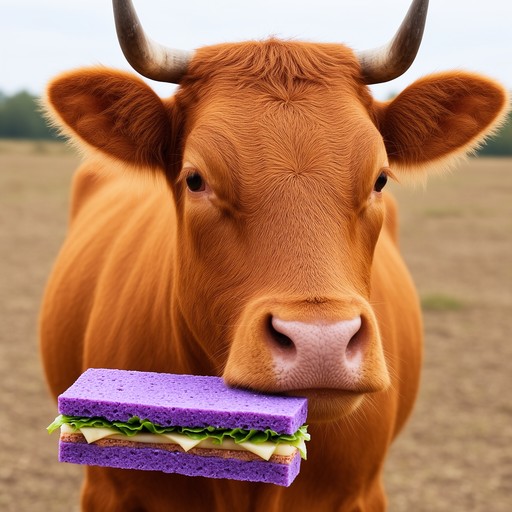} & \rimg[width=0.077\textwidth]{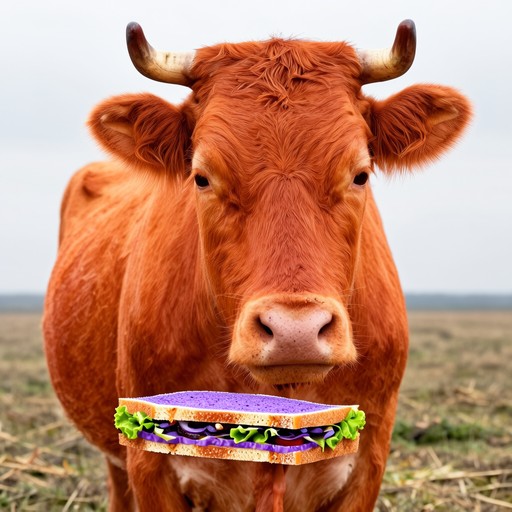} & \rimg[width=0.077\textwidth]{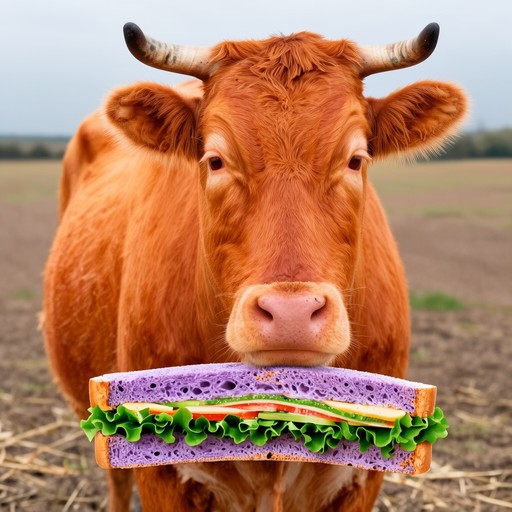} \\
  \multicolumn{3}{c}{\parbox[t]{0.23\textwidth}{\centering\scriptsize\textcolor{qmute}{\textit{``a photo of a purple computer keyboard and a blue scissors''}}}} & \multicolumn{3}{c}{\parbox[t]{0.23\textwidth}{\centering\scriptsize\textcolor{qmute}{\textit{``a photo of a red giraffe and a black cell phone''}}}} & \multicolumn{3}{c}{\parbox[t]{0.23\textwidth}{\centering\scriptsize\textcolor{qmute}{\textit{``a photo of a brown giraffe and a white stop sign''}}}} & \multicolumn{3}{c}{\parbox[t]{0.23\textwidth}{\centering\scriptsize\textcolor{qmute}{\textit{``a photo of an orange cow and a purple sandwich''}}}} \\[3pt]
  \end{tabular}\end{adjustbox}\end{qualband}
  \caption{\textbf{GenEval prompts on FLUX.2 [klein]} (same seed).}
  \label{fig:supp_flux_gen}
\end{figure*}

\Cref{fig:supp_flux_gen} shows the same on FLUX.2 [klein].

\subsection{Training seeds}
\label{sec:supp_seeds}
\Cref{tab:supp_seeds} repeats the training of \method and of the SFT control with two further seeds (2026011016 and
2026011017; the main paper uses 2026011015), each with its own LoRA initialisation, data order and noise, and with
everything else unchanged. \method is ahead of the SFT control with the same seed on every seed ($+0.9$, $+0.4$ and $+0.1$
GenEval; $73.4$ against $72.9$ on average), and the gain stays in the categories our descriptions state most often
(colours in $99\%$ and positions in $83\%$ of descriptions, \cref{sec:supp_data}): colour attribution ($65.0$ against $63.1$ on average, ahead on two of three seeds), position ($28.3$ against $27.3$) and
colours ($85.9$ against $85.2$). The GenEval advantage is therefore consistent across training seeds.

\begin{table}[H]
\caption{\textbf{GenEval over three training seeds}, \method and the SFT control.}
\label{tab:supp_seeds}
\centering\footnotesize
\setlength{\tabcolsep}{3pt}
\begin{adjustbox}{max width=\linewidth}
\begin{tabular}{llccccccc}
\toprule
Model & Seed & Overall & 1 obj & 2 obj & Count & Col. & Pos. & Attr. \\
\midrule
SFT control & 2026011015 & 73.4 & 99.1 & 90.9 & 72.8 & 84.8 & 30.5 & 62.3 \\
            & 2026011016 & 72.6 & 99.4 & 88.6 & 72.2 & 86.2 & 25.8 & 63.3 \\
            & 2026011017 & 72.8 & 99.4 & 89.4 & 73.8 & 84.6 & 25.8 & 63.8 \\
            & mean $\pm$ sd & $72.9 \pm 0.4$ & 99.3 & 89.7 & 72.9 & 85.2 & 27.3 & 63.1 \\
\midrule
\rowcolor{sdours}\textbf{\method} & 2026011015 & 74.3 & 99.7 & 89.4 & 71.6 & 85.6 & 32.5 & 67.0 \\
\rowcolor{sdours}            & 2026011016 & 72.9 & 99.7 & 87.1 & 73.4 & 86.7 & 26.0 & 64.5 \\
\rowcolor{sdours}            & 2026011017 & 72.9 & 99.4 & 89.9 & 72.5 & 85.4 & 26.5 & 63.5 \\
\rowcolor{sdours}            & mean $\pm$ sd & $\mathbf{73.4 \pm 0.8}$ & 99.6 & 88.8 & 72.5 & 85.9 & 28.3 & 65.0 \\
\bottomrule
\end{tabular}
\end{adjustbox}
\end{table}

\subsection{Significance of the GenEval gains}
\label{sec:supp_sig}
GenEval scores every prompt at four sampling seeds, and \method and the SFT control are sampled with the same seeds, so
the two can be compared image by image. We resample the $553$ prompts with replacement ($10{,}000$ times; each prompt
keeps its four images from both models) and recompute the score difference. On FLUX.2 [klein] \method's overall gain over
the SFT control is significant: $+1.86$ points, $95\%$ interval $[+0.59, +3.11]$, and \method is ahead on each of the four
sampling seeds. On both backbones the gain is significant in colour attribution, $+4.75$ and $+5.25$ points (paired
McNemar test on the images: $p=0.037$ and $p=0.019$), the category that the descriptions bind most directly
(\cref{sec:why}). Against the base models the gain is significant on both backbones ($+5.9$ and $+2.7$ points, intervals
$[+4.0, +7.8]$ and $[+0.8, +4.6]$). \Cref{tab:supp_sig} gives the differences per category.

\begin{table}[H]
\caption{\textbf{GenEval gains of \method over the SFT control}: score difference (points) and its $95\%$ paired
bootstrap interval over prompts ($10{,}000$ resamples), and the number of sampling seeds on which \method is ahead.}
\label{tab:supp_sig}
\centering\footnotesize
\setlength{\tabcolsep}{3pt}
\begin{adjustbox}{max width=\linewidth}
\begin{tabular}{lcccc}
\toprule
 & \multicolumn{2}{c}{SD3.5} & \multicolumn{2}{c}{FLUX.2 [klein]} \\
 & diff. & $95\%$ interval & diff. & $95\%$ interval \\
\midrule
Overall & $+0.90$ & $[-0.42, +2.26]$ & $+1.86$ & $[+0.59, +3.11]$ \\
Single object & $+0.62$ & $[+0.00, +1.87]$ & $+0.00$ & $[+0.00, +0.00]$ \\
Two objects & $-1.52$ & $[-5.05, +1.77]$ & $+2.02$ & $[-0.76, +5.05]$ \\
Counting & $-1.25$ & $[-5.62, +3.12]$ & $+0.63$ & $[-3.12, +4.06]$ \\
Colours & $+0.80$ & $[-1.06, +2.93]$ & $+0.53$ & $[-1.06, +2.13]$ \\
Position & $+2.00$ & $[-1.25, +5.25]$ & $+2.75$ & $[-1.25, +6.75]$ \\
Colour attribution & $+4.75$ & $[+0.25, +9.00]$ & $+5.25$ & $[+1.00, +9.50]$ \\
\midrule
Seeds with \method ahead & \multicolumn{2}{c}{$3$ of $4$} & \multicolumn{2}{c}{$4$ of $4$} \\
\bottomrule
\end{tabular}
\end{adjustbox}
\end{table}

\begin{table*}[!tp]
\caption{\textbf{GenEval of generation-only and unified models.} Baselines are quoted from their papers.}
\label{tab:unified}
\centering
\footnotesize
\setlength{\tabcolsep}{3pt}
\renewcommand{\arraystretch}{1.06}
\begin{tabularx}{\textwidth}{l>{\raggedright\arraybackslash}Xccccccccc}
\toprule
Type & Model & Params & Training budget & Single & Two & Count. & Colours & Pos. & \mbox{Col.\ attr.} & Overall$\uparrow$ \\
\midrule
\multirow{3}{*}{\shortstack{\textit{Gen.}\\\textit{only}}}
& SDXL~\cite{podell2024sdxl}          & 2.6B & -- & 0.98 & 0.74 & 0.39 & 0.85 & 0.15 & 0.23 & 0.55 \\
& Hunyuan-DiT~\cite{li2024hunyuan}    & 1.5B & -- & 0.97 & 0.77 & 0.71 & 0.88 & 0.13 & 0.30 & 0.63 \\
& SD3 Medium~\cite{esser2024scaling}  & 2.0B & -- & 0.98 & 0.74 & 0.63 & 0.67 & 0.34 & 0.36 & 0.62 \\
\midrule
\multirow{11}{*}{\textit{Unified}}
& \multicolumn{10}{l}{\textit{Fully discrete}} \\
& LWM~\cite{liu2025world}             & 7B   & -- & 0.93 & 0.41 & 0.46 & 0.79 & 0.09 & 0.15 & 0.47 \\
& Show-o~\cite{xie2025show}           & 1.3B & -- & 0.95 & 0.52 & 0.49 & 0.82 & 0.11 & 0.28 & 0.53 \\
& Emu3-Gen~\cite{wang2024emu3}        & 8B   & -- & 0.98 & 0.71 & 0.34 & 0.81 & 0.17 & 0.21 & 0.54 \\
& Muddit~\cite{shi2025muddit} (from Meissonic) & 1B & 10M pairs & 0.98 & 0.72 & 0.54 & 0.82 & 0.19 & 0.41 & 0.61 \\
& UniDisc~\cite{swerdlow2025unified}  & 1.4B & -- & 0.92 & 0.47 & 0.15 & 0.67 & 0.13 & 0.19 & 0.42 \\
\cmidrule(lr){2-11}
& \multicolumn{10}{l}{\textit{Hybrid discrete--continuous}} \\
& Transfusion~\cite{zhou2025transfusion} & 7B & 1T tokens & -- & -- & -- & -- & -- & -- & 0.63 \\
& JanusFlow~\cite{ma2025janusflow}    & 1.3B & -- & 0.97 & 0.59 & 0.45 & 0.83 & 0.53 & 0.42 & 0.63 \\
& OmniFlow~\cite{li2025omniflow} (from SD3-M) & 3.4B & 30M pairs & -- & -- & -- & -- & -- & -- & 0.62 \\
& D-DiT~\cite{li2025dual} (from SD3-M) & 2B   & 40M pairs & 0.97 & 0.80 & 0.54 & 0.76 & 0.32 & 0.50 & 0.65 \\
\midrule
& SD3.5 Medium (base, our measurement) & 2.5B & -- & 0.99 & 0.84 & 0.64 & 0.83 & 0.27 & 0.54 & 0.68 \\
& SFT control (no stream)              & 2.5B & 0.45B image tok. & 0.99 & 0.91 & 0.73 & 0.85 & 0.31 & 0.62 & 0.73 \\
\rowcolor{sdours}\cellcolor{white}\multirow{-3}{*}{\shortstack{\textit{Ours}\\(SD3.5-M)}}& \textbf{\method}                     & 2.5B$^{\dagger}$ & 13M text + 0.45B image tok. & 1.00 & 0.89 & 0.72 & 0.86 & 0.33 & 0.67 & 0.74 \\
\midrule
& FLUX.2 [klein] (base, our measurement) & 4B & -- & 0.99 & 0.89 & 0.72 & 0.88 & 0.52 & 0.59 & 0.76 \\
& SFT control (no stream)              & 4B & 0.45B image tok. & 0.99 & 0.87 & 0.81 & 0.89 & 0.43 & 0.64 & 0.77 \\
\rowcolor{fluxours}\cellcolor{white}\multirow{-3}{*}{\shortstack{\textit{Ours}\\(FLUX.2)}}& \textbf{\method}                     & 4B$^{\dagger}$ & 13M text + 0.45B image tok. & 0.99 & 0.89 & 0.82 & 0.90 & 0.45 & 0.69 & 0.79 \\
\bottomrule
\end{tabularx}
\\[2pt]
{\footnotesize $^{\dagger}$At inference; the text-only MLPs (597M on SD3.5, 566M on FLUX.2 [klein]; trained with LoRA of 47M and 48M) and the read-out head are not used with the stream off.}
\end{table*}

\begin{table*}[!tp]
\caption{\textbf{DPG-Bench of generation-only and unified models.} Baselines are quoted from \cite{mmflow}.}
\label{tab:dpg}
\centering
\footnotesize
\setlength{\tabcolsep}{6pt}
\renewcommand{\arraystretch}{1.06}
\begin{adjustbox}{max width=\textwidth}
\begin{tabular}{lccccccc}
\toprule
Model & Params & Global & Entity & Attr. & Relation & Other & Overall$\uparrow$ \\
\midrule
\multicolumn{8}{@{}l}{\textit{Generation only}} \\
SDXL~\cite{podell2024sdxl}          & 2.6B & 83.27 & 82.43 & 80.91 & 86.76 & 80.41 & 74.65 \\
Hunyuan-DiT~\cite{li2024hunyuan}    & 1.5B & 84.59 & 80.59 & 88.01 & 74.36 & 86.41 & 78.87 \\
SD3 Medium~\cite{esser2024scaling}  & 2.0B & \textbf{87.90} & 91.01 & 88.83 & 80.70 & \textbf{88.68} & 84.08 \\
\midrule
\multicolumn{8}{@{}l}{\textit{Unified}} \\
Show-o~\cite{xie2025show}           & 1.3B & 79.33 & 75.44 & 78.02 & 84.45 & 60.80 & 67.27 \\
Janus~\cite{wu2025janus}            & 1.3B & 82.33 & 87.38 & 87.70 & 85.46 & 86.41 & 79.68 \\
Emu3-Gen~\cite{wang2024emu3}        & 8B   & 85.21 & 86.68 & 86.84 & 90.22 & 83.15 & 80.60 \\
JanusFlow~\cite{ma2025janusflow}    & 1.3B & 87.03 & 87.31 & 87.39 & 89.79 & 88.10 & 80.09 \\
\midrule
\rowcolor{sdband}\multicolumn{8}{l}{\textbf{Ours (SD3.5 Medium)}} \\
SD3.5 Medium (base, our measurement)  & 2.5B & 83.89 & 90.34 & 88.40 & 93.27 & 81.20 & 84.20 \\
SFT control (no stream)             & 2.5B & 82.98 & 91.11 & 89.50 & 93.69 & 80.00 & 84.58 \\
\rowcolor{sdours}\textbf{\method}                    & 2.5B$^{\dagger}$ & 82.37 & 91.11 & 89.42 & 93.46 & 83.20 & 85.25 \\
\midrule
\rowcolor{fluxband}\multicolumn{8}{l}{\textbf{Ours (FLUX.2 [klein])}} \\
FLUX.2 [klein] (base, our measurement) & 4B & 82.07 & 90.13 & 89.48 & 93.50 & 77.20 & 83.15 \\
SFT control (no stream)             & 4B & 83.89 & 91.84 & 89.40 & 94.43 & 80.80 & 85.12 \\
\rowcolor{fluxours}\textbf{\method}                    & 4B$^{\dagger}$ & 83.59 & \textbf{91.97} & \textbf{89.80} & \textbf{94.51} & 81.20 & \textbf{85.36} \\
\bottomrule
\end{tabular}
\end{adjustbox}
\\[2pt]
{\footnotesize $^{\dagger}$At inference, as in \cref{tab:unified}.}
\end{table*}

\section{Judge and human evaluation}
\label{sec:supp_g_eval}
The blind MLLM judge's prompts, the fine-detail study, the human study and the ambiguous prompts.

\subsection{Judge prompts}
\label{sec:supp_judge}
\begin{figure*}[!p]
\centering
\begin{tcbraster}[raster columns=2,raster equal height,raster column skip=5pt,raster left skip=0pt,raster right skip=0pt]
\begin{tcolorbox}[judgebox=sdours]
{\small\bfseries\color{sdcol}Fine detail}\\[3pt]
\qchip{sdband}{Criterion}\\[2pt]
{\footnotesize\itshape The user asked an image generator for the request. Compare the two
images ONLY on refinement: how crisp, realistic and well resolved the fine textures and small parts are, for example fur,
hair and feathers, skin, fabric and stitching, wood, metal, glass, plastic and food surfaces, foliage, edges, buttons and
other small parts. Do NOT judge whether the image follows the request (which objects, counts, colours, positions), and do NOT
judge structural correctness (shape, anatomy, missing, merged or duplicated parts), composition, framing or general
aesthetics and style preference. Choose the image whose fine detail is better rendered. If you see no difference, pick the
one with slightly better fine detail and give close scores. Scores = refinement quality (10 = excellent fine detail).}\\[4pt]
\qchip{sdband}{Example sheet}\\[2pt]
\rimg[width=\linewidth]{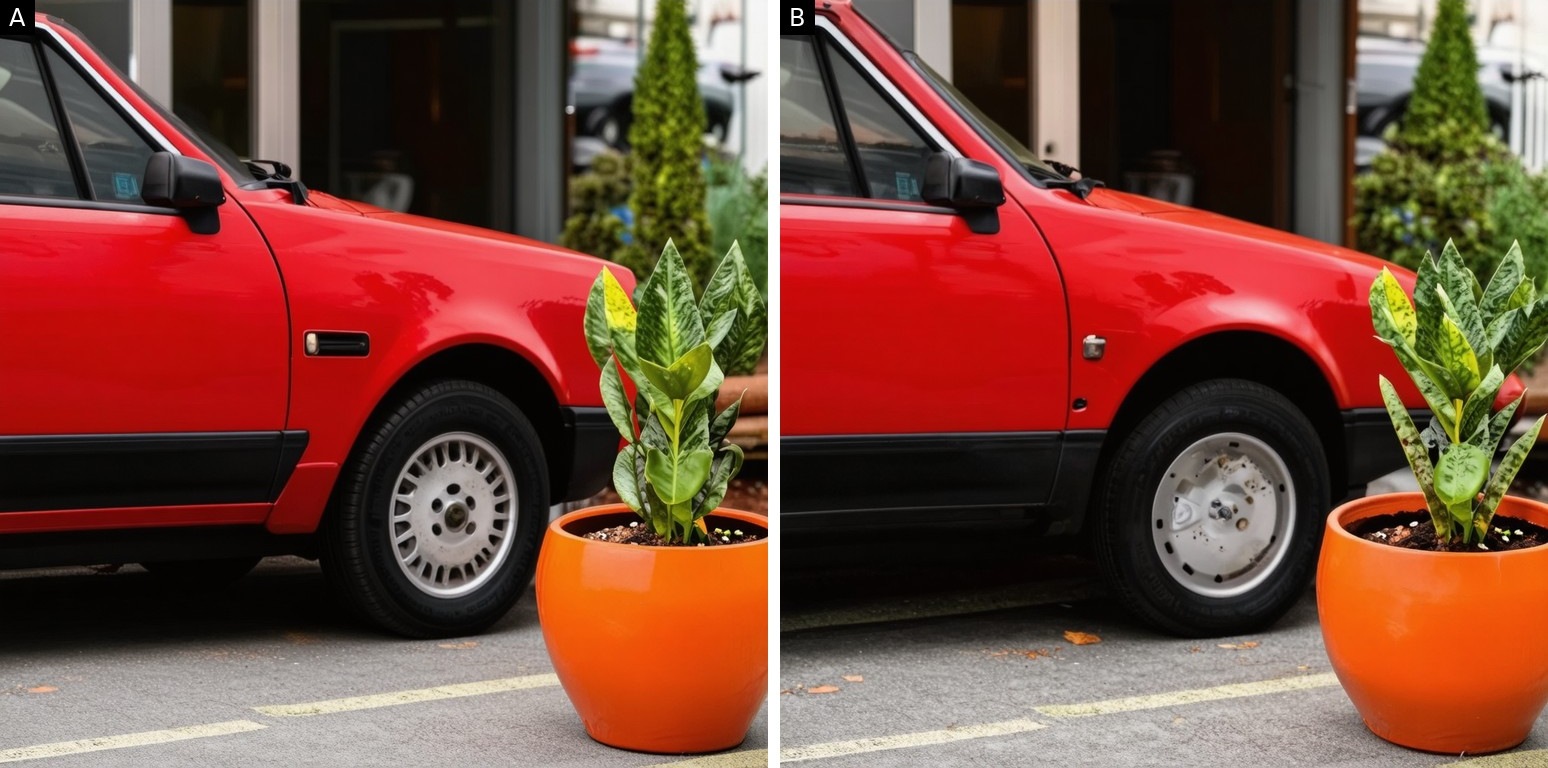}\\[2pt]
\qchip{sdband}{Request}\ {\footnotesize a photo of a red car and an orange potted plant}\\[3pt]
\qchip{sdband}{Judgement}\ {\footnotesize\ttfamily best: A, scores A 7, B 6}\\[1pt]
{\footnotesize ``A's alloy wheel spokes and tire tread are crisper; B's wheel is smudgy.''}\\[3pt]
{\scriptsize\color{qmute}Hidden from the judge: A is \method, B the SFT control.}
\end{tcolorbox}
\begin{tcolorbox}[judgebox=sdours]
{\small\bfseries\color{sdcol}Ambiguous prompts}\\[3pt]
\qchip{sdband}{Criterion}\\[2pt]
{\footnotesize\itshape The user gave an image generator a short, open-ended request. The two
images were generated for this request. The request leaves almost everything unspecified, so the generator has to decide what
to show. Choose the image that is the better result for this request: it depicts the request clearly and recognisably (for an
abstract word or a number, a meaningful, evocative visual interpretation rather than an arbitrary scene), makes sensible and
rich choices for what was left unspecified, and is coherent in its details, without malformed or nonsensical parts. Scores =
how good a result for the request each image is (10 = excellent).}\\[4pt]
\qchip{sdband}{Example sheet}\\[2pt]
\rimg[width=\linewidth]{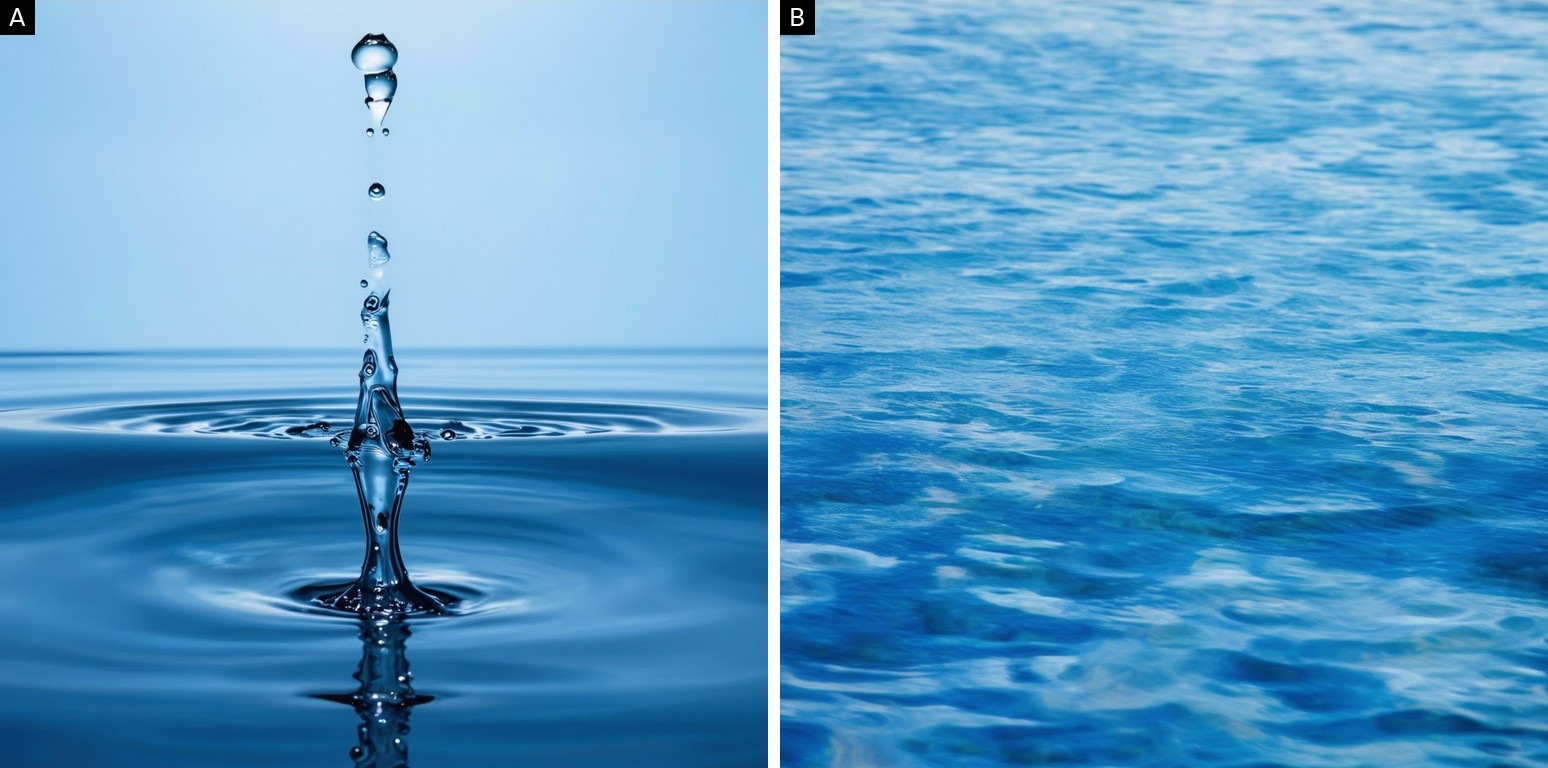}\\[2pt]
\qchip{sdband}{Request}\ {\footnotesize water}\\[3pt]
\qchip{sdband}{Judgement}\ {\footnotesize\ttfamily best: A, scores A 8, B 6}\\[1pt]
{\footnotesize ``A is a crisp macro of a water droplet splash with ripples, a striking and clear depiction of water. B is a
plain rippled blue surface texture, correct but generic.''}\\[3pt]
{\scriptsize\color{qmute}Hidden from the judge: A is \method, B SD3.5 Medium.}
\end{tcolorbox}
\end{tcbraster}

\caption{\textbf{Judge prompts}, with one real SD3.5 sheet and judgement per study.}
\label{fig:supp_judge}
\end{figure*}

\begin{figure*}[!tp]
\centering
\begin{tcbraster}[raster columns=2,raster equal height,raster column skip=5pt,raster left skip=0pt,raster right skip=0pt]
\begin{tcolorbox}[judgebox=sdours]
{\small\bfseries\color{sdcol}Fine detail with a tie option}\\[3pt]
\qchip{sdband}{Criterion}\\[2pt]
{\footnotesize\itshape The user asked an image generator for the request. Compare the two
images ONLY on refinement: how crisp, realistic and well resolved the fine textures and small parts are, for example fur,
hair and feathers, skin, fabric and stitching, wood, metal, glass, plastic and food surfaces, foliage, edges, buttons and
other small parts. Do NOT judge whether the image follows the request (which objects, counts, colours, positions), and do NOT
judge structural correctness (shape, anatomy, missing, merged or duplicated parts), composition, framing or general
aesthetics and style preference. Choose the image whose fine detail is better rendered. \qhl{If both are equally good in
refinement, answer "tie".} Scores = refinement quality (10 = excellent fine detail).}\\[4pt]
\qchip{sdband}{Answer}\\[2pt]
{\scriptsize\ttfamily "best": "A", "B" \qhl{or "tie"}}\\[1pt]
{\footnotesize\itshape \qhl{Choose A, B, or "tie" (Both are equally good).}}
\end{tcolorbox}
\begin{tcolorbox}[judgebox=sdours]
\qchip{sdband}{Example sheet}\\[2pt]
\rimg[width=\linewidth]{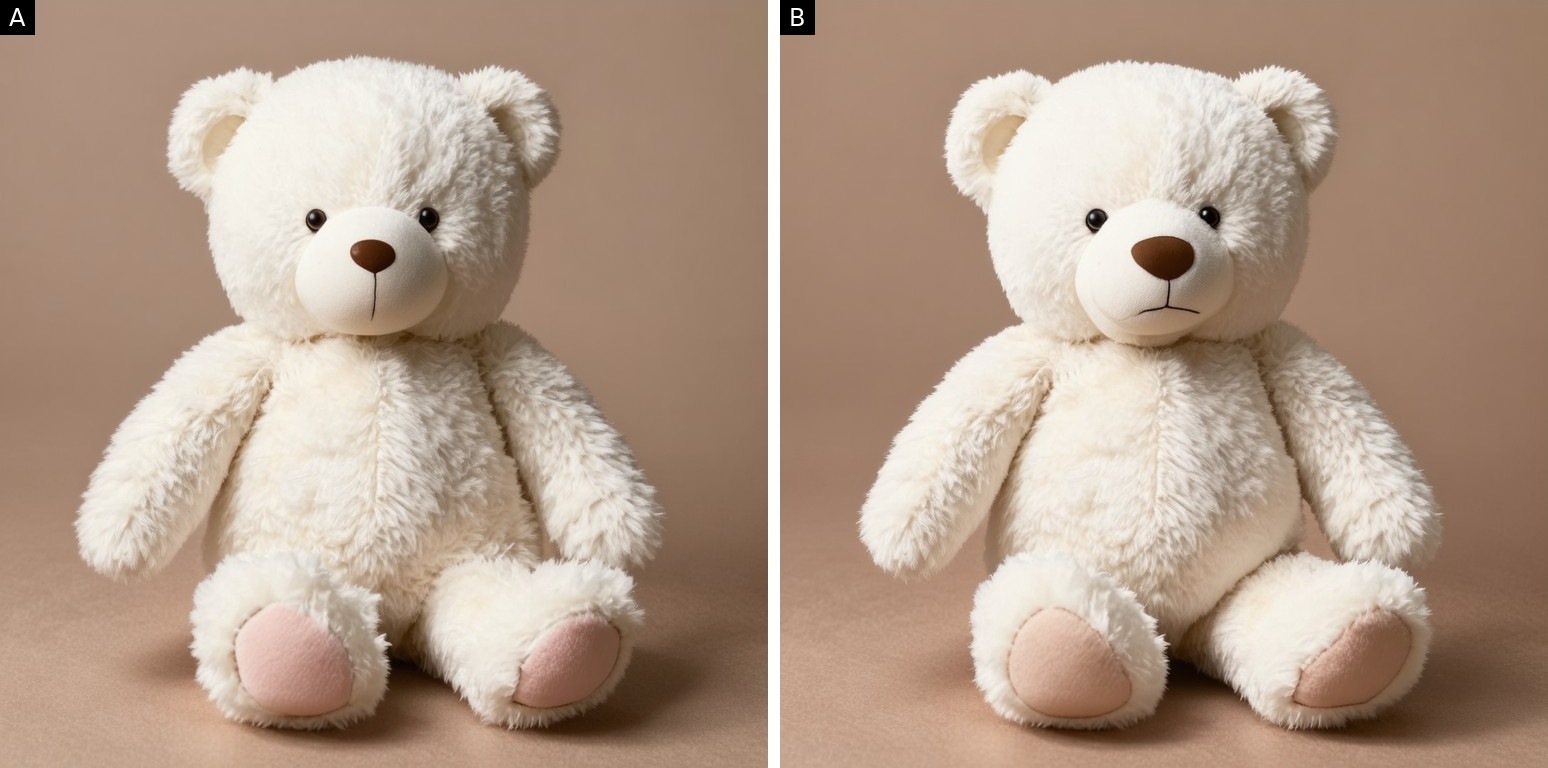}\\[2pt]
\qchip{sdband}{Request}\ {\footnotesize a photo of a white teddy bear}\\[3pt]
\qchip{sdband}{Judgement}\ {\footnotesize\ttfamily best: tie, scores A 8, B 8}\\[1pt]
{\footnotesize ``Both render fluffy plush fur and felt paw pads with similar fidelity.''}
\end{tcolorbox}
\end{tcbraster}
\caption{\textbf{Fine-detail judge with a tie option}, with one real sheet and judgement. \textcolor{accent}{\textbf{Highlighted}} text is what
changes. In the binary comparison (\cref{fig:supp_judge}) the judge must choose A or B, even when it sees no difference. A forced choice
can bias the judge, so here it may answer ``tie'' when the two images are close.}
\label{fig:supp_judge_tie}
\end{figure*}

\begin{figure*}[!tp]
\centering
\begin{tcbraster}[raster columns=2,raster equal height,raster column skip=5pt,raster left skip=0pt,raster right skip=0pt]
\begin{tcolorbox}[judgebox=sdours]
{\small\bfseries\color{sdcol}Single-image rating}\\[3pt]
\qchip{sdband}{Setup}\\[2pt]
{\footnotesize\itshape The image was generated for the request by an image generator; you are not told which.
Open and look at it yourself, and rate it against the scale below. Use
the whole scale: keep 5 for images that truly have no flaw of this kind, and give low scores when they are deserved.}\\[4pt]
\qchip{sdband}{Criterion}\\[2pt]
{\footnotesize\itshape Rate ONLY refinement: how crisp, realistic and well resolved the fine textures and small parts are, for
example fur, hair and feathers, skin, fabric and stitching, wood, metal, glass, plastic and food surfaces, foliage, edges,
buttons and other small parts. For drawn or stylised images, judge how clean and well resolved the lines, shading and small
details are for that style. Do NOT judge whether the image follows the request, and do NOT judge structural correctness (shape,
anatomy, missing, merged or duplicated parts), composition, framing or general aesthetics and style preference.}\\[4pt]
\qchip{sdband}{Scale}\\[2pt]
{\footnotesize\itshape 5 = fine detail is crisp and convincing everywhere, including small parts and edges.\\
4 = crisp overall, with slight softness or smearing in a few small areas.\\
3 = acceptable, but textures are noticeably soft, waxy, plasticky or smudged in places.\\
2 = textures are mostly smeared or waxy, and small parts are poorly resolved.\\
1 = very blurry or mushy, with almost no fine detail.\\
0 = no usable detail (noise, extreme blur or heavy artefacts).}
\end{tcolorbox}
\begin{tcolorbox}[judgebox=sdours]
\qchip{sdband}{Example image}\\[2pt]
\centerline{\rimg[width=0.72\linewidth]{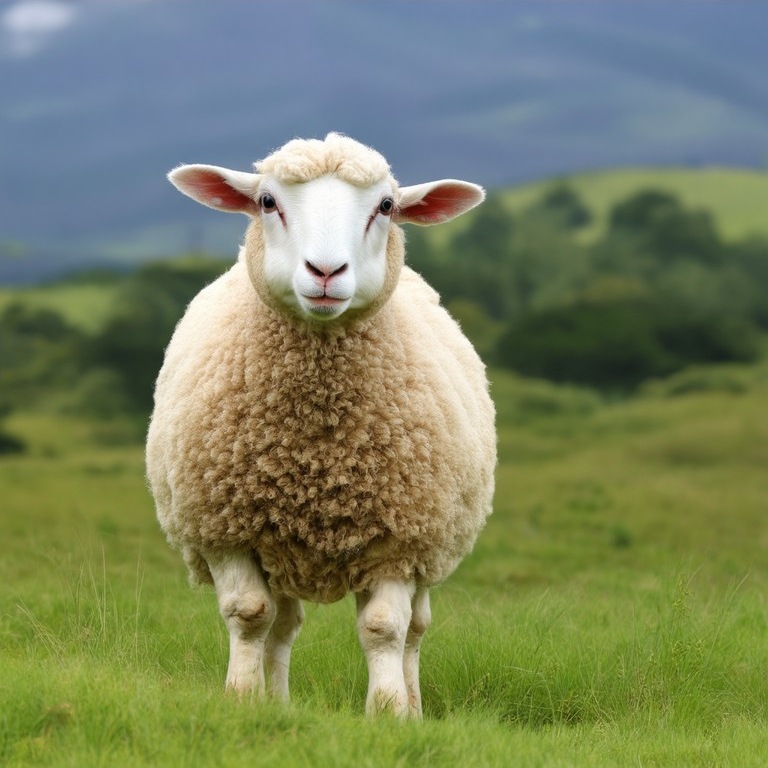}}\\[2pt]
\qchip{sdband}{Request}\ {\footnotesize a photo of a sheep}\\[3pt]
\qchip{sdband}{Judgement}\ {\footnotesize\ttfamily score: 4}\\[1pt]
{\footnotesize ``Wool curls detailed; face slightly smooth.''}\\[3pt]
{\scriptsize\color{qmute}Hidden from the judge: the image is from SD3.5 Medium.}
\end{tcolorbox}
\end{tcbraster}
\caption{\textbf{Single-image rating judge}, with one real image and judgement. Unlike the binary comparison (\cref{fig:supp_judge}),
this is not a choice between two images. An image can have good parts and bad parts, so the judge sees one image, with no pair,
and gives it a score from 0 to 5; every image is rated by its own fresh judge, which sees no other image, and the scores of the three
models are compared to show which model generates better images overall.}
\label{fig:supp_judge_rate}
\end{figure*}

All preference results of the main paper come from blind pairwise judgements by Claude Opus 5.5; \cref{sec:supp_refine} repeats the
fine-detail studies with an open-weight judge, Qwen3-VL-32B-Instruct. Each judge is a fresh session
with no other context, no memory and no other sample: it sees exactly one sheet. A sheet is one pair of images side by side, labelled A and B, generated for the
same prompt and seed by two models. Which model is on which side is randomised, and the key is kept away from the judge. For
every sheet the judge picks the better image, scores both from 0 to 10 and gives one to three sentences of reasoning. \Cref{fig:supp_judge} gives the instructions,
leaving out file handling, with one real sheet per study, and \cref{fig:supp_judge_tie,fig:supp_judge_rate} give the two variants of the
fine-detail judge. The fine-detail judges see every GenEval prompt that all compared models get right at all four sampling seeds (216 on
SD3.5, 259 on FLUX.2 [klein]), each comparison with its own sheets and A/B order, and every sheet with its own judge; the ambiguous
prompts are the 100 prompts of \cref{sec:supp_ambig} at three seeds.

\subsection{Fine detail of correct images}
\label{sec:supp_refine}
\begin{figure}[H]
  \centering
  \includegraphics[width=\linewidth]{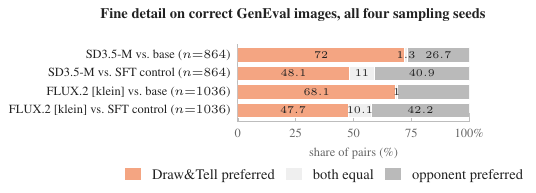}
  \caption{\textbf{Fine detail with a ``both are equally good'' option, averaged over all four sampling seeds} (the GenEval
  prompts all three models get right at every seed, at all four seeds): share of pairs.}
  \label{fig:supp_tie}
\end{figure}

\paragraph{Average over all four sampling seeds.} The main paper reports the fine-detail study with a ``both are equally good''
option at one sampling seed per backbone, the seed with the largest margin of \method's wins over its losses across both
comparisons (seed 3 for SD3.5, seed 2 for FLUX.2 [klein]). The study takes, per backbone, every GenEval prompt that the
base model, the SFT control and \method all get right at all four sampling seeds: 216 prompts on SD3.5 and 259 on FLUX.2
[klein], so 864 and 1{,}036 prompt--seed samples, the same prompts at every seed. \Cref{fig:supp_tie} and the last column of
\cref{tab:supp_tie_seeds} give the result \emph{averaged over all four seeds}, that is, pooled over all samples; the other
columns give each seed on its own. Averaged over the four seeds, the judge sees no difference in $1\%$ of pairs against the
base model on both backbones and in $11\%$ (SD3.5) and $10\%$ (FLUX.2 [klein]) against the SFT control. Among the pairs it
decides, \method is preferred to the base model in $72.9\%$ ($853$ decided pairs $=$ $622$ wins $+$ $231$ losses) and $68.7\%$ ($1{,}026$ decided pairs $=$ $705$ wins $+$ $321$ losses)
and to the SFT control in $54.1\%$ ($769$ decided pairs $=$ $416$ wins $+$ $353$ losses) and $53.1\%$ ($931$ decided pairs $=$ $494$ wins $+$ $437$ losses). The judge picks the left-hand image slightly more often ($54\%$ of the pairs it decides);
\method is on either side about equally often, so this favours neither model.

\begin{table*}[t]
\caption{\textbf{Fine detail with a tie option, by sampling seed}: share of pairs (\%) in which \method is preferred (win), both images are judged equally good (tie) and the opponent is preferred (loss), for Claude Opus 5.5 and the open-weight judge Qwen3-VL-32B on the same sheets; the last group is the average over all four seeds (pooled). Shaded: against the SFT control. $216$ pairs per seed on SD3.5 and $259$ on FLUX.2 [klein] ($864$ and $1{,}036$ pooled).}
\label{tab:supp_tie_seeds}
\centering\scriptsize
\setlength{\tabcolsep}{2.8pt}
\begin{adjustbox}{max width=\textwidth}
\begin{tabular}{@{}ll@{\hspace{5pt}}ccc@{\hspace{5pt}}ccc@{\hspace{5pt}}ccc@{\hspace{5pt}}ccc@{\hspace{7pt}}ccc@{}}
\toprule
& & \multicolumn{3}{c}{seed 0} & \multicolumn{3}{c}{seed 1} & \multicolumn{3}{c}{seed 2} & \multicolumn{3}{c}{seed 3} & \multicolumn{3}{c}{\textbf{all four}} \\
\cmidrule(lr){3-5}\cmidrule(lr){6-8}\cmidrule(lr){9-11}\cmidrule(lr){12-14}\cmidrule(lr){15-17}
Backbone & Comparison & Win & Tie & Loss & Win & Tie & Loss & Win & Tie & Loss & Win & Tie & Loss & Win & Tie & Loss \\
\midrule
\multicolumn{17}{@{}l}{\textbf{Claude Opus 5.5}} \\[1pt]
SD3.5-M & vs.\ base & 70.4 & 2.8 & 26.9 & 69.4 & 0.5 & 30.1 & 74.5 & 0.9 & 24.5 & 73.6 & 0.9 & 25.5 & \textbf{72.0} & 1.3 & 26.7 \\
\rowcolor{sdours} & vs.\ SFT control & 53.7 & 6.5 & 39.8 & 45.4 & 13.4 & 41.2 & 44.0 & 12.0 & 44.0 & 49.5 & 12.0 & 38.4 & \textbf{48.1} & 11.0 & 40.9 \\
FLUX.2 [klein] & vs.\ base & 69.5 & 0.8 & 29.7 & 68.0 & 0.8 & 31.3 & 63.7 & 1.5 & 34.7 & 71.0 & 0.8 & 28.2 & \textbf{68.1} & 1.0 & 31.0 \\
\rowcolor{fluxours} & vs.\ SFT control & 46.7 & 9.3 & 44.0 & 41.3 & 13.1 & 45.6 & 55.2 & 8.5 & 36.3 & 47.5 & 9.7 & 42.9 & \textbf{47.7} & 10.1 & 42.2 \\
\midrule
\multicolumn{17}{@{}l}{\textbf{Qwen3-VL-32B}} \\[1pt]
SD3.5-M & vs.\ base & 60.6 & 0.0 & 39.4 & 63.4 & 0.0 & 36.6 & 69.4 & 0.0 & 30.6 & 63.0 & 0.0 & 37.0 & \textbf{64.1} & 0.0 & 35.9 \\
\rowcolor{sdours} & vs.\ SFT control & 54.6 & 0.0 & 45.4 & 56.9 & 0.5 & 42.6 & 50.5 & 0.9 & 48.6 & 55.1 & 0.0 & 44.9 & \textbf{54.3} & 0.3 & 45.4 \\
FLUX.2 [klein] & vs.\ base & 54.4 & 0.0 & 45.6 & 57.5 & 0.0 & 42.5 & 53.3 & 0.0 & 46.7 & 56.0 & 0.0 & 44.0 & \textbf{55.3} & 0.0 & 44.7 \\
\rowcolor{fluxours} & vs.\ SFT control & 54.4 & 0.0 & 45.6 & 47.5 & 0.8 & 51.7 & 52.9 & 0.4 & 46.7 & 51.4 & 0.0 & 48.6 & \textbf{51.5} & 0.3 & 48.2 \\
\bottomrule
\end{tabular}
\end{adjustbox}
\end{table*}

\begin{figure}[H]
  \centering
  \includegraphics[width=\linewidth]{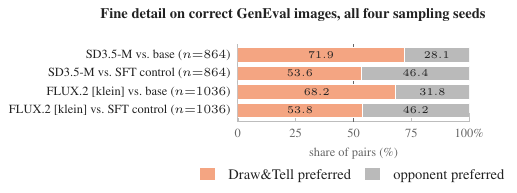}
  \caption{\textbf{Fine detail with forced choice, averaged over all four sampling seeds} (the same samples as \cref{fig:supp_tie}):
  share of pairs in which each image is preferred.}
  \label{fig:supp_refine_all}
\end{figure}

\paragraph{Forced choice.} A tie is an easy answer for a judge looking at two similar images. We therefore ask what happens when
the judge has to choose: the same samples with the options A and B only (\cref{fig:supp_judge}). Even then
\method stays ahead of both models. Averaged over the four seeds (\cref{fig:supp_refine_all,tab:supp_sr_seeds}) it is preferred to
the base model in $71.9\%$ (SD3.5) and $68.2\%$ (FLUX.2 [klein]) of pairs and to the SFT control in $53.6\%$ and $53.8\%$.

\begin{table*}[t]
\caption{\textbf{Fine detail with forced choice, by sampling seed}: share of pairs (\%) in which \method is preferred (win) and the opponent is preferred (loss), for Claude Opus 5.5 and the open-weight judge Qwen3-VL-32B on the same sheets; the last group is the average over all four seeds (pooled). Shaded: against the SFT control. $216$ pairs per seed on SD3.5 and $259$ on FLUX.2 [klein] ($864$ and $1{,}036$ pooled).}
\label{tab:supp_sr_seeds}
\centering\scriptsize
\setlength{\tabcolsep}{2.8pt}
\begin{adjustbox}{max width=\textwidth}
\begin{tabular}{@{}ll@{\hspace{5pt}}cc@{\hspace{5pt}}cc@{\hspace{5pt}}cc@{\hspace{5pt}}cc@{\hspace{7pt}}cc@{}}
\toprule
& & \multicolumn{2}{c}{seed 0} & \multicolumn{2}{c}{seed 1} & \multicolumn{2}{c}{seed 2} & \multicolumn{2}{c}{seed 3} & \multicolumn{2}{c}{\textbf{all four}} \\
\cmidrule(lr){3-4}\cmidrule(lr){5-6}\cmidrule(lr){7-8}\cmidrule(lr){9-10}\cmidrule(lr){11-12}
Backbone & Comparison & Win & Loss & Win & Loss & Win & Loss & Win & Loss & Win & Loss \\
\midrule
\multicolumn{12}{@{}l}{\textbf{Claude Opus 5.5}} \\[1pt]
SD3.5-M & vs.\ base & 71.3 & 28.7 & 69.0 & 31.0 & 73.1 & 26.9 & 74.1 & 25.9 & \textbf{71.9} & 28.1 \\
\rowcolor{sdours} & vs.\ SFT control & 57.9 & 42.1 & 53.2 & 46.8 & 52.3 & 47.7 & 50.9 & 49.1 & \textbf{53.6} & 46.4 \\
FLUX.2 [klein] & vs.\ base & 70.3 & 29.7 & 70.7 & 29.3 & 64.5 & 35.5 & 67.6 & 32.4 & \textbf{68.2} & 31.8 \\
\rowcolor{fluxours} & vs.\ SFT control & 55.6 & 44.4 & 49.4 & 50.6 & 58.7 & 41.3 & 51.4 & 48.6 & \textbf{53.8} & 46.2 \\
\midrule
\multicolumn{12}{@{}l}{\textbf{Qwen3-VL-32B}} \\[1pt]
SD3.5-M & vs.\ base & 61.1 & 38.9 & 60.6 & 39.4 & 67.1 & 32.9 & 61.1 & 38.9 & \textbf{62.5} & 37.5 \\
\rowcolor{sdours} & vs.\ SFT control & 52.8 & 47.2 & 57.9 & 42.1 & 51.4 & 48.6 & 58.8 & 41.2 & \textbf{55.2} & 44.8 \\
FLUX.2 [klein] & vs.\ base & 54.8 & 45.2 & 55.6 & 44.4 & 54.4 & 45.6 & 57.9 & 42.1 & \textbf{55.7} & 44.3 \\
\rowcolor{fluxours} & vs.\ SFT control & 55.2 & 44.8 & 47.1 & 52.9 & 56.4 & 43.6 & 52.9 & 47.1 & \textbf{52.9} & 47.1 \\
\bottomrule
\end{tabular}
\end{adjustbox}
\end{table*}

\paragraph{Open-weight judge.} To check that the preference is not particular to one judge, we repeat both fine detail
studies on the same samples with an open-weight judge, Qwen3-VL-32B-Instruct, with the same instructions,
sheets and A/B order, and again a separate request for every sheet (the sheet is attached to the request instead of being
opened by the judge). The open-weight judge also favours \method, in every pooled comparison
(Qwen3-VL rows of \cref{tab:supp_tie_seeds,tab:supp_sr_seeds}). With the tie option \method is preferred, among the pairs
decided, to the base model in $64.1\%$ (SD3.5; $864$ decided pairs $=$ $554$ wins $+$ $310$ losses) and $55.3\%$ (FLUX.2 [klein]; $1{,}036$ decided pairs $=$ $573$ wins $+$ $463$ losses) and to the
SFT control in $54.5\%$ ($861$ decided pairs $=$ $469$ wins $+$ $392$ losses) and $51.7\%$ ($1{,}033$ decided pairs $=$ $534$ wins $+$ $499$ losses);
with forced choice, to the base model in $62.5\%$ and $55.7\%$ and to the SFT control in $55.2\%$ and $52.9\%$. Qwen3-VL almost
never answers tie (6 of 3{,}800 pairs) and favours the right-hand image ($72\%$ of the pairs it decides); \method is on each side
in exactly half of the pairs, so this favours neither model.

\paragraph{SFT control against the base model.} To separate the effect of the data from that of writing, we also compare
the SFT control with the base model on the same samples, with the same instruction and sheet layout (A/B balanced, every sheet
with its own fresh judge). \Cref{tab:supp_sftbase_seeds} gives the result. Both judges prefer the SFT control to the base model
on both backbones: averaged over the four seeds, among the pairs decided, in $71.8\%$ (SD3.5; $856$ decided pairs $=$ $615$ wins $+$ $241$ losses) and
$67.8\%$ (FLUX.2 [klein]; $1{,}023$ decided pairs $=$ $694$ wins $+$ $329$ losses) for Claude Opus 5.5 and in $58.2\%$ ($864$ decided pairs $=$ $503$ wins $+$ $361$ losses) and $52.4\%$
($1{,}036$ decided pairs $=$ $543$ wins $+$ $493$ losses) for Qwen3-VL, close to \method's shares against the base model ($72.9\%$ and $68.7\%$;
$64.1\%$ and $55.3\%$, \cref{tab:supp_tie_seeds}). Gains over the base model can thus also come from the quality of the training data. However,
\method outperforms the SFT control (\cref{tab:supp_tie_seeds}), which sees the same data, batches and noise and differs
only in that it does not write. We conclude that writing helps: learning to draw and write is better than learning to draw
alone.

\begin{table*}[t]
\caption{\textbf{SFT control against the base model}, fine detail with a tie option, by sampling seed: share of pairs (\%) in which the SFT control is preferred (win), both images are judged equally good (tie) and the base model is preferred (loss), on the samples of \cref{tab:supp_tie_seeds}, for Claude Opus 5.5 and Qwen3-VL-32B; the last group is the average over all four seeds (pooled).}
\label{tab:supp_sftbase_seeds}
\centering\scriptsize
\setlength{\tabcolsep}{2.8pt}
\begin{adjustbox}{max width=\textwidth}
\begin{tabular}{@{}ll@{\hspace{5pt}}ccc@{\hspace{5pt}}ccc@{\hspace{5pt}}ccc@{\hspace{5pt}}ccc@{\hspace{7pt}}ccc@{}}
\toprule
& & \multicolumn{3}{c}{seed 0} & \multicolumn{3}{c}{seed 1} & \multicolumn{3}{c}{seed 2} & \multicolumn{3}{c}{seed 3} & \multicolumn{3}{c}{\textbf{all four}} \\
\cmidrule(lr){3-5}\cmidrule(lr){6-8}\cmidrule(lr){9-11}\cmidrule(lr){12-14}\cmidrule(lr){15-17}
Backbone & Comparison & Win & Tie & Loss & Win & Tie & Loss & Win & Tie & Loss & Win & Tie & Loss & Win & Tie & Loss \\
\midrule
\multicolumn{17}{@{}l}{\textbf{Claude Opus 5.5}} \\[1pt]
\rowcolor{sdours}SD3.5-M & SFT vs.\ base & 71.3 & 1.4 & 27.3 & 70.4 & 0.5 & 29.2 & 70.4 & 0.9 & 28.7 & 72.7 & 0.9 & 26.4 & \textbf{71.2} & 0.9 & 27.9 \\
\rowcolor{fluxours}FLUX.2 [klein] & SFT vs.\ base & 66.4 & 2.3 & 31.3 & 68.3 & 1.5 & 30.1 & 61.8 & 0.8 & 37.5 & 71.4 & 0.4 & 28.2 & \textbf{67.0} & 1.3 & 31.8 \\
\midrule
\multicolumn{17}{@{}l}{\textbf{Qwen3-VL-32B}} \\[1pt]
\rowcolor{sdours}SD3.5-M & SFT vs.\ base & 54.2 & 0.0 & 45.8 & 57.9 & 0.0 & 42.1 & 60.6 & 0.0 & 39.4 & 60.2 & 0.0 & 39.8 & \textbf{58.2} & 0.0 & 41.8 \\
\rowcolor{fluxours}FLUX.2 [klein] & SFT vs.\ base & 53.3 & 0.0 & 46.7 & 52.1 & 0.0 & 47.9 & 49.4 & 0.0 & 50.6 & 54.8 & 0.0 & 45.2 & \textbf{52.4} & 0.0 & 47.6 \\
\bottomrule
\end{tabular}
\end{adjustbox}
\end{table*}

\begin{figure}[!t]
  \centering
  \includegraphics[width=\linewidth]{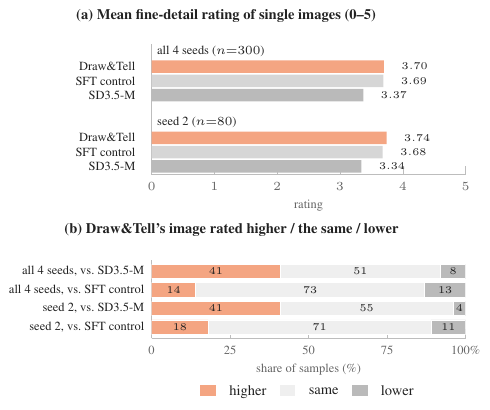}
  \caption{\textbf{SD3.5 fine detail rated one image at a time} (0--5 scale, 300 random samples correct for
  all three models, one seed per prompt, and seed 2 alone). (a)~Mean rating. (b)~Per sample,
  \method's image rated higher, the same or lower than the opponent's.}
  \label{fig:supp_rate}
\end{figure}

\paragraph{Rating each image on its own.} The quality of an image is not a binary decision: an image can have good parts and bad parts,
which a choice between two images hides. We therefore also analyse each image on its own (\cref{fig:supp_judge_rate}) to see
which model generates better images overall, rating every image alone,
on a 0--5 fine-detail scale, without its pair and without the model's name: 300 SD3.5 samples drawn at random among those correct for all three
models (one sampling seed per prompt), all three
models' images, each image rated by its own fresh judge that sees no other image (\cref{fig:supp_rate}). At the best
sampling seed (seed 2, 80 samples), \method rates highest: $3.74$, against $3.67$ for the SFT control and $3.34$ for SD3.5
Medium. Its image rates higher than the base model's on $41\%$ of samples and lower on only $4\%$ ($+0.40$ on average), and
higher than the SFT control's on $18\%$ against lower on $11\%$ ($+0.06$). Averaged over all four seeds the order is the
same ($3.70$, $3.69$ and $3.37$): against the base model \method rates higher on $41\%$ of samples and lower on $8\%$
($+0.33$), and against the SFT control the two images get the same score on $73\%$ of samples, with \method higher on
$14\%$ and lower on $13\%$ ($+0.01$). A six-point scale used one image at a time is coarser than a side-by-side choice, so most of the
difference to the control that the pairwise judge resolves falls within one score step here.

\begin{figure}[!t]
  \centering
  \includegraphics[width=\linewidth]{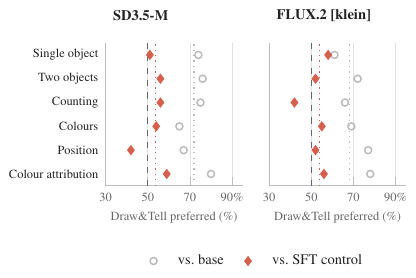}
  \caption{\textbf{Fine detail by GenEval category} (forced choice, the samples of \cref{fig:supp_refine_all}):
  share of pairs in which \method is preferred. Dotted lines are the overall shares, the dashed line is parity.}
  \label{fig:supp_refine_cat}
\end{figure}

\paragraph{By category.} \Cref{fig:supp_refine_cat} splits the forced-choice fine-detail study, averaged over all four seeds, by
GenEval category. Against the base model \method is ahead in every category on both backbones, most clearly on colour
attribution ($80\%$ on SD3.5, $78\%$ on FLUX.2 [klein]). Against the SFT control it is ahead in every category but position
($42\%$, 12 pairs) on SD3.5, clearest on colour attribution ($59\%$), and in every category but counting ($42\%$) on FLUX.2
[klein], clearest on single objects ($58\%$). Some categories are small (12 to 304 pairs).

\begin{figure*}[t]
  \centering
  \setlength{\tabcolsep}{0.5pt}
  \renewcommand{\arraystretch}{0.6}
  \begin{qualband}{SD3.5 Medium}\begin{adjustbox}{max width=\linewidth}%
  \begin{tabular}{@{}ccc@{\hspace{5pt}}ccc@{\hspace{5pt}}ccc@{\hspace{5pt}}ccc@{}}
  \tiny SD3.5-M & \tiny SFT control & \tiny \textcolor{sdcol}{\textbf{\method}} & \tiny SD3.5-M & \tiny SFT control & \tiny \textcolor{sdcol}{\textbf{\method}} & \tiny SD3.5-M & \tiny SFT control & \tiny \textcolor{sdcol}{\textbf{\method}} & \tiny SD3.5-M & \tiny SFT control & \tiny \textcolor{sdcol}{\textbf{\method}} \\
  \rimg[width=0.081\linewidth]{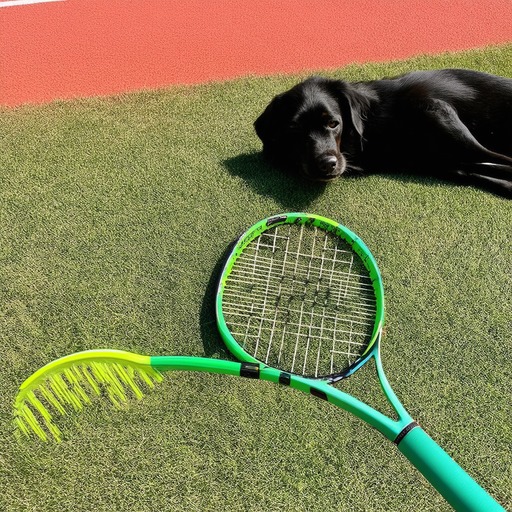} & \rimg[width=0.081\linewidth]{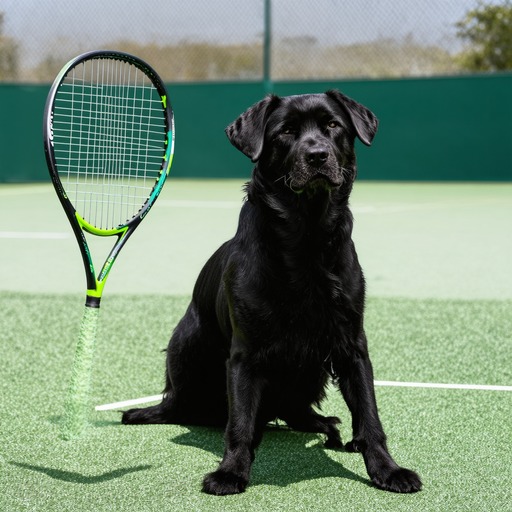} & \rimg[width=0.081\linewidth]{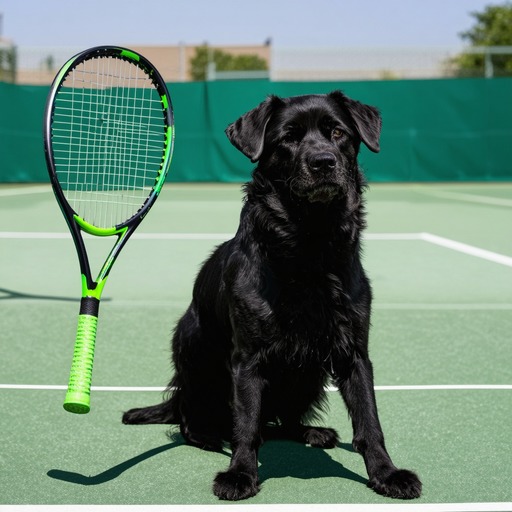} & \rimg[width=0.081\linewidth]{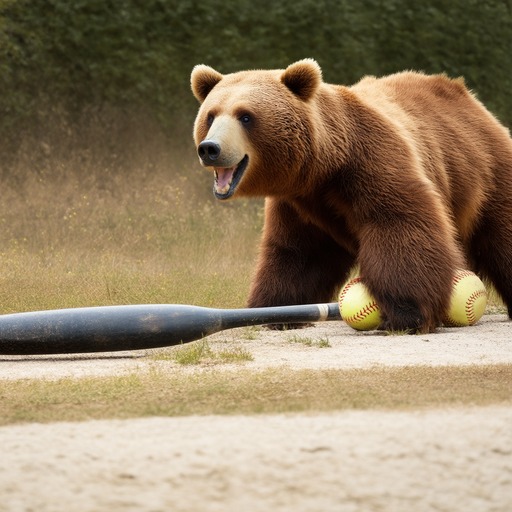} & \rimg[width=0.081\linewidth]{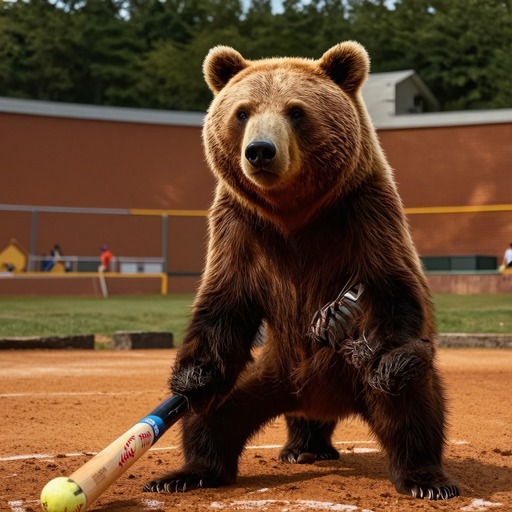} & \rimg[width=0.081\linewidth]{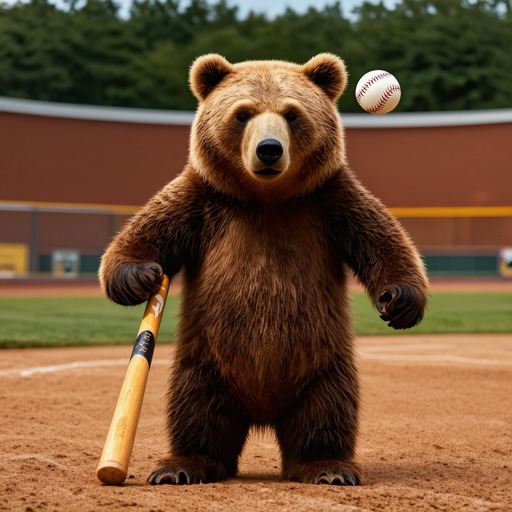} & \rimg[width=0.081\linewidth]{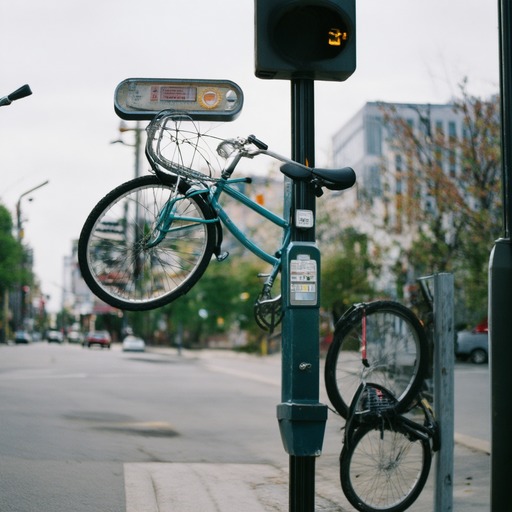} & \rimg[width=0.081\linewidth]{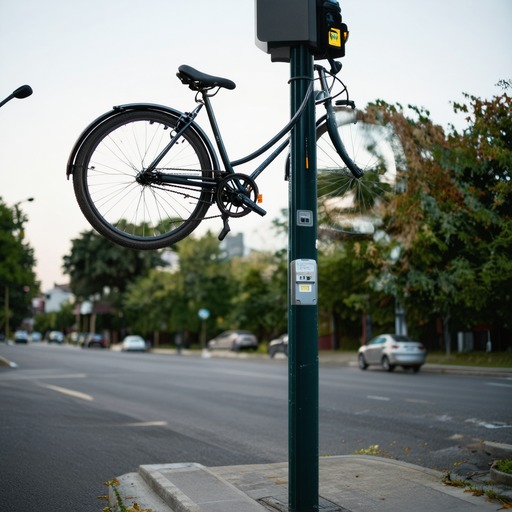} & \rimg[width=0.081\linewidth]{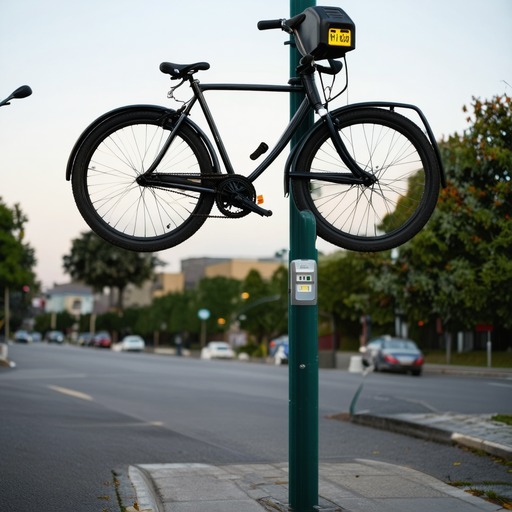} & \rimg[width=0.081\linewidth]{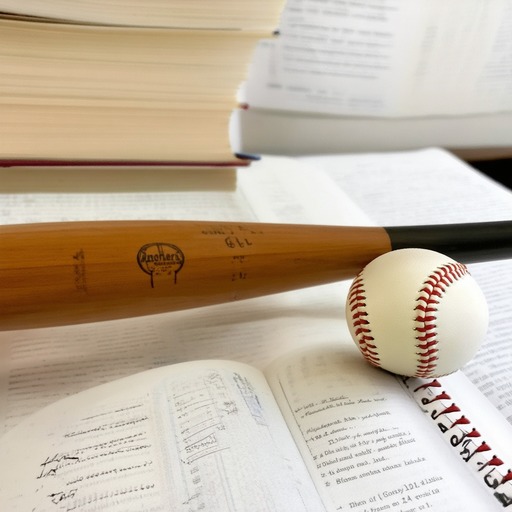} & \rimg[width=0.081\linewidth]{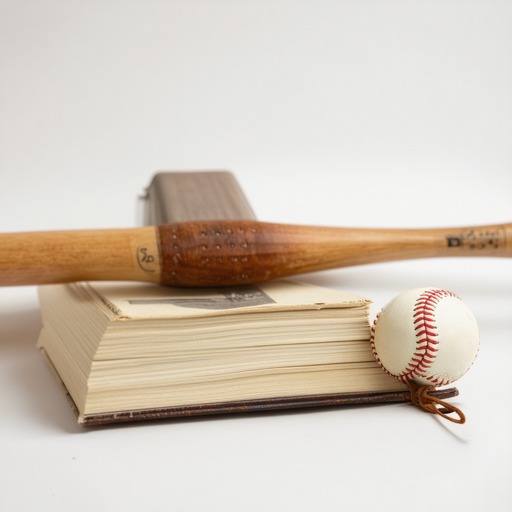} & \rimg[width=0.081\linewidth]{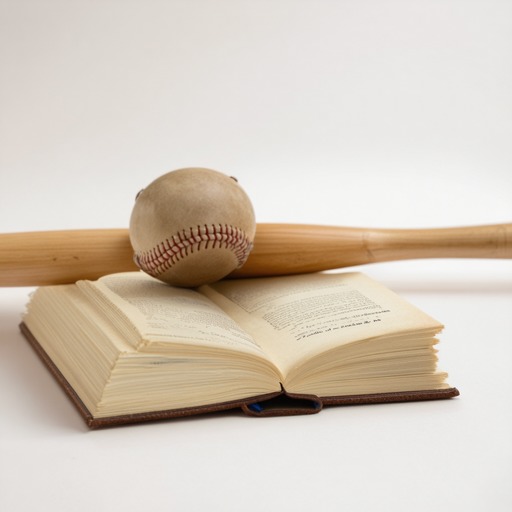} \\
  \multicolumn{3}{c}{\parbox[t]{0.243\linewidth}{\centering\scriptsize\textcolor{qmute}{\textit{``a photo of a green tennis racket and a black dog''}}{\tiny\ (GenEval)}}} & \multicolumn{3}{c}{\parbox[t]{0.243\linewidth}{\centering\scriptsize\textcolor{qmute}{\textit{``a photo of a baseball bat and a bear''}}{\tiny\ (GenEval)}}} & \multicolumn{3}{c}{\parbox[t]{0.243\linewidth}{\centering\scriptsize\textcolor{qmute}{\textit{``a photo of a bicycle above a parking meter''}}{\tiny\ (GenEval)}}} & \multicolumn{3}{c}{\parbox[t]{0.243\linewidth}{\centering\scriptsize\textcolor{qmute}{\textit{``a photo of a book and a baseball bat''}}{\tiny\ (GenEval)}}} \\[3pt]
  \rimg[width=0.081\linewidth]{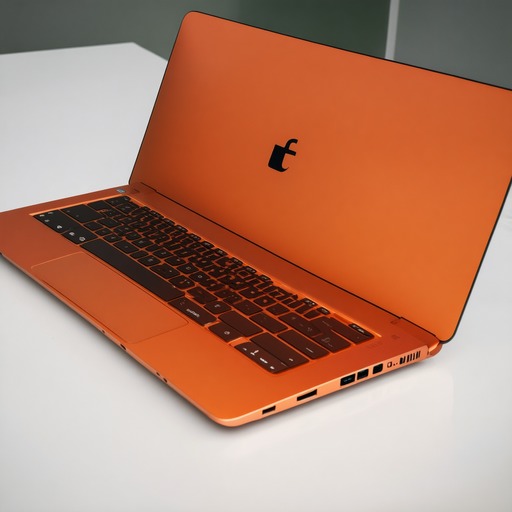} & \rimg[width=0.081\linewidth]{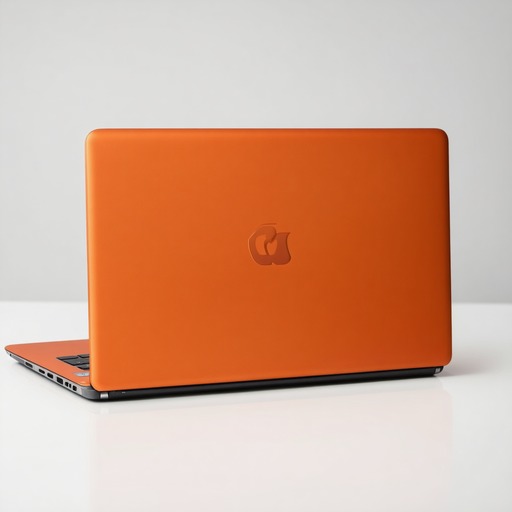} & \rimg[width=0.081\linewidth]{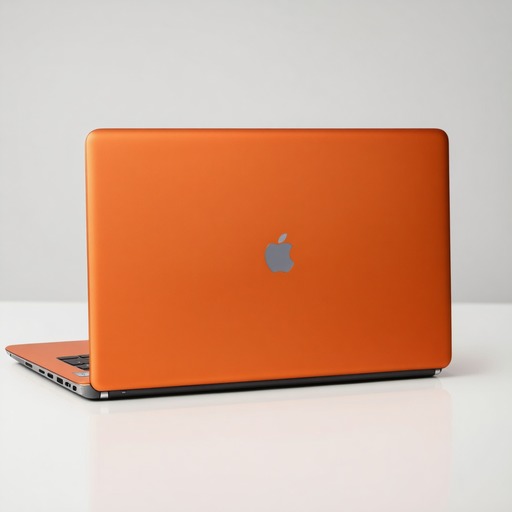} & \rimg[width=0.081\linewidth]{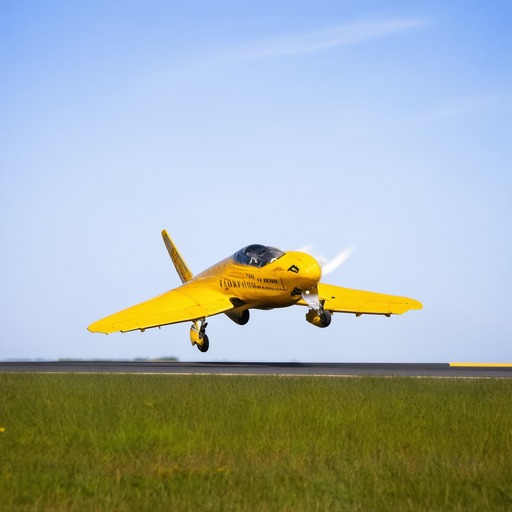} & \rimg[width=0.081\linewidth]{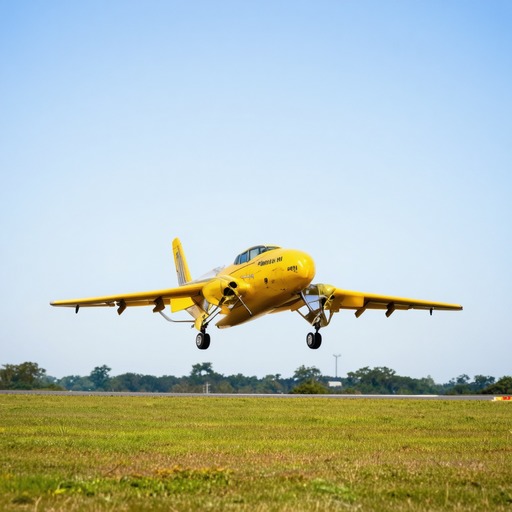} & \rimg[width=0.081\linewidth]{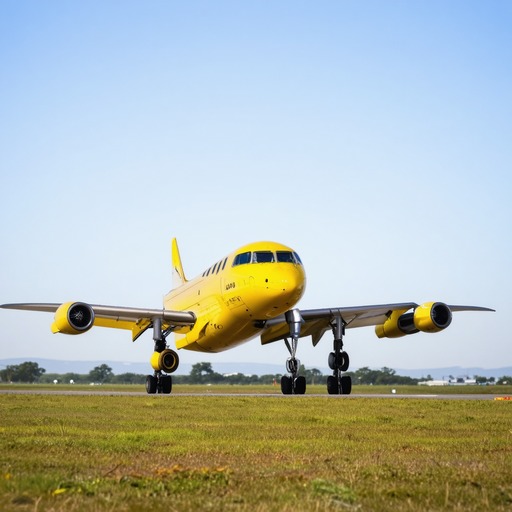} & \rimg[width=0.081\linewidth]{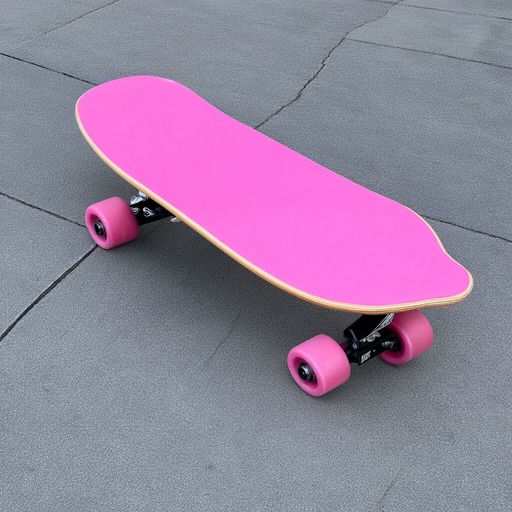} & \rimg[width=0.081\linewidth]{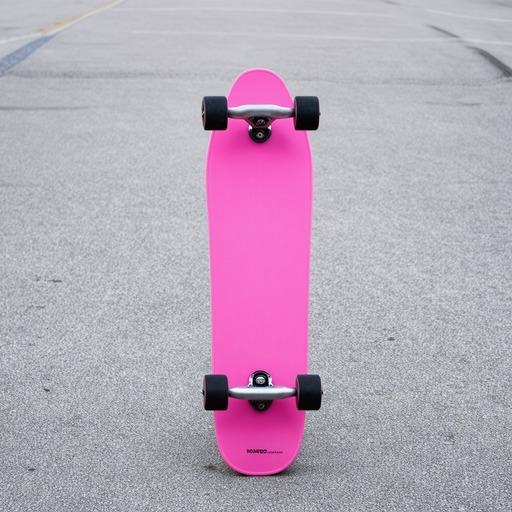} & \rimg[width=0.081\linewidth]{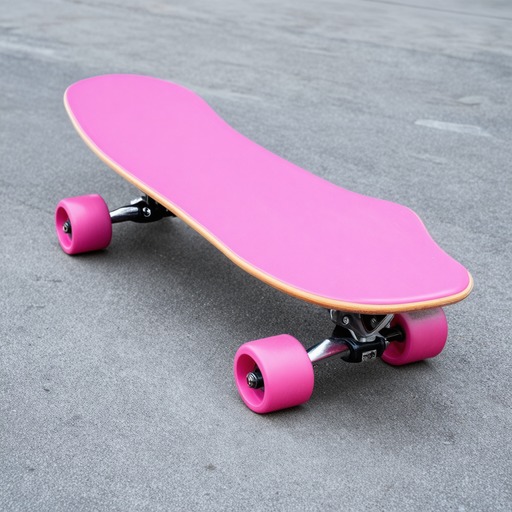} & \rimg[width=0.081\linewidth]{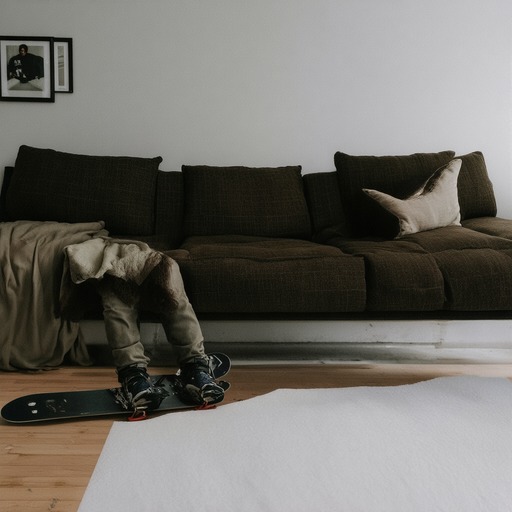} & \rimg[width=0.081\linewidth]{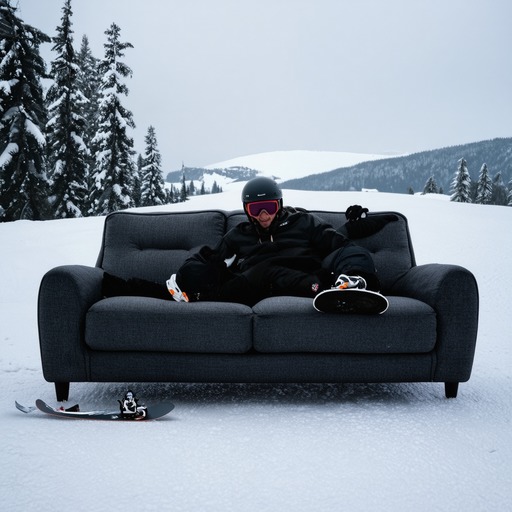} & \rimg[width=0.081\linewidth]{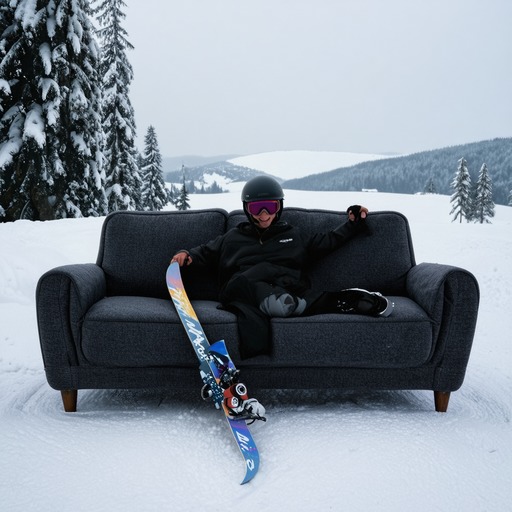} \\
  \multicolumn{3}{c}{\parbox[t]{0.243\linewidth}{\centering\scriptsize\textcolor{qmute}{\textit{``a photo of an orange laptop''}}{\tiny\ (GenEval)}}} & \multicolumn{3}{c}{\parbox[t]{0.243\linewidth}{\centering\scriptsize\textcolor{qmute}{\textit{``a photo of a yellow airplane''}}{\tiny\ (GenEval)}}} & \multicolumn{3}{c}{\parbox[t]{0.243\linewidth}{\centering\scriptsize\textcolor{qmute}{\textit{``a photo of a pink skateboard''}}{\tiny\ (GenEval)}}} & \multicolumn{3}{c}{\parbox[t]{0.243\linewidth}{\centering\scriptsize\textcolor{qmute}{\textit{``a photo of a couch and a snowboard''}}{\tiny\ (GenEval)}}} \\[3pt]
  \end{tabular}\end{adjustbox}\end{qualband}
  \caption{\textbf{Fine detail} (continued; GenEval, same prompt and seed). \method's fine texture and small parts are sharper and more complete.}
  \label{fig:supp_refine}
\end{figure*}

\paragraph{Examples.} \Cref{fig:supp_refine} continues the qualitative figure of the main paper with samples on which the
difference in fine detail is visible to the naked eye.

\begin{figure*}[!tp]
  \centering
  \setlength{\tabcolsep}{0.5pt}
  \renewcommand{\arraystretch}{0.6}
  \begin{qualband}{FLUX.2 [klein]}\begin{adjustbox}{max width=\linewidth}%
  \begin{tabular}{@{}ccc@{\hspace{5pt}}ccc@{\hspace{5pt}}ccc@{\hspace{5pt}}ccc@{}}
  \scriptsize FLUX.2 & \scriptsize SFT control & \scriptsize \textcolor{sdcol}{\textbf{\method}} & \scriptsize FLUX.2 & \scriptsize SFT control & \scriptsize \textcolor{sdcol}{\textbf{\method}} & \scriptsize FLUX.2 & \scriptsize SFT control & \scriptsize \textcolor{sdcol}{\textbf{\method}} & \scriptsize FLUX.2 & \scriptsize SFT control & \scriptsize \textcolor{sdcol}{\textbf{\method}} \\
  \rimg[width=0.077\textwidth]{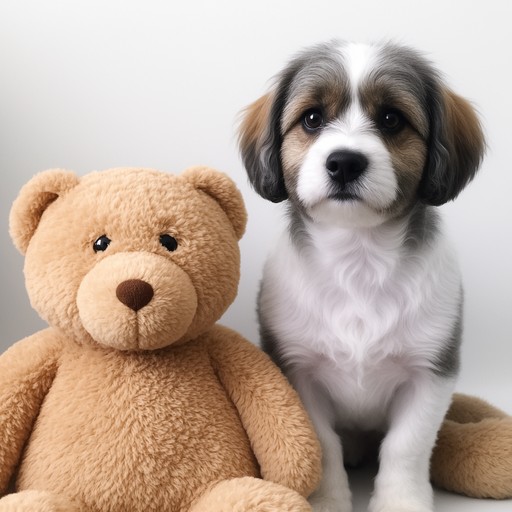} & \rimg[width=0.077\textwidth]{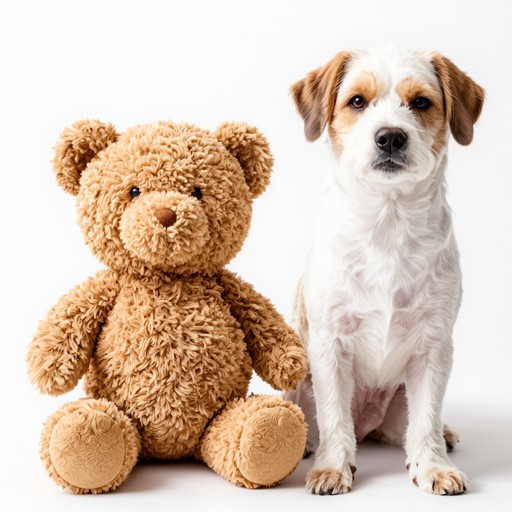} & \rimg[width=0.077\textwidth]{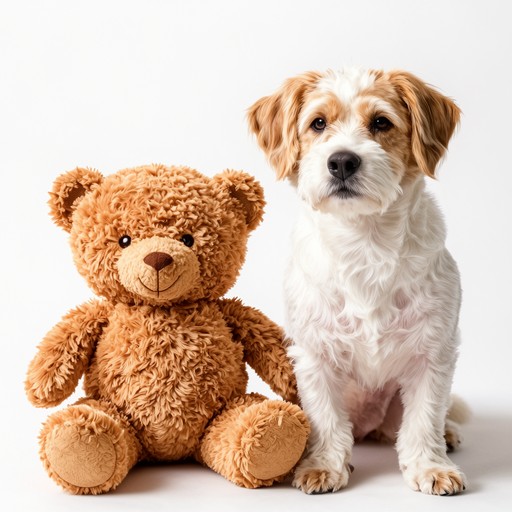} & \rimg[width=0.077\textwidth]{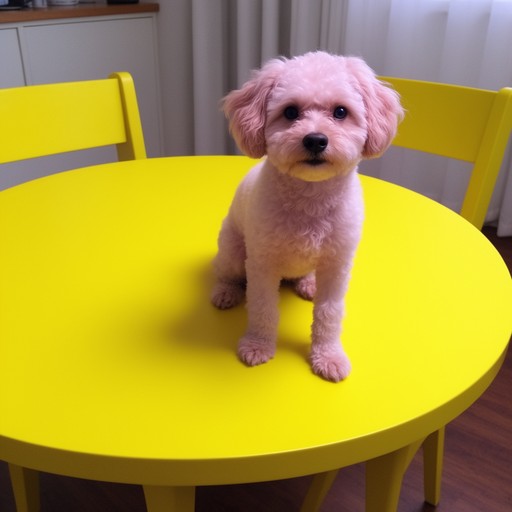} & \rimg[width=0.077\textwidth]{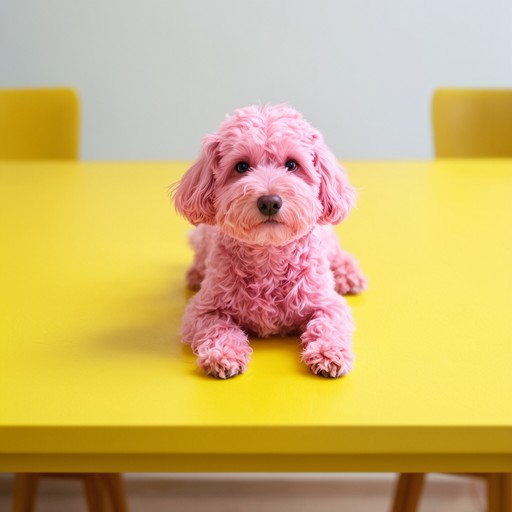} & \rimg[width=0.077\textwidth]{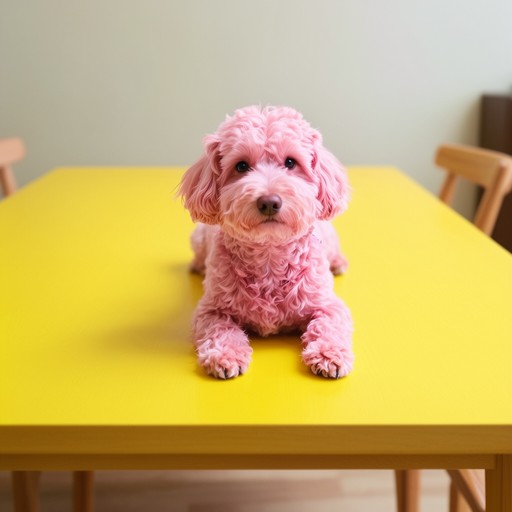} & \rimg[width=0.077\textwidth]{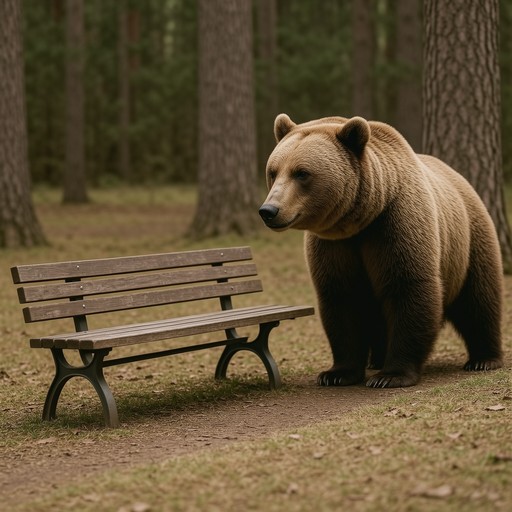} & \rimg[width=0.077\textwidth]{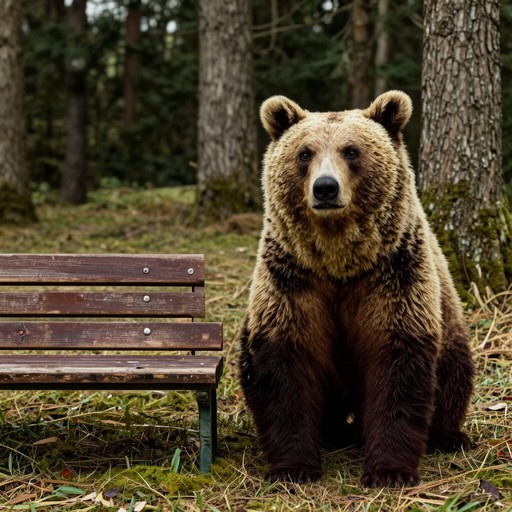} & \rimg[width=0.077\textwidth]{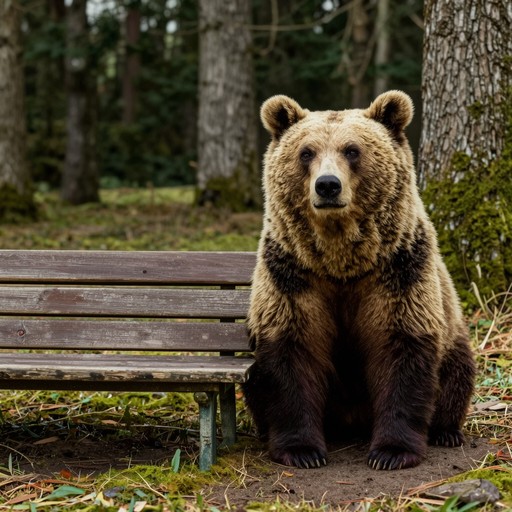} & \rimg[width=0.077\textwidth]{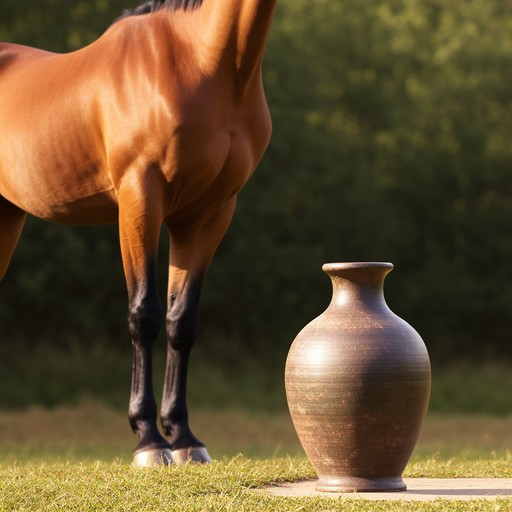} & \rimg[width=0.077\textwidth]{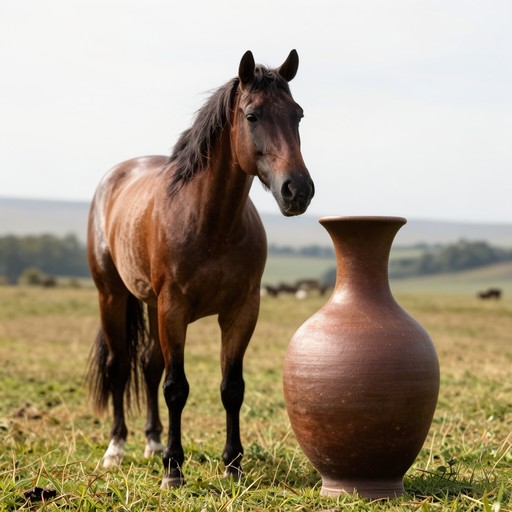} & \rimg[width=0.077\textwidth]{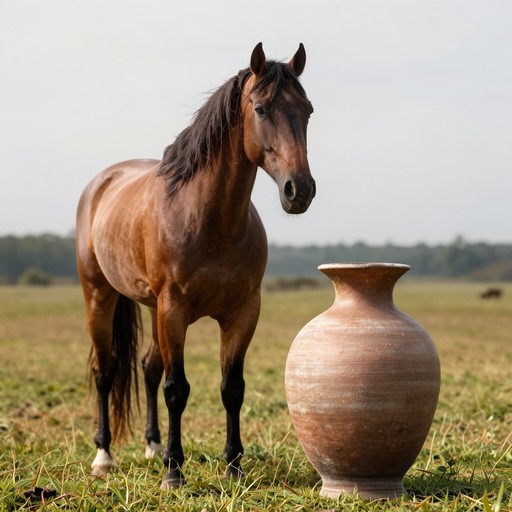} \\
  \multicolumn{3}{c}{\parbox[t]{0.23\textwidth}{\centering\scriptsize\textcolor{qmute}{\textit{``a photo of a dog right of a teddy bear''}}}} & \multicolumn{3}{c}{\parbox[t]{0.23\textwidth}{\centering\scriptsize\textcolor{qmute}{\textit{``a photo of a yellow dining table and a pink dog''}}}} & \multicolumn{3}{c}{\parbox[t]{0.23\textwidth}{\centering\scriptsize\textcolor{qmute}{\textit{``a photo of a bench left of a bear''}}}} & \multicolumn{3}{c}{\parbox[t]{0.23\textwidth}{\centering\scriptsize\textcolor{qmute}{\textit{``a photo of a vase right of a horse''}}}} \\[3pt]
  \rimg[width=0.077\textwidth]{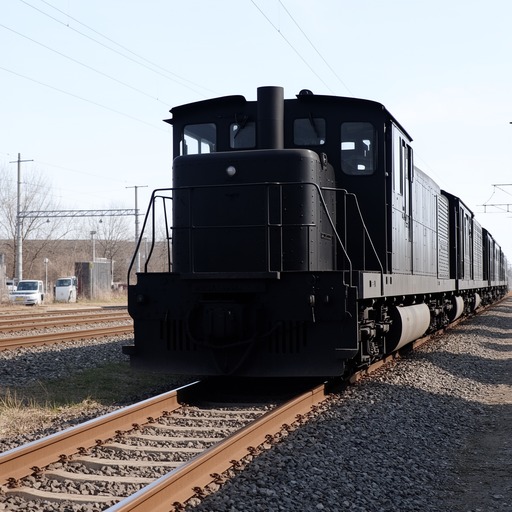} & \rimg[width=0.077\textwidth]{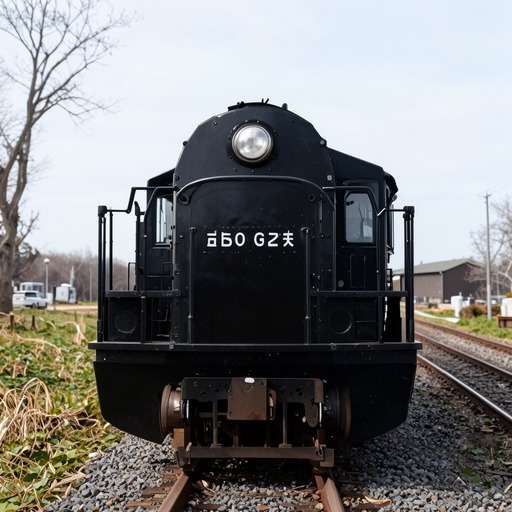} & \rimg[width=0.077\textwidth]{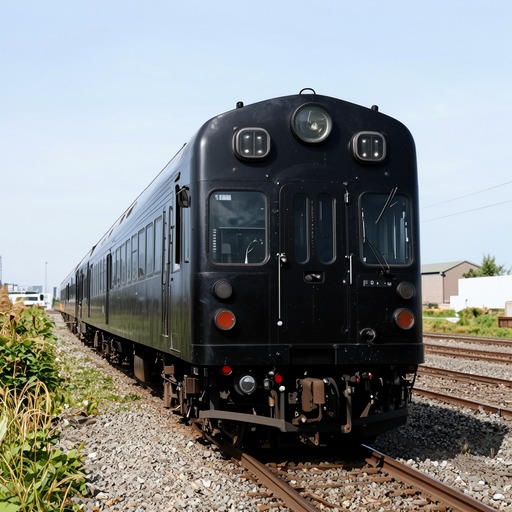} & \rimg[width=0.077\textwidth]{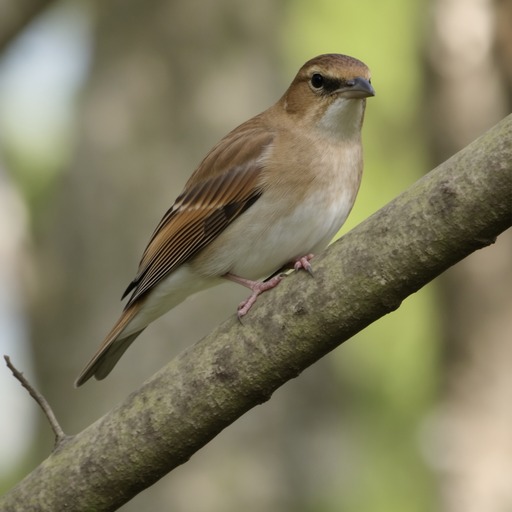} & \rimg[width=0.077\textwidth]{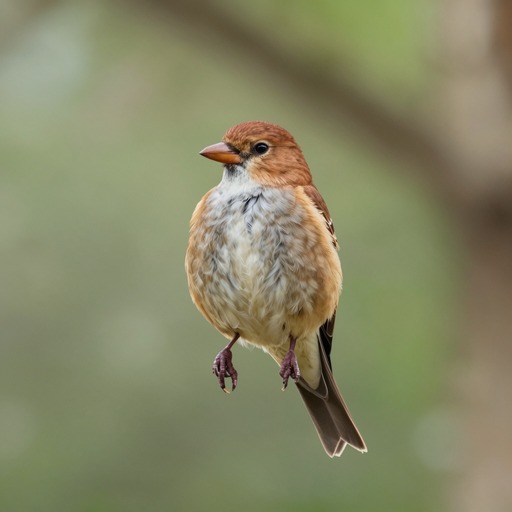} & \rimg[width=0.077\textwidth]{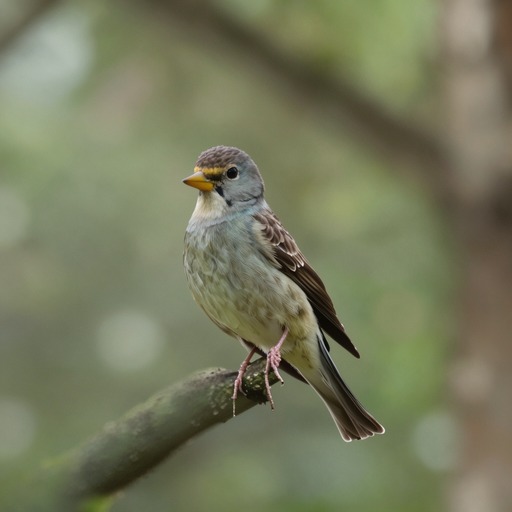} & \rimg[width=0.077\textwidth]{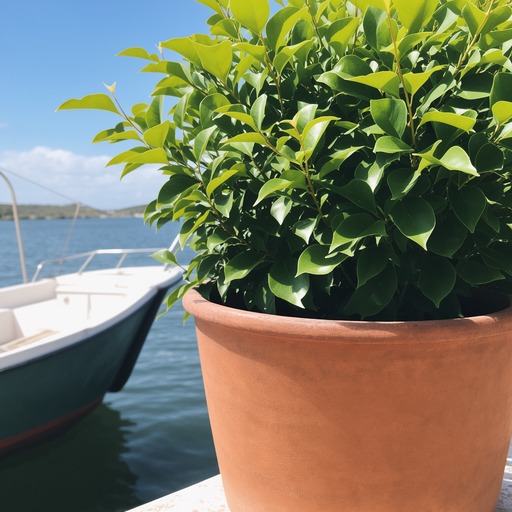} & \rimg[width=0.077\textwidth]{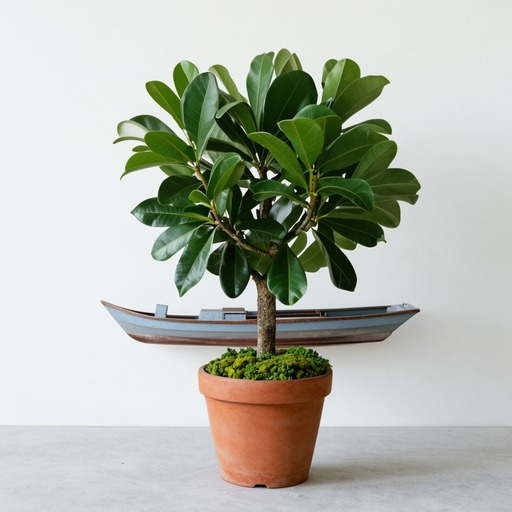} & \rimg[width=0.077\textwidth]{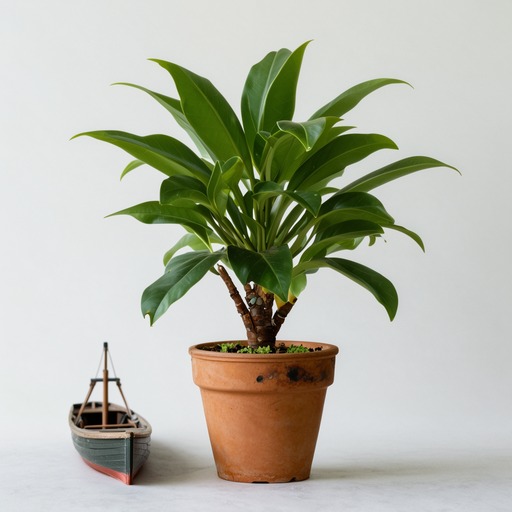} & \rimg[width=0.077\textwidth]{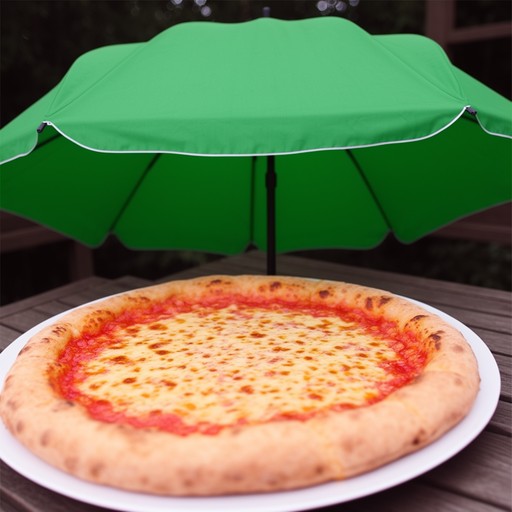} & \rimg[width=0.077\textwidth]{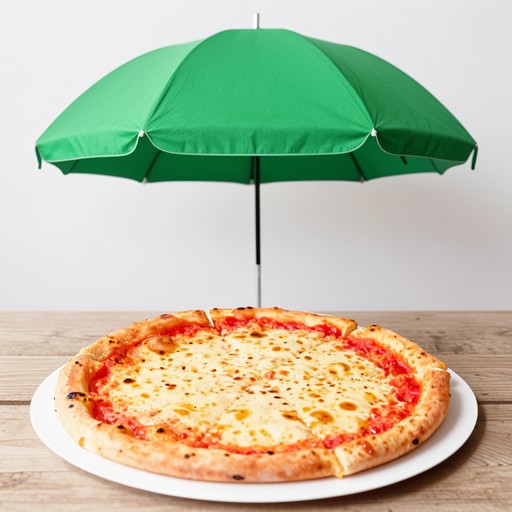} & \rimg[width=0.077\textwidth]{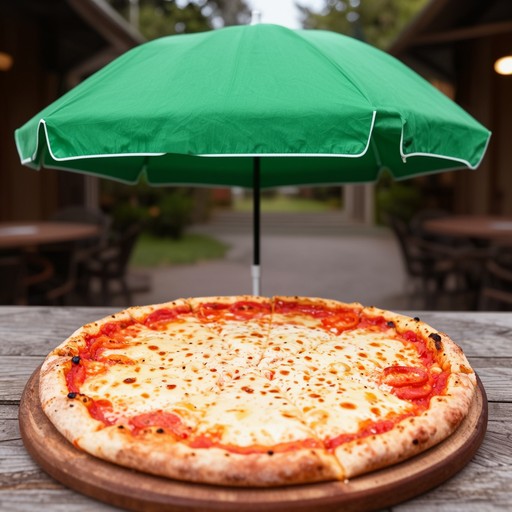} \\
  \multicolumn{3}{c}{\parbox[t]{0.23\textwidth}{\centering\scriptsize\textcolor{qmute}{\textit{``a photo of a black train''}}}} & \multicolumn{3}{c}{\parbox[t]{0.23\textwidth}{\centering\scriptsize\textcolor{qmute}{\textit{``a photo of a bird''}}}} & \multicolumn{3}{c}{\parbox[t]{0.23\textwidth}{\centering\scriptsize\textcolor{qmute}{\textit{``a photo of a potted plant and a boat''}}}} & \multicolumn{3}{c}{\parbox[t]{0.23\textwidth}{\centering\scriptsize\textcolor{qmute}{\textit{``a photo of a white pizza and a green umbrella''}}}} \\[3pt]
  \end{tabular}\end{adjustbox}\end{qualband}
  \caption{\textbf{Fine detail on FLUX.2 [klein]} (GenEval, same prompt and seed).}
  \label{fig:supp_flux_refine}
\end{figure*}

\Cref{fig:supp_flux_refine} shows the same on FLUX.2 [klein].

\subsection{Human study}
\label{sec:supp_human}
\paragraph{Participants.} To check that the blind judge's fine-detail choices reflect what people see, we shared an
online survey openly with the members and alumni of our lab; the set of participants was not fixed in advance, and
anyone who received the link could take part. Twelve people volunteered and gave $233$ answers.
Participants were not told that the study concerned our method, nor which models produced the images, and were not
told its purpose beyond the task itself; model names were never shown, and answers were stored without names.

\paragraph{Image selection.} The images come from the fine-detail study of the main paper (\cref{sec:refine}), so
participants and the judge see the same images. We selected them in three steps.
\begin{enumerate}
\item \emph{Source set.} For each backbone we take the GenEval prompts that the base model, the SFT control and \method
all get right at all four sampling seeds ($216$ on SD3.5, $259$ on FLUX.2 [klein]), so the images differ in how the
content is drawn, not in whether it is there, and use the images of the sampling seed shown in the main paper (seed 3 on
SD3.5, seed 2 on FLUX.2 [klein]). A pair is one prompt: the image of \method and the image of the opponent, base model
or SFT control, for the same prompt and seed.
\item \emph{Pool.} For each of the four comparisons (SD3.5 and FLUX.2 [klein], each against the base model and against
the SFT control) we draw $20$ pairs at random, stratified by the blind judge's choice: as many pairs the judge gave to
\method as pairs it gave to the opponent ($8$ and $8$), and $4$ it found equal ($9$, $9$ and $2$ for SD3.5 against the
base model, where only $2$ equal pairs exist). Prompts are distinct within a backbone. Stratifying by the judge's
choice keeps the pool from favouring \method: with a random draw, most pairs would be ones the judge gave to \method.
The draw never looked at participants' answers; \cref{tab:supp_human_pairs} lists all $80$ pairs.
\item \emph{Per participant.} Each participant sees $20$ pairs, $5$ from each comparison, drawn at random from the
pool without repeats, in random order and with the two images in random left/right positions. $74$ of the $80$ pairs
were drawn for at least one participant.
\end{enumerate}

\begin{figure*}[t]
\centering
\begin{tcbraster}[raster columns=2,raster equal height,raster column skip=5pt,raster left skip=0pt,raster right skip=0pt]
\begin{tcolorbox}[judgebox=sdours]
{\small\bfseries\color{sdcol}Instructions to participants}\\[3pt]
{\footnotesize You will see 20 pairs of images. Both images of a pair were generated from the same text prompt, shown above
them, and both already contain what the prompt asks for.\\[3pt]
\textbf{What to judge.} Compare only how well the fine details are drawn: how sharp, realistic and well resolved the small
textures and parts are, for example fur, hair and feathers, skin, fabric and stitching, wood, metal, glass, plastic and food
surfaces, leaves, edges, buttons and other small parts.\\[3pt]
\textbf{What to ignore.} Which image follows the prompt better (objects, counts, colours, positions); shape or anatomy
errors, such as missing, merged or extra parts; layout, framing and composition; which style you personally like more.\\[3pt]
\textbf{How to answer.} Click an image to see it full size; there you can switch between the two images in place, which
makes small differences easier to spot. Answer \emph{About the same} only if you cannot tell them apart on fine detail.
Your answers are stored without your name. Please answer on your own.}
\end{tcolorbox}
\begin{tcolorbox}[judgebox=sdours]
{\small\bfseries\color{sdcol}One screen of the study}\\[3pt]
\qchip{sdband}{Prompt}\ {\footnotesize a photo of a fork and a book}\\[3pt]
\rimg[width=\linewidth]{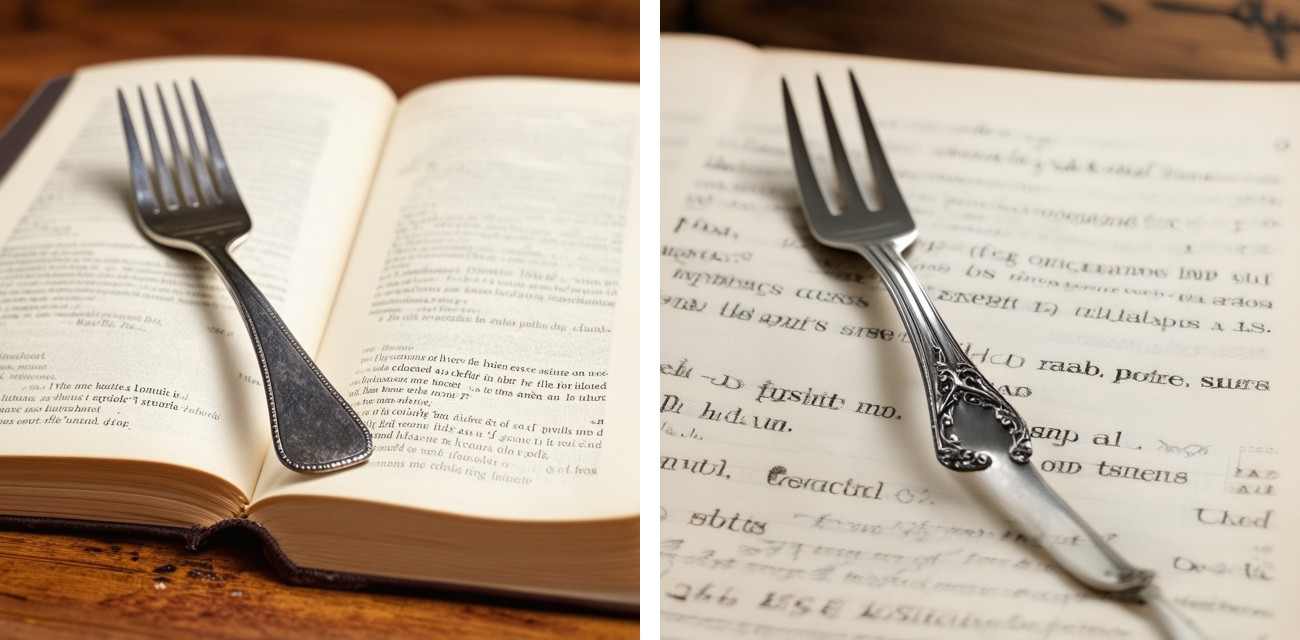}\\[1pt]
{\scriptsize\hspace*{0.2\linewidth}Image 1\hspace{0.42\linewidth}Image 2}\\[3pt]
\qchip{sdband}{Question}\ {\footnotesize Which image has the better fine detail? (only fine detail: not prompt match,
structure, layout or style)}\\[2pt]
\qchip{sdband}{Options}\ {\footnotesize Image 1 \ \textbar\ \ Image 2 \ \textbar\ \ About the same}\\[3pt]
\qchip{sdband}{Answers}\ {\footnotesize Image 1 from all four participants who saw this pair; the blind judge also chose
it.}\\[3pt]
{\scriptsize\color{qmute}Hidden from participants: Image 1 is \method, Image 2 the base model (SD3.5-M). Sides are randomised
for every participant.}
\end{tcolorbox}
\end{tcbraster}
\caption{\textbf{Human study}: the instructions shown to participants and one screen of the study.}
\label{fig:supp_human_ui}
\end{figure*}

\paragraph{Procedure.} For each pair, participants answer which image has the better fine detail, or ``About the
same''. The instructions give the judge's fine-detail criterion in plain words: compare only how sharp, realistic and well
resolved fine textures and small parts are, and ignore prompt following, shape or anatomy errors, layout and style. Each
image can be opened at full size and switched in place with the other image of its pair.
\Cref{fig:supp_human_ui} gives the instructions, word for word, and one screen of the study.

\begin{figure}[H]
  \centering
  \includegraphics[width=\linewidth]{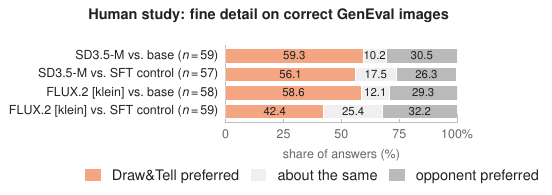}
  \caption{\textbf{Human study}: share of answers in which each image has the better fine detail, per comparison
  ($n$ = number of answers). The pool is balanced by the judge's choice.}
  \label{fig:supp_human}
\end{figure}

\paragraph{Results.} Participants prefer \method in every
comparison (\cref{fig:supp_human}). Among the answers that pick a side, \method is preferred in $66.0\%$ (SD3.5-M against
the base model), $68.1\%$ (against the SFT control), $66.7\%$ (FLUX.2 [klein] against the base model) and $56.8\%$
(against the control), $64.6\%$ overall ($p{=}5{\cdot}10^{-5}$, sign test). Because the pool is balanced by the judge's
choice, these shares likely understate the preference on the full sets: even on the pairs the judge gave to the
opponent, participants chose \method in $37.8\%$ of answers and the opponent in $42.9\%$. A single answer matches the
judge's choice in $47.6\%$ of cases, and two participants answering the same pair agree in $48.9\%$, so the judge agrees
with people about as often as people agree with each other. In more than half of the disagreements ($56\%$), either the participant or
the judge answered ``about the same'' while the other picked an image; when both pick an image, participant and judge agree
in $66.0\%$ of cases.

\paragraph{Per-image answers.} \Cref{tab:supp_human_cross} crosses every answer with the judge's choice on its pair, and
\cref{tab:supp_human_pairs} lists every pair of the pool with its answers. When the judge gave the pair to \method,
participants chose \method in $63$ of $93$ answers and the opponent in $17$; when the judge gave it to the opponent,
the opponent in $42$ of $98$ and \method in $37$; pairs the judge found equal went to \method in $26$ of $42$ answers.

\begin{table}[H]
\caption{\textbf{Human study: participants against the judge}. Every answer ($233$), by the image the blind judge chose for
its pair (rows) and the participant's answer (columns).}
\label{tab:supp_human_cross}
\centering
\footnotesize
\begin{tabular}{lcccc}
\toprule
 & \multicolumn{3}{c}{Participant chose} & \\
Judge chose & \method & Same & Opponent & Answers \\
\midrule
\method & $\mathbf{63}$ & $13$ & $17$ & $93$ \\
Equal & $26$ & $6$ & $10$ & $42$ \\
Opponent & $37$ & $19$ & $\mathbf{42}$ & $98$ \\
\midrule
All & $126$ & $38$ & $69$ & $233$ \\
\bottomrule
\end{tabular}
\end{table}

\begin{table*}[t]
\caption{\textbf{Human study, every pair}: the $80$ pairs of the pool, per comparison, ordered by the blind judge's choice, with the number of participants who chose \method, the opponent (base model or SFT control) or \emph{About the same}. Pairs with no answers were not drawn for any participant.}
\label{tab:supp_human_pairs}
\centering
\scriptsize
\setlength{\tabcolsep}{3pt}
\begin{adjustbox}{max width=\linewidth}
\begin{tabular}{@{}llcccc@{}}
\toprule
Comparison & Prompt (\emph{a photo of} \dots) & Judge & \method & Opp. & Same \\
\midrule
SD3.5 vs base & a fork and a book & \method & 4 & 0 & 0 \\
 & a stop sign and a bottle & \method & 3 & 0 & 0 \\
 & two toothbrushs & \method & 0 & 1 & 0 \\
 & a motorcycle & \method & 2 & 1 & 1 \\
 & two toilets & \method & 2 & 1 & 1 \\
 & a green motorcycle & \method & 1 & 1 & 0 \\
 & a green bus & \method & 3 & 0 & 1 \\
 & a broccoli & \method & 2 & 0 & 0 \\
 & an umbrella & \method & 2 & 1 & 0 \\
 & a tv and a carrot & equal & 1 & 0 & 0 \\
 & a bowl & equal & 3 & 0 & 1 \\
 & a giraffe and a computer mouse & opponent & 0 & 1 & 0 \\
 & three zebras & opponent & 3 & 2 & 1 \\
 & a purple potted plant & opponent & 3 & 0 & 0 \\
 & an orange computer mouse & opponent & 1 & 2 & 0 \\
 & a frisbee & opponent & 0 & 2 & 1 \\
 & a purple elephant and a brown sports ball & opponent & 0 & 1 & 0 \\
 & a blue clock and a white cup & opponent & 0 & 3 & 0 \\
 & a white wine glass and a brown giraffe & opponent & 3 & 0 & 0 \\
 & an airplane & opponent & 2 & 2 & 0 \\
\midrule
SD3.5 vs SFT & a skateboard & \method & 3 & 0 & 1 \\
 & a suitcase and a dining table & \method & 3 & 0 & 0 \\
 & a car & \method & 0 & 0 & 2 \\
 & a yellow elephant & \method & 3 & 1 & 1 \\
 & a pink skateboard & \method & 0 & 1 & 0 \\
 & a black bicycle & \method & 3 & 0 & 0 \\
 & an orange scissors & \method & 2 & 1 & 0 \\
 & a chair & \method & 2 & 1 & 2 \\
 & a potted plant and a backpack & equal & 2 & 0 & 0 \\
 & a chair and a laptop & equal & 2 & 2 & 0 \\
 & a black donut & equal & 2 & 0 & 0 \\
 & a scissors & equal & 0 & 2 & 0 \\
 & a zebra & opponent & 0 & 0 & 0 \\
 & a yellow tv remote & opponent & 0 & 0 & 0 \\
 & a hair drier & opponent & 1 & 2 & 1 \\
 & a fork & opponent & 2 & 0 & 0 \\
 & a tennis racket & opponent & 1 & 3 & 0 \\
 & a bench and a sports ball & opponent & 2 & 0 & 0 \\
 & a couch and a wine glass & opponent & 2 & 2 & 0 \\
 & a book and a laptop & opponent & 2 & 0 & 3 \\
\bottomrule
\end{tabular}
\hspace{6pt}
\begin{tabular}{@{}llcccc@{}}
\toprule
Comparison & Prompt (\emph{a photo of} \dots) & Judge & \method & Opp. & Same \\
\midrule
FLUX.2 vs base & a pizza and a bench & \method & 3 & 1 & 0 \\
 & a person and a bear & \method & 2 & 1 & 1 \\
 & a fire hydrant and a tennis racket & \method & 2 & 0 & 0 \\
 & a toothbrush and a bench & \method & 1 & 0 & 0 \\
 & a blue umbrella & \method & 2 & 0 & 0 \\
 & a toilet & \method & 2 & 0 & 0 \\
 & a truck & \method & 2 & 1 & 0 \\
 & an umbrella & \method & 1 & 0 & 0 \\
 & a bowl & equal & 1 & 1 & 1 \\
 & a dog right of a teddy bear & equal & 3 & 0 & 1 \\
 & a sink & equal & 2 & 1 & 1 \\
 & a hair drier and a cake & equal & 3 & 0 & 1 \\
 & a broccoli and a parking meter & opponent & 3 & 1 & 0 \\
 & a blue cow & opponent & 3 & 1 & 0 \\
 & a purple hair drier & opponent & 0 & 4 & 0 \\
 & a blue toilet & opponent & 1 & 1 & 1 \\
 & a green clock & opponent & 1 & 2 & 0 \\
 & a pink parking meter & opponent & 1 & 1 & 1 \\
 & a bottle & opponent & 0 & 0 & 0 \\
 & a horse & opponent & 1 & 2 & 0 \\
\midrule
FLUX.2 vs SFT & a fork and a baseball glove & \method & 2 & 0 & 0 \\
 & three refrigerators & \method & 3 & 1 & 0 \\
 & a white scissors & \method & 2 & 0 & 1 \\
 & a green vase & \method & 0 & 0 & 2 \\
 & a white dog & \method & 2 & 2 & 0 \\
 & a skis & \method & 0 & 2 & 0 \\
 & a frisbee & \method & 4 & 0 & 0 \\
 & a couch and a horse & \method & 0 & 0 & 0 \\
 & a parking meter and a teddy bear & equal & 0 & 2 & 0 \\
 & an orange orange & equal & 0 & 0 & 0 \\
 & a brown toaster & equal & 5 & 2 & 0 \\
 & a red umbrella and a blue couch & equal & 2 & 0 & 1 \\
 & a zebra and a bed & opponent & 3 & 0 & 2 \\
 & a surfboard and a suitcase & opponent & 0 & 0 & 4 \\
 & four donuts & opponent & 1 & 1 & 3 \\
 & three birds & opponent & 0 & 0 & 0 \\
 & a red apple & opponent & 0 & 2 & 1 \\
 & a pink cell phone & opponent & 1 & 0 & 0 \\
 & an orange tennis racket and a yellow sports ball & opponent & 0 & 2 & 1 \\
 & a frisbee and a couch & opponent & 0 & 5 & 0 \\
\bottomrule
\end{tabular}
\end{adjustbox}
\end{table*}

\subsection{Ambiguous prompts}
\label{sec:supp_ambig}
\paragraph{Prompts.} We use the 100 shortest PartiPrompts: 52 prompts of one word and 48 of two, listed below. A short prompt leaves
the generator to decide what to show: an abstract word or a number must be given a meaning, and a single object must be given a
setting, a style and a composition. Every model generates each prompt at three seeds (0, 1, 2), with the same seeds for all models.

{\footnotesize\sloppy\noindent\textit{One word (52):} bond; element; molecule; life; protein; yin-yang; wood; metal; space; air; fire; water; earth; force; motion; inertia; energy; gravity; peace; fairness; gender; intelligence; bias; hate; anger; emotion; feeling; love; 42; 0; infinity; 300; 101; commonsense; happiness; hope; insight; inspiration; derision; robots; food; teacup; chair; waterfall; cash; sneakers; beer; supercalifragilisticexpialidocious; Incomprehensibilities; Pneumonoultramicroscopicsilicovolcanoconiosis; bottles; children.\par\smallskip\noindent\textit{Two words (48):} an ostrich; a key; U.S. 101; a screwdriver; a chemtrail; orange juice; a tree; a TV; a doorknocker; a ladder; a circle; a flower; an owl; a banana; a girl; a panda; a squirrel; a motorcycle; a cow; a pineapple; a city; a hedgehog; a bottle; a book; a windmill; a plate; a bookshelf; a toaster; a street; a rabbit; an avocado; a statue; a box; toy cars; two cats; an orange; concentric circles; two chairs; a lizard; Times Square; artificial intelligence; Salvador Dal\'{\i}; a Stegasaurus; orange jello; lily pads; five chairs; a rowboat; a road.\par}

\begin{figure*}[t]
  \centering
  \setlength{\tabcolsep}{0.5pt}
  \renewcommand{\arraystretch}{0.6}
  \begin{qualband}{SD3.5 Medium}\begin{adjustbox}{max width=\linewidth}%
  \begin{tabular}{@{}ccc@{\hspace{5pt}}ccc@{\hspace{5pt}}ccc@{}}
  \scriptsize SD3.5-M & \scriptsize SFT control & \scriptsize \textcolor{sdcol}{\textbf{\method}} & \scriptsize SD3.5-M & \scriptsize SFT control & \scriptsize \textcolor{sdcol}{\textbf{\method}} & \scriptsize SD3.5-M & \scriptsize SFT control & \scriptsize \textcolor{sdcol}{\textbf{\method}} \\
  \rimg[width=0.104\textwidth]{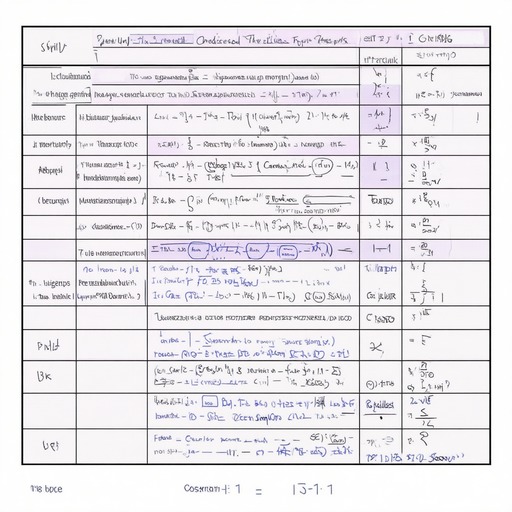} & \rimg[width=0.104\textwidth]{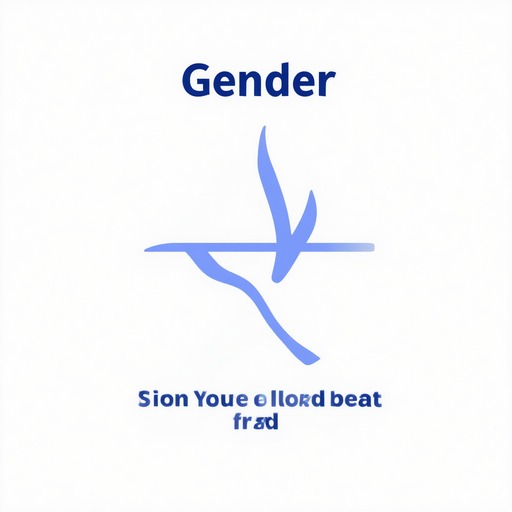} & \rimg[width=0.104\textwidth]{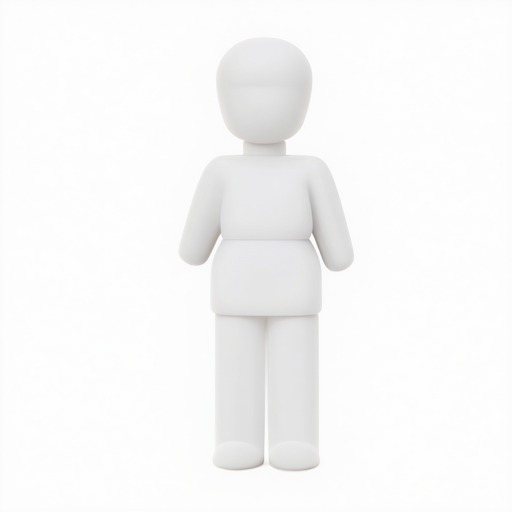} & \rimg[width=0.104\textwidth]{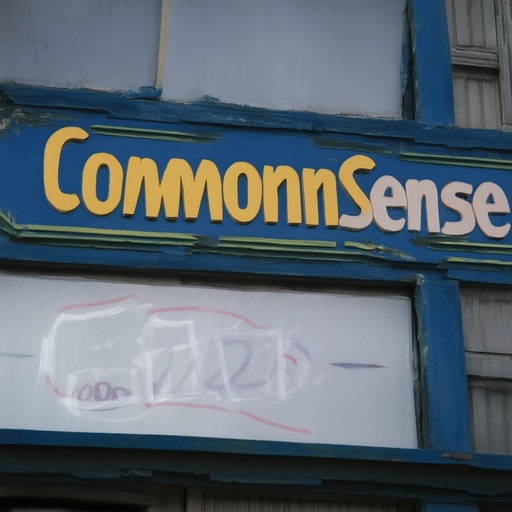} & \rimg[width=0.104\textwidth]{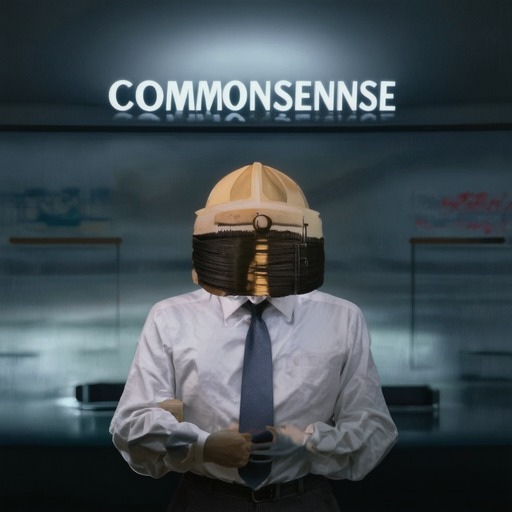} & \rimg[width=0.104\textwidth]{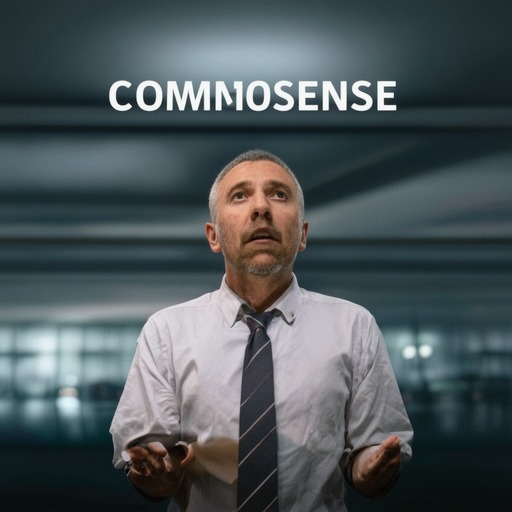} & \rimg[width=0.104\textwidth]{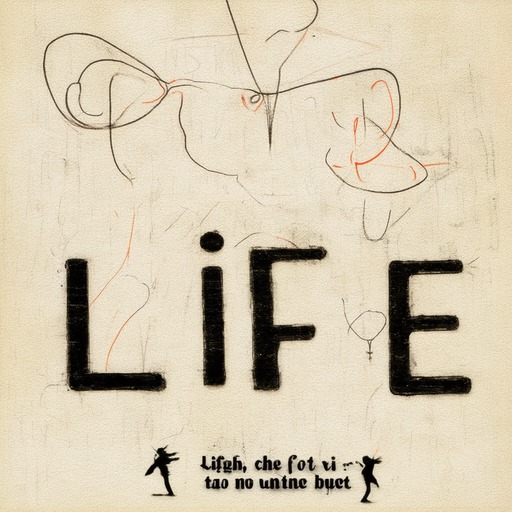} & \rimg[width=0.104\textwidth]{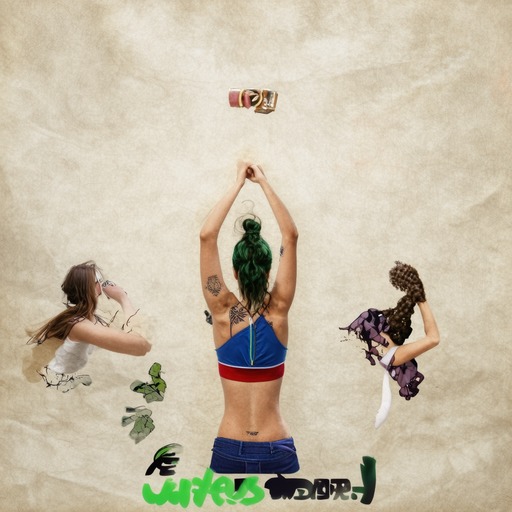} & \rimg[width=0.104\textwidth]{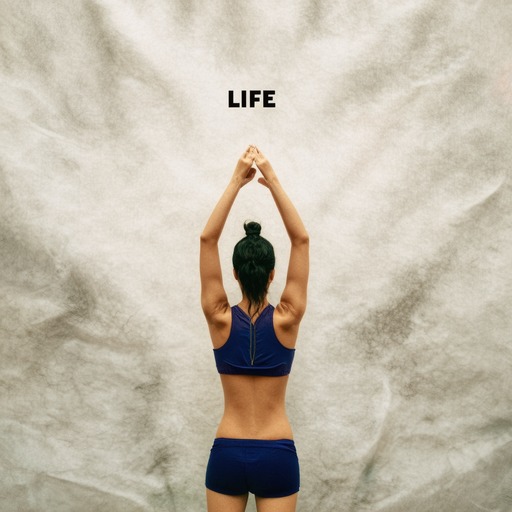} \\
  \multicolumn{3}{c}{\scriptsize\textcolor{qmute}{\textit{``gender''}}} & \multicolumn{3}{c}{\scriptsize\textcolor{qmute}{\textit{``commonsense''}}} & \multicolumn{3}{c}{\scriptsize\textcolor{qmute}{\textit{``life''}}} \\[3pt]
  \rimg[width=0.104\textwidth]{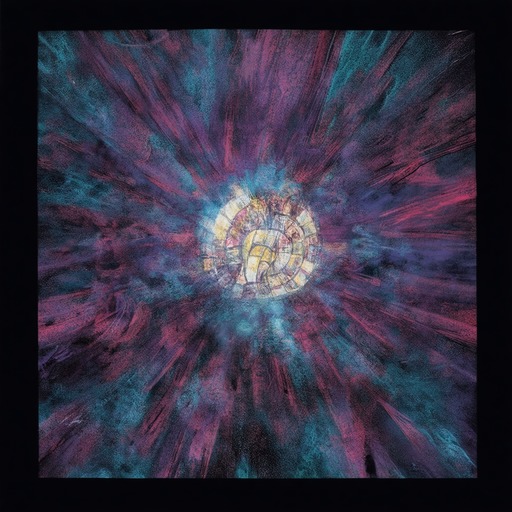} & \rimg[width=0.104\textwidth]{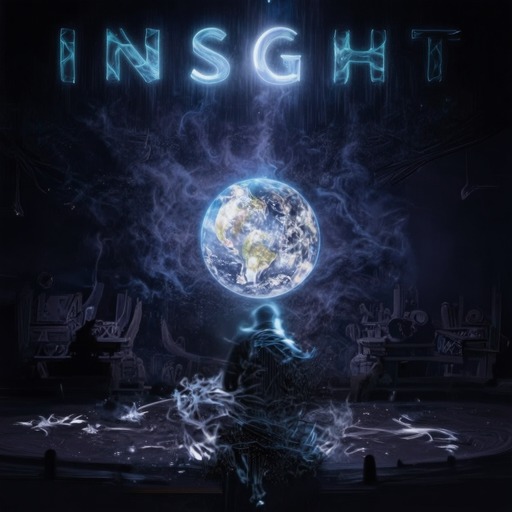} & \rimg[width=0.104\textwidth]{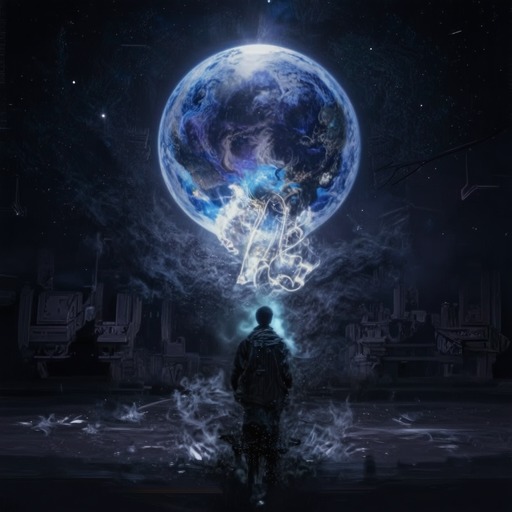} & \rimg[width=0.104\textwidth]{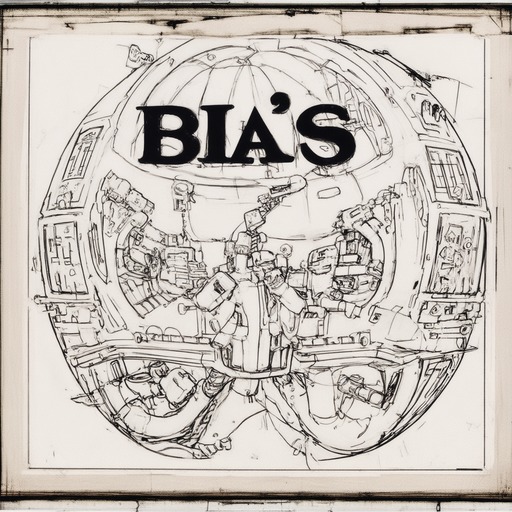} & \rimg[width=0.104\textwidth]{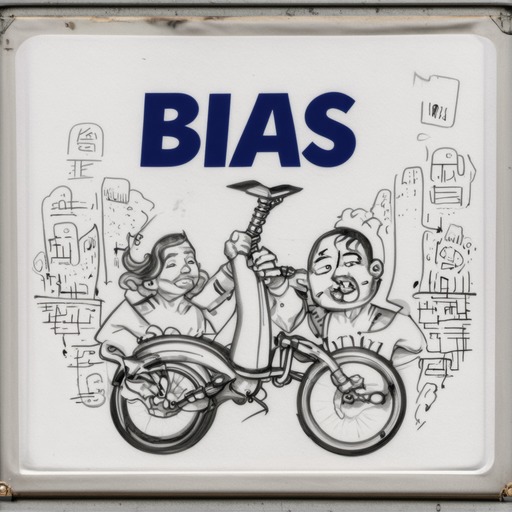} & \rimg[width=0.104\textwidth]{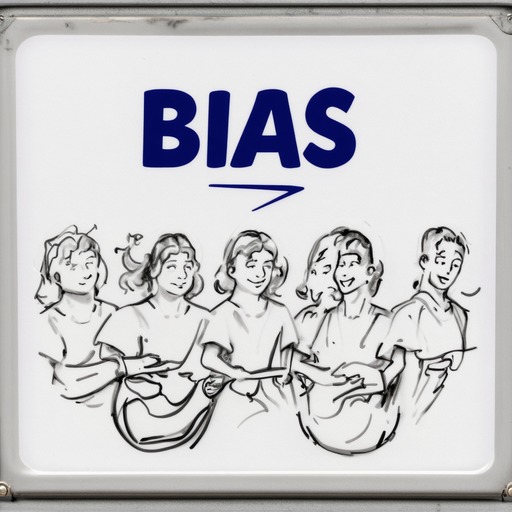} & \rimg[width=0.104\textwidth]{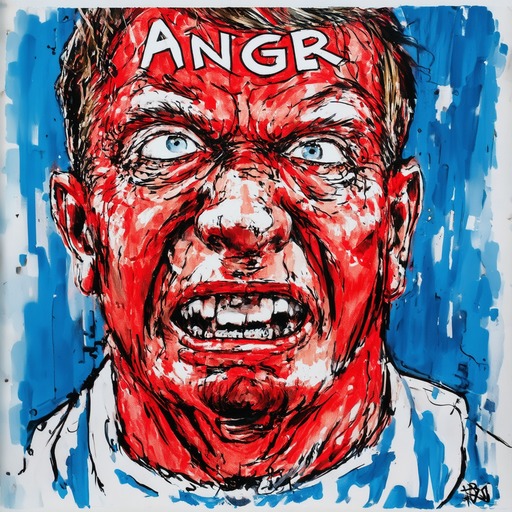} & \rimg[width=0.104\textwidth]{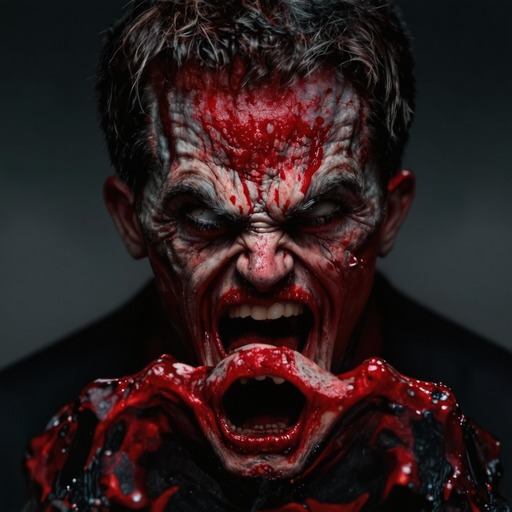} & \rimg[width=0.104\textwidth]{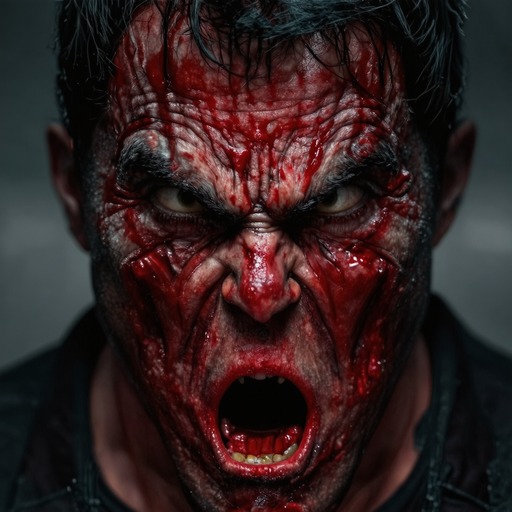} \\
  \multicolumn{3}{c}{\scriptsize\textcolor{qmute}{\textit{``insight''}}} & \multicolumn{3}{c}{\scriptsize\textcolor{qmute}{\textit{``bias''}}} & \multicolumn{3}{c}{\scriptsize\textcolor{qmute}{\textit{``anger''}}} \\[3pt]
  \end{tabular}\end{adjustbox}\end{qualband}
  \caption{\textbf{Ambiguous prompts} (same seed).}
  \label{fig:supp_ambig}
\end{figure*}

\begin{figure*}[t]
  \centering
  \setlength{\tabcolsep}{0.5pt}
  \renewcommand{\arraystretch}{0.6}
  \begin{qualband}{FLUX.2 [klein]}\begin{adjustbox}{max width=\linewidth}%
  \begin{tabular}{@{}ccc@{\hspace{5pt}}ccc@{\hspace{5pt}}ccc@{}}
  \scriptsize FLUX.2 & \scriptsize SFT control & \scriptsize \textcolor{sdcol}{\textbf{\method}} & \scriptsize FLUX.2 & \scriptsize SFT control & \scriptsize \textcolor{sdcol}{\textbf{\method}} & \scriptsize FLUX.2 & \scriptsize SFT control & \scriptsize \textcolor{sdcol}{\textbf{\method}} \\
  \rimg[width=0.104\textwidth]{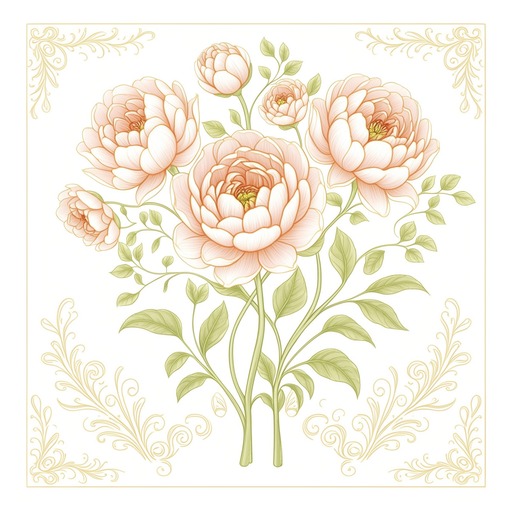} & \rimg[width=0.104\textwidth]{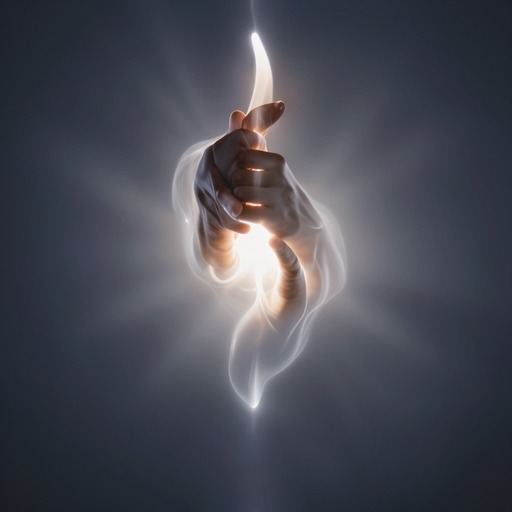} & \rimg[width=0.104\textwidth]{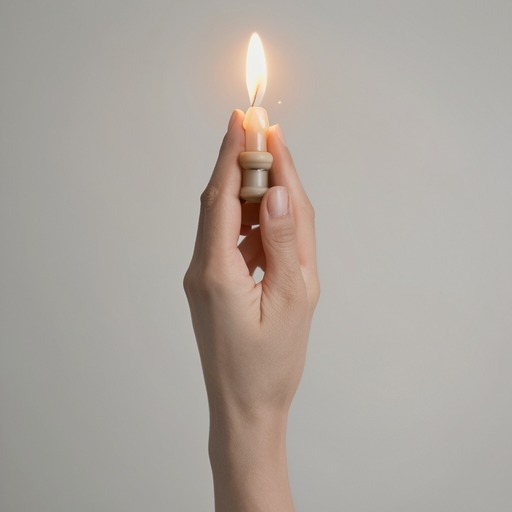} & \rimg[width=0.104\textwidth]{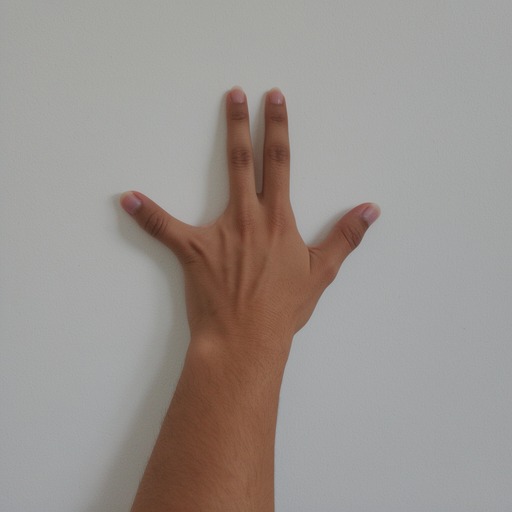} & \rimg[width=0.104\textwidth]{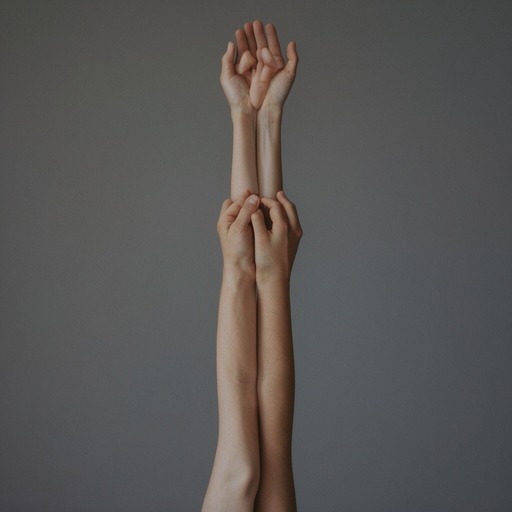} & \rimg[width=0.104\textwidth]{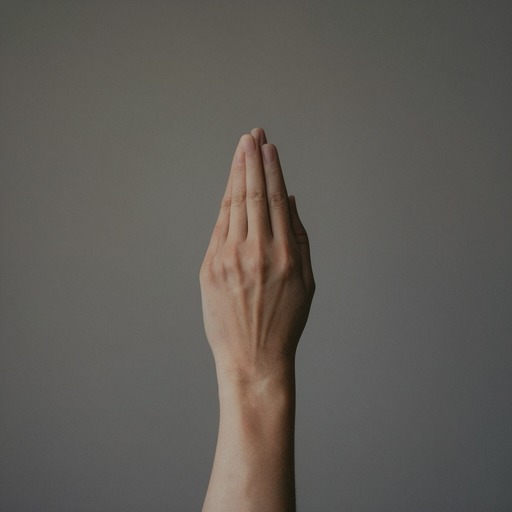} & \rimg[width=0.104\textwidth]{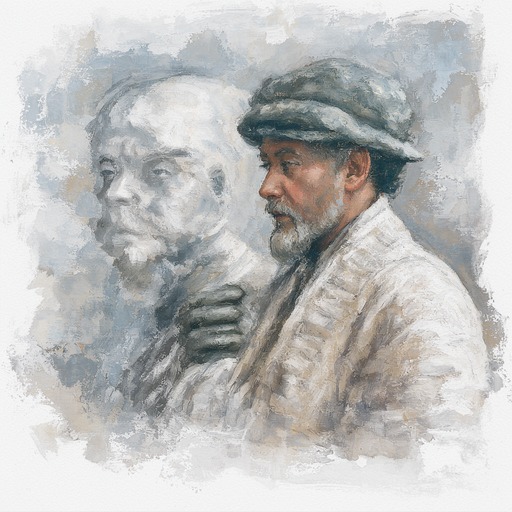} & \rimg[width=0.104\textwidth]{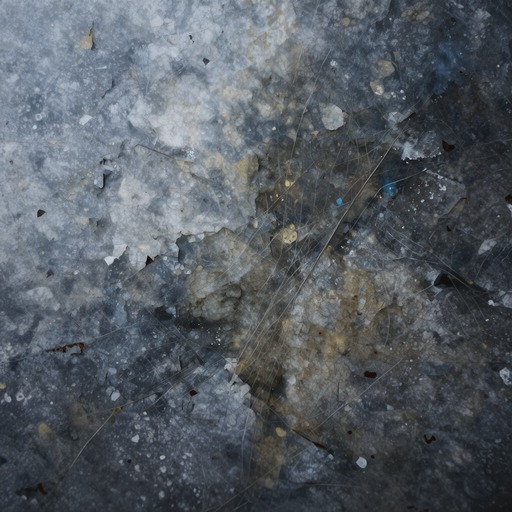} & \rimg[width=0.104\textwidth]{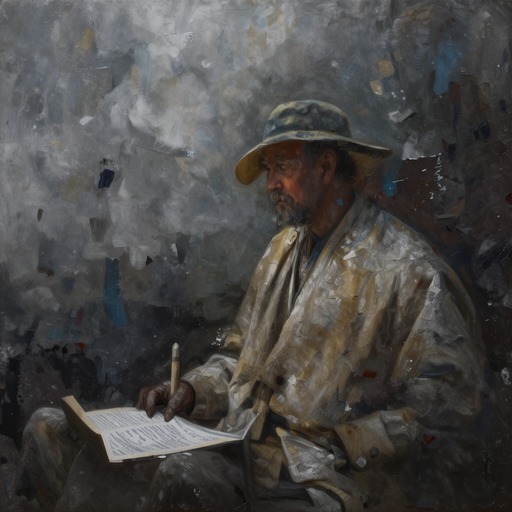} \\
  \multicolumn{3}{c}{\scriptsize\textcolor{qmute}{\textit{``inspiration''}}} & \multicolumn{3}{c}{\scriptsize\textcolor{qmute}{\textit{``force''}}} & \multicolumn{3}{c}{\scriptsize\textcolor{qmute}{\textit{``insight''}}} \\[3pt]
  \rimg[width=0.104\textwidth]{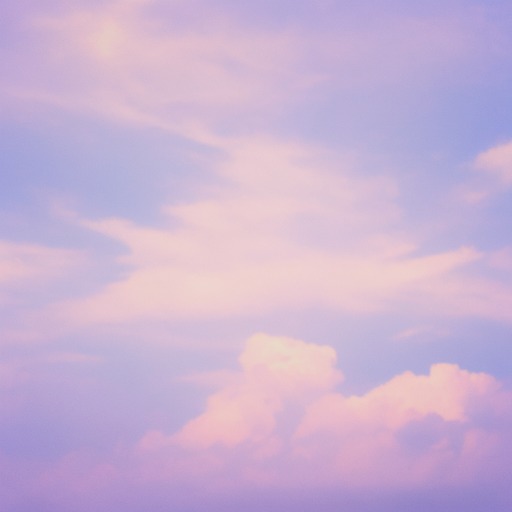} & \rimg[width=0.104\textwidth]{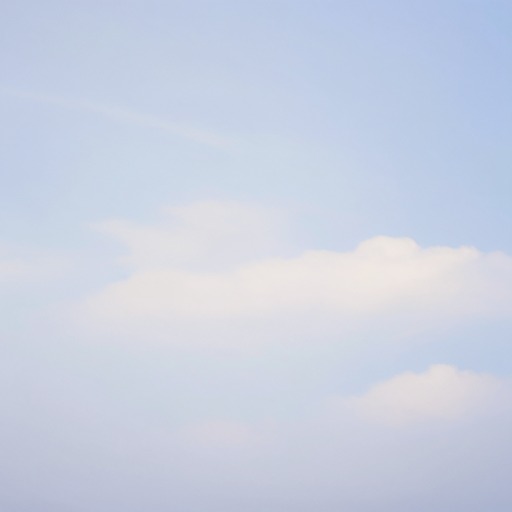} & \rimg[width=0.104\textwidth]{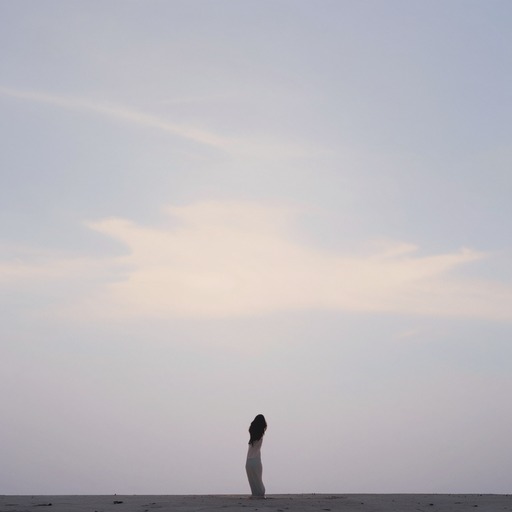} & \rimg[width=0.104\textwidth]{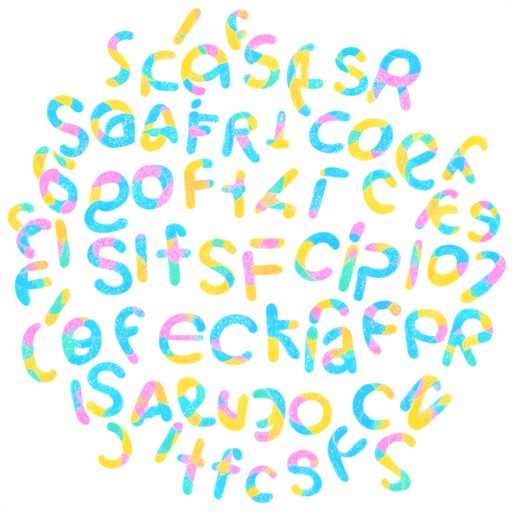} & \rimg[width=0.104\textwidth]{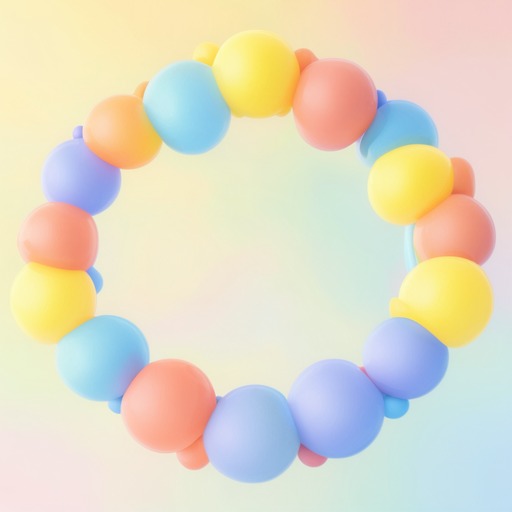} & \rimg[width=0.104\textwidth]{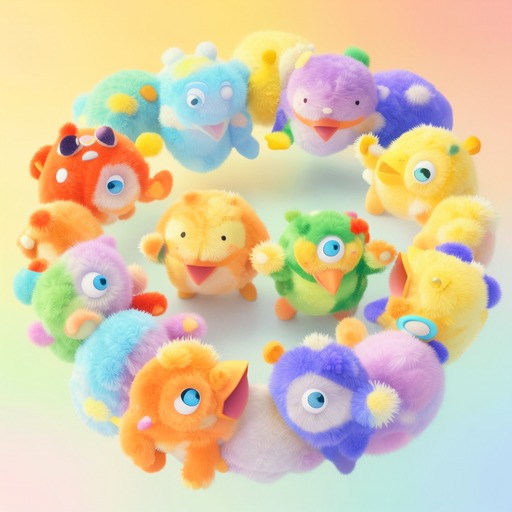} & \rimg[width=0.104\textwidth]{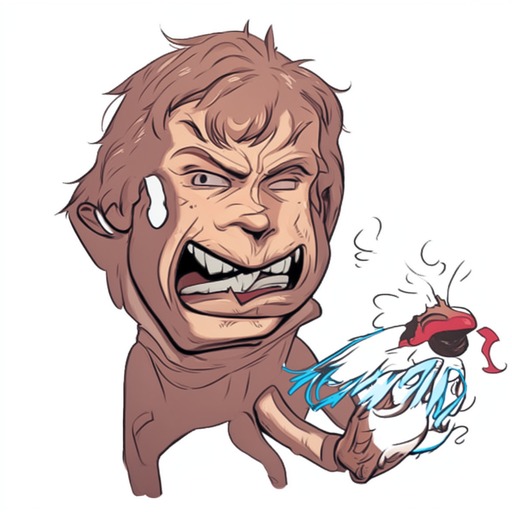} & \rimg[width=0.104\textwidth]{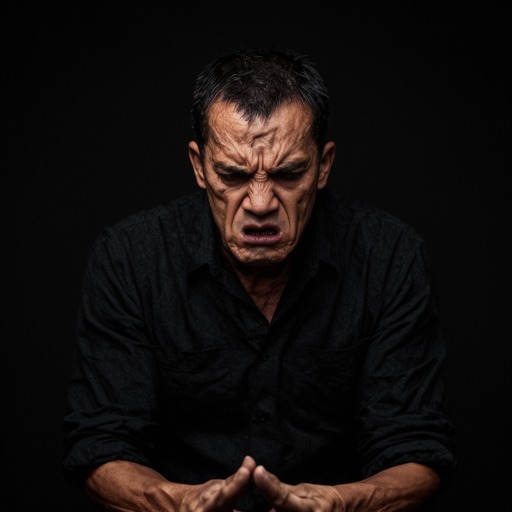} & \rimg[width=0.104\textwidth]{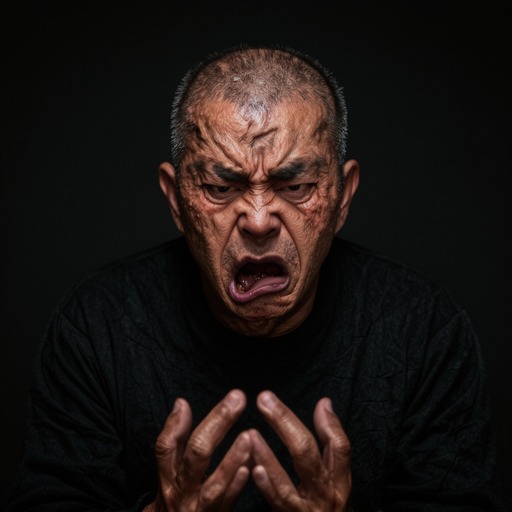} \\
  \multicolumn{3}{c}{\scriptsize\textcolor{qmute}{\textit{``hope''}}} & \multicolumn{3}{c}{\scriptsize\textcolor{qmute}{\textit{``supercalifragilisticexpialidocious''}}} & \multicolumn{3}{c}{\scriptsize\textcolor{qmute}{\textit{``anger''}}} \\[3pt]
  \end{tabular}\end{adjustbox}\end{qualband}
  \caption{\textbf{Ambiguous prompts on FLUX.2 [klein]} (same seed).}
  \label{fig:supp_flux_ambig}
\end{figure*}

\paragraph{Results.} The main paper reports the study with a ``both are equally good'' option: the instruction of
\cref{sec:supp_judge} with the sentence ``If both are equally good results for this request, answer tie'' added.
\Cref{tab:supp_ambig_split} gives the result per seed and the mean over the three seeds. Each seed is a separate comparison of
100 pairs with its own sheets and A/B order, every sheet with its own fresh judge. The main paper shows the best seed of each backbone, the one with the
largest margin of \method wins over losses across both comparisons: seed 1 for both SD3.5-M and FLUX.2 [klein].
\Cref{fig:supp_ambig,fig:supp_flux_ambig} show ambiguous-prompt qualitatives.

\begin{table*}[t]
\caption{\textbf{Ambiguous prompts with a tie option}, by seed: share of pairs (\%) in which \method is preferred (win), both images are judged equally good (tie) and the opponent is preferred (loss), for Claude Opus 5.5; the last group is the average over the three seeds (pooled). Shaded: against the SFT control. $100$ pairs per seed ($300$ pooled).}
\label{tab:supp_ambig_split}
\centering\scriptsize
\setlength{\tabcolsep}{2.8pt}
\begin{adjustbox}{max width=\textwidth}
\begin{tabular}{@{}ll@{\hspace{5pt}}ccc@{\hspace{5pt}}ccc@{\hspace{5pt}}ccc@{\hspace{7pt}}ccc@{}}
\toprule
& & \multicolumn{3}{c}{seed 0} & \multicolumn{3}{c}{seed 1} & \multicolumn{3}{c}{seed 2} & \multicolumn{3}{c}{\textbf{all three}} \\
\cmidrule(lr){3-5}\cmidrule(lr){6-8}\cmidrule(lr){9-11}\cmidrule(lr){12-14}
Backbone & Comparison & Win & Tie & Loss & Win & Tie & Loss & Win & Tie & Loss & Win & Tie & Loss \\
\midrule
\multicolumn{14}{@{}l}{\textbf{Claude Opus 5.5}} \\[1pt]
SD3.5-M & vs.\ base & 67.0 & 8.0 & 25.0 & 70.0 & 2.0 & 28.0 & 63.0 & 0.0 & 37.0 & \textbf{66.7} & 3.3 & 30.0 \\
\rowcolor{sdours} & vs.\ SFT control & 40.0 & 27.0 & 33.0 & 43.0 & 28.0 & 29.0 & 43.0 & 19.0 & 38.0 & \textbf{42.0} & 24.7 & 33.3 \\
FLUX.2 [klein] & vs.\ base & 64.0 & 2.0 & 34.0 & 63.0 & 4.0 & 33.0 & 56.0 & 7.0 & 37.0 & \textbf{61.0} & 4.3 & 34.7 \\
\rowcolor{fluxours} & vs.\ SFT control & 36.0 & 29.0 & 35.0 & 40.0 & 25.0 & 35.0 & 36.0 & 27.0 & 37.0 & \textbf{37.3} & 27.0 & 35.7 \\
\bottomrule
\end{tabular}
\end{adjustbox}
\end{table*}

\paragraph{SFT control against the base model.} As for fine detail, we compare the SFT control with the base model on the
same 100 prompts and three seeds (\cref{tab:supp_ambig_sftbase}). It is preferred to the base model about as often as \method:
among the pairs Claude Opus 5.5 decides, in $68.6\%$ (SD3.5-M; $293$ decided pairs $=$ $201$ wins $+$ $92$ losses) and $64.4\%$ (FLUX.2 [klein]; $284$ decided pairs $=$ $183$ wins $+$ $101$ losses), against $69.0\%$ ($290$ decided pairs $=$ $200$ wins $+$ $90$ losses) and $63.8\%$ ($287$ decided pairs $=$ $183$ wins $+$ $104$ losses) for \method
(\cref{tab:supp_ambig_split}). Here too, gains over the base model can also come from the quality of the training data; but
\method, which writes, outperforms the SFT control, which does not, so writing helps on ambiguous prompts as well.

\begin{table*}[t]
\caption{\textbf{Ambiguous prompts, SFT control against the base model}, with a tie option, by seed: share of pairs (\%) in which the SFT control is preferred (win), both images are judged equally good (tie) and the base model is preferred (loss), for Claude Opus 5.5 and Qwen3-VL-32B; the last group is the average over the three seeds (pooled). $100$ pairs per seed ($300$ pooled).}
\label{tab:supp_ambig_sftbase}
\centering\scriptsize
\setlength{\tabcolsep}{2.8pt}
\begin{adjustbox}{max width=\textwidth}
\begin{tabular}{@{}ll@{\hspace{5pt}}ccc@{\hspace{5pt}}ccc@{\hspace{5pt}}ccc@{\hspace{7pt}}ccc@{}}
\toprule
& & \multicolumn{3}{c}{seed 0} & \multicolumn{3}{c}{seed 1} & \multicolumn{3}{c}{seed 2} & \multicolumn{3}{c}{\textbf{all three}} \\
\cmidrule(lr){3-5}\cmidrule(lr){6-8}\cmidrule(lr){9-11}\cmidrule(lr){12-14}
Backbone & Comparison & Win & Tie & Loss & Win & Tie & Loss & Win & Tie & Loss & Win & Tie & Loss \\
\midrule
\multicolumn{14}{@{}l}{\textbf{Claude Opus 5.5}} \\[1pt]
\rowcolor{sdours}SD3.5-M & SFT vs.\ base & 72.0 & 2.0 & 26.0 & 67.0 & 3.0 & 30.0 & 62.0 & 2.0 & 36.0 & \textbf{67.0} & 2.3 & 30.7 \\
\rowcolor{fluxours}FLUX.2 [klein] & SFT vs.\ base & 67.0 & 3.0 & 30.0 & 58.0 & 5.0 & 37.0 & 58.0 & 8.0 & 34.0 & \textbf{61.0} & 5.3 & 33.7 \\
\midrule
\multicolumn{14}{@{}l}{\textbf{Qwen3-VL-32B}} \\[1pt]
\rowcolor{sdours}SD3.5-M & SFT vs.\ base & 68.0 & 0.0 & 32.0 & 68.0 & 1.0 & 31.0 & 71.0 & 0.0 & 29.0 & \textbf{69.0} & 0.3 & 30.7 \\
\rowcolor{fluxours}FLUX.2 [klein] & SFT vs.\ base & 56.0 & 3.0 & 41.0 & 52.0 & 4.0 & 44.0 & 55.0 & 1.0 & 44.0 & \textbf{54.3} & 2.7 & 43.0 \\
\bottomrule
\end{tabular}
\end{adjustbox}
\end{table*}

\paragraph{Forced choice.} We run the comparison without the tie option, on the same $1{,}200$ sheets ($100$ prompts,
three seeds, two backbones, two opponents) and A/B order: the judge always chooses A or B and, if it sees no difference, picks
the image that is the slightly better result for the request. \Cref{tab:supp_ambig_forced} gives the result. \method stays ahead of
both models, and the shares match those of the tie-option study on the pairs the judge decides
(\cref{tab:supp_ambig_split}) within about one point: averaged over the three seeds, \method is preferred to the base
model in $70.3\%$ (SD3.5-M) and $63.7\%$ (FLUX.2 [klein]) of pairs and to the SFT control in $54.7\%$ and $52.0\%$ (sign test over the $300$ pairs against the base model: $p{=}10^{-12}$ and $2{\cdot}10^{-6}$).

\begin{table}[H]
\caption{\textbf{Ambiguous prompts with forced choice}, by seed: share of pairs (\%) in which \method is preferred (win) and the opponent is preferred (loss), for Claude Opus 5.5; the last group is the average over the three seeds (pooled). Shaded: against the SFT control. $100$ pairs per seed ($300$ pooled).}
\label{tab:supp_ambig_forced}
\centering\scriptsize
\setlength{\tabcolsep}{2.8pt}
\begin{adjustbox}{max width=\linewidth}
\begin{tabular}{@{}ll@{\hspace{5pt}}cc@{\hspace{5pt}}cc@{\hspace{5pt}}cc@{\hspace{7pt}}cc@{}}
\toprule
& & \multicolumn{2}{c}{seed 0} & \multicolumn{2}{c}{seed 1} & \multicolumn{2}{c}{seed 2} & \multicolumn{2}{c}{\textbf{all three}} \\
\cmidrule(lr){3-4}\cmidrule(lr){5-6}\cmidrule(lr){7-8}\cmidrule(lr){9-10}
Backbone & Comparison & Win & Loss & Win & Loss & Win & Loss & Win & Loss \\
\midrule
\multicolumn{10}{@{}l}{\textbf{Claude Opus 5.5}} \\[1pt]
SD3.5-M & vs.\ base & 73.0 & 27.0 & 72.0 & 28.0 & 66.0 & 34.0 & \textbf{70.3} & 29.7 \\
\rowcolor{sdours} & vs.\ SFT control & 52.0 & 48.0 & 59.0 & 41.0 & 53.0 & 47.0 & \textbf{54.7} & 45.3 \\
FLUX.2 [klein] & vs.\ base & 64.0 & 36.0 & 65.0 & 35.0 & 62.0 & 38.0 & \textbf{63.7} & 36.3 \\
\rowcolor{fluxours} & vs.\ SFT control & 51.0 & 49.0 & 53.0 & 47.0 & 52.0 & 48.0 & \textbf{52.0} & 48.0 \\
\bottomrule
\end{tabular}
\end{adjustbox}
\end{table}

\section{The read-out decoder}
\label{sec:supp_g_readout}
Descriptions decoded from generated streams, in domain and out of domain, and ablations at the level of the read-out: direct mapping against the language-model decoder, and the choice of decoder.

\subsection{Held-out examples with decoded descriptions}
\label{sec:supp_readout}
\begin{figure*}[!tp]
  \centering
  \setlength{\tabcolsep}{2pt}
  \begin{qualband}{SD3.5 Medium}\begin{adjustbox}{max width=\linewidth}%
  \begin{tabular}{@{}cccp{0.52\textwidth}@{}}
  \scriptsize SD3.5-M & \scriptsize SFT control & \scriptsize \textcolor{sdcol}{\textbf{\method}} & \scriptsize prompt and the description \method decodes for its image \\
  \raisebox{-\height}{\rimg[width=0.145\textwidth]{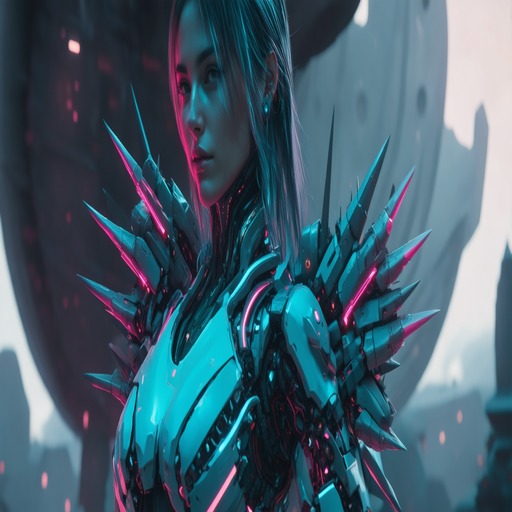}} & \raisebox{-\height}{\rimg[width=0.145\textwidth]{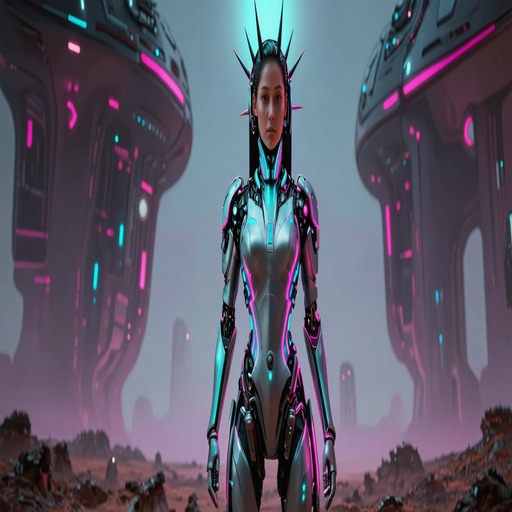}} & \raisebox{-\height}{\rimg[width=0.145\textwidth]{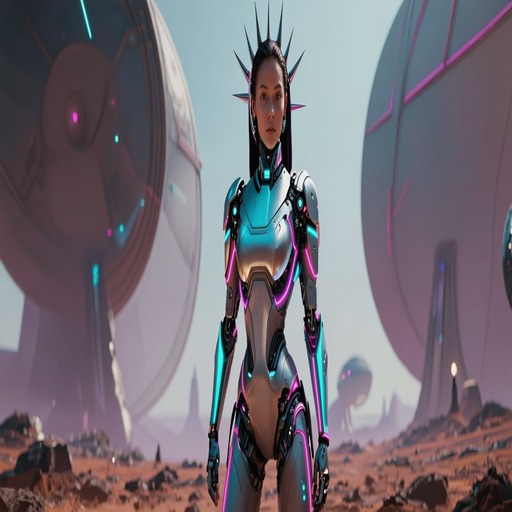}} & \vspace{0pt}\scriptsize\qplabel{PROMPT} {\color{qmute}\itshape The image features a futuristic woman in a modular mechanical armored suit, with sharp spines protruding from her upper back. The suit has a mixed [\dots]}\newline\qdlabel{DESCRIPTION} Female robot: The arms are mechanical structure, with a metallic sheen. Arm guards on the shoulders and chest; both hands (one at the joints), in an abstract circuit board-like device is shown by her headgear or gear structures visible). She has arm naturally hanging down to the sides of the body is made of silver-gray-white metal armor, including cyan-blue glowing lines. \\[3pt]
  \raisebox{-\height}{\rimg[width=0.145\textwidth]{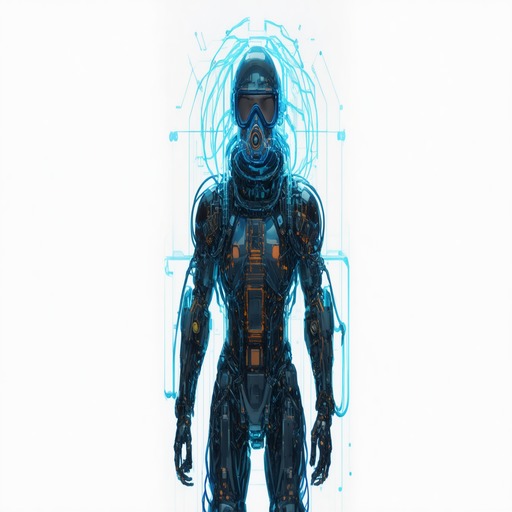}} & \raisebox{-\height}{\rimg[width=0.145\textwidth]{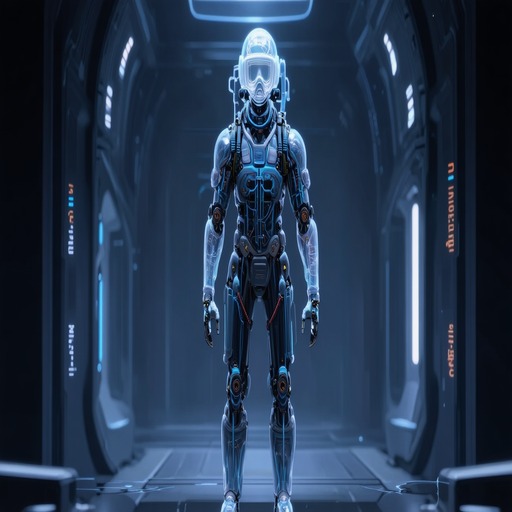}} & \raisebox{-\height}{\rimg[width=0.145\textwidth]{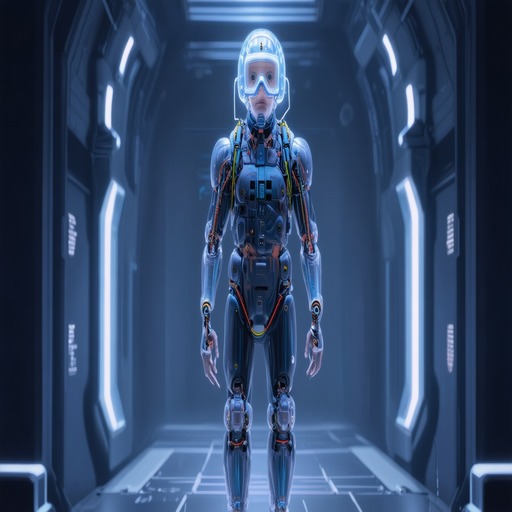}} & \vspace{0pt}\scriptsize\qplabel{PROMPT} {\color{qmute}\itshape Concept art, neurosync diver character for deep cognition vault world, full-body, wearing translucent cortex mask and mnemonic wiring, standing [\dots]}\newline\qdlabel{DESCRIPTION} Robot: Human robot, with a transparent visor; the torso and is made of cyan light effects. The human skeleton structure has an internal mechanical device or robotic meshed form (resembling circuit board), in its body shape, emitting orange be controlled panels at both ends). A humanoid creature's head is suspended from inside to it as if he towards the joints are connected by her arm structures. \\[3pt]
  \raisebox{-\height}{\rimg[width=0.145\textwidth]{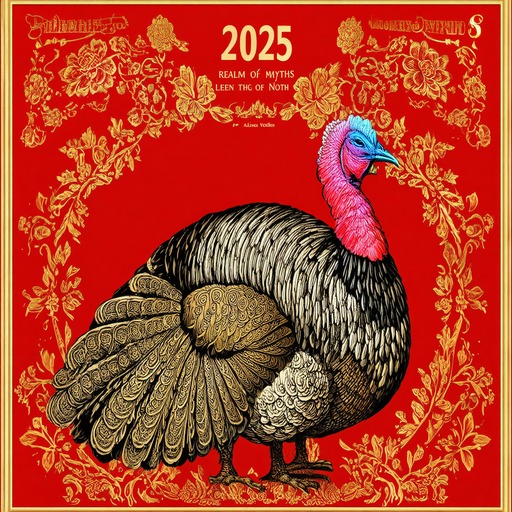}} & \raisebox{-\height}{\rimg[width=0.145\textwidth]{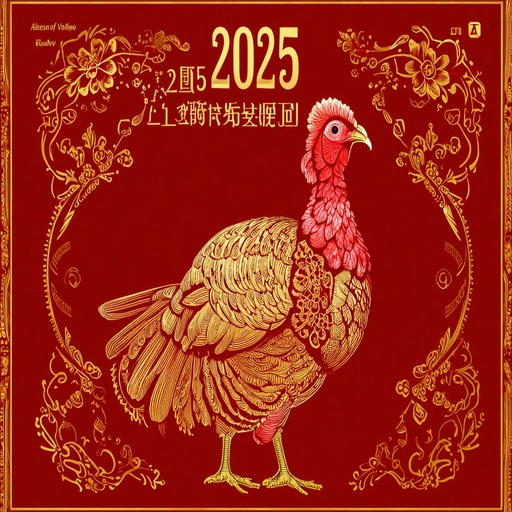}} & \raisebox{-\height}{\rimg[width=0.145\textwidth]{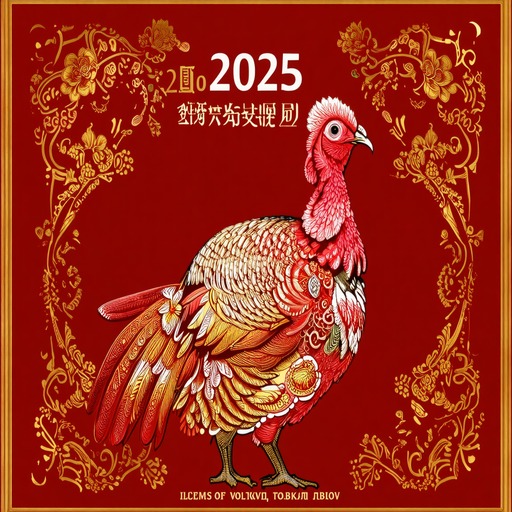}} & \vspace{0pt}\scriptsize\qplabel{PROMPT} {\color{qmute}\itshape Poster design, with the main visual featuring a turkey pattern rich in traditional Russian cultural elements. The background is dominated by red, [\dots]}\newline\qdlabel{DESCRIPTION} Bird: Bird feathers, with a golden feather-like patterned decorations on the chest. The bird's body is in gold and red. \\[3pt]
  \raisebox{-\height}{\rimg[width=0.145\textwidth]{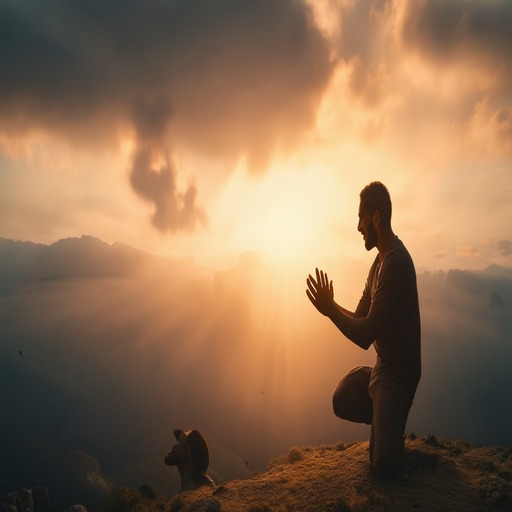}} & \raisebox{-\height}{\rimg[width=0.145\textwidth]{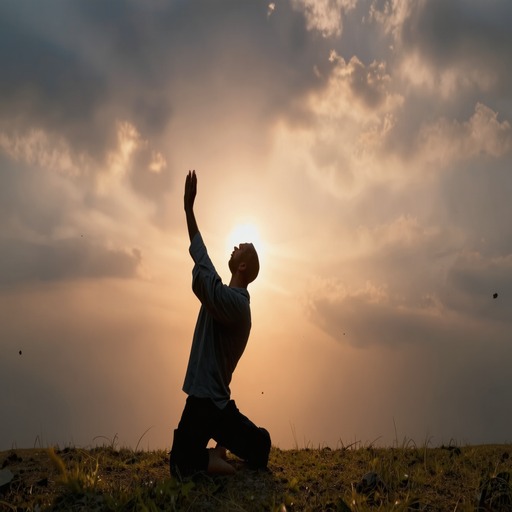}} & \raisebox{-\height}{\rimg[width=0.145\textwidth]{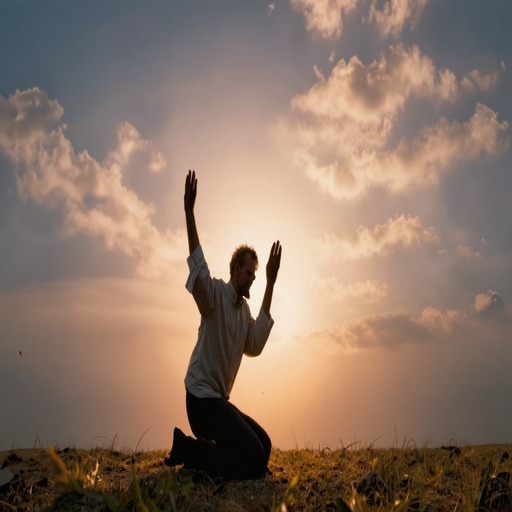}} & \vspace{0pt}\scriptsize\qplabel{PROMPT} {\color{qmute}\itshape Make a man adoring God with his hands raised to the sky, with the man on his knees praying.}\newline\qdlabel{DESCRIPTION} An elderly man: An older male, with short hair slightly disheveled and his face; wearing a white shirt (with the hood), dark pants. His right hand is naturally resting on an office bench of him or kneeling position. \\[3pt]
  \raisebox{-\height}{\rimg[width=0.145\textwidth]{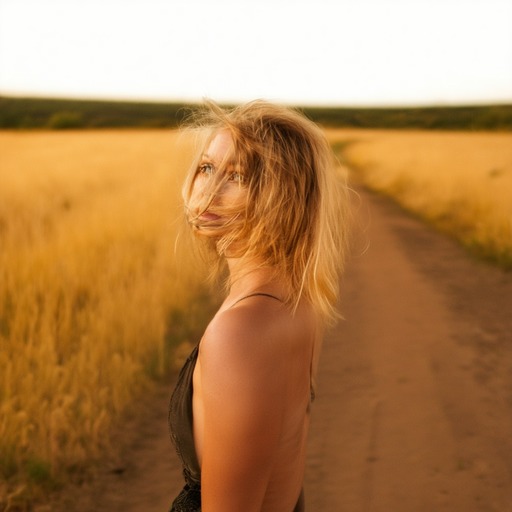}} & \raisebox{-\height}{\rimg[width=0.145\textwidth]{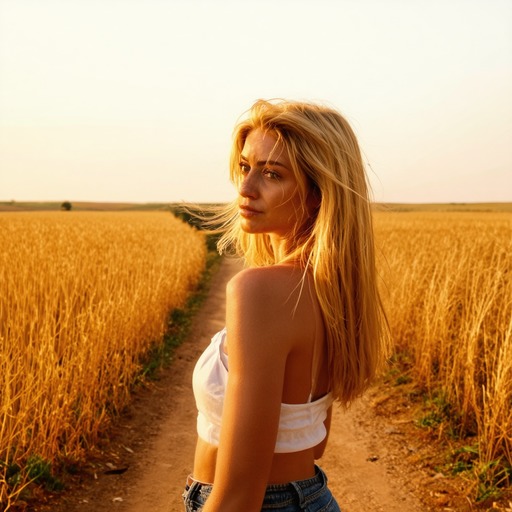}} & \raisebox{-\height}{\rimg[width=0.145\textwidth]{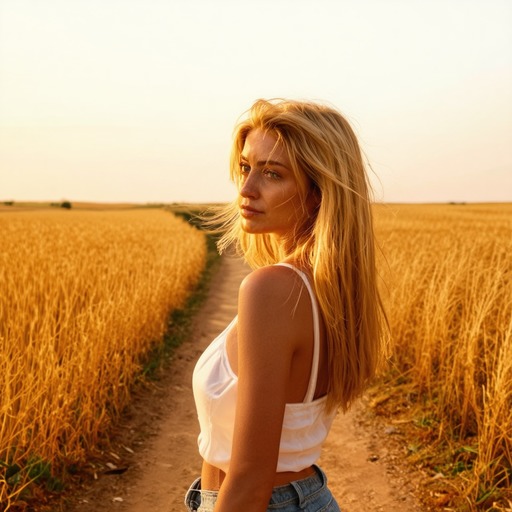}} & \vspace{0pt}\scriptsize\qplabel{PROMPT} {\color{qmute}\itshape A woman, with sun-kissed blonde hair and a thoughtful expression, pauses on a dirt path in an expansive, golden-hued field. She looks back over her [\dots]}\newline\qdlabel{DESCRIPTION} Young woman: 1, with long hair flowing in the wind. Wearing a white V-neck design (with golden yellow-brown), light orange and blue color scheme of denim skirt; no facial features or missing limbs are visible. \\[3pt]
  \raisebox{-\height}{\rimg[width=0.145\textwidth]{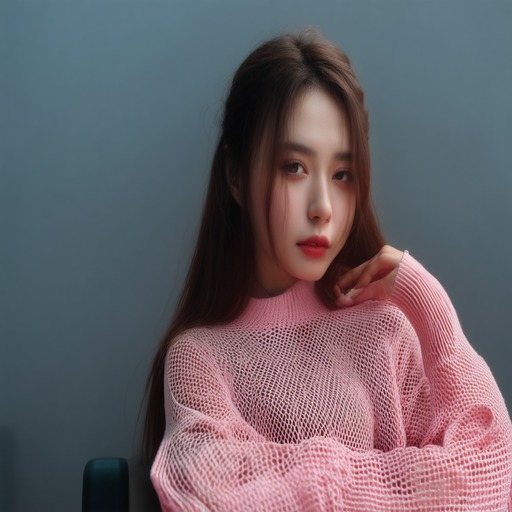}} & \raisebox{-\height}{\rimg[width=0.145\textwidth]{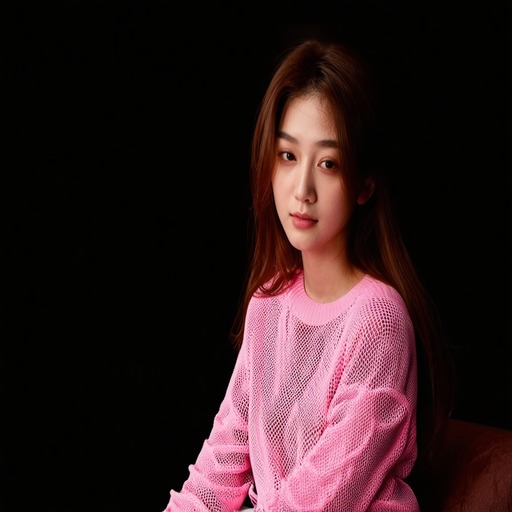}} & \raisebox{-\height}{\rimg[width=0.145\textwidth]{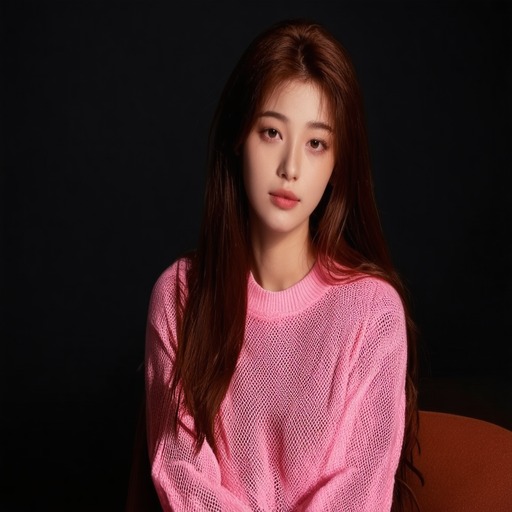}} & \vspace{0pt}\scriptsize\qplabel{PROMPT} {\color{qmute}\itshape Realistic photo, 25-year-old Chinese beauty,hazy dark tone background scene, lazily sitting. Wearing a lazy-style pink net fashion sweater, chestnut [\dots]}\newline\qdlabel{DESCRIPTION} Young woman: East Asian features, oval face shape (slightly messy), with a brownish-brown hair strands flowing to the side of her eyes. She is wearing an orange knitted top and long lashes; hands are naturally hanging down at both sides in front of her body, looking up at the bottom. \\[3pt]
  \end{tabular}\end{adjustbox}\end{qualband}
  \caption{\textbf{Held-out prompts with decoded descriptions.} Descriptions are shown in full.}
  \label{fig:supp_ho1}
\end{figure*}

\Cref{fig:supp_ho1} shows prompts from our held-out set: the image of SD3.5 Medium, the SFT control and \method at
the same seed, and the description that \method generates together with its image. The description stream is sampled jointly
with the image from noise (28 steps, guidance 4); the image tokens never read it, so the image is the same with or without it.
$\phi$ decodes it greedily (repetition penalty 1.3, no repeated 3-grams). Descriptions are shown in full and unedited, except that the
trailing Chinese full stop of our description format is printed as a period. We show six samples.
On held-out prompts the SFT control and \method produce similar
images at the same seed, so these examples show what the model says, not a difference in images.

\begin{figure*}[!tp]
  \centering
  \setlength{\tabcolsep}{2pt}
  \begin{qualband}{FLUX.2 [klein]}\begin{adjustbox}{max width=\linewidth}%
  \begin{tabular}{@{}cccp{0.52\textwidth}@{}}
  \scriptsize FLUX.2 [klein] & \scriptsize SFT control & \scriptsize \textcolor{sdcol}{\textbf{\method}} & \scriptsize prompt and the description \method decodes for its image \\
  \raisebox{-\height}{\rimg[width=0.145\textwidth]{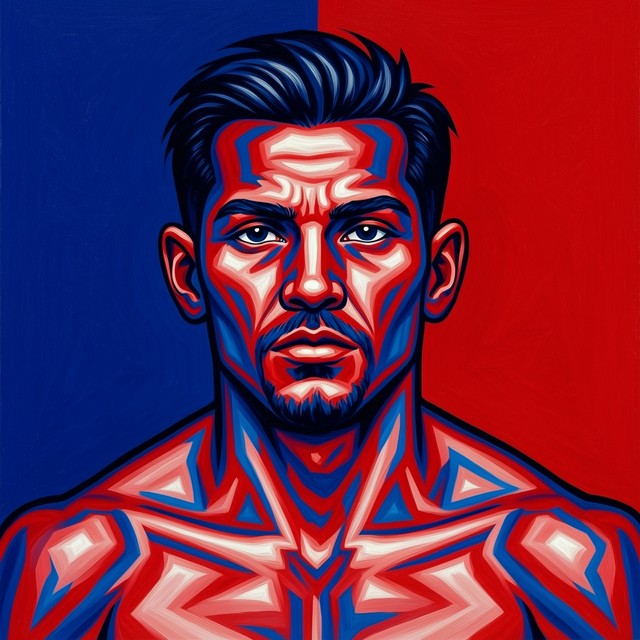}} & \raisebox{-\height}{\rimg[width=0.145\textwidth]{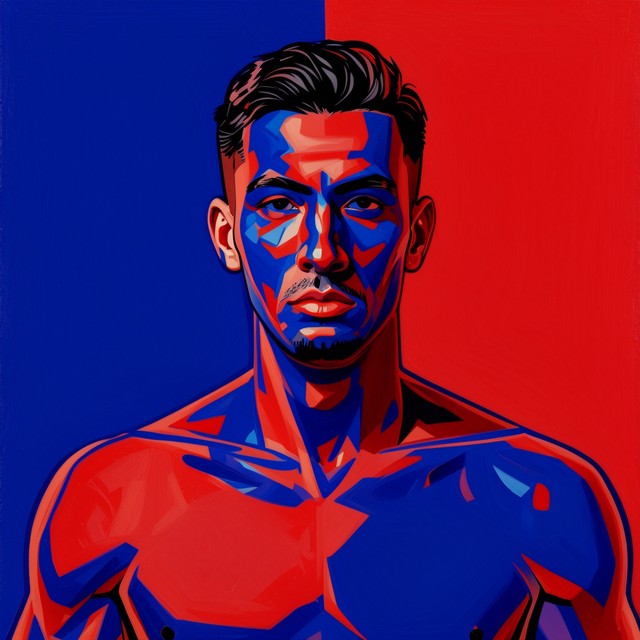}} & \raisebox{-\height}{\rimg[width=0.145\textwidth]{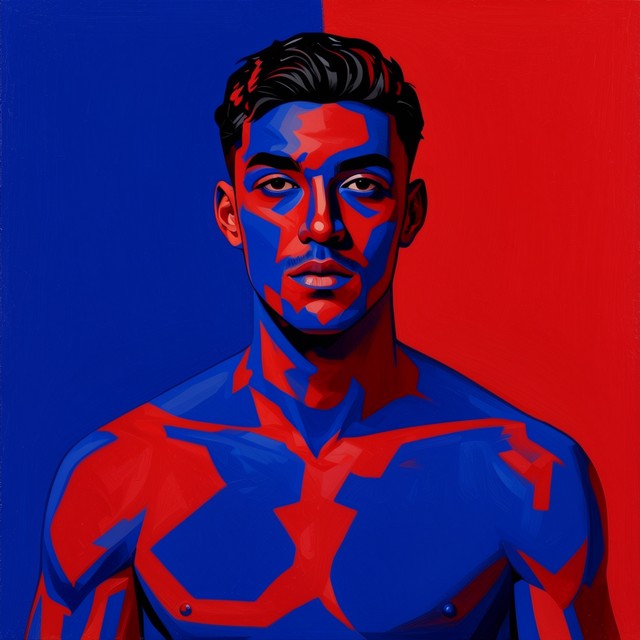}} & \vspace{0pt}\scriptsize\qplabel{PROMPT} {\color{qmute}\itshape Graphic art, portrait of a man, background in deep blue and red, intense color contrasts on the face and torso, stark and expressive facial features [\dots]}\newline\qdlabel{DESCRIPTION} male: tall, robust physique with a clear facial contour (no beard or mustache), broad nose bridge. Hair is short and slightly parted in the middle of the face; no visible accessories are held items present. \\[3pt]
  \raisebox{-\height}{\rimg[width=0.145\textwidth]{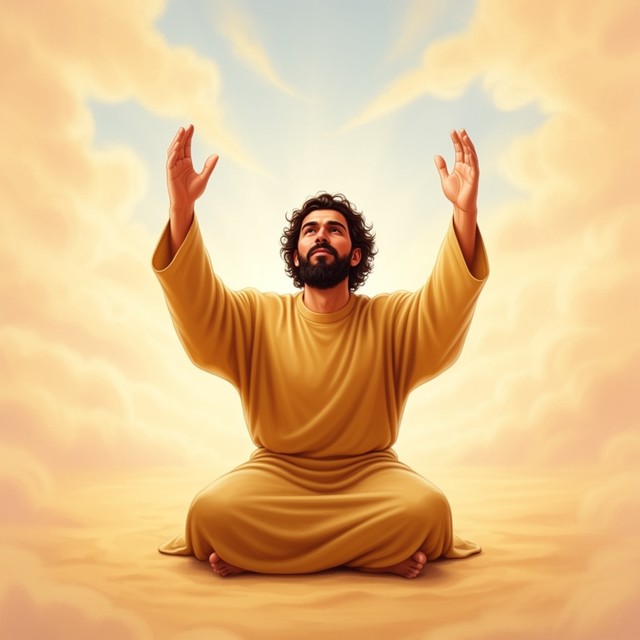}} & \raisebox{-\height}{\rimg[width=0.145\textwidth]{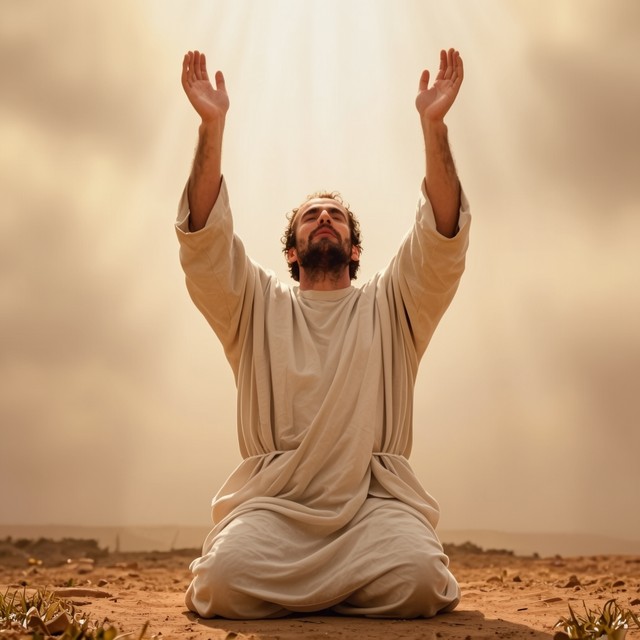}} & \raisebox{-\height}{\rimg[width=0.145\textwidth]{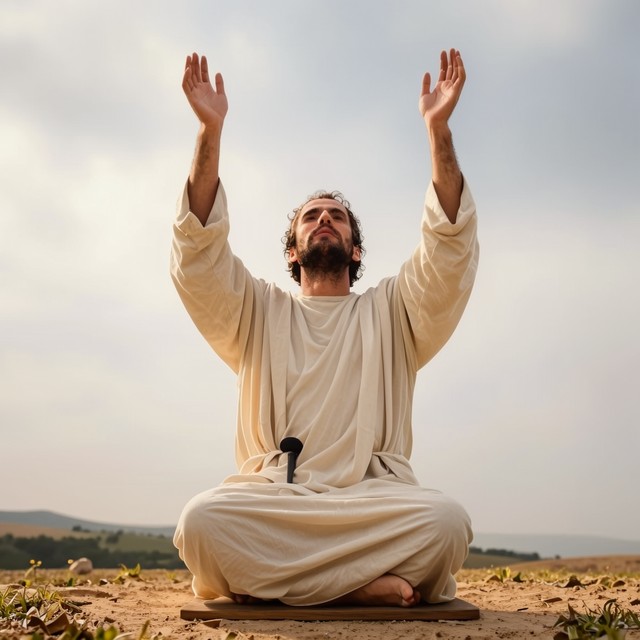}} & \vspace{0pt}\scriptsize\qplabel{PROMPT} {\color{qmute}\itshape Make a man adoring God with his hands raised to the sky, with the man on his knees praying.}\newline\qdlabel{DESCRIPTION} male: dark skin, short hair (dark brown), medium-length curly beard. He is wearing a white monk's robe with light beige trim and matching color; his hands are visible, each holding an orange-yellow lantern in both hands. \\[3pt]
  \raisebox{-\height}{\rimg[width=0.145\textwidth]{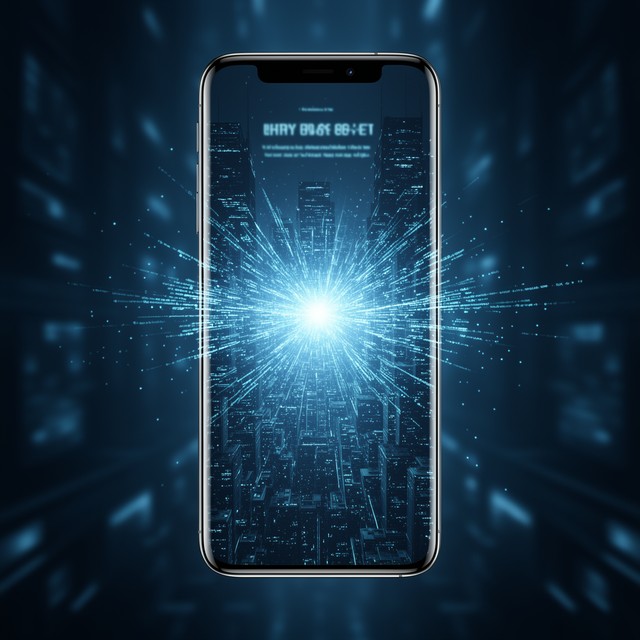}} & \raisebox{-\height}{\rimg[width=0.145\textwidth]{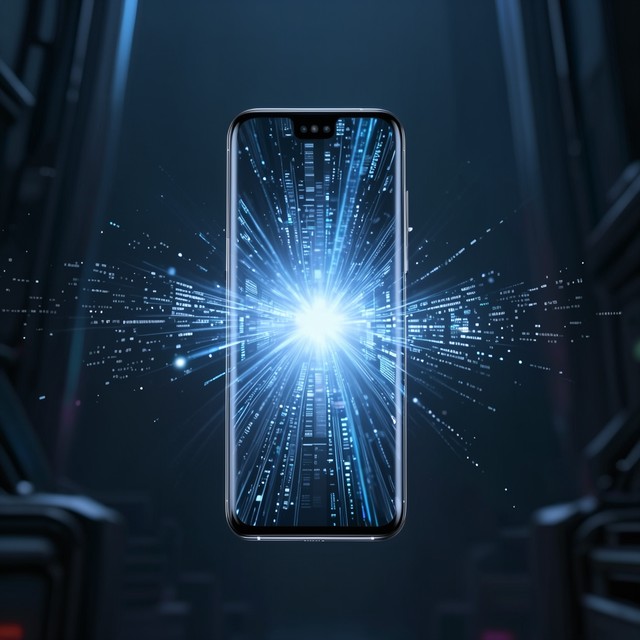}} & \raisebox{-\height}{\rimg[width=0.145\textwidth]{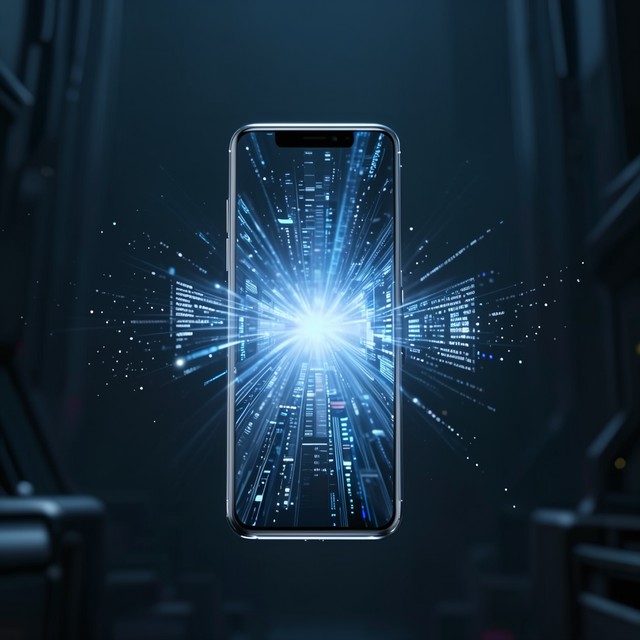}} & \vspace{0pt}\scriptsize\qplabel{PROMPT} {\color{qmute}\itshape Product design shot, a sleek flagship smartphone floating in a dark, cyberpunk-inspired void. Holographic data streams and light particles mainly [\dots]}\newline\qdlabel{DESCRIPTION} silver-gray metallic texture, with a matte finish and high-gloss reflective material; screen displays the interface. The display area of varying sizes are visible in the frame, including 3-40 characters "Z" (such as lines), white glowing lights or light effects inside it is illuminated by smoke rising from within. \\[3pt]
  \raisebox{-\height}{\rimg[width=0.145\textwidth]{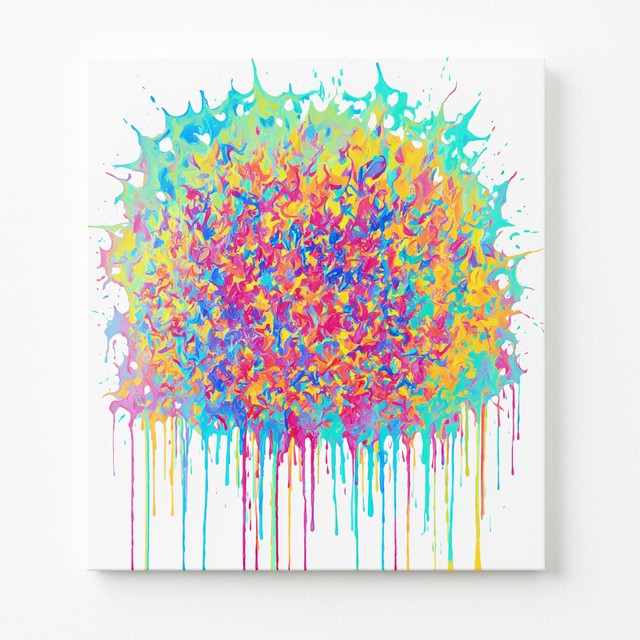}} & \raisebox{-\height}{\rimg[width=0.145\textwidth]{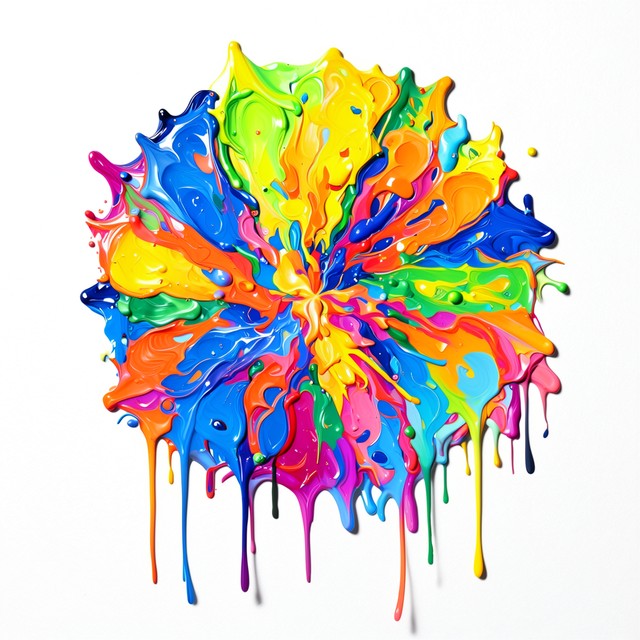}} & \raisebox{-\height}{\rimg[width=0.145\textwidth]{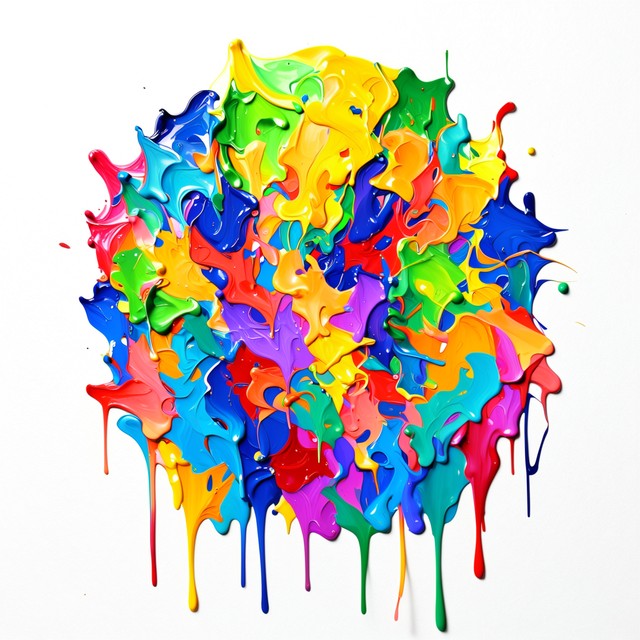}} & \vspace{0pt}\scriptsize\qplabel{PROMPT} {\color{qmute}\itshape A beautiful painting on a white background, dripping paint, abstract art, colorful.}\newline\qdlabel{DESCRIPTION} paint, colors including orange-red (red), yellow, cyan blue, purple). The color blocks of varying sizes are scattered. The material exhibits a liquid or fluid texture resembling the paper; the brushstrokes is painted with numerous red and white splatter effects on its surface. - Figure: A thick black ink wash flow in various shades of deep green and brown tones interwoven together, some areas showing distinct layers of water droplet particles. \\[3pt]
  \raisebox{-\height}{\rimg[width=0.145\textwidth]{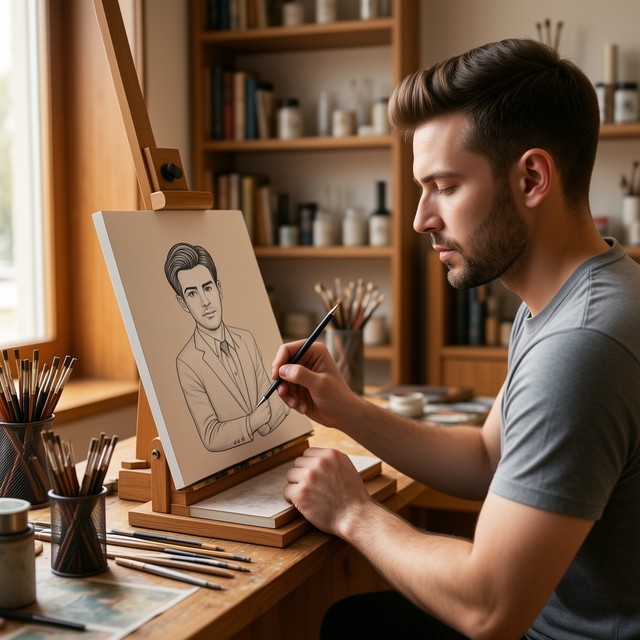}} & \raisebox{-\height}{\rimg[width=0.145\textwidth]{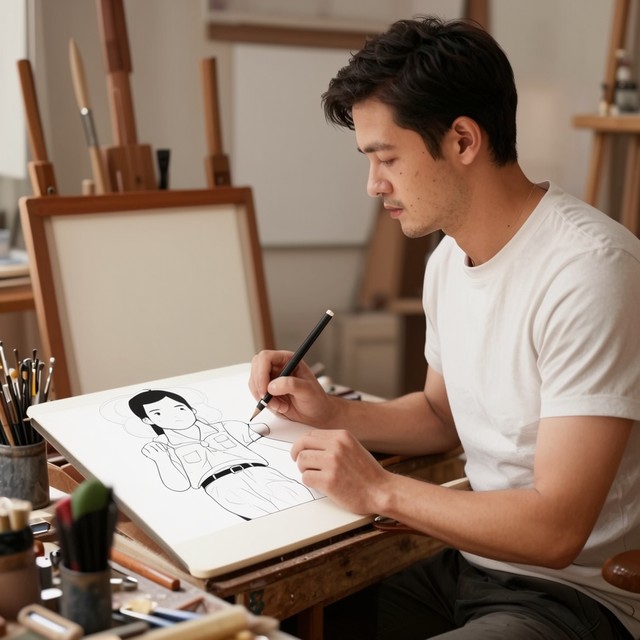}} & \raisebox{-\height}{\rimg[width=0.145\textwidth]{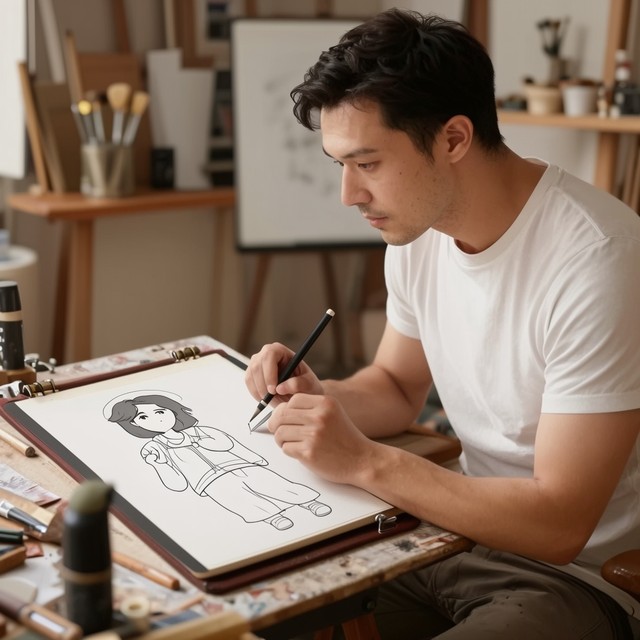}} & \vspace{0pt}\scriptsize\qplabel{PROMPT} {\color{qmute}\itshape A handsome man - he is an illustrator. Sitting in his studio, he is using a drawing pad and drawing a character. On the screen we see his [\dots]}\newline\qdlabel{DESCRIPTION} male: dark short hair, clear facial contour (dark brown), light skin tone; medium build. He is holding a book with his right hand gripping the handle of him are visible. - A white paper pen and placed on it. \\[3pt]
  \raisebox{-\height}{\rimg[width=0.145\textwidth]{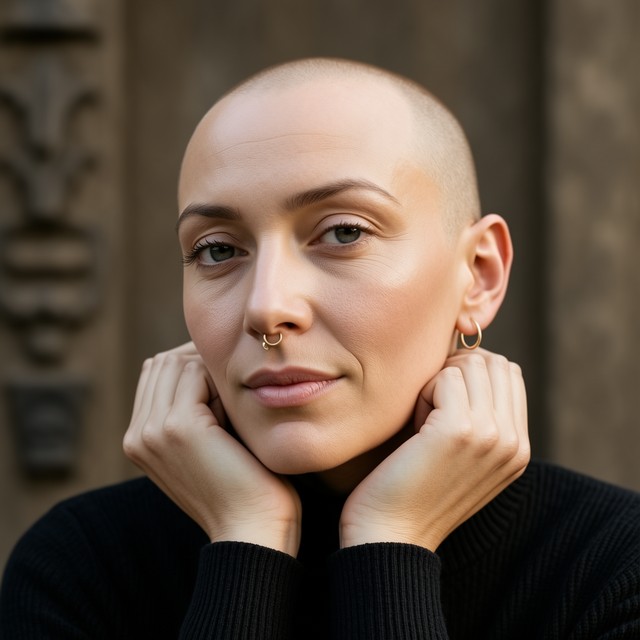}} & \raisebox{-\height}{\rimg[width=0.145\textwidth]{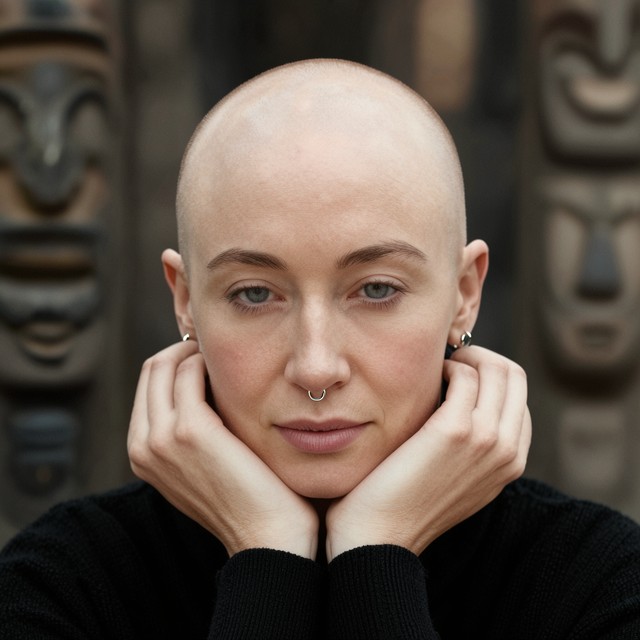}} & \raisebox{-\height}{\rimg[width=0.145\textwidth]{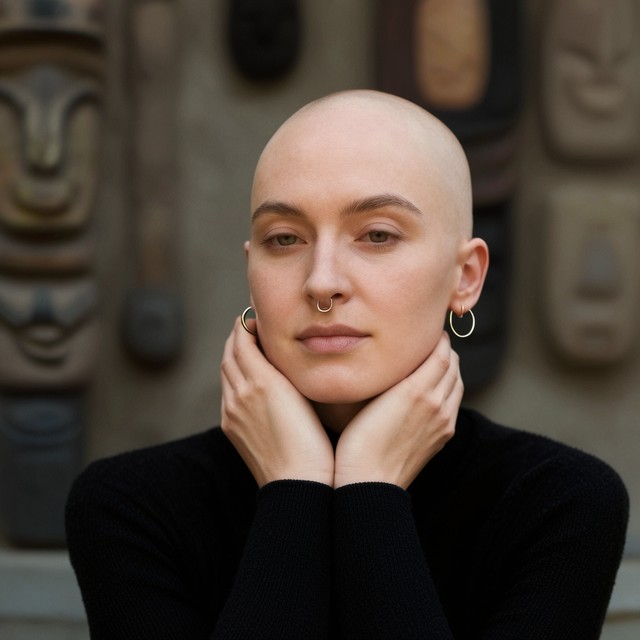}} & \vspace{0pt}\scriptsize\qplabel{PROMPT} {\color{qmute}\itshape A captivating portrait of a bald woman with delicate facial features, wearing minimalist jewelry (such as a single hoop earring or septum ring), in a [\dots]}\newline\qdlabel{DESCRIPTION} A young woman: clear facial contour, almond-shaped eyes with dark brown pupils; deep eye sockets and nose bridge. Clear lip shape is a light beige round-neck short-sleeve top (light-colored) around the neck of her chest), fair skin tone, slender figure. Her hands are clasped together in front of her body. She has long fingers, naturally shaped as if draped to thin metal ring on both sides. \\[3pt]
  \end{tabular}\end{adjustbox}\end{qualband}
  \caption{\textbf{Held-out prompts with decoded descriptions on FLUX.2 [klein].} Descriptions are shown in full.}
  \label{fig:supp_flux_ho}
\end{figure*}

\Cref{fig:supp_flux_ho} shares held-out qualitatives on FLUX.2 [klein]; leading punctuation is removed.

\subsection{Out-of-domain examples: FLUX-Reason-6M}
\label{sec:supp_ood}
\begin{figure*}[!tp]
  \centering
  \setlength{\tabcolsep}{2pt}
  \begin{qualband}{SD3.5 Medium}\begin{adjustbox}{max width=\linewidth}%
  \begin{tabular}{@{}cccp{0.52\textwidth}@{}}
  \scriptsize SD3.5-M & \scriptsize SFT control & \scriptsize \textcolor{sdcol}{\textbf{\method}} & \scriptsize prompt and the description \method decodes for its image \\
  \raisebox{-\height}{\rimg[width=0.145\textwidth]{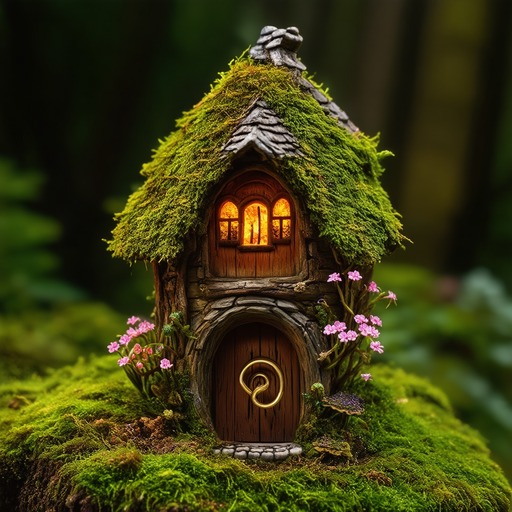}} & \raisebox{-\height}{\rimg[width=0.145\textwidth]{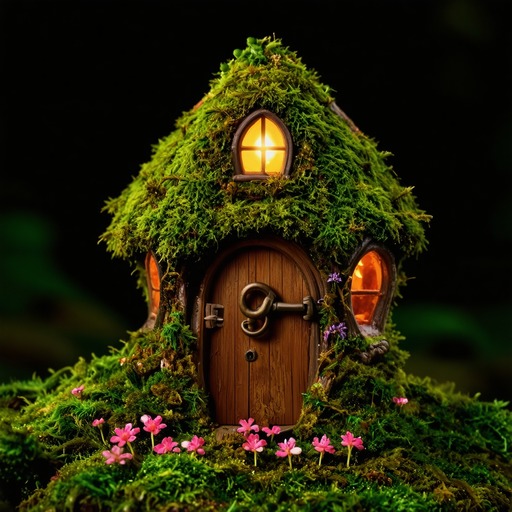}} & \raisebox{-\height}{\rimg[width=0.145\textwidth]{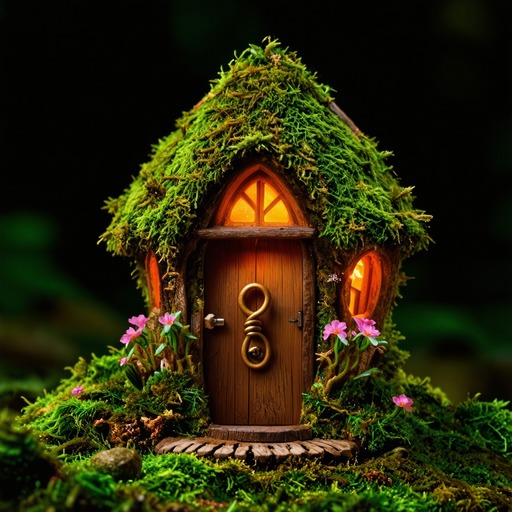}} & \vspace{0pt}\scriptsize\qplabel{PROMPT} {\color{qmute}\itshape A whimsical fairy cottage emerges from a patch of vibrant green moss, adorned with glowing windows, a wooden door with a spiral knocker, and [\dots]}\newline\qdlabel{DESCRIPTION} A small fairy ball: the house, with a white fluffy and purple at its top; on both sides of the body is covered in green moss. The base has pinkish-brown shade that emits light points are visible (1): A wide open book with black spire-like protrusions), hanging down to the right side of the frame, there is an orange-yellow cone-shaped [\dots] \\[3pt]
  \raisebox{-\height}{\rimg[width=0.145\textwidth]{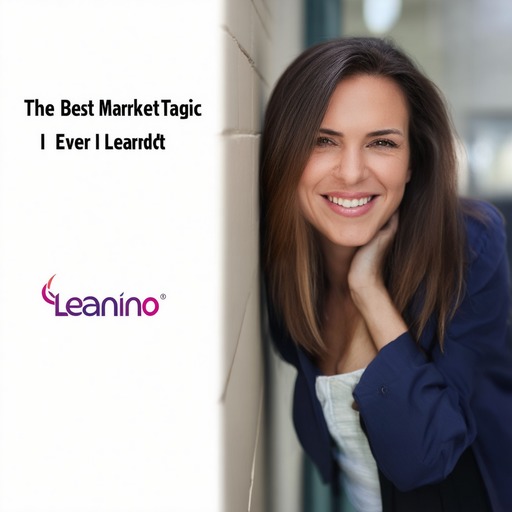}} & \raisebox{-\height}{\rimg[width=0.145\textwidth]{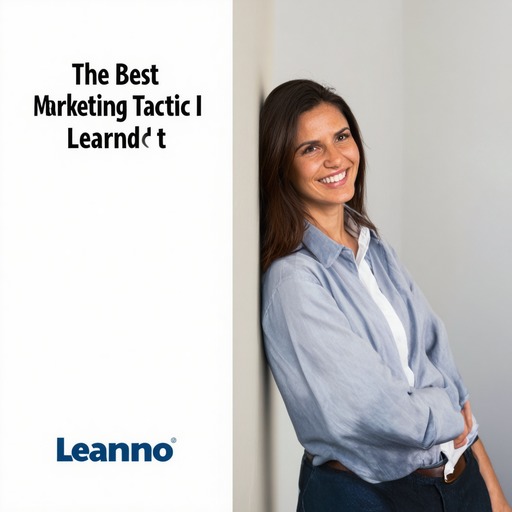}} & \raisebox{-\height}{\rimg[width=0.145\textwidth]{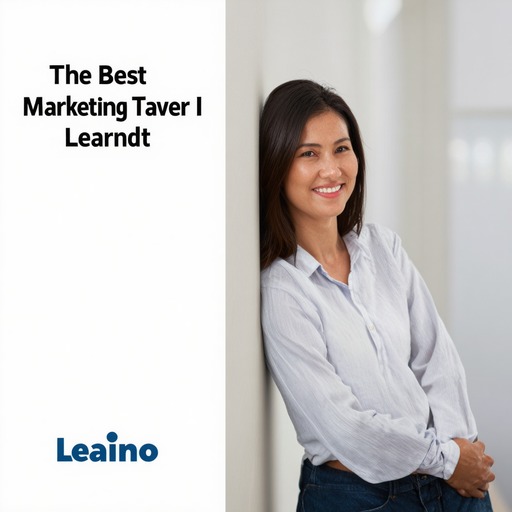}} & \vspace{0pt}\scriptsize\qplabel{PROMPT} {\color{qmute}\itshape A woman leans against a wall with a smile, while text on the left reads "The Best Marketing Tactic I Ever Learndt" and the [\dots]}\newline\qdlabel{DESCRIPTION} A woman: light skin tone, long hair (dark brown), wearing a dark blue denim jacket with white and pleated skirt; holding the left hand is naturally hanging down. -shoulder top of the image contains only one in her right hand's fingers are visible at his back. 2. Head: has deep green body, looking upward to the middle-right) on it: 10. [\dots] \\[3pt]
  \raisebox{-\height}{\rimg[width=0.145\textwidth]{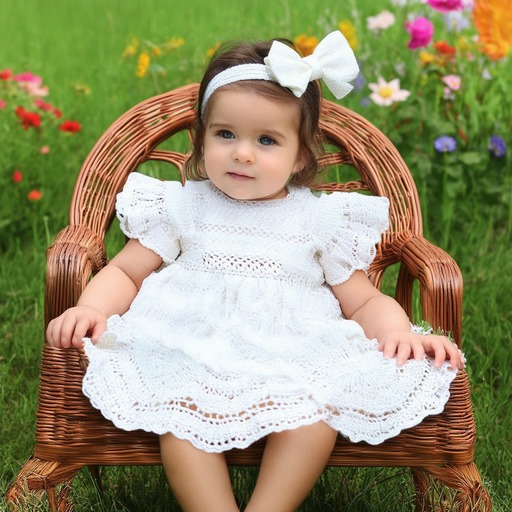}} & \raisebox{-\height}{\rimg[width=0.145\textwidth]{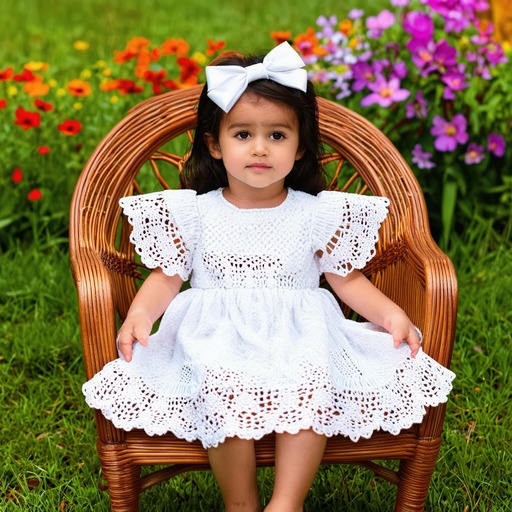}} & \raisebox{-\height}{\rimg[width=0.145\textwidth]{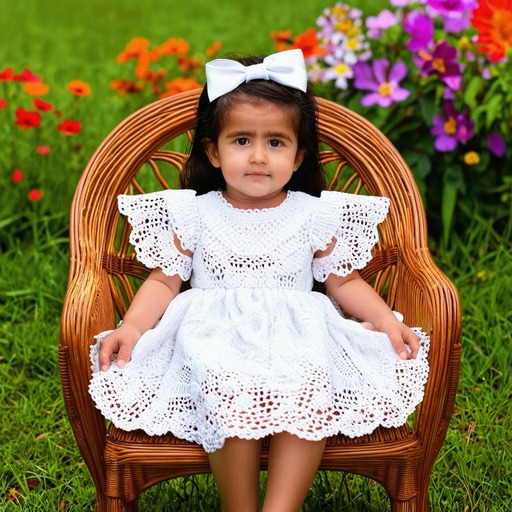}} & \vspace{0pt}\scriptsize\qplabel{PROMPT} {\color{qmute}\itshape Child sitting gracefully on a wicker chair, wearing a delicate white crochet dress with scalloped edges and flutter sleeves, complemented by a matching bow [\dots]}\newline\qdlabel{DESCRIPTION} A woman: one, with a white flower-shaped hair (a pink/white and light beige), fair skin; wearing a sleeveless top hat decorated with floral patterns on the skirt. She is dressed in a white lace dress made of blue silk-like trim at her back, she wears earrings or puffed sleeves are tied around it, revealing an inner garment composed of multiple [\dots] \\[3pt]
  \end{tabular}\end{adjustbox}\end{qualband}
  \par\vspace{4pt}
  \begin{qualband}{FLUX.2 [klein]}\begin{adjustbox}{max width=\linewidth}%
  \begin{tabular}{@{}cccp{0.52\textwidth}@{}}
  \scriptsize FLUX.2 [klein] & \scriptsize SFT control & \scriptsize \textcolor{sdcol}{\textbf{\method}} & \scriptsize prompt and the description \method decodes for its image \\
  \raisebox{-\height}{\rimg[width=0.145\textwidth]{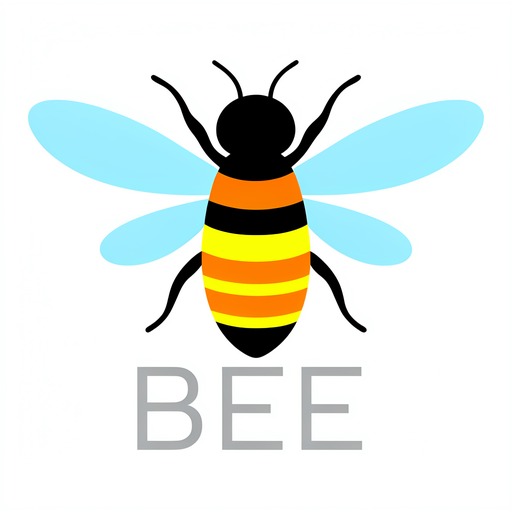}} & \raisebox{-\height}{\rimg[width=0.145\textwidth]{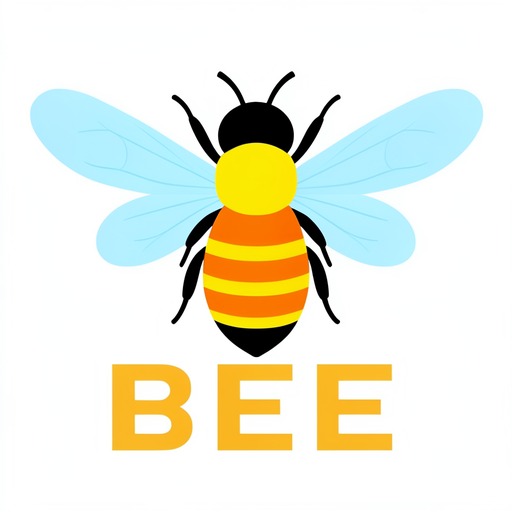}} & \raisebox{-\height}{\rimg[width=0.145\textwidth]{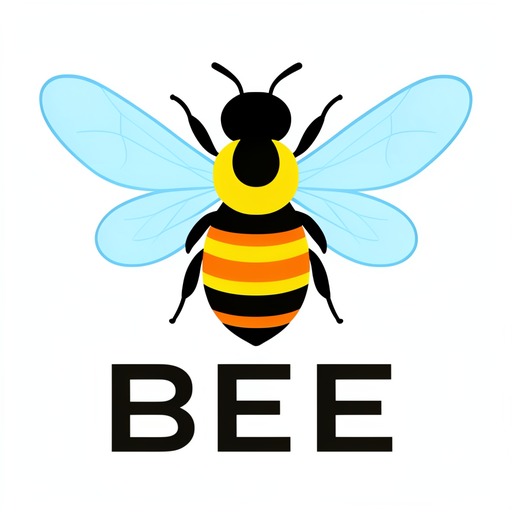}} & \vspace{0pt}\scriptsize\qplabel{PROMPT} {\color{qmute}\itshape A vibrant flat-style illustration of a bee with bold, solid colors featuring a black head, yellow-and-orange striped body, and light blue wings, accompanied by [\dots]}\newline\qdlabel{DESCRIPTION} A cartoon-style white head (or a small black dot), 1 large, round eyes with an oval shape; the mouth is slightly open. The body has two curved lines of yellow and orange-yellow stripes on its surface. - Left: A pair of semi-transparent wings are fully visible, located at different sizes in the upper right area below each side, one left [\dots] \\[3pt]
  \raisebox{-\height}{\rimg[width=0.145\textwidth]{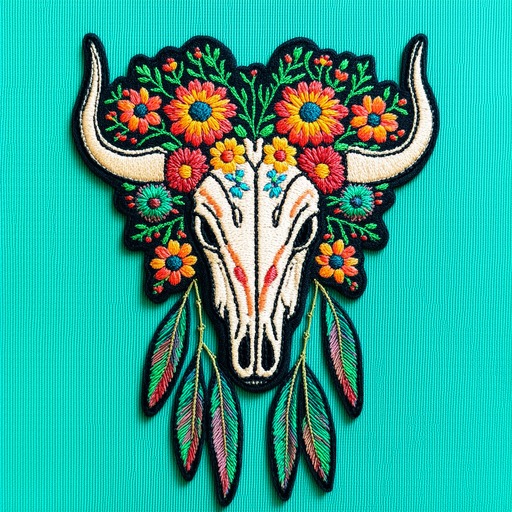}} & \raisebox{-\height}{\rimg[width=0.145\textwidth]{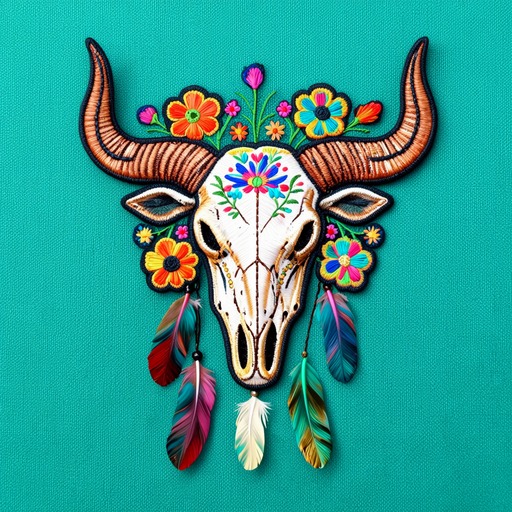}} & \raisebox{-\height}{\rimg[width=0.145\textwidth]{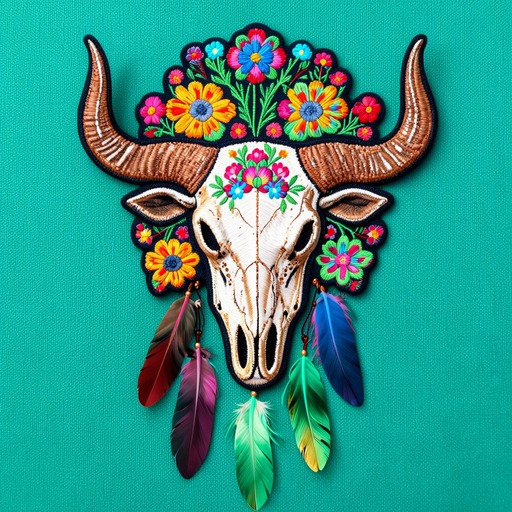}} & \vspace{0pt}\scriptsize\qplabel{PROMPT} {\color{qmute}\itshape A vibrant embroidered patch featuring a stylized steer skull adorned with colorful floral designs and dangling feathers, showcasing a folk-art aesthetic with bold, handcrafted [\dots]}\newline\qdlabel{DESCRIPTION} a deep blue bow, decorated with white and green leaves; the surface of fabric is made up to brass or woven rope. The head has two large areas of red roses (with embroidered patterns), yellow flower-shaped decorations in colors such as decoration. - Text: Located at the center top left side, there are multiple layers of colorful flowers arranged along [\dots] \\[3pt]
  \raisebox{-\height}{\rimg[width=0.145\textwidth]{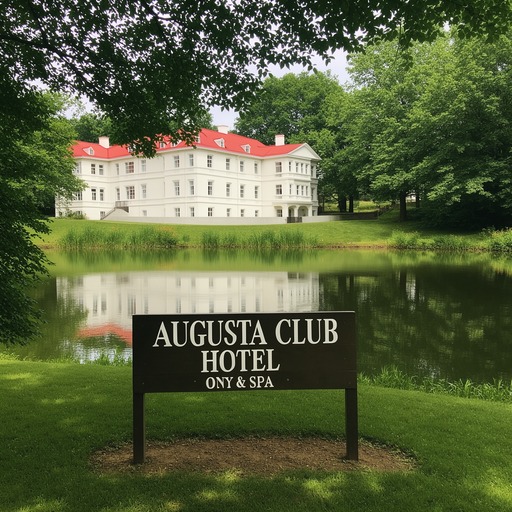}} & \raisebox{-\height}{\rimg[width=0.145\textwidth]{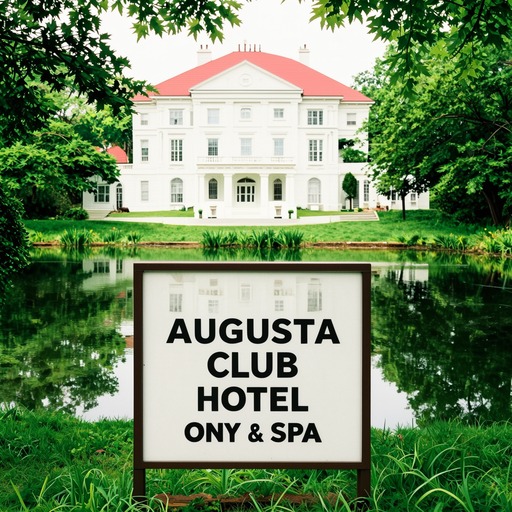}} & \raisebox{-\height}{\rimg[width=0.145\textwidth]{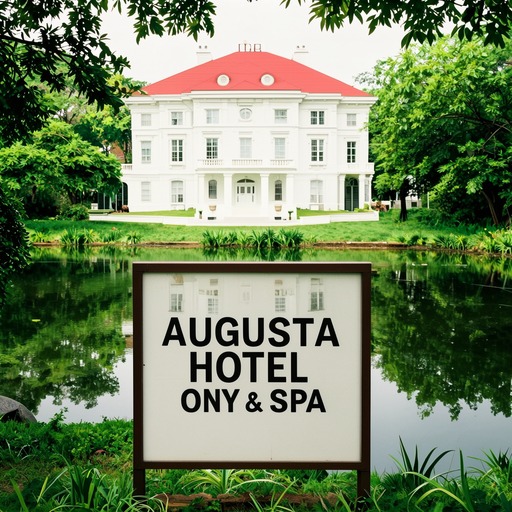}} & \vspace{0pt}\scriptsize\qplabel{PROMPT} {\color{qmute}\itshape A sign reading "AUGUSTA CLUB HOTEL $\cdot$ ONY \& SPA" stands prominently in front of a serene pond reflecting a large, elegant white building [\dots]}\newline\qdlabel{DESCRIPTION} a red brick building with white text at the top, and two-story structure (one on each side), one is an arched entrance. The house has three-dimensional relief-style architectural structures; in front of the roof are visible: 10-35 o'clock, square-shaped decorative elements inside which there is a rectangular window frame containing several small windows lights. 2. A large number of dark-colored [\dots] \\[3pt]
  \end{tabular}\end{adjustbox}\end{qualband}
  \caption{\textbf{FLUX-Reason-6M (out-of-domain) prompts with decoded descriptions}, models trained on our corpus (same seed).}
  \label{fig:supp_ood}
\end{figure*}

\Cref{fig:supp_ood} repeats the held-out examples of \cref{sec:supp_readout} on prompts of FLUX-Reason-6M
\cite{fang2025flux}, which our corpus does not contain and whose style differs from it: long, often imaginative captions.
All models are the ones trained on our corpus (as in the rest of the paper), not the FLUX-Reason runs of
\cref{sec:supp_open}; the prompts are held out from those runs. We show examples of both backbones. Out of domain the descriptions
still name the subject and some attributes (a moss-covered fairy house, a woman with long dark hair, a child in a white
lace dress, a striped bee with semi-transparent wings, roses embroidered on a skull patch, a building with an arched
entrance) but drift sooner.

\subsection{Direct read-out against the decoder}
\label{sec:supp_direct}
\Cref{fig:supp_direct} decodes held-out prompts with the two read-outs of \cref{sec:abl_direct} of the main paper. Both
streams are sampled jointly with the image from noise (10 steps, guidance 4, seed $4321{+}j$ for sample $j$); the
decoder of \method decodes greedily (repetition penalty 1.3, no repeated 3-grams), and the direct mapping takes the most
likely token at every stream position. The image is the one \method draws with its description; the direct-read-out
model draws its own image, which is not shown. The direct read-out recovers content words (a turtle's round head and
claws, a young woman with flowers, a white building with windows), but not their order; the decoder writes sentences,
with errors in detail.

\paragraph{The words are there.} We read three streams with the direct mapping of \cref{sec:abl_direct} (stream trained
towards Qwen3-4B states, position $i$ read as token $i{+}1$) on $400$ held-out prompts (\cref{tab:supp_direct}): the
true stream, the Qwen3-4B encoding of the reference description, in which position $i$ holds the state of token $i$;
the generated stream, sampled jointly with the image; and a completely random stream drawn from $\mathcal{N}(0,I)$, the
state the sampler starts from. From the random stream the direct mapping produces no words of the reference (ROUGE-1
$2.8$). From the generated stream it recovers many of them: ROUGE-1 $28.9$, ten times the random stream and more than
half of the $50.2$ it reaches on the true stream. Their order is lost: ROUGE-2 falls from $17.3$ to $2.1$, BLEU-4 from
$11.5$ to $0.0$, and the perplexity of the reference rises from $17.0$ to $3{,}526$. The generated stream therefore
carries the content of the description, but the direct mapping cannot read it in order.

\begin{table}[t]
\caption{\textbf{Direct read-out}: the direct mapping reading a completely random stream ($\mathcal{N}(0,I)$), the
stream generated jointly with the image, and the true stream (the encoding of the reference description); $400$
held-out prompts, decoded text against the reference description.}
\label{tab:supp_direct}
\centering
\footnotesize
\setlength{\tabcolsep}{5pt}
\begin{tabular}{lccc}
\toprule
Direct mapping reads & Random & Generated & True \\
\midrule
ROUGE-1$\uparrow$ & $2.8$ & $28.9$ & $50.2$ \\
ROUGE-2$\uparrow$ & $0.0$ & $2.1$ & $17.3$ \\
ROUGE-L$\uparrow$ & $2.2$ & $17.1$ & $39.8$ \\
BLEU-4$\uparrow$ & $0.0$ & $0.0$ & $11.5$ \\
Perplexity$\downarrow$ & $806{,}484$ & $3{,}526$ & $17.0$ \\
\bottomrule
\end{tabular}
\end{table}

\paragraph{The content is in the wrong place.} To see why, we perturb the true stream in three ways and read it with
the direct mapping and with a language-model decoder, each trained on the same Qwen3-4B stream ($200$ held-out prompts,
perplexity of the reference; \cref{tab:supp_probe}). \emph{Shuffling} the stream within windows of $2$, $5$ or $10$
positions keeps its content and only moves it; \emph{shifting} the whole stream by one position moves every state to
its neighbour's slot; \emph{noise} adds $\mathcal{N}(0,I)$ scaled to the stream's own root-mean-square. The direct
mapping depends on position alone: swapping neighbouring states raises its perplexity from $16.7$ to $262$, and a shift
by one position to over $5{,}800$. Noise as large as the stream itself barely affects it ($21.3$). The decoder is
indifferent to all three: shuffled within up to ten positions or shifted, the stream reads at $25.3$ to $25.4$, as
unperturbed ($25.4$), and full noise only raises it to $27.6$. The generated stream differs from the true stream by $1.38$ times the stream's
root-mean-square, more than the largest noise we add, yet that noise leaves both read-outs near their ceiling while the
generated stream raises the direct mapping's perplexity to $3{,}467$. Its error is therefore not noise around the right
states: it is content in the wrong positions. Shuffling the generated stream does not hurt the direct mapping any further
($3{,}467$ to $3{,}645$), because its order is already lost, and no single shift recovers it ($2{,}851$ at the best shift
per prompt). The decoder reads the same kind of generated stream at $34.3$, and the decoder of \method reads its own
generated T5 stream at $42.0$ with the same indifference to shuffling and shifting ($12.7$ on the true stream, $12.8$
shuffled within ten positions or shifted by one).

\begin{table}[t]
\caption{\textbf{Reading a perturbed stream}: perplexity of the reference description read by the direct mapping and by
a Qwen3-4B decoder, both trained on the Qwen3-4B stream, $200$ held-out prompts. Noise is scaled to the stream's
root-mean-square.}
\label{tab:supp_probe}
\centering
\footnotesize
\setlength{\tabcolsep}{4pt}
\begin{adjustbox}{max width=\linewidth}
\begin{tabular}{lcc}
\toprule
Stream & Direct mapping & Decoder \\
\midrule
True & $16.7$ & $25.4$ \\
\quad shuffled within 2 / 5 / 10 positions & $262$ / $2{,}432$ / $7{,}268$ & $25.4$ / $25.3$ / $25.4$ \\
\quad shifted by $-1$ / $+1$ position & $7{,}194$ / $5{,}867$ & $25.3$ / $25.4$ \\
\quad noise $0.25\times$ / $1\times$ & $17.0$ / $21.3$ & $25.4$ / $27.6$ \\
Generated & $3{,}467$ & $34.3$ \\
\quad shuffled within 10 positions & $4{,}037$ & $34.4$ \\
\quad best shift per prompt ($-3$ to $+3$) & $2{,}851$ & $34.0$ \\
\bottomrule
\end{tabular}
\end{adjustbox}
\end{table}

\paragraph{Why.} The stream is generated like the image: all positions are denoised together, in parallel, over a few
sampling steps. Nothing makes position $i$ commit to a word given the word at position $i{-}1$, so where in the stream a
piece of content lands, and how two plausible orderings mix, is left open. Read position by position, such a stream
gives the failure known from non-autoregressive text generation \cite{gu2018nat} and from rounding continuous diffusion
language models position by position \cite{shen2026codar}: the right words, out of order, with
frequent words repeated (\emph{with with with}, \emph{black black black}; \cref{fig:supp_direct}). The decoder of \method does not need the
content at a fixed position. It attends over the whole stream, wherever the content lies, and writes autoregressively,
each token conditioned on those it has already written, so it commits to one ordering and keeps to it. This is also
why a light pre-trained language-model layer is enough to read the stream (\cref{sec:supp_dec}): it needs to arrange
the content, not to locate it.

\begin{figure*}[!tp]
  \centering
  \setlength{\tabcolsep}{3pt}
  \begin{qualband}{SD3.5 Medium}\begin{adjustbox}{max width=\linewidth}%
  \begin{tabular}{@{}cp{0.40\textwidth}p{0.40\textwidth}@{}}
  \multicolumn{2}{c}{\scriptsize\textcolor{sdcol}{\textbf{Qwen3-4B decoder (\method)}}} & \multicolumn{1}{c}{\scriptsize\textbf{Direct mapping}} \\[3pt]
  \multicolumn{3}{@{}p{0.98\textwidth}@{}}{\scriptsize\qplabel{PROMPT} {\color{qmute}\itshape 6D little turtle with sunglasses, anthropomorphic, cute and adorable expression, background is a green tunnel}} \\[2pt]
  \raisebox{-\height}{\rimg[width=0.13\textwidth]{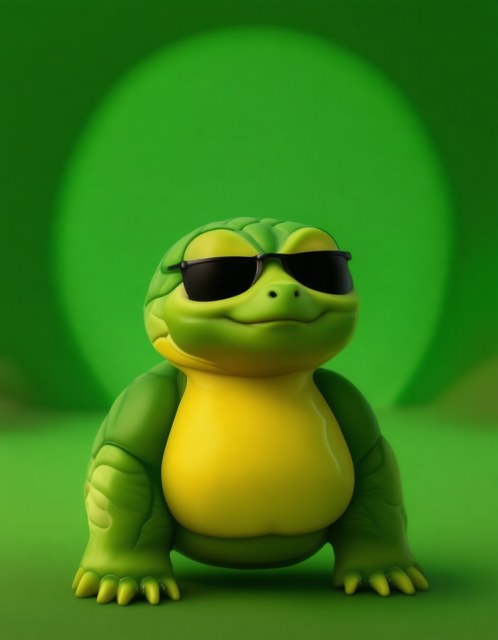}} & \vspace{0pt}\scriptsize A cartoon-style animal: 1, with a white face and black eyes; nose is light yellow in color. The head has two large ears (black pupils), orange circular eye shape of the abdomen are gray-white beak, green mouth open revealing teeth on each side); wearing dark blue shorts, round cap as if back to both sides.. & \vspace{0pt}\scriptsize A1omorphic1 head body skin head, round round, with with with head round roundak round large with with featuresises mouth round body two green and large round round large large body round on roundak round with round round two yellow eye with nostr paw large and and eyes. no features-like two circular, claw claw. round with and round and and eye orangeera head claw pupils round. on tail round-like two claw.. \dots \\[7pt]
  \multicolumn{3}{@{}p{0.98\textwidth}@{}}{\scriptsize\qplabel{PROMPT} {\color{qmute}\itshape The Light Cascade of the Greenhouse: The main content features a sunlit plant greenhouse with a vibrant young girl. She is wearing a light yellow dress, adorned with a wide-brimmed grass hat decorated with slender grass leaves, and carrying a\dots}} \\[2pt]
  \raisebox{-\height}{\rimg[width=0.13\textwidth]{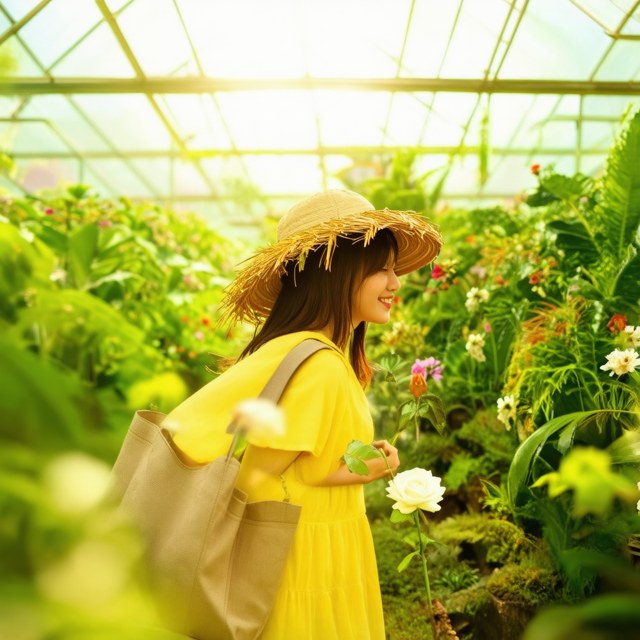}} & \vspace{0pt}\scriptsize Young woman: 1, with a light beige long hair (flowing to the side of the picture), fair skin; wearing white off-shoulder dress in pink and yellow flowers. Holding an umbrella-shaped bouquet made up from bright green leaves are visible on each hand's left ear is holding a flower basket containing small flowers at her back, right hand naturally hanging down, facing away from it.. & \vspace{0pt}\scriptsize Young: the flowers young woman a fair soft white on top flowers a and beige hair,,,, orange white, the long brown, wearing is light green a her, white a long-s light kn a with, flowerneck a top white with bare, pink, with a ple her scarf wearing young white flowers silk boy, flower, sitting sideways a. a; a bow slender kn,, right bobold flowerlee, on,, kn.,, a of,flower woman, with \dots \\[7pt]
  \multicolumn{3}{@{}p{0.98\textwidth}@{}}{\scriptsize\qplabel{PROMPT} {\color{qmute}\itshape A DJ wearing a blue T-shirt adjusting turntables and faders on a Pioneer equipment setup on a beach at sunset with palm trees and buildings in the background, illuminated by strong backlight and lens flare, standing to the right of\dots}} \\[2pt]
  \raisebox{-\height}{\rimg[width=0.13\textwidth]{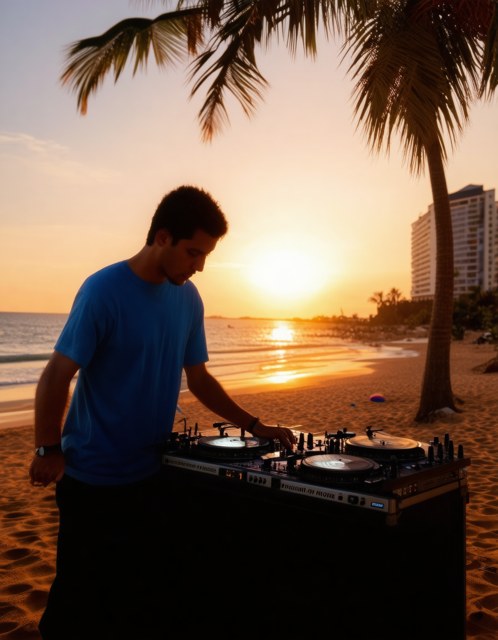}} & \vspace{0pt}\scriptsize Young male: 1, short black hair (black), dark skin tone. Wearing a blue hoodie and denim shorts on the desk of his hands in front; holding an electric guitar device with both hands visible at each hand pressing against him. & \vspace{0pt}\scriptsize Female sportsA male the build, with suit male wearing short, sports,blue, short wearing,-length a sports sportsble a a backpack black is backpack black black sportsblack Electric sports computer wearing sports suit sports black black black, black black tank,epad camera. holding,with. \\[7pt]
  \multicolumn{3}{@{}p{0.98\textwidth}@{}}{\scriptsize\qplabel{PROMPT} {\color{qmute}\itshape A hotel in Miami called KMP}} \\[2pt]
  \raisebox{-\height}{\rimg[width=0.13\textwidth]{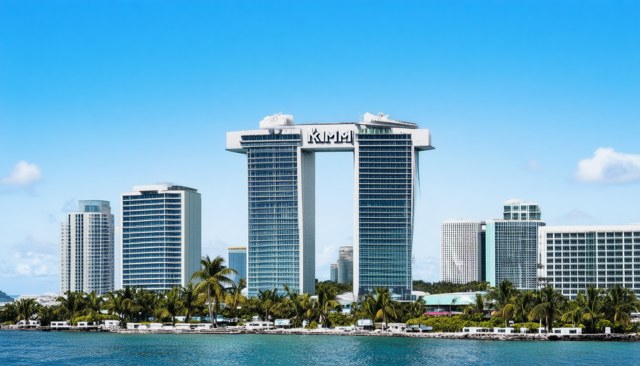}} & \vspace{0pt}\scriptsize The main structure is a white vertical strip-like building, with the interior and an outdoor facade. The first floor are visible in the foreground of the frame: on top (the left side) has a high-rise buildings or glass curtain wall; below the right edge of the image contains multiple tall towers as street lamps. - Floor - Top 10 to 3D245 people at the bottom), one located above each other \dots & \vspace{0pt}\scriptsize The :: building,:-story in with-white white-story tower with: windows, a blue), the the and the building-body with roof, deep tall white. building building at white. the a center approximately pink white building building the In,, Theat and. white. building building skys the high locatedcrap white modern modern arranged: on street a is window street city buildings structure1 building:, the the,ighthouse modern the skys a structure in building the Building building \dots \\[7pt]
  \end{tabular}\end{adjustbox}\end{qualband}
  \caption{\textbf{Decoder against direct read-out}, held-out prompts (same seed per row). The image is the one \method
  draws with its description; the direct-read-out model draws its own image, not shown. Text unedited except that the
  trailing Chinese full stop is printed as a period and long outputs are cut.}
  \label{fig:supp_direct}
\end{figure*}

\subsection{Ablating the read-out decoder}
\label{sec:supp_dec}
The stream is trained towards T5-XXL encoder states and read by the last layer of Qwen3-4B. We replace the read-out by
the last layer of the larger Qwen3-8B and by the T5 v1.1 XXL decoder, which is pre-trained to read exactly these
states (\cref{tab:abl_dec}). A larger or T5-native read-out reads the stream back better (perplexity $27.9$ and
$22.3$ against $46.3$), but the image gets worse (GenEval $73.6$ and $72.3$
against $74.3$). With a lighter read-out the backbone itself has to make its states carry the language, and the image
backbone benefits from learning to process it.

\begin{table}[H]
\caption{\textbf{Read-out decoder.} In-domain read-back: perplexity per token (each read-out's own tokens).}
\label{tab:abl_dec}
\centering
\footnotesize
\setlength{\tabcolsep}{4pt}
\begin{tabular}{lcc}
\toprule
Read-out & GenEval & ppl$\downarrow$ \\
\midrule
\rowcolor{sdours}Qwen3-4B last layer (\textbf{\method}) & \textbf{74.3} & 46.3 \\
Qwen3-8B last layer & 73.6 & 27.9 \\
T5-XXL decoder & 72.3 & \textbf{22.3} \\
\bottomrule
\end{tabular}
\end{table}

\begin{figure*}[p]
  \centering
  \setlength{\tabcolsep}{2pt}
  \begin{qualband}{SD3.5 Medium}\begin{adjustbox}{max width=\linewidth}%
  \begin{tabular}{@{}cp{0.21\textwidth}cp{0.21\textwidth}cp{0.21\textwidth}@{}}
  \multicolumn{2}{c}{\scriptsize\textcolor{sdcol}{\textbf{Qwen3-4B (\method)}}} & \multicolumn{2}{c}{\scriptsize\textbf{Qwen3-8B}} & \multicolumn{2}{c}{\scriptsize\textbf{T5-XXL decoder}} \\[3pt]
  \multicolumn{6}{@{}p{0.98\textwidth}@{}}{\scriptsize\qplabel{PROMPT} {\color{qmute}\itshape an adorable plush pumpkin monster, fluffy and huggable, cute big eyes}} \\[2pt]
  \raisebox{-\height}{\rimg[width=0.11\textwidth]{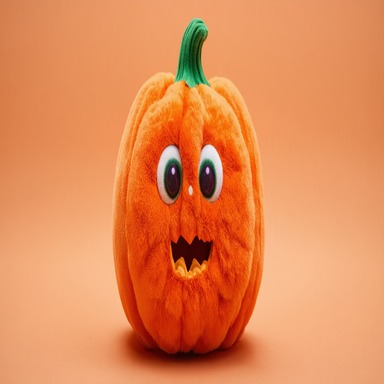}} & \vspace{0pt}\scriptsize A anthropomorphic cartoon character: the head is a black, with an orange and white fur on its surface; two large round eyes (black pupils), pink nose, mouth slightly open revealing sharp teeth. The body has dark green color blocks of red in yellow-brown from inside to be made up of light brown-black or rock-like texture. - Head: A small dog's face shape, pointed ears are visible at each end); wearing \dots & \raisebox{-\height}{\rimg[width=0.11\textwidth]{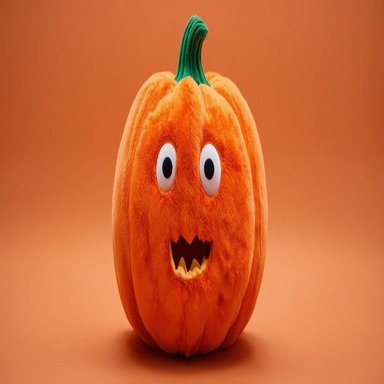}} & \vspace{0pt}\scriptsize A cartoon-style small round in shape, with a \dok{rounded body}. The character is located at the center of an \dok{orange spherical head} (a large and \dbad{two circular paws}), covered entirely made up of \dbad{light beige plush material}; on both sides are visible: \dbad{dark short hair} slightly fluffy fur around its neck, wearing a \dbad{black top hat} adorned with multiple pockets or knitted cap-like decorations on their surfaces. 2. A transparent \dots & \raisebox{-\height}{\rimg[width=0.11\textwidth]{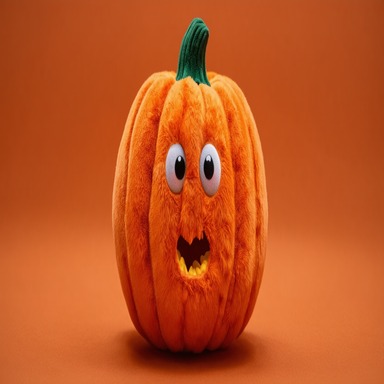}} & \vspace{0pt}\scriptsize A cartoon-style \dok{stuffed animal}: the main subject is an exaggerated version of the "scroll" \dbad{motif}, with a \dok{round shape} and a cylindrical body. The surface shows a mixture of \dok{orange plush texture} and yellowish fabric; there are \dbad{two black buttons on top of each eye} (one visible at the center), and \dbad{three white circular eyes}, all with \dbad{golden or bright yellow enamel dots} as their surfaces. In addition to these \dots \\[7pt]
  \multicolumn{6}{@{}p{0.98\textwidth}@{}}{\scriptsize\qplabel{PROMPT} {\color{qmute}\itshape Lion resting cartoon, lion resting near river}} \\[2pt]
  \raisebox{-\height}{\rimg[width=0.11\textwidth]{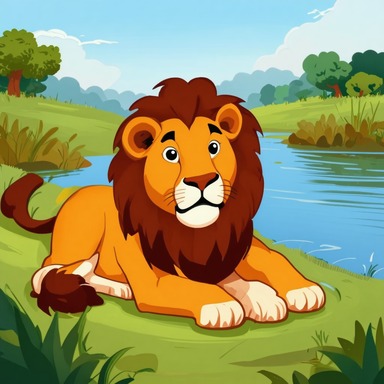}} & \vspace{0pt}\scriptsize A dog: the tiger's body is in a cat. The lion of its mane and tail are present on each side, with two front limbs (one one orange) visible at the end; hind paws clearly defined as an S-shapezman or animal area. Golden horse has dark brown stripes, fur color being recognizable 1.. & \raisebox{-\height}{\rimg[width=0.11\textwidth]{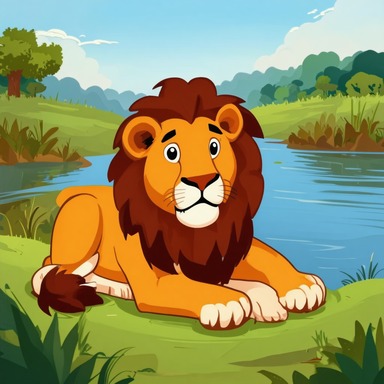}} & \vspace{0pt}\scriptsize \dbad{A large brown bear}: medium-sized, with a fluffy fur (paws) on the front of its head; two hind paws are visible. \dbad{The cat's tail} is in an inverted V shape and round body shapes \dbad{resembling tiger stripes} or animal patterns), \dok{orange-brown mane} \dbad{standing posture}, limbs naturally bent joints, \dok{no extra animals present}. & \raisebox{-\height}{\rimg[width=0.11\textwidth]{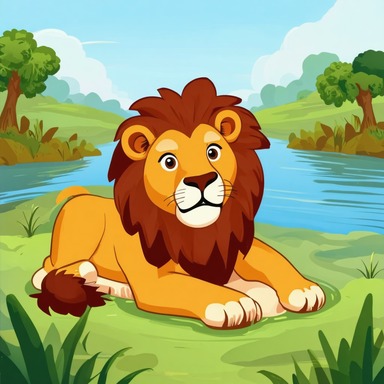}} & \vspace{0pt}\scriptsize A slender and distinct one: an adult, with \dbad{dark brown short hair} (brown), \dbad{feathered neck}, ears, and face; a facial expression is a \dok{mixture of orange and yellow}. The main body is arranged in a horizontally split style, including a top-of-the-head light brown fur, a head covered by \dbad{two black bows}, and \dbad{pinkish-brown legs}. It also has the following features: it is surrounded by a \dbad{circular pattern of running feet}.. \\[7pt]
  \multicolumn{6}{@{}p{0.98\textwidth}@{}}{\scriptsize\qplabel{PROMPT} {\color{qmute}\itshape Grigori Rasputin next to a noblewoman}} \\[2pt]
  \raisebox{-\height}{\rimg[width=0.11\textwidth]{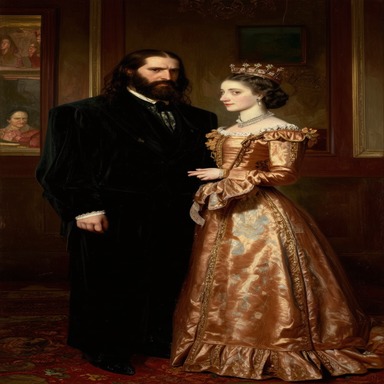}} & \vspace{0pt}\scriptsize The main subject is a female figure, with dark hair and beard; wearing a white base of deep red velvet robe (with gold trim on the inner side), long black curly hair. He has golden crown adorned with patterns in his right hand: an elderly woman's headband, holding a silver-white pearl necklace around her waist, eyes looking at the center of the frame. In front of them are seated beside \dots & \raisebox{-\height}{\rimg[width=0.11\textwidth]{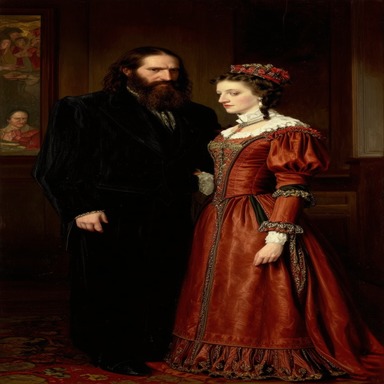}} & \vspace{0pt}\scriptsize \dbad{An elderly woman}: with a \dbad{white hair}, pale face covered by the head and shoulders. Facial features include black eyes, straight nose bridge; slightly parted lips in a natural expression of calm (a), \dok{fair skin tone}, wearing an ancient-style \dok{long robe} or dark patterned clothing underneath). Wearing a \dbad{light-colored high-collared shirt}, holding a \dbad{red silk-like object} on his left hand to support the right side is visible behind her figure's \dots & \raisebox{-\height}{\rimg[width=0.11\textwidth]{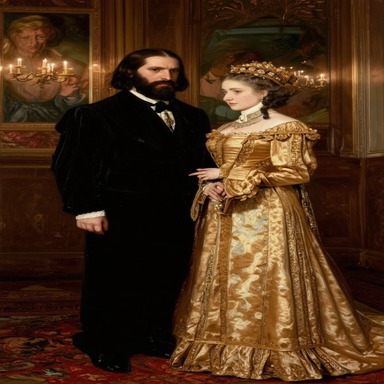}} & \vspace{0pt}\scriptsize A short portrait of a \dbad{female samurai}: The subject is an \dbad{Asian male with long hair}, wearing a \dbad{black-and-red striped garment} and accessories. In the background, there are \dok{two people in the scene}, one on the left and one behind; both have distinct gazes but their eyes are not visible. The subject has a \dok{dark brown curly hairstyle}, clear facial features, and a \dbad{thin metal frame around his eyes}. He \dots \\[7pt]
  \multicolumn{6}{@{}p{0.98\textwidth}@{}}{\scriptsize\qplabel{PROMPT} {\color{qmute}\itshape Young attractive woman with wavy brown hair and a dangerous handsome mafia man, romantic embrace, romance cover, romantic city background}} \\[2pt]
  \raisebox{-\height}{\rimg[width=0.11\textwidth]{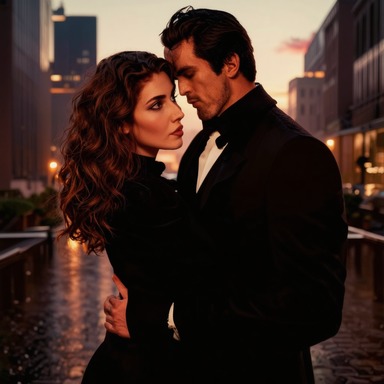}} & \vspace{0pt}\scriptsize Young woman: 1, with a black long hair (black), dark eyes and straight nose bridge; wearing an olive green suit jacket. In the background on the left is a male in his right hand) are naturally hanging down at the bottom of the picture, there are two women's men. Male: facing away from the camera directly to the man has deep brown curly hair, top hat, looking toward the back \dots & \raisebox{-\height}{\rimg[width=0.11\textwidth]{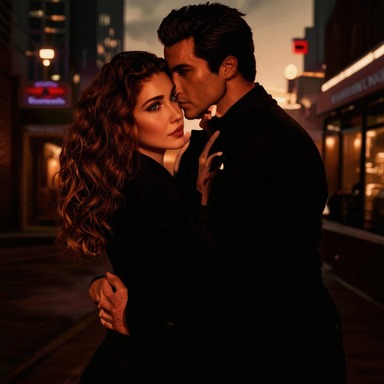}} & \vspace{0pt}\scriptsize \dok{A young male}: \dbad{Asian face}, \dok{short hair (black)}, wearing a \dbad{black high-necked knitted sweater}; the man's left shoulder and right hands are visible. The woman is in \dbad{dark skin tone with freckles} on his chin, looking at her gaze forward to \dbad{hold a watch} or mouth.. & \raisebox{-\height}{\rimg[width=0.11\textwidth]{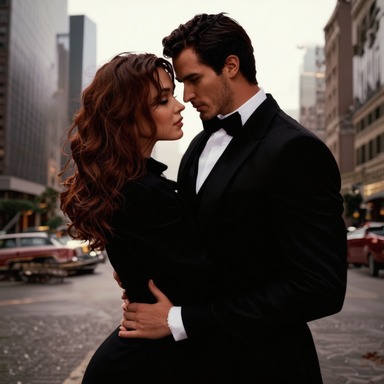}} & \vspace{0pt}\scriptsize Young woman: Approximately 20 years old, with \dok{long dark hair} (complimentary to the left side of her face), clear facial features, no obvious expression; wearing a black suit jacket and a \dbad{white knitted sweater}. She is \dbad{sitting on a dark-colored couch in front of a window}. Her hands are naturally placed in front, but not visible as part of 'sleeveless' clothing. In the background, there is a blurred image of \dots \\[7pt]
  \end{tabular}\end{adjustbox}\end{qualband}
  \caption{\textbf{Read-out decoders, held-out prompts (1/2)} (same seed per row). \dok{Blue}: differs from Qwen3-4B and is correct for that model's image; \dbad{red}: differs and is wrong.}
  \label{fig:supp_dec1}
\end{figure*}

\begin{figure*}[p]
  \centering
  \setlength{\tabcolsep}{2pt}
  \begin{qualband}{SD3.5 Medium}\begin{adjustbox}{max width=\linewidth}%
  \begin{tabular}{@{}cp{0.21\textwidth}cp{0.21\textwidth}cp{0.21\textwidth}@{}}
  \multicolumn{2}{c}{\scriptsize\textcolor{sdcol}{\textbf{Qwen3-4B (\method)}}} & \multicolumn{2}{c}{\scriptsize\textbf{Qwen3-8B}} & \multicolumn{2}{c}{\scriptsize\textbf{T5-XXL decoder}} \\[3pt]
  \multicolumn{6}{@{}p{0.98\textwidth}@{}}{\scriptsize\qplabel{PROMPT} {\color{qmute}\itshape Nezuko Kamado, shrunk down to adorable chibi proportions, beams brightly while perched atop the head of a colossal, happy Totoro. Sunlight dapples through the leaves of the camphor tree, illuminating Nezuko's pink\dots}} \\[2pt]
  \raisebox{-\height}{\rimg[width=0.11\textwidth]{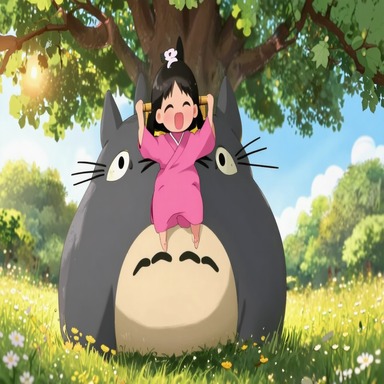}} & \vspace{0pt}\scriptsize Cartoon anthropomorphic cat: 1, with black and white fur on the face; large eyes (black pupils), pink nose, small mouth in a light gray-white cartoon character. Black eyebrows are short brown-paved monkey's head is holding an orange kitten-shaped object of cloud-like creature (with two paws) at the top. - Head: A yellow rabbit cartcat to dark grey-black hair, wearing a green fox hat that has red-orange snowflake patterns underneath, both \dots & \raisebox{-\height}{\rimg[width=0.11\textwidth]{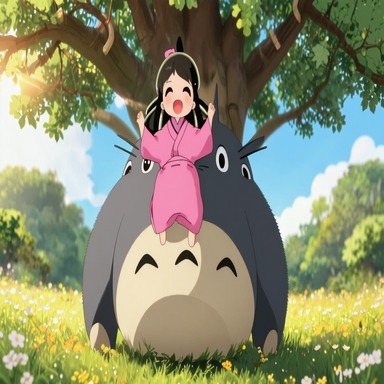}} & \vspace{0pt}\scriptsize A anthropomorphic cartoon character (1): \dok{round body}, large in size and \dok{fluffy fur}. Appearance features: wearing a \dbad{light gray-brown long coat} with white spots on the back of its head; facial expression is calm, slightly plump figure, \dok{sitting posture}. Cartoon Bear - \dbad{A small dog} located at the center of the picture to be a cat's tail behind it.. & \raisebox{-\height}{\rimg[width=0.11\textwidth]{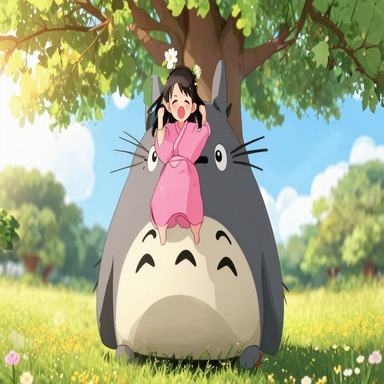}} & \vspace{0pt}\scriptsize \dbad{Young male}: one person, with a small and simple facial appearance; \dok{black hair}, in a \dok{bun or ponytail style}, with clear facial features. She wears a \dok{purple pink-gray (Pinkish) plum dress}, the top is covered by a \dbad{red cape/hoodie}, with an \dbad{orange scarf} around her neck, \dbad{holding a pair of dark brown boots} below. The background contains \dbad{two blue cartoon characters}: the left has a white face, with the right \dots \\[7pt]
  \multicolumn{6}{@{}p{0.98\textwidth}@{}}{\scriptsize\qplabel{PROMPT} {\color{qmute}\itshape Bitten cheeseburger on a plate}} \\[2pt]
  \raisebox{-\height}{\rimg[width=0.11\textwidth]{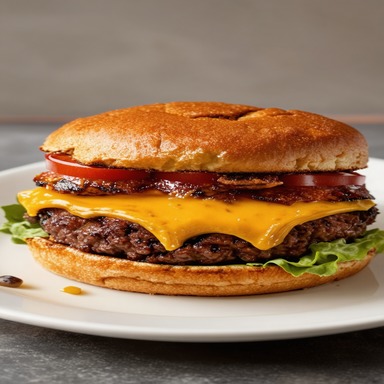}} & \vspace{0pt}\scriptsize A white pepper (with visible amount of the flesh); a sauce plate containing 1, with dark brown slices and yellowed meat; below is in an orange-red cheese: fresh green sesame seeds are scattered on top. The crust has black onion bread slice to be identifiable, including red tomato slices, plump body covered which have spots at the bottom), sprinkled sweat droplets or food traces inside. Includes: contains several chocolate lemon \dots & \raisebox{-\height}{\rimg[width=0.11\textwidth]{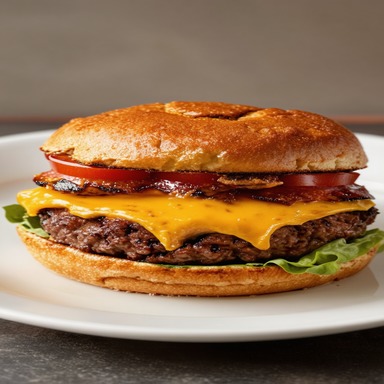}} & \vspace{0pt}\scriptsize \dbad{A bowl of cooked meat}: \dok{a tomato slice} (with dark brown sauce) and \dok{cheese slices}, with the surface showing distinct flesh texture; inside contains food is fresh green pepper granular seeds on top. The bread has \dbad{dried chocolate pieces} are cut off-white in color from yellow to orange-yellow skin or white sesame kernels (suspected to be garlic), covered entirely irregularly torn potato slices); visible \dok{lettuce leaves} (1 piece): made up \dots & \raisebox{-\height}{\rimg[width=0.11\textwidth]{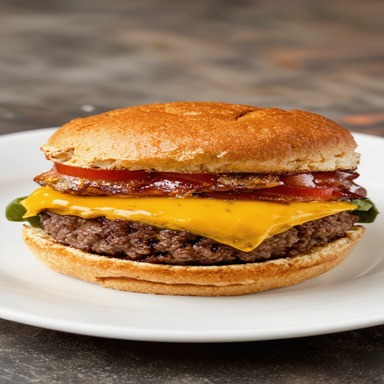}} & \vspace{0pt}\scriptsize A smashed yellow-brown meat, with a deep red glaze on the surface, featuring \dbad{black watermarks} and \dbad{a white circular hole for a straw}. The flesh is covered with an irregular brownish-yellow sticky web, including multiple layers of dark gray and black peppercorns; the shell has a \dbad{golden honeycomb} embedded in it, along with \dok{melted cheese}, and there are several \dbad{small white feathers} (on top of each egg) as well as \dots \\[7pt]
  \multicolumn{6}{@{}p{0.98\textwidth}@{}}{\scriptsize\qplabel{PROMPT} {\color{qmute}\itshape A dragon in a heroic pose with geometric and mechanical motifs, in black and white line work with stippling, to serve as a back tattoo design, avant-garde style with minimal ink splatters and geometric fragments}} \\[2pt]
  \raisebox{-\height}{\rimg[width=0.11\textwidth]{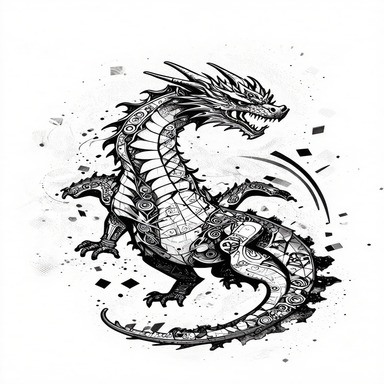}} & \vspace{0pt}\scriptsize A anthropomorphic dragon: the head, with a tail of sharp and horn-like mane. The horse is in an exaggerated animal image composed of multiple horns (with two pointed claws), each clawed wings extending from its back to be made up of black scales; the mouth has long whiskers, also covered by thick furrowed limbs extended forward, facing upward. The main subject as the muzzle. Each creature's front limb structures are \dots & \raisebox{-\height}{\rimg[width=0.11\textwidth]{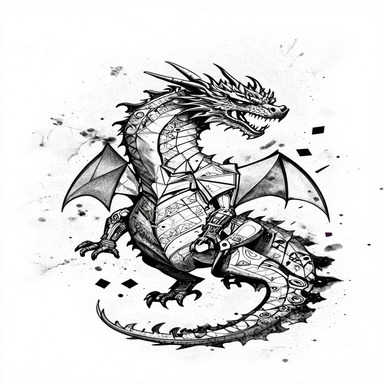}} & \vspace{0pt}\scriptsize Dragon: A large in size, with a sharp horn on the head and back. The dragon's body is covered by an \dbad{eagle or skull-like tail fin}; its \dok{wings are spread out} to both sides of the horns (similar from the wing), each claw has two black pointed snout at the end of the sword, there are \dok{three-dimensional mechanical structures} resembling a snake creature that extends backward-curving forward-winged into a complex \dots & \raisebox{-\height}{\rimg[width=0.11\textwidth]{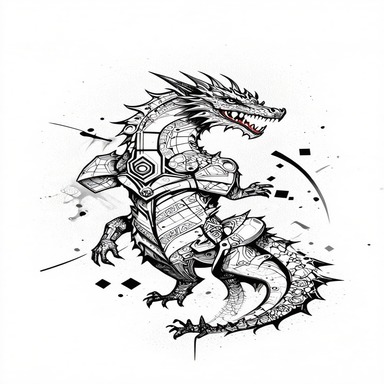}} & \vspace{0pt}\scriptsize \dbad{A bald dragon}: the surface of the dragon's body is covered with gray-white scales, and it has an \dok{intricate pattern on its face}. The head displays a black outline, with white spots for eyes, and a skull featuring \dbad{purple-red eye patterns}. \dbad{Its wings are spread out in multiple layers}, including one at the base of the spine, which contains a \dok{spiraling geometric design}. On top of the Dragon's shoulders, there \dots \\[7pt]
  \multicolumn{6}{@{}p{0.98\textwidth}@{}}{\scriptsize\qplabel{PROMPT} {\color{qmute}\itshape Vibrant colors, healing style, warm atmosphere, asymmetrical composition, perfect details, masterpiece. High-quality details, best quality, high saturation, contrast of light and shadow, cinematic texture, smooth 3D\dots}} \\[2pt]
  \raisebox{-\height}{\rimg[width=0.11\textwidth]{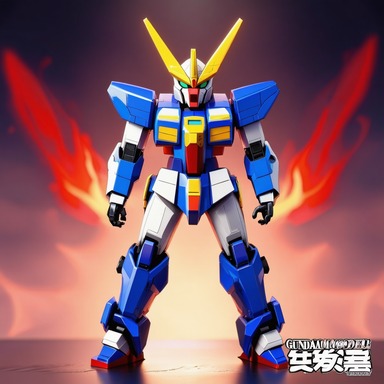}} & \vspace{0pt}\scriptsize A white robot: the main body is a silver helmet, with red and yellow wings. The overall design has an orange mechanical armor featuring golden circuit patterns (with two circular holes), in blue visor; below it includes multiple horns on both sides of the chest plate arm guards are visible at the joints. Each character has sharp wing structures located to the right side of the head features a shield-shaped \dots & \raisebox{-\height}{\rimg[width=0.11\textwidth]{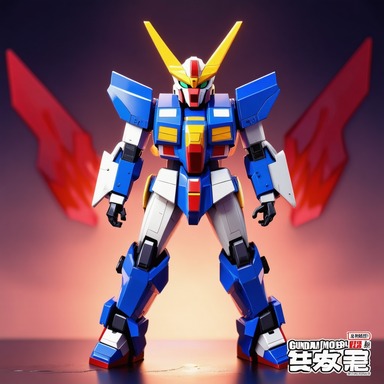}} & \vspace{0pt}\scriptsize A complex mechanical device: the main body is a \dbad{golden rectangular structure}, with an orange base featuring red and gold color scheme. The overall design has a \dok{metallic texture} (with two silver-gray metal rivets on each side), its head shaped like a large number of \dok{futuristic-style helmet}; it wears \dok{multiple armor plates} in \dbad{yellow-green spacesuit} or antenna-like structures at the joints). A cylindrical robot-shaped weapon system as if there are \dots & \raisebox{-\height}{\rimg[width=0.11\textwidth]{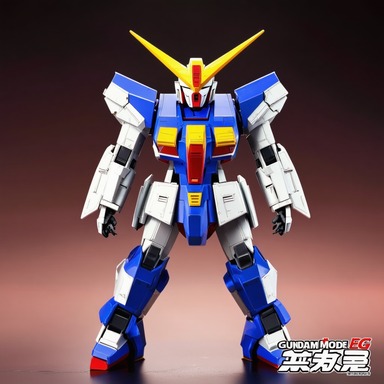}} & \vspace{0pt}\scriptsize \dbad{A 3D animated GIF}: The main subject is an \dbad{inflatable} white "S"-shaped robot, with a gray texture in its surface. Its upper body (the top) has a blue faceplate and \dbad{yellow fist}; the bottom of the middle unit also features a bright orange ring, which shows a clear internal structure. On both sides of the robot are \dbad{two golden gears}, each featuring a circular silver plate, as well as black \dots \\[7pt]
  \end{tabular}\end{adjustbox}\end{qualband}
  \caption{\textbf{Read-out decoders, held-out prompts (2/2)} (same seed per row). \dok{Blue}: differs from Qwen3-4B and is correct for that model's image; \dbad{red}: differs and is wrong.}
  \label{fig:supp_dec2}
\end{figure*}

Each read-out is its own training run with everything else as in \method. \Cref{fig:supp_dec1,fig:supp_dec2}
compare the three on held-out prompts. All three
use the protocol of \cref{sec:supp_readout}: the description stream is sampled jointly with the image from noise
(28 steps, guidance 4, the same seed for the three models), and the read-out decodes it greedily (repetition penalty 1.3,
no repeated 3-grams). Descriptions are unedited except that the trailing Chinese full stop of our description format is
printed as a period and long ones are cut.
Both alternatives lower GenEval mostly on position ($28.5$ and $25.3$ against $32.5$) and colour attribution ($64.5$
and $62.0$ against $67.0$; \cref{tab:supp_abl_comp}). All three read-outs name the subject and its main attributes, and
the T5 decoder writes the longest descriptions (97 words on average, against 69 for \method and 74 for Qwen3-8B). The
Qwen3-8B read-out is not more faithful than the smaller one: it names a bear for a lion and puts a top hat on a plush pumpkin. The T5 decoder also invents the most, such as straw holes and feathers on a cheeseburger, a samurai next to Rasputin or a
second pair of eyes on the pumpkin; its errors are fluent sentences, where those of the Qwen read-outs are more often
broken phrases.

The read-out decoder ablation was run on SD3.5 only, so these figures have no FLUX.2 [klein] counterpart.

\section{Further ablations}
\label{sec:supp_g_abl}
Components per GenEval category, the read-out loss, where to add capacity, attention to the description at inference and the image-loss weight.

\subsection{Component ablation by GenEval category}
\label{sec:supp_abl_comp}
\Cref{tab:supp_abl_comp} gives the per-category GenEval scores of the component ablation of the main paper, which
reports the overall score only.

\begin{table}[H]
\caption{\textbf{Component ablation by GenEval category} (stream off). Each row changes one component of \method.}
\label{tab:supp_abl_comp}
\centering
\footnotesize
\setlength{\tabcolsep}{3pt}
\resizebox{\linewidth}{!}{%
\begin{tabular}{lccccccc}
\toprule
Training & Overall & 1 obj & 2 obj & Count & Col. & Pos. & Attr. \\
\midrule
SFT control (no stream) & 73.4 & 99.1 & 90.9 & 72.8 & 84.8 & 30.5 & 62.3 \\
\midrule
\rowcolor{sdours}\textbf{\method} & 74.3 & 99.7 & 89.4 & 71.6 & 85.6 & 32.5 & 67.0 \\
\midrule
no description flow loss (read-out loss only) & 74.1 & 99.1 & 90.4 & 72.5 & 86.7 & 28.8 & 67.2 \\
no text-only MLPs & 72.0 & 99.7 & 87.4 & 71.9 & 87.5 & 23.2 & 62.2 \\
text-only MLPs as LoRA (rank 32) & 73.7 & 99.4 & 91.2 & 75.3 & 84.6 & 27.8 & 64.2 \\
full attention (image and prompt read the stream) & 73.1 & 99.4 & 89.1 & 73.1 & 84.0 & 29.2 & 63.8 \\
\midrule
read-out: Qwen3-8B last layer & 73.6 & 99.4 & 89.6 & 73.4 & 85.9 & 28.5 & 64.5 \\
read-out: T5-XXL decoder & 72.3 & 99.4 & 88.9 & 70.0 & 88.3 & 25.3 & 62.0 \\
\bottomrule
\end{tabular}}
\end{table}

\subsection{Removing the read-out loss}
\label{sec:supp_abl_noce}
Row (c) of \cref{tab:abl_comp} removes the description flow loss and keeps the read-out cross-entropy. \Cref{tab:supp_abl_noce}
gives the converse ablation: \method without the read-out cross-entropy, so the description stream is trained by flow
matching alone, with everything else unchanged (same SFT checkpoint at step 800, seed, data, image-loss weight and schedule;
the read-out still exists but receives no gradient, and the stream is off at inference). On our corpus it was trained on our
local cluster (H200, like the SFT control; the \method row was trained on GH200, \cref{sec:supp_train}) on 4 GPUs with half
the gradient accumulation, so the global batch is unchanged. It reaches $73.2$ GenEval, below the SFT control ($73.4$) and
$1.1$ below \method ($74.3$). In the categories our descriptions state most often, colour attribution keeps part of the
gain ($64.5$, against $62.3$ for the control and $67.0$ for \method) and position falls below the control ($26.8$, against
$30.5$ and $32.5$). On
FLUX-Reason-6M (\cref{sec:supp_open}) the same ablation reaches $67.5$ (\cref{tab:supp_abl_noce_open}), level with the SFT control ($67.6$) and $2.6$ below
\method ($70.1$), with colour attribution at $52.5$ against $61.5$. Flow matching on the description tokens alone thus
does not lift the image: the gain needs the token-level read-out loss, in line with \cref{sec:supp_theory_write}, where a
regression lets the states hedge on an unresolved fact and the cross-entropy does not.

\begin{table}[H]
\caption{\textbf{Removing the read-out loss} (our corpus, GenEval, stream off). \method trained without the read-out
cross-entropy (description flow loss only). $^\dagger$4 GPUs with half the gradient accumulation, same global batch.}
\label{tab:supp_abl_noce}
\centering
\footnotesize
\setlength{\tabcolsep}{3pt}
\resizebox{\linewidth}{!}{%
\begin{tabular}{lccccccc}
\toprule
Training & Overall & 1 obj & 2 obj & Count & Col. & Pos. & Attr. \\
\midrule
SFT control (no stream) & 73.4 & 99.1 & \textbf{90.9} & 72.8 & 84.8 & 30.5 & 62.3 \\
\rowcolor{sdours}\textbf{\method} & \textbf{74.3} & 99.7 & 89.4 & 71.6 & 85.6 & \textbf{32.5} & \textbf{67.0} \\
no read-out loss$^\dagger$ & 73.2 & 99.7 & 88.6 & \textbf{73.4} & \textbf{86.4} & 26.8 & 64.5 \\
\bottomrule
\end{tabular}}
\end{table}

\begin{table}[H]
\caption{\textbf{Removing the read-out loss on FLUX-Reason-6M} (\cref{sec:supp_open}; SD3.5 Medium, GenEval, stream off).
$^\dagger$4 GPUs with half the gradient accumulation, same global batch.}
\label{tab:supp_abl_noce_open}
\centering
\footnotesize
\setlength{\tabcolsep}{3pt}
\resizebox{\linewidth}{!}{%
\begin{tabular}{lccccccc}
\toprule
Training & Overall & 1 obj & 2 obj & Count & Col. & Pos. & Attr. \\
\midrule
SFT control (no stream) & 67.6 & 98.4 & 81.6 & 61.9 & \textbf{86.2} & 23.5 & 54.3 \\
\rowcolor{sdours}\textbf{\method} & \textbf{70.1} & \textbf{100.0} & \textbf{82.1} & \textbf{67.5} & 84.6 & \textbf{25.3} & \textbf{61.5} \\
no read-out loss$^\dagger$ & 67.5 & 98.1 & 79.3 & 66.9 & 84.3 & 23.8 & 52.5 \\
\bottomrule
\end{tabular}}
\end{table}

\subsection{Where to add capacity}
\label{sec:supp_capacity}
\method trains the backbone with LoRA and adds new capacity only in the text-only MLPs (\cref{sec:method}). We test
two ways of giving it more: training every weight of the backbone on SD3.5, and adding text-only MLPs to every block of
FLUX.2 [klein] (\cref{tab:supp_capacity}). Each run changes only that, and its control changes the same thing.

\paragraph{Full fine-tuning (SD3.5).} We train the SFT control and \method with every backbone weight trainable in
place of LoRA, for the same 1800 steps. Both lose GenEval: the control drops from $73.4$ to $72.7$ and \method from
$74.3$ to $72.3$, now $0.4$ below its own control, mostly on position ($26.5$ against $32.5$ with LoRA) and two objects.
The text gains little: read-back perplexity is $44.8$ against $46.3$. A frozen backbone with low-rank updates is thus not what limits \method; the extra freedom
lets the image drift from the pre-trained model without making the description more faithful.

\paragraph{Text-only MLPs in every block (FLUX.2 [klein]).} \method adds text-only MLPs after the 5 double-stream
blocks only, the blocks with separate text weights. We also add one after each of the 20 single-stream blocks, each
initialised from the MLP of its block, which raises the new parameters from 566M to 2.89B. GenEval falls from $79.0$ to
$78.1$, mostly on counting ($78.1$ against $81.6$), and stays above the SFT control ($77.2$). The text-only MLPs of the
double-stream blocks are therefore enough, and five times more of them does not draw better.

\begin{table}[H]
\caption{\textbf{Where to add capacity} (GenEval, stream off). Read-back perplexity (in-domain, 28 steps) is for the
description \method writes; the SFT control writes none.}
\label{tab:supp_capacity}
\centering
\footnotesize
\setlength{\tabcolsep}{3pt}
\resizebox{\linewidth}{!}{%
\begin{tabular}{llccccccccc}
\toprule
& Training & New params & Overall & 1 obj & 2 obj & Count & Col. & Pos. & Attr. & ppl$\downarrow$ \\
\midrule
\multicolumn{11}{l}{\emph{SD3.5}} \\
(a) & SFT control, LoRA & -- & 73.4 & 99.1 & \textbf{90.9} & 72.8 & 84.8 & 30.5 & 62.3 & -- \\
\rowcolor{sdours}(b) & \textbf{\method}, LoRA & 597M & \textbf{74.3} & 99.7 & 89.4 & 71.6 & \textbf{85.6} & \textbf{32.5} & \textbf{67.0} & 46.3 \\
(c) & SFT control, full fine-tuning & -- & 72.7 & \textbf{100.0} & 87.6 & 69.1 & 82.7 & 30.3 & 66.3 & -- \\
(d) & \method, full fine-tuning & 597M & 72.3 & 99.4 & 85.6 & \textbf{74.4} & 82.7 & 26.5 & 65.3 & \textbf{44.8} \\
\midrule
\multicolumn{11}{l}{\emph{FLUX.2 [klein]}} \\
(e) & SFT control & -- & 77.2 & \textbf{99.4} & 87.4 & 80.9 & 89.4 & 42.5 & 63.5 & -- \\
\rowcolor{fluxours}(f) & \textbf{\method}, text-only MLPs in 5 double-stream blocks & 566M & \textbf{79.0} & \textbf{99.4} & \textbf{89.4} & \textbf{81.6} & \textbf{89.9} & \textbf{45.3} & \textbf{68.8} & 26.2 \\
(g) & \method, text-only MLPs in all 25 blocks & 2.89B & 78.1 & 98.4 & 88.9 & 78.1 & \textbf{89.9} & 44.8 & 68.5 & -- \\
\bottomrule
\end{tabular}}
\end{table}

\subsection{REPA hyperparameters}
\label{sec:supp_repa_sweep}
The REPA baselines of \cref{tab:abl_repa} align the prompt or the image tokens after block 8 to the frozen T5-XXL
features of the description with weight 0.5. To test whether they are held back by this choice, we change one of the two
for each source: an earlier block (4) at the same weight, and a larger weight (1.0) at the same block
(\cref{tab:supp_repa_sweep}). Everything else is as in \cref{tab:abl_repa}: the SFT control's checkpoint at step 800 as
the start, the same data order and 1800 steps in all. These runs use 2 GPUs with twice the gradient accumulation, so the
global batch (128) is unchanged.

No setting reaches the SFT control ($73.4$) or \method ($74.3$). The best, prompt tokens at block 4, scores $72.5$, $0.5$
above the paper's prompt row and $0.3$ below its image row; it gains on colour attribution ($68.3$, above \method's
$67.0$) but loses on counting and position. A larger weight on the prompt tokens lowers GenEval ($71.6$ against
$72.0$), and so does aligning the image tokens earlier or with a larger weight ($70.8$ and $71.9$ against $72.8$), mostly
on counting and position. Aligning to the description is thus not short of
\method because of the block or the weight; reading the description out of the model, rather than writing it, does not
give the same gain.

\begin{table}[H]
\caption{\textbf{REPA block and weight} (SD3.5, GenEval, stream off, 4 seeds). Rows (c) and (d) are the REPA rows of
\cref{tab:abl_repa}; each other REPA row changes the block or the weight of the row of the same source. $^\dagger$2 GPUs
with twice the gradient accumulation (same global batch).}
\label{tab:supp_repa_sweep}
\centering
\footnotesize
\setlength{\tabcolsep}{3pt}
\resizebox{\linewidth}{!}{%
\begin{tabular}{rlccccccccc}
\toprule
& Training & Block & Weight & Overall & 1 obj. & 2 obj. & Count. & Colours & Pos. & Col.\ attr. \\
\midrule
(a) & SFT control & -- & -- & 73.4 & 99.1 & 90.9 & \textbf{72.8} & 84.8 & 30.5 & 62.3 \\
\midrule
(c) & REPA: prompt tokens $\rightarrow$ description & 8 & 0.5 & 72.0 & 99.7 & 88.4 & 70.9 & 83.8 & 28.0 & 61.3 \\
(c1) & REPA: prompt tokens $\rightarrow$ description$^\dagger$ & 4 & 0.5 & 72.5 & 99.4 & 89.9 & 66.6 & 83.0 & 27.8 & \textbf{68.3} \\
(c2) & REPA: prompt tokens $\rightarrow$ description$^\dagger$ & 8 & 1.0 & 71.6 & 99.4 & 90.2 & 67.2 & 85.1 & 25.8 & 61.8 \\
\midrule
(d) & REPA: image tokens $\rightarrow$ description & 8 & 0.5 & 72.8 & 99.7 & 87.6 & 70.6 & 84.6 & 30.8 & 63.8 \\
(d1) & REPA: image tokens $\rightarrow$ description$^\dagger$ & 4 & 0.5 & 70.8 & 98.8 & 90.2 & 62.5 & 84.6 & 25.3 & 63.8 \\
(d2) & REPA: image tokens $\rightarrow$ description$^\dagger$ & 8 & 1.0 & 71.9 & \textbf{100.0} & \textbf{91.7} & 66.9 & \textbf{85.6} & 24.5 & 62.8 \\
\midrule
\rowcolor{sdours}(e) & \textbf{\method} & -- & -- & \textbf{74.3} & 99.7 & 89.4 & 71.6 & \textbf{85.6} & \textbf{32.5} & 67.0 \\
\bottomrule
\end{tabular}}
\end{table}

\subsection{Inference-time attention to the description}
\label{sec:supp_attn}
\method is trained with the image and the prompt blind to the description stream: the stream reads both, but neither
reads the stream. We compare it with a model trained with full attention, in which the image, the prompt and the stream
all attend to each other, otherwise trained identically (same SFT checkpoint, data order, noise, loss weights and seed as
\method). At inference, with the stream on 
(description guidance 2, 140 stream tokens, 28 steps, image guidance 4), the image
and prompt tokens are blocked from the stream on the sampler steps of a window and may read it on all other steps.
\Cref{tab:supp_attn} gives GenEval for windows from never blocked to always blocked; ``always'' gives the same image as the
stream off.

\begin{table}[H]
\caption{\textbf{Inference-time attention to the description} (GenEval). In the window the image and prompt do not read
the stream.}
\label{tab:supp_attn}
\centering\footnotesize
\setlength{\tabcolsep}{3pt}
\begin{adjustbox}{max width=\linewidth}
\begin{tabular}{@{}lcccccccccc@{}}
\toprule
& \multicolumn{5}{>{\columncolor{sdours}}c}{\textbf{\method} (trained blind)} & \multicolumn{5}{c}{Trained with full attention} \\
\cmidrule(lr){2-6}\cmidrule(lr){7-11}
Blocked on steps & All & 2 obj & Count & Pos. & Attr. & All & 2 obj & Count & Pos. & Attr. \\
\midrule
never & 54.8 & 60.4 & 46.6 & 12.8 & 33.5 & 72.1 & 88.9 & 70.3 & 29.2 & 61.8 \\
18--26 & 57.7 & 66.9 & 48.1 & 14.5 & 39.8 & 72.4 & 88.9 & 71.2 & 29.5 & 62.2 \\
8--14 & 60.8 & 74.2 & 51.9 & 15.5 & 42.0 & 72.7 & 89.9 & 70.3 & 30.0 & 63.8 \\
8--26 & 63.7 & 81.6 & 53.1 & 18.5 & 45.8 & 72.9 & 90.2 & 71.9 & 30.2 & 62.0 \\
0--8 & 68.4 & 81.8 & 60.0 & 27.0 & 58.5 & 71.9 & 87.6 & 69.7 & 28.2 & 62.8 \\
always (= stream off) & \textbf{74.3} & 89.4 & 71.6 & \textbf{32.5} & \textbf{67.0} & 73.1 & 89.1 & 73.1 & 29.2 & 63.8 \\
\bottomrule
\end{tabular}
\end{adjustbox}
\end{table}

When the image interacts with the description it is co-denoising, the result is worse, even for the model trained to do
so. For \method, which never reads the stream in training, letting the image read it costs up to $19.5$ points, most on
two objects, counting, position and colour attribution. The model trained with full attention, whose image reads the
stream throughout training, does not gain from it either: every window stays within about a point ($71.9$--$73.1$), and
its best setting is below \method with the stream off ($74.3$). Keeping the image blind to the description, in training
and at inference, is therefore not only cheaper but better: the description is a training signal for the image, not an
input to it.

\subsection{Descriptions: \method against image-loss weight 1}
\label{sec:supp_text_h2h}
We compare the descriptions of \method with those of the ablation that keeps the image-loss weight at $1$ (main paper,
Ablating the loss weights), head to head on the same 300 held-out and 100 FLUX-Reason-6M (out-of-domain, \cref{sec:supp_ood}) prompts and seeds. Each model
generates its image and description together; a sheet shows both images and both descriptions, labelled X and Y in random
order, and a blind judge (a fresh judge for every sheet) grades each description for fluency (0--10) and picks the
better description.

\begin{table}[H]
\caption{\textbf{Blind head-to-head of the descriptions}, \method against image-loss weight $1$.}
\label{tab:supp_text_h2h}
\centering\scriptsize
\setlength{\tabcolsep}{3pt}
\begin{adjustbox}{max width=\linewidth}
\begin{tabular}{@{}llcc@{}}
\toprule
& & \method higher / lower / tie & mean diff. \\
\midrule
Fluency & all & \textbf{95 / 61 / 244} & $+0.08$ \\
& held-out & 80 / 55 / 165 & $+0.08$ \\
& FLUX-Reason-6M & 15 / 6 / 79 & $+0.09$ \\
\midrule
& & \method--ablation & win \% \\
\midrule
Better description & all & 216--184 & 54.0 \\
\bottomrule
\end{tabular}
\end{adjustbox}
\end{table}

\Cref{tab:supp_text_h2h} gives the result. \method's descriptions are more fluent: where the judge separates them on
fluency it prefers \method $95$ to $61$, in domain ($80$--$55$) and out of domain ($15$--$6$). The overall
preference is $54.0\%$: level in domain ($154$--$146$) and ahead out of domain ($62$--$38$, $p{=}0.02$). The judge prefers the left
description ($58\%$ for \method on the left, $50\%$ on the right); \method is on the left in $214$ of $400$ sheets, so the
order-balanced preference is $54\%$.

\section{Theoretical explanation}
\label{sec:supp_theory}
We formalise two observations of the paper: writing the description improves the image (\cref{sec:why}), and the
generated stream must be read by a language model in
context, not position by position (\cref{sec:abl_direct}). Both explanations are simplified: they isolate one discrete fact
or one position and assume the network can represent the Bayes-optimal predictor, so they describe what each loss asks
of the states, not how training reaches it.

\subsection{Why writing improves drawing}
\label{sec:supp_theory_write}
\paragraph{Setting.} Fix a prompt and a time $s$. Let $A\in\{0,1\}$ be a discrete fact the image must settle: which of
two objects is yellow, whether a cup is the second or the third, whether the dog is left of the cat. Let $h$ be the
states of the image and prompt tokens in every block, from which the image velocity $\hat v_x$ is computed and which the
description stream reads through attention (\cref{sec:stream}); since the image does not attend to the stream, $h$ is a
function of $(x_s,c)$ alone. Let $p=P(A{=}1\mid h)$ measure how far $h$ has settled $A$. Write
$\mu_a=\mathbb E[x_0\mid h,A{=}a]$ for the expected clean latent under each outcome and $\Delta=\mu_1-\mu_0$; $\Delta$ is
non-zero only on the latent positions (``pixels'') that $A$ changes; we call this region $S$ and write $|S|$ for its number
of positions, the area the fact occupies. Throughout,
$H(A\mid Z)=-\mathbb E\big[\log P(A\mid Z)\big]$ denotes the conditional Shannon entropy (in nats) of $A$ given any
variable $Z$ (below, $Z$ is $(x_s,c)$, $h$ or $(h,e_s)$), which is zero exactly when $Z$ determines $A$. We take $A$ to be fixed by the
prompt and $x_s$, $H(A\mid x_s,c)=0$, as for the facts GenEval checks; any $p\in(0,1)$ is then a shortfall of the states
that training can remove. For a fact the prompt leaves open, $H(A\mid x_s,c)>0$ at high noise and only the part of each
excess above this floor can be trained away. Following \cref{sec:objective}, the image loss is a regression; with
$v_x=(x_s-x_0)/s$ it is $\|\hat x_0-x_0\|^2/s^2$ for $\hat x_0=x_s-s\hat v_x$.

\paragraph{Proposition 1 (the image loss weighs a fact by its pixels).} Among predictors that read $h$, the lowest
image loss exceeds the lowest image loss of predictors that also know $A$ by
\begin{equation}
\frac{1}{s^2}\,\mathbb E\big[p(1-p)\,\|\Delta\|^2\big]\;\le\;\frac{|S|\,\delta^2}{s^2}\,\mathbb E\big[p(1-p)\big],
\label{eq:th_img}
\end{equation}
where $\delta$ bounds the norm of $\Delta$ at a single position.
\par\noindent\emph{Proof.} The optimal predictor is the conditional
mean, so the lowest loss is the conditional variance (summed over entries); by the law of total variance,
$\mathbb E[\mathrm{Var}(x_0\mid h)]-\mathbb E[\mathrm{Var}(x_0\mid h,A)]=\mathbb E[\mathrm{Var}(\mathbb E[x_0\mid h,A]\mid h)]
=\mathbb E[p(1-p)\|\Delta\|^2]$. $\square$

The optimum under uncertainty is the blend $p\,\mu_1+(1-p)\,\mu_0$: a colour between two objects, a half-present third
cup. Its cost is at most $|S|\delta^2p(1-p)/s^2$, and proportional to the area $|S|$ when the fact changes each of its
pixels by about $\delta$, so a fact that decides correctness but covers few pixels (an attribute of a small object, a
part, a texture) carries little weight in $\Limg$.

\paragraph{Proposition 2 (the language loss weighs a fact by the words that state it).} Let the description state $A$
in a word, $y_j=w_A$, where $w_a$ is the word that states outcome $a$ (e.g.\ \emph{yellow} and \emph{purple}),
$w_0\neq w_1$. The read-out sees only $\hat e_0$, which is computed from $h$ and the noised
stream $e_s$, and the earlier words $y_{<j}$; assume these words carry no information on $A$ beyond $(h,e_s)$,
$A\perp y_{<j}\mid(h,e_s)$. Let $\tilde p=P(A{=}1\mid h,e_s)$. Then every read-out has
\begin{equation}
\mathbb E\big[-\log p_\phi(y_j\mid\cdot)\big]\;\ge\;H(A\mid h,e_s)=\mathbb E\big[H_b(\tilde p)\big],
\label{eq:th_txt}
\end{equation}
where $H_b(p)=-p\log p-(1-p)\log(1-p)$ (natural log), with equality for the Bayes read-out (the one that outputs the true conditional
probability of $y_j$), while a read-out of states
that settle $A$ pays nothing on $y_j$. At $s=1$, $e_s=\eta$ is independent of $A$, so $\tilde p=p$ and the bound is
$H(A\mid h)=\mathbb E[H_b(p)]$.
\par\noindent\emph{Proof.} The read-out's input $(\hat e_0,y_{<j})$ is a function of $(h,e_s,y_{<j})$; the cross-entropy is at least the conditional
entropy of $y_j$ given this input, which is at least $H(y_j\mid h,e_s,y_{<j})=H(A\mid h,e_s,y_{<j})=H(A\mid h,e_s)$,
since $y_j$ and $A$ determine each other and by the assumption. $\square$

The bound does not depend on $|S|$: one word costs the same whether it names a small object or a large one. The
assumption on $y_{<j}$ fails for a description that states a fact twice; the bound then holds for its first mention.

\paragraph{Corollary (where the two losses differ).} Fix $h$ at a noise level high enough that $e_s$ carries little
about $A$, so that $\tilde p\approx p$ (exactly at $s=1$). With the loss weights of \cref{sec:objective}, the ratio of the
language to the image excess is
\begin{equation}
\frac{\lambda_{\mathrm{txt}}}{w_{\mathrm{img}}}\cdot\frac{H_b(p)}{p(1-p)}\cdot\frac{s^2}{\|\Delta\|^2},
\label{eq:th_ratio}
\end{equation}
where $H_b(p)/\big(p(1-p)\big)\ge 4\ln 2$, with equality at $p=\tfrac12$, and this middle factor grows without bound as
$p\to0$ or $p\to1$. The constant in front depends on how each loss is normalised (a mean over latent entries, a mean over
tokens) and on the scale of $x_0$, so only the dependence on $p$ and $\|\Delta\|$ carries meaning.
Writing therefore adds weight where the image loss has least: on facts with small $\|\Delta\|^2$ (small area, a fine
difference) and on facts that are nearly but not fully settled, where the regression's cost $p(1-p)$ vanishes faster
than the entropy. As $s$ decreases, $e_s$ itself increasingly reveals $A$, $H(A\mid h,e_s)$ falls below $H(A\mid h)$,
and the read-out can pay its loss from $e_s$ rather than from $h$: the language loss presses on the image states mainly
at high noise. Since the description is written from $h$, this weight reaches the blocks the image is drawn with, and it
remains when the stream is switched off.

Writing adds no weight on how much of an object is coloured, an area no
word states, so the explanation predicts no gain from writing on colour coverage. It does not predict the control's edge
there (coverage is the main reason for $49$ of the control's $111$ wins on SD3.5 and $60$ of $109$ on FLUX.2 [klein],
against $35$ of $106$ and $28$ of $114$ for \method; \cref{tab:why,tab:supp_cawhy}). Under the Bayes-optimal
assumption the two losses never compete, since states that settle $A$ lower both; the edge needs finite capacity, where
the shared weights of \method also serve the description losses and the image loss receives a smaller share of each
update. Finally, it predicts the sign
of each effect and where it falls, not its size.

\subsection{Why the stream must be read in context}
\label{sec:supp_theory_read}
\paragraph{Setting.} Let $e(y)\in\mathbb R^{L_e\times d_e}$ be the encoding of a description $y$, with position $i$
holding the state $e_i(y)$ of its $i$-th token $y_i$. The prompt and the image settle what a description says far more
than how it says it: many descriptions fit the same image. We write $y^{(1)},\dots,y^{(K)}$ for these $K$ descriptions,
so that the superscript $(k)$ indexes a whole description and the subscript $i$ a token within it ($y^{(k)}_{i}$ is the
$i$-th token of the $k$-th description). They share their content but word it differently, and one added word shifts
every later word by a position. The stream loss $\Ldesc$ is a regression, so by the argument of
Proposition 1 its optimum is the posterior mean
\begin{equation}
\hat e_0=\mathbb E[e_0\mid e_s,x_s,c]=\textstyle\sum_k\pi_k\,e\big(y^{(k)}\big),
\label{eq:th_blend}
\end{equation}
where $\pi_k=P(y=y^{(k)}\mid e_s,x_s,c)$ is the probability of the $k$-th description given what the network sees, and
$\hat e_{0,i}=\sum_k\pi_k\,e_i(y^{(k)})$ is its $i$-th row: a blend of wordings whose weights $\pi_k$ the sampler sharpens
over its steps. What the sampler leaves unsettled
remains a blend. The read-out loss, trained jointly, also acts on $\hat e_0$ and moves it from this optimum; we analyse
the optimum of $\Ldesc$.

\paragraph{Proposition 3 (a position-wise read-out of a blend).} Let the direct mapping read position $i$ with logits
linear in the state, $\ell_i=W\hat e_{0,i}$, where $W$ maps a state to one score per vocabulary token, and let it read
every single wording cleanly: $W e_i(y^{(k)})=a\,\mathbf 1_{y^{(k)}_{i+1}}+r_{k,i}$, where $\mathbf 1_w$ is the one-hot
vector of token $w$, $a>0$ is the margin by which the correct token's score exceeds the others, and the residual
$r_{k,i}$ has $\|r_{k,i}\|_\infty\le\rho$. Then
\begin{equation}
\ell_i[w]=a\sum_k\pi_k\,\mathbf 1\big[y^{(k)}_{i+1}=w\big]+r_i[w],
\label{eq:th_pos}
\end{equation}
with $|r_i[w]|\le\rho$, so whenever the leading word wins by more than $2\rho/a$ in $\pi$-mass, position $i$ reads the
word that the $\pi$-weighted majority of wordings put at $i{+}1$.
\par\noindent\emph{Proof.} Linearity of $\ell_i$ in
\cref{eq:th_blend}, with $r_i=\sum_k\pi_k r_{k,i}$ and $\sum_k\pi_k=1$. $\square$

Four consequences match \cref{tab:supp_direct,tab:supp_probe}. (a) Every word that wins its position by this margin
belongs to some description of the image: the content survives (ROUGE-1 $28.9$, against $2.8$ from a random stream). (b) When the wordings are offset,
neighbouring positions are won by different wordings, so neighbouring words come from different sentences: bigrams and
longer n-grams of the output rarely occur in any single description (ROUGE-2 $2.1$, BLEU-4 $0.0$), and a word that two offset
wordings place one position apart is read twice (\emph{black black}, \cref{fig:supp_direct}). (c) A blend of wordings
offset by different amounts has no single shift that aligns it (best shift per prompt: $2{,}851$). (d) With one wording
($K{=}1$, the true stream) the read-out is exact up to $\rho$, and additive noise changes it only once it overturns the
margin $a$ (true stream $16.7$; with noise of the stream's own size $21.3$; generated stream $3{,}467$).

\paragraph{Proposition 4 (a read-out in context reads content wherever it lies).} Let a read-out pool
$u(e)=\sum_i\alpha_i(e)\,Ve_i$, where $V$ is a linear value projection shared by all positions (as in attention) and
the weights $\alpha_i$ depend on a state's content and not on its position.
(a) If $e'$ permutes the positions of $e$, then $u(e')=u(e)$. (b) If, further, the weights are uniform,
$\alpha_i=1/L_e$, then $u$ is linear, $u(\hat e_0)=\sum_k\pi_k\,u\big(e(y^{(k)})\big)$, so wordings whose states are
permutations of one another, and every blend of them \cref{eq:th_blend}, pool to the same $u$: the offsets that defeat
Proposition 3 do not reach $u$.
\par\noindent\emph{Proof.} (a) Permuting the positions permutes the $\alpha_i$ with them, and a sum
over positions is invariant to their order. (b) Linearity of $u$ and \cref{eq:th_blend}, then (a). $\square$

Part (b) is an idealisation in two ways. Softmax attention is not linear in the blend, and the contextual encodings of
two wordings are not exact permutations of each other (an inserted word adds a state and changes its neighbours). We
therefore test the claim directly. A language-model decoder also encodes positions, so it is not exactly invariant, but
it behaves so: the decoder of \cref{tab:supp_probe} reads the true stream shuffled within ten positions or shifted by one
at $25.3$ to $25.4$, against $25.4$, and the decoder of \method reads its own true stream at $12.7$ and shuffled or
shifted at $12.8$ (\cref{sec:supp_direct}). It then writes autoregressively, each word conditioned on those already
written, and so commits to one wording of the content it has pooled.

The two propositions also tie back to \cref{sec:supp_theory_write}. Both read-outs pay for the wording the stream leaves
open: an autoregressive read-out in context has cross-entropy at least $H(y\mid\hat e_0)$, which includes the entropy of
the wording given the content, a cost the backbone cannot remove by drawing better. A position-wise read-out pays more,
\begin{equation}
\begin{split}
\textstyle\sum_i H\big(y_{i+1}\mid\hat e_{0,i}\big)
&\ge\textstyle\sum_i H\big(y_{i+1}\mid\hat e_0,y_{\le i}\big)\\
&=H(y\mid\hat e_0),
\end{split}
\label{eq:th_poswise}
\end{equation}
a gap that \cite{shen2026codar} also derive for decoding continuous diffusion language models, and its excess depends on how the wordings in the blend are aligned, so it is lowered most directly by committing $\hat e_0$
to one wording. A position-wise loss would thus spend the backbone on the wording, which the image does not determine; a
read-out in context charges the backbone only for missing \emph{content}, the facts the image must settle. Reading the generated stream at inference also lowers GenEval
(\cref{tab:attn}): \method is trained with the image never attending to the stream, so at inference the stream is an
input the image pathway has not learnt to read.

\section{Why does writing help?}
\label{sec:supp_g_why}
Pairwise analyses of colour-attribution and position wins, and where the description's content comes from.

\subsection{Why one image wins a colour-attribution pair}
\label{sec:supp_cawhy}
This section details the blind judge study behind \cref{tab:why} of the main paper (\cref{sec:why}). Colour attribution
is \method's largest GenEval gain on both backbones; the study asks \emph{how} the images of \method and of the SFT
control differ when one of them is the better match, and checks the judge against the official scorer's own outputs.

\paragraph{Pairs.} We take every GenEval colour-attribution prompt at all four sampling seeds (400 prompt-seed pairs per
backbone) and pair the image of \method with the image of the SFT control for the same prompt and seed. The pairs fall
into two groups, judged with two variants of the same instruction:
\begin{itemize}
\item \emph{GenEval disagrees} (75 pairs on SD3.5, 73 on FLUX.2 [klein]): the official scorer accepts exactly one of the
two images; we call the other one, which the scorer classifies as incorrect, the \emph{rejected image}. The judge is told which image the scorer accepted, compares the two, and says whether it agrees that the
accepted image is the better match. A pair counts as won by a model only when the judge agrees that its image is better.
The judge agrees on 37 of 75 pairs on SD3.5 and on 42 of 73 on FLUX.2 [klein].
\item \emph{GenEval agrees} (325 and 327 pairs; both images accepted, or both rejected): the judge is not told anything
about the scorer and picks the better match, or answers \emph{equal} (145 and 146 pairs).
\end{itemize}
In both variants the judge compares the two images only on how well each shows the two requested objects in their requested
colours, ignoring image quality and style, explains its choice in one or two sentences, and tags the differences from the
better image's side with every tag that applies, naming one as the main reason. The tags are fixed in advance and are the
same in both variants (\cref{tab:supp_cawhy}). As in all our judge studies (\cref{sec:supp_judge}), every pair is one side-by-side
sheet with sides balanced, the model names are hidden, and every sheet gets its own fresh Claude Opus 5.5 session.

\paragraph{Wins.} Pooled over both groups, the two models win about equally often: \method $106$ and the control $111$
pairs on SD3.5, $114$ and $109$ on FLUX.2 [klein] (with the scorer disagreeing: $22$ to $15$ and $26$ to $16$).
The study is therefore not about \emph{how often} each model wins but \emph{why}.

\paragraph{Reasons.} \Cref{tab:supp_cawhy} gives, per backbone and model, how often each tag is the main reason for a
win and how often it is named at all. The main-paper table groups the main reasons into two families:
\emph{colour coverage} (the colour covers more of the object; fewer distracting colours around it) and \emph{objects and
their colours} (all other tags: the objects are present, recognisable, separated and prominent, each carries its own,
pure colour, and that colour does not leak onto the other object or revert to the object's natural colour). \method's wins
fall mostly in the second family ($71$ of $106$ on SD3.5, $86$ of $114$ on FLUX.2 [klein]), while the control's wins lean
towards colour coverage ($49$ of $111$ and $60$ of $109$, against $35$ and $28$ for \method).
On FLUX.2 [klein] the judge most often prefers \method because the
object is more recognisable ($36$ main reasons against $13$) or both objects are present at all ($10$ against $3$);
the control's wins are dominated by colour coverage ($57$ of $109$).

\paragraph{Colour leakage on SD3.5.} On SD3.5, the judge names the absence of colour leakage, the colour of one object
spilling onto the other, as the main reason for $13$ of \method's wins and $6$ of the control's, and among all reasons for
$20\%$ and $14\%$ of their wins; \method's images are also more often preferred for a purer, more saturated colour
($20$ main reasons against $5$). These are the hedged renderings the introduction describes: a colour that is neither
one object's nor the other's costs a pixel loss little, while a description must assign each colour to one noun.
On FLUX.2 [klein] the judge rarely names leakage as the main reason ($4$ and $6$) and credits \method's wins to presence
and recognisability instead; the scorer itself, however, records leakage on FLUX.2 [klein] as well (next paragraph).

\paragraph{What the scorer itself measures.} The judge's reasons are a reading of the images. GenEval's own outputs give
a second account that involves no judge: for each image the official scorer detects the requested objects
(Mask2Former), assigns each detected object a colour (CLIP), and records why it rejects the image. For the $75$ and $73$
pairs on which the scorer accepts exactly one image, we sort the rejected image by the scorer's record into three causes:
a requested object is not detected; both are detected but one carries the colour requested for the \emph{other} object
(leakage); or both are detected and one carries some other wrong colour. \Cref{tab:supp_cadirect} gives the counts.
On both backbones the control loses mainly through wrong colours: $35$ and $31$ of its rejected images against $18$ and
$21$ of \method's, and leakage alone is about twice as frequent for the control ($13$ against $6$ on SD3.5, $17$
against $8$ on FLUX.2 [klein]). Missing objects depend on the backbone: level on SD3.5 ($12$ and $10$), but $16$
against $5$ on FLUX.2 [klein], the backbone on which the judge also credits \method with both objects being present
($10$ main reasons against $3$; $2$ and $2$ on SD3.5).

These counts support the wording of \cref{sec:why}. \emph{Binding each colour to its object} is what the leakage count
measures directly: the control more often gives one object the colour requested for the other, the hedged rendering
that a pixel loss tolerates and a description, which assigns each colour word to one noun, does not. That \method wins
\emph{mainly on the presence of the objects and their colours} holds in the scorer too: through correct colours on both
backbones, and through the presence of both objects on FLUX.2 [klein]. Colour coverage, the control's most frequent
winning reason, has no counterpart among the scorer's outputs, so that part of \cref{tab:why} rests on the judge alone.

\begin{table}[t]
\caption{\textbf{Why the rejected image of a GenEval colour-attribution pair fails, according to the official scorer}
(no judge). Pairs on which the scorer accepts exactly one of \method and the SFT control; causes read from the scorer's
detections and colour assignments. \emph{Control wrong}: pairs \method wins; \emph{\method wrong}: pairs the control wins.}
\label{tab:supp_cadirect}
\centering
\footnotesize
\setlength{\tabcolsep}{3pt}
\begin{adjustbox}{max width=\linewidth}
\begin{tabular}{lcccccc}
\toprule
 & \multicolumn{2}{c}{SD3.5} & \multicolumn{2}{c}{FLUX.2 [klein]} & \multicolumn{2}{c}{Pooled} \\
Cause (scorer) & control & \method & control & \method & control & \method \\
 & wrong & wrong & wrong & wrong & wrong & wrong \\
\midrule
other object's colour (leakage) & 13 & 6 & 17 & 8 & 30 & 14 \\
other wrong colour & 22 & 12 & 14 & 13 & 36 & 25 \\
\rowcolor{sdours}\textbf{wrong colour, total} & \textbf{35} & 18 & \textbf{31} & 21 & \textbf{66} & 39 \\
\midrule
object not detected & 12 & 10 & 16 & 5 & 28 & 15 \\
\midrule
Total & 47 & 28 & 47 & 26 & 94 & 54 \\
\bottomrule
\end{tabular}
\end{adjustbox}
\end{table}

\begin{table*}[t]
\caption{\textbf{Why the blind judge prefers one image of a GenEval colour-attribution pair}, per backbone and model.
\emph{Main}: number of won pairs for which the tag is the judge's main reason; \emph{any}: share of won pairs for which the
tag is named among the reasons. The family column gives the grouping of \cref{tab:why} in the main paper.}
\label{tab:supp_cawhy}
\centering
\footnotesize
\setlength{\tabcolsep}{4pt}
\begin{adjustbox}{max width=\linewidth}
\begin{tabular}{ll>{\columncolor{sdours}}c>{\columncolor{sdours}}ccc>{\columncolor{sdours}}c>{\columncolor{sdours}}ccc}
\toprule
 & & \multicolumn{4}{c}{SD3.5} & \multicolumn{4}{c}{FLUX.2 [klein]} \\
 & & \multicolumn{2}{>{\columncolor{sdours}}c}{\textbf{\method}} & \multicolumn{2}{c}{SFT} & \multicolumn{2}{>{\columncolor{sdours}}c}{\textbf{\method}} & \multicolumn{2}{c}{SFT} \\
Tag (from the winner's side) & Family & main & any & main & any & main & any & main & any \\
\midrule
object more recognisable & objects & 17 & 40\% & \textbf{33} & 44\% & \textbf{36} & 48\% & 13 & 29\% \\
colour more saturated or pure & objects & \textbf{20} & 32\% & 5 & 22\% & \textbf{15} & 27\% & 10 & 28\% \\
other object's colour not leaking & objects & \textbf{13} & 20\% & 6 & 14\% & 4 & 9\% & \textbf{6} & 18\% \\
object larger or more prominent & objects & \textbf{12} & 29\% & 2 & 14\% & 6 & 18\% & 6 & 26\% \\
objects more clearly separated & objects & 6 & 24\% & \textbf{9} & 18\% & \textbf{13} & 37\% & 5 & 28\% \\
both objects present & objects & 2 & 3\% & 2 & 3\% & \textbf{10} & 11\% & 3 & 3\% \\
unusual colour kept (not reverted to natural) & objects & 1 & 11\% & \textbf{5} & 18\% & 2 & 11\% & \textbf{6} & 15\% \\
colour covers more of the object & coverage & 31 & 48\% & \textbf{45} & 60\% & 26 & 39\% & \textbf{57} & 65\% \\
fewer distracting colours around & coverage & 4 & 24\% & 4 & 27\% & 2 & 9\% & \textbf{3} & 31\% \\
\midrule
\emph{objects and their colours} & & \textbf{71} & & 62 & & \textbf{86} & & 49 & \\
\emph{colour coverage} & & 35 & & \textbf{49} & & 28 & & \textbf{60} & \\
pairs won & & 106 & & 111 & & 114 & & 109 & \\
\bottomrule
\end{tabular}
\end{adjustbox}
\end{table*}

\subsection{Why one image wins a position pair}
\label{sec:supp_poswhy}
Position is \method's second GenEval gain over the SFT control on both backbones ($+2.0$ on SD3.5, $+2.8$ on FLUX.2
[klein]). We ask, as for colour attribution (\cref{sec:supp_cawhy}), what the images of the two models do differently.

\paragraph{Pairs.} We take every GenEval position prompt at all four sampling seeds (400 prompt-seed pairs per backbone)
and keep the pairs on which the official scorer accepts exactly one of the two images: $24$ pairs where \method is
right and the control wrong and $16$ the other way round on SD3.5, $29$ and $18$ on FLUX.2 [klein] ($53$ to $34$
pooled).

\paragraph{Judge.} For every pair the \emph{wrong} image is shown alone, under a neutral id, to a fresh blind Claude Opus 5.5
Opus 5.5 session, with the request. Rejected images of both models are mixed and shuffled, so the judge does not know which
model it is looking at, nor that the other model got the prompt right. The judge is told that the image does not fully
match the request, judges only what it shows (objects, colours, number, positions; not style or quality), names every
error type that applies and a main one from a fixed list (colour on the wrong object, other wrong colour, missing object,
extra object, merged or fused objects, wrong count, wrong position, object unrecognisable, other), and explains the
error in one sentence.

\paragraph{From the judge's explanations to categories.} The judge's own label is too coarse for position: \emph{wrong
position} is the main error for $32$ of the control's $53$ rejected images and $17$ of \method's $34$, and covers very
different failures. We therefore had Claude Opus 5.5 read all $87$ one-sentence explanations, with the request, and sort
each rejected image into one of eight categories by what the explanation says is wrong; the categories were formed from the
explanations, not fixed in advance. The eight categories are grouped by what they fail at:
\begin{itemize}
\item \emph{Separating the two objects}: one object stands in front of or behind the other and overlaps it instead of
being beside, above or below it (\emph{depth or overlap}); one object is drawn as part of the other, \eg a zebra-print
chair for \emph{a chair left of a zebra} or a toothbrush on a kite's string (\emph{fused}); one object is on, in or leaning
on the other, \eg a suitcase on the boat's deck or a pizza on the keyboard (\emph{touching}).
\item \emph{Getting each object right}: a requested object is absent or replaced by another (a calf for the dog, a
skateboard for the snowboard); an object appears more than once; an object is malformed or unrecognisable.
\item \emph{Getting the direction right}: the two objects are separate but the relation is wrong (left for right,
reversed above/below).
\end{itemize}
In two cases the judge found no clear error; we count them separately. The judge never saw these categories, and every
assignment, with the judge's explanation, is released with the code.

\paragraph{Result.} \Cref{tab:supp_poswhy} gives the counts. Almost all of \method's net gain comes from separating
the two objects: $35$ of the control's rejected images fail there against $17$ of \method's ($p=0.02$), a net of $18$ of
the $19$ pairs. Getting each object right is level ($12$ and $12$): \method has fewer missing objects but more
malformed ones. The direction is level as well ($5$ and $4$): the control knows what \emph{left of} means; what it fails to
do is place two separate objects so that the relation can be read at all. The largest single category is depth or overlap
($19$ against $9$): the control's typical answer to \emph{a bottle right of a train} is a bottle \emph{in front of} the
train, half beside it and half on it.

\paragraph{What the scorer itself measures.} As for colour attribution, GenEval's own outputs give a judge-free check.
The scorer detects the two objects of \emph{a $\langle$named object$\rangle$ right of a $\langle$reference$\rangle$} and
tests the requested relation between their boxes. We sort each rejected image by the scorer's record: the reference
not detected, the named object not detected, or both detected and the relation wrong (\cref{tab:supp_posdirect}).
On both backbones the control loses mainly with both objects present and the relation wrong: $20$ against $12$ on
SD3.5 and $19$ against $10$ on FLUX.2 [klein], $17$ of the pooled net of $19$. Missing objects depend on the backbone
and are level when pooled ($14$ and $12$): the control misses the named object more often on FLUX.2 [klein] ($8$
against $0$), \method misses the reference more often there ($8$ against $2$), and SD3.5 shows the opposite on a few
images. The judge's \emph{getting each object right} group is level as well ($12$ and $12$).
The scorer's boxes also show what a wrong relation looks like. In the $32$ and $29$ pairs in which both images contain
both objects, we measure how much the two boxes overlap, as the share of the smaller box covered by the other ($0$: apart;
$1$: one inside the other). The rejected image's boxes overlap far more than those of the accepted image of the same
pair (median $0.65$ against $0.21$ on SD3.5, $0.66$ against $0.22$ on FLUX.2 [klein]; more in $22$ and $24$ pairs, less
in $6$ and $5$). With both objects present, a rejected image has typically drawn them over one another rather than on
the wrong side, which is the \emph{depth or overlap} and \emph{fused} failures of \cref{tab:supp_poswhy}. Both accounts
agree: \method's gain is in getting the two objects and their relation right.

\begin{table}[t]
\caption{\textbf{Why the rejected image of a GenEval position pair fails, according to the official scorer} (no judge).
Pairs on which the scorer accepts exactly one of \method and the SFT control. \emph{Named object}: the object to be
placed (\emph{bottle} in \emph{a bottle right of a train}); \emph{reference}: the object it is placed relative to.}
\label{tab:supp_posdirect}
\centering
\footnotesize
\setlength{\tabcolsep}{3pt}
\begin{adjustbox}{max width=\linewidth}
\begin{tabular}{lcccccc}
\toprule
 & \multicolumn{2}{c}{SD3.5} & \multicolumn{2}{c}{FLUX.2 [klein]} & \multicolumn{2}{c}{Pooled} \\
Cause (scorer) & control & \method & control & \method & control & \method \\
 & wrong & wrong & wrong & wrong & wrong & wrong \\
\midrule
\rowcolor{sdours}\textbf{both present, wrong relation} & \textbf{20} & 12 & \textbf{19} & 10 & \textbf{39} & 22 \\
\midrule
reference not detected & 4 & 1 & 2 & 8 & 6 & 9 \\
named object not detected & 0 & 3 & 8 & 0 & 8 & 3 \\
\midrule
Total & 24 & 16 & 29 & 18 & 53 & 34 \\
\bottomrule
\end{tabular}
\end{adjustbox}
\end{table}

This is the same commitment as in colour attribution (\cref{sec:why}), now binding each object to its own place. A pixel loss tolerates a blend: a colour between
two objects, or one object half covering the other. A description of the scene names two separate things and the relation
between them, and for the language loss to be low each must be readable from the states the image is drawn from, which
pushes the image to commit to two objects in two places.

\begin{table*}[t]
\caption{\textbf{Why the rejected image of a GenEval position pair fails}, for the pairs on which the official scorer
accepts exactly one of \method and the SFT control. A blind judge explains each rejected image; we sort the explanations into
eight categories and three groups (\cref{sec:supp_poswhy}). \emph{Control wrong}: pairs \method wins;
\emph{\method wrong}: pairs the control wins. Net: control wrong minus \method wrong.}
\label{tab:supp_poswhy}
\centering
\footnotesize
\setlength{\tabcolsep}{4pt}
\begin{adjustbox}{max width=\linewidth}
\begin{tabular}{llcccccc}
\toprule
 & & \multicolumn{2}{c}{SD3.5} & \multicolumn{2}{c}{FLUX.2 [klein]} & \multicolumn{2}{c}{Pooled} \\
Group & Category & control wrong & \method wrong & control wrong & \method wrong & control / \method & net \\
\midrule
\multirow{3}{*}{Separating the two objects} & depth or overlap & 6 & 2 & 13 & 7 & 19 / 9 & $+10$ \\
 & fused & 6 & 4 & 3 & 0 & 9 / 4 & $+5$ \\
 & touching (on, in, leaning on) & 3 & 2 & 4 & 2 & 7 / 4 & $+3$ \\
\midrule
\multirow{3}{*}{Getting each object right} & missing or substituted & 2 & 1 & 5 & 3 & 7 / 4 & $+3$ \\
 & extra instance & 4 & 2 & 0 & 1 & 4 / 3 & $+1$ \\
 & unrecognisable or malformed & 0 & 3 & 1 & 2 & 1 / 5 & $-4$ \\
\midrule
Getting the direction right & wrong direction, objects separate & 2 & 2 & 3 & 2 & 5 / 4 & $+1$ \\
\midrule
-- & no clear error & 1 & 0 & 0 & 1 & 1 / 1 & $0$ \\
\midrule
\rowcolor{sdours}\multicolumn{2}{l}{\textbf{Separating the two objects}} & 15 & 8 & 20 & 9 & \textbf{35 / 17} & $\mathbf{+18}$ \\
\multicolumn{2}{l}{Getting each object right} & 6 & 6 & 6 & 6 & 12 / 12 & $0$ \\
\multicolumn{2}{l}{Getting the direction right} & 2 & 2 & 3 & 2 & 5 / 4 & $+1$ \\
\multicolumn{2}{l}{Total} & 24 & 16 & 29 & 18 & 53 / 34 & $+19$ \\
\bottomrule
\end{tabular}
\end{adjustbox}
\end{table*}

\subsection{Where the description comes from}
\label{sec:supp_blind}
During training and sampling the description stream attends to both the prompt and the image. To see which of the two
the generated description draws on, we cut the stream's attention to the prompt tokens or to the image tokens,
at every sampling step, at inference only. The image and prompt never attend to the stream, so the image is unchanged.
\Cref{tab:supp_blind} reports the read-out perplexity of the generated description on the held-out set of the
main paper (400 samples, 28 steps, the same noise seeds).

Without the image the description is read back almost as well as normally ($48.1$ vs.\ $46.3$). Without the prompt it is clearly worse ($59.8$). The description is
therefore written mainly from the prompt rather than copied from the image: the prompt supplies its content, the image
adds a little.
Without the prompt it also drifts away from the target as sampling proceeds ($50$ at step 0, $57.5$ at the end).

\begin{table}[H]
\caption{\textbf{Read-out perplexity when the stream cannot attend to the image or the prompt.}}
\label{tab:supp_blind}
\centering
\footnotesize
\begin{adjustbox}{max width=\linewidth}
\begin{tabular}{lc}
\toprule
Stream attends to & Perplexity \\
\midrule
\rowcolor{sdours}prompt and image (\textbf{\method}) & \textbf{46.3} \\
prompt only & 48.1 \\
image only & 59.8 \\
\midrule
uniform guess (analytical) & 151{,}936 \\
\bottomrule
\end{tabular}
\end{adjustbox}
\end{table}

\section{Failure cases of the generated descriptions}
\label{sec:supp_failures}
\begin{figure*}[!t]
  \centering
  \setlength{\tabcolsep}{4pt}
  \begin{qualband}{SD3.5 Medium}\begin{adjustbox}{max width=\linewidth}%
  \begin{tabular}{@{}cp{0.33\textwidth}cp{0.33\textwidth}@{}}
  \raisebox{-\height}{\rimg[width=0.14\textwidth]{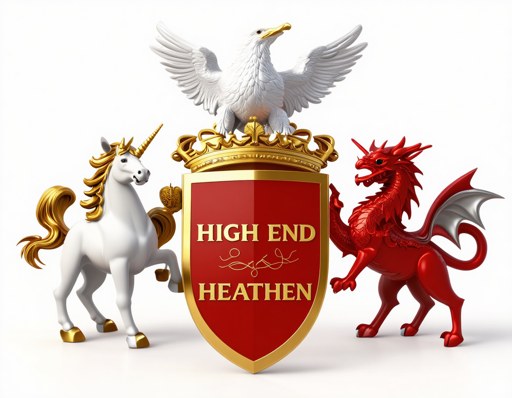}} &
  \vspace{0pt}\scriptsize{\color{qmute}\itshape 3D rendered image of a coat of arms shield. Writing in gold ``HIGH END HEATHEN'' centered on the shield. On the left side a white and gold unicorn \dots}\par\smallskip
  The main subject is a \dbad{horse}, \dok{with golden mane and gold decorations}. Horse: the tail are in size (gold) on its head, facing to the right of the \dok{eagle}'s \dok{dragon}). The \dbad{lion has white hooves} at the top; the horses have symmetrical scale patterns visible. &
  \raisebox{-\height}{\rimg[width=0.14\textwidth]{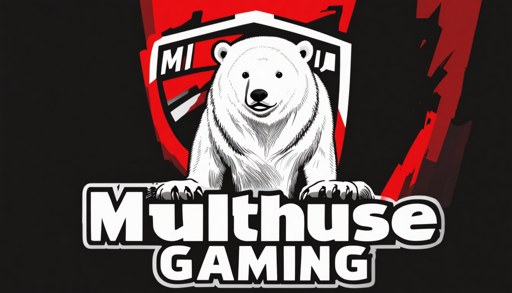}} &
  \vspace{0pt}\scriptsize{\color{qmute}\itshape Black, white, and red team logo for Mulhouse Gaming esports team with a drawing of a polar bear}\par\smallskip
  The main subject is a \dok{white fur} with \dbad{black and white spots}, the mouth area in an \dbad{orange coat}. The \dbad{dog} has \dok{two large ears (white on its head)}, as shadow; \dbad{tail: gray-white mane are dark brown} at the top of the image) \\
  \end{tabular}\end{adjustbox}\end{qualband}
  \begin{qualband}{FLUX.2 [klein]}\begin{adjustbox}{max width=\linewidth}%
  \begin{tabular}{@{}cp{0.33\textwidth}cp{0.33\textwidth}@{}}
  \raisebox{-\height}{\rimg[width=0.14\textwidth]{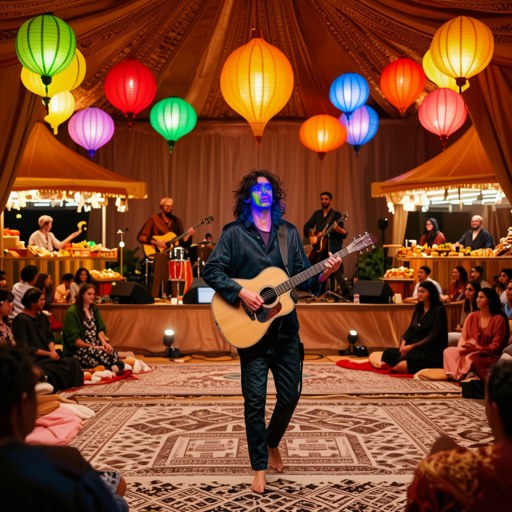}} &
  \vspace{0pt}\scriptsize{\color{qmute}\itshape Arabian tent themed outdoor concert stage with live music bands, food festival, the guitar in the central musician's hand has an irregular shape \dots}\par\smallskip
  \dok{male}: dark skin, \dok{black} \dbad{short} \dok{hair}, wearing a \dbad{white T-shirt and blue jeans}; \dok{holding the guitar in both hands}. The man's left hand is raised to \dbad{touch an open book} with his right arm visible on its back. Male 20th \dok{A young woman} with light-colored long-sleeved shirt underneath, \dok{sitting cross-legged} on her legs standing straight, facing each other. In front of him are \dok{multiple people} walking or dancing instruments. &
  \raisebox{-\height}{\rimg[width=0.14\textwidth]{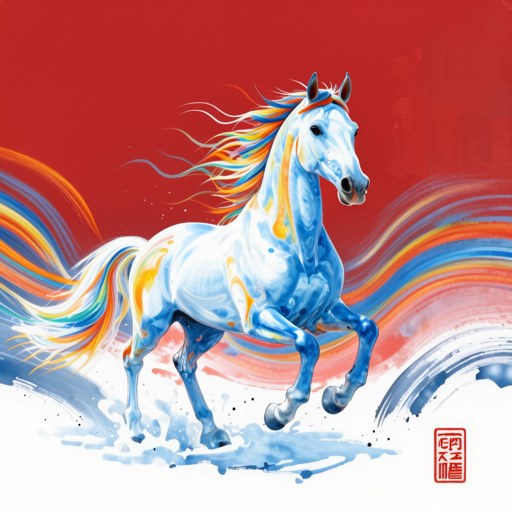}} &
  \vspace{0pt}\scriptsize{\color{qmute}\itshape CG illustration, enamel-colored galloping horse, colorful lines flowing backward as afterimages, elegant, masterful composition \dots}\par\smallskip
  : \dok{white, light gray-white color tones} (with clear texture), \dok{mane and tail flowing backward}. The horse is \dok{in a running posture}; the left side of the picture's body has four legs visible, \dbad{hoof-like structure on its back} to \dok{right front leg raised high upward}, with clearly defined muscle lines outlining the outline are drawn or spread across her limbs, but slightly blurred due to movement. Horse: head shape, \dok{long ears upright}, \dbad{black hooves} facing forward. \\
  \end{tabular}\end{adjustbox}\end{qualband}
  \caption{\textbf{Failure cases of the generated descriptions.} Held-out prompts (grey, cut), the image \method generates and
  the description it writes with it (default decoding, text unedited). \dok{Blue}: matches the image; \dbad{red}: an error.
  The samples are typical rather than the worst: each names the right subject and has fewer errors than the average
  description. Three kinds of error recur: things named as something else (a unicorn as a
  horse, a polar bear as a dog), objects or parts that are not in the image (a lion, an open book, a hoof on the horse's
  back), and wrong colours (white hooves for gold, white T-shirt and blue jeans for black clothes).}
  \label{fig:supp_failures}
\end{figure*}

The descriptions name the right kind of scene but are not yet accurate in their details (\cref{sec:supp_reference}).
\Cref{fig:supp_failures} shows four representative samples on the held-out prompts, two per backbone, with the errors
marked. They are typical rather than the worst: each description names the subject of its image and has fewer errors than
the average one. Three kinds of error recur. The description names an object as something else (a unicorn as a horse, a
polar bear as a dog), it adds objects or parts that are not in the image (a lion, an open book, a hoof on the horse's
back), and it gives wrong colours (white hooves for gold ones, a white T-shirt and blue jeans for black clothes). Broken
phrases and repetitions also remain, since the read-out is a single language-model layer trained on the stream of a
generator.

\section{Extending to an open-source dataset}
\label{sec:supp_open}
Our corpus pairs each image with a description written for that image. We test whether an open-source image--text dataset
can supply the descriptions instead. We use FLUX-Reason-6M \cite{fang2025flux}: $5.9$M generated images at
$1024{\times}1024$, each with captions for six characteristics (composition, entity, text rendering, imagination, style and
affection, each flagged as valid or not) and a generation chain-of-thought (GCoT), a long reasoning-style account of the
image.

\begin{table*}[!t]
\caption{\textbf{T2I-CompBench per category after training on FLUX-Reason-6M} ($\times 100$; SD3.5 Medium, GCoT rewritten in
our corpus' form), under the protocol of \cref{tab:supp_t2icomp}: official scorers, $300$ validation prompts per category, ten
images per prompt, stream off. For comparison, the last row is \method trained on our corpus (\cref{tab:supp_t2icomp}).}
\label{tab:supp_open_t2i}
\centering
\footnotesize
\setlength{\tabcolsep}{4pt}
\begin{tabular}{lccccccccc}
\toprule
& \multicolumn{3}{c}{Attribute binding} & \multicolumn{3}{c}{Object relationship} & & & \\
\cmidrule(lr){2-4}\cmidrule(lr){5-7}
Training & Colour & Shape & Texture & 2D spatial & 3D spatial & Non-spatial & Numeracy & Complex & Avg.$\uparrow$ \\
\midrule
base model & 78.26 & \textbf{57.36} & 70.86 & 28.06 & 37.98 & 31.32 & 59.84 & 36.41 & 50.01 \\
SFT control, FLUX-Reason-6M & 79.55 & 56.81 & \textbf{71.03} & 30.72 & 39.94 & 31.45 & 61.08 & 37.06 & 50.95 \\
\rowcolor{sdours}\method, FLUX-Reason-6M & \textbf{79.86} & 56.77 & 70.50 & \textbf{32.15} & \textbf{40.47} & \textbf{31.52} & \textbf{62.29} & \textbf{37.42} & \textbf{51.37} \\
\midrule
\method, our corpus & 82.12 & 57.76 & 71.59 & 35.02 & 42.66 & 31.32 & 63.02 & 37.19 & 52.58 \\
\bottomrule
\end{tabular}
\end{table*}

\begin{table*}[t]
\caption{\textbf{Ablation: the form of the description} on FLUX-Reason-6M ($114$k images, SD3.5 Medium). \method writes
either the GCoT as it is or the GCoT rewritten into our corpus' form; the SFT control writes nothing. Both are trained from
the base model on the same images; the base model is in \cref{tab:main} of the main paper. GenEval (stream off) and DPG-Bench.}
\label{tab:supp_open_form}
\centering
\scriptsize
\setlength{\tabcolsep}{4pt}
\begin{tabular}{lccccccc@{\hspace{10pt}}cccccc}
\toprule
 & \multicolumn{7}{c}{GenEval} & \multicolumn{6}{c}{DPG-Bench} \\
\cmidrule(lr){2-8}\cmidrule(lr){9-14}
Training & Overall & 1 obj & 2 obj & Count & Col. & Pos. & Attr. & Overall & Global & Entity & Attribute & Relation & Other \\
\midrule
SFT control (no description) & 67.6 & 98.4 & 81.6 & 61.9 & 86.2 & 23.5 & 54.3 & 83.9 & 80.9 & 89.9 & 87.1 & 92.3 & 81.6 \\
\method, GCoT as it is & \textbf{70.1} & 99.4 & \textbf{82.6} & 65.9 & \textbf{87.0} & 24.0 & \textbf{61.5} & 84.0 & 81.5 & \textbf{90.5} & 87.8 & 92.7 & \textbf{84.4} \\
\rowcolor{sdours} \method, GCoT in our form & \textbf{70.1} & \textbf{100.0} & 82.1 & \textbf{67.5} & 84.6 & \textbf{25.3} & \textbf{61.5} & \textbf{84.3} & \textbf{82.1} & 90.3 & \textbf{88.1} & \textbf{93.2} & 82.8 \\
\bottomrule
\end{tabular}
\end{table*}

\paragraph{Corpus.} We draw $19{,}000$ images at random from each characteristic, among the images whose caption for it is
valid, with no image drawn twice: $114{,}000$ images, of which $1{,}165$ are held out. A sample's prompt is the caption of
the characteristic it was drawn for, and its description is its GCoT, rewritten into the form of our descriptions as below.
Images are used at their native $1024{\times}1024$.

\paragraph{Training.} Unlike the runs on our corpus, both models are trained from the base model on this corpus alone. The
SFT control is trained from SD3.5 Medium to step $1{,}800$; \method starts from its checkpoint at step $800$ and is trained
to step $1{,}800$ with the recipe of the main paper. From step $800$ the two see the same images and prompts in the same
order, with the same batches and noise, and the SFT control reads no description. Both use 2 GPUs with the batch of $128$
of the main paper.

\paragraph{How our descriptions are structured.} The description \method writes on our corpus is the subject field of a
structured description (\cref{sec:supp_data}). It is written by a vision-language model that sees only the image, never the
prompt, so it states what is drawn rather than what was asked for. It is organised by entity: one block per object, person,
animal or group, introduced by a short label, each with its count, appearance, colours and materials, any written text,
its position in the frame and its relations to the other entities. Setting, lighting, camera and style go to other fields,
so the description states the composition and nothing else, in plain present-tense English of median $86$ words. For
example (\cref{fig:supp_data_ex}): \emph{``Three peonies clustered together: Left peony: deep red, petals layered tightly
in a spherical shape, with golden yellow stamens; Top peony: pink, petals wide and spread out, edges wavy, with golden yellow
stamens; Right peony: light pink, \dots; Green leaves: multiple green leaves, palmately lobed, with clear veins, surrounding
the flowers; Stem: light yellow-green, slender, partially visible.''} The GCoT is different in kind: it is written as
reasoning about how to compose the image, mixes the content with lighting, mood, style and intent, does not go through the
image entity by entity and is about twice as long (median $170$ words). For example: \emph{``Amber-colored ceramic teapot
set featuring a central teapot adorned with vibrant green leaf and red berry motifs, surrounded by six matching cups, each
delicately decorated with subtle leaf patterns. The teapot, with its glossy finish and elegant handle, stands prominently in
the center, while the cups are arranged symmetrically around it, creating a balanced and harmonious composition. \dots\ Each
piece is meticulously crafted, showcasing hand-painted details that evoke a vintage-inspired style, blending natural
elements with a touch of elegance. The overall arrangement highlights the cohesive design, where the amber hue and botanical
motifs create a serene and inviting atmosphere, perfect for a tranquil tea experience.''}

\paragraph{From GCoT to our descriptions.} We rewrite each GCoT into the form of our descriptions with Qwen3.5-35B-A3B, the
model that wrote our corpus' descriptions, from the text alone: the model sees no image. It is told to state only the
composition, one block per object, person, animal or group introduced by a short label (\emph{Young woman: \dots; Wooden
table: \dots}), with counts, colours, materials, written text, positions in the frame and relations stated explicitly; to
leave out lighting, atmosphere, mood, camera terms, style and any interpretation; to use only facts the input states and
never add an object, count, colour or position; and to write about $50$ to $150$ words. It also sees three example
descriptions of other images in our corpus' style and the sample's prompt (temperature $0.3$, no thinking). The rewrites have a median of $76$ words, close to our
corpus' $86$. The GCoT above becomes: \emph{``Central teapot: amber-colored ceramic with a glossy finish, featuring vibrant
green leaf and red berry motifs, equipped with an elegant handle. Six matching cups: amber-colored ceramic with subtle leaf
patterns, arranged symmetrically around the central teapot. All items rest on a clean white surface.''} The entities, the
count and the layout stay; the composition talk, the craftsmanship and the atmosphere go.

\paragraph{Results.} With the rewritten descriptions \method gains $2.5$ points of GenEval over the SFT control trained on
the same images (\cref{tab:supp_open_form}, GCoT in our form), mostly in colour attribution ($+7.2$), the category in which it also gains most on our corpus
(\cref{sec:supp_sig}), and counting ($+5.6$). The gain over the SFT control is significant (paired bootstrap over prompts as in \cref{sec:supp_sig}):
$+2.5$ points, $95\%$ interval $[+1.1, +3.9]$, with \method ahead on all four sampling seeds. The SFT control ends $0.8$ points below the base model, while
\method ends $1.7$ points above it. On DPG-Bench \method is again above the SFT control ($84.3$ against $83.9$) and slightly
above the base model ($84.2$). \Cref{fig:supp_open_desc} shares qualitative samples of the three models on held-out
FLUX-Reason-6M prompts, with the descriptions \method decodes for its images; the descriptions follow our labelled form
and name the main subject. Read back from the stream on the held-out
samples, the descriptions reach a perplexity of $40.1$ against the rewritten GCoT, and the run that writes the GCoT as it is
reaches $37.7$ against the GCoT, about the level of the models
trained on our corpus on their own held-out set ($46.3$; \cref{tab:supp_open_ppl}).
 Against the SFT control it is ahead in every DPG category: global ($+1.2$), entity ($+0.4$),
attribute ($+1.0$), relation ($+0.8$) and other, which holds counting and text ($+1.2$). Against the base model it is ahead
overall and in other ($+1.6$).

\begin{table}[t]
\caption{\textbf{Read-out perplexity after training on FLUX-Reason-6M} (SD3.5 Medium, $400$ held-out samples, $28$ steps,
guidance $4$): the description generated with the image, read back against the text the model was trained to write, and
the read-out of the true description (ceiling). Our corpus: \method on its own held-out set.}
\label{tab:supp_open_ppl}
\centering\footnotesize
\begin{tabular}{lcc}
\toprule
Training data & Generated & True \\
\midrule
FLUX-Reason-6M, GCoT in our form & 40.1 & 10.9 \\
FLUX-Reason-6M, GCoT as it is & 37.7 & 22.8 \\
Our corpus & 46.3 & 13.5 \\
\bottomrule
\end{tabular}
\end{table}

\paragraph{T2I-CompBench.} On T2I-CompBench (\cref{tab:supp_open_t2i}) \method trained on FLUX-Reason-6M has the highest
average of the three models ($51.37$, against $50.95$ for the SFT control and $50.01$ for the base model) and is above the SFT
control in six of the eight categories, most in 2D spatial ($+1.4$) and numeracy ($+1.2$), as on our corpus, where 2D spatial
also gains most. It is slightly below the SFT control in shape ($-0.04$) and texture ($-0.5$). Trained on our corpus, \method
reaches $52.58$: above the FLUX-Reason-6M run in six of the eight categories (all but non-spatial and complex), consistent with the GenEval gap between the two
corpora.

\paragraph{Ablation: the form of the description.} To see whether the form of the text matters, we train \method a second
time on the same images with the GCoT used as it is, everything else unchanged (\cref{tab:supp_open_form}). Both forms are
well above the SFT control, so writing helps with either. On GenEval they are on par overall ($70.1$ each) and equal in
colour attribution ($61.5$), but they differ by category: our form is ahead in counting ($67.5$ against $65.9$), position
($25.3$ against $24.0$) and single objects ($100.0$ against $99.4$), the GCoT in colours ($87.0$ against $84.6$) and two
objects ($82.6$ against $82.1$). Our form states counts and positions explicitly for every entity, while the GCoT names
colours often but places objects only loosely. On DPG-Bench, whose prompts are long and list many entities, attributes and
relations, our form is ahead overall ($84.3$ against $84.0$) and in global ($+0.6$), attribute ($+0.3$) and relation
($+0.5$), and only with it does \method pass the base model; the GCoT is ahead in entity ($+0.2$) and other ($+1.6$, from
counting, where it scores $82$ against $80$). The gaps are small, but most point the same way as the design of our corpus:
a description that goes through the image entity by entity and states counts, positions and relations explicitly is a
better thing to write than reasoning-style prose that mixes them with mood and style.

\paragraph{Does the rewrite stay with the GCoT?} The rewrite should change the form of the description, not its content. We
check this with Claude Opus 5.5 on $300$ training samples drawn at random: each judge sees one GCoT and its rewrite, in a fresh session,
and labels every statement of the rewrite (an object, an attribute such as colour, material, count or written text, a position
or a relation) as \emph{supported} by the GCoT, \emph{added} (a detail the GCoT does not state, or states less
specifically) or \emph{contradicted} (incompatible with the GCoT). \Cref{fig:supp_open_prompts} gives the
instructions of the rewrite and of this check. \Cref{tab:supp_open_fid} gives the result. $85.4\%$ of the $4{,}672$
statements are supported by the GCoT and only $0.4\%$ contradict it. The rest ($14.2\%$) add or sharpen a small detail, most often a
position the house style asks for and the GCoT leaves open, or a more specific word: a cremation urn the GCoT does not place is
\emph{positioned centrally}, and the \emph{dark specks} of a cheese become \emph{black specks}. A typical contradiction turns
``a structure with two garage doors'' into \emph{two garages}. The rewrite thus stays faithful to the GCoT: it restructures
and shortens it (median $170$ to $76$ words), adds little and almost never contradicts it.

\begin{table}[H]
\caption{\textbf{Does the rewrite stay with the GCoT?} Claude Opus 5.5 labels every statement of $300$ random rewrites against the
GCoT it was written from: share of the $4{,}672$ statements.}
\label{tab:supp_open_fid}
\centering
\scriptsize
\setlength{\tabcolsep}{3pt}
\begin{tabular}{lc}
\toprule
Statement of the rewrite & Statements (\%) \\
\midrule
Supported by the GCoT & 85.4 \\
Added (not in the GCoT, or more specific) & 14.2 \\
Contradicts the GCoT & \textbf{0.4} \\
\bottomrule
\end{tabular}
\end{table}

\paragraph{Towards a more faithful conversion.} Our rewrite restates the GCoT: it does not correct what the GCoT leaves out or what the conversion may
not capture fully. A capable closed model that sees the image together with the GCoT
could write the description from the image itself, using the GCoT only as a guide, and check every object, count, colour
and position. Such descriptions would match the image more closely than either the GCoT or our
text-only rewrite, and since \method learns to write what it draws, we expect it to gain from them. We leave this to
future work.

\begin{figure*}[!p]
  \centering
  \setlength{\tabcolsep}{2pt}
  \begin{qualband}{SD3.5 Medium, trained on FLUX-Reason-6M}\begin{adjustbox}{max width=\linewidth}%
  \begin{tabular}{@{}cccp{0.52\textwidth}@{}}
  \scriptsize SD3.5-M & \scriptsize SFT control & \scriptsize \textcolor{sdcol}{\textbf{\method}} & \scriptsize prompt and the description \method decodes for its image \\
  \raisebox{-\height}{\rimg[width=0.145\textwidth]{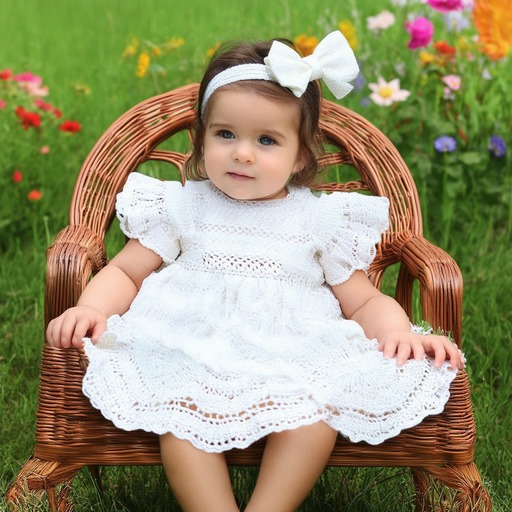}} & \raisebox{-\height}{\rimg[width=0.145\textwidth]{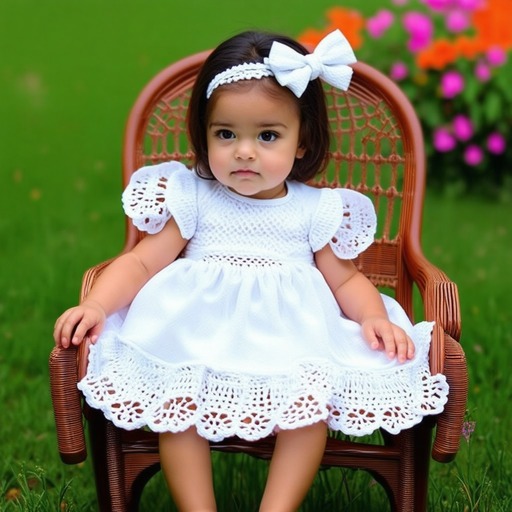}} & \raisebox{-\height}{\rimg[width=0.145\textwidth]{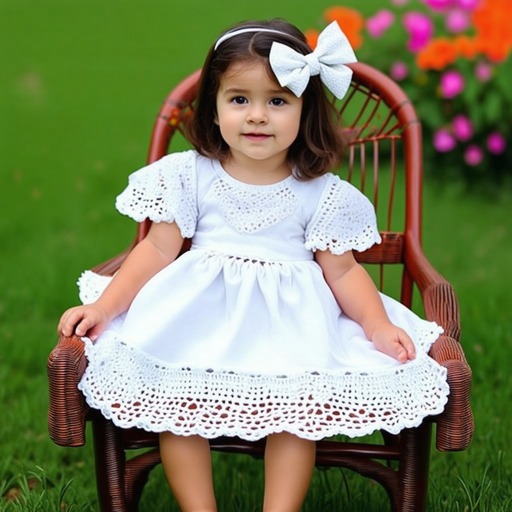}} & \vspace{0pt}\scriptsize\qplabel{PROMPT} {\color{qmute}\itshape Child sitting gracefully on a wicker chair, wearing a delicate white crochet dress with scalloped edges and flutter sleeves, complemented by a matching bow [\dots]}\newline\qdlabel{DESCRIPTION} Young girl: positioned in the center, featuring a vibrant red dress with blue accents and white lace sleeves. A pink ribbon is located at the bottom of the frame; holding an identical heart-shaped arrangement on her headboard stand. Lush greenery: arranged around the base of the image, displaying soft pastel hues along its front hem, flared skirt reaching upward into [\dots] \\[3pt]
  \raisebox{-\height}{\rimg[width=0.145\textwidth]{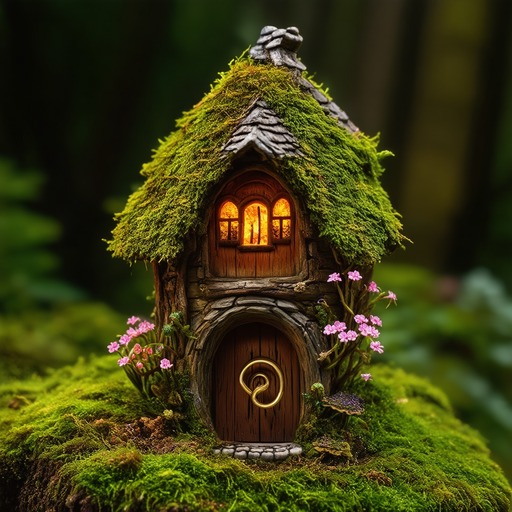}} & \raisebox{-\height}{\rimg[width=0.145\textwidth]{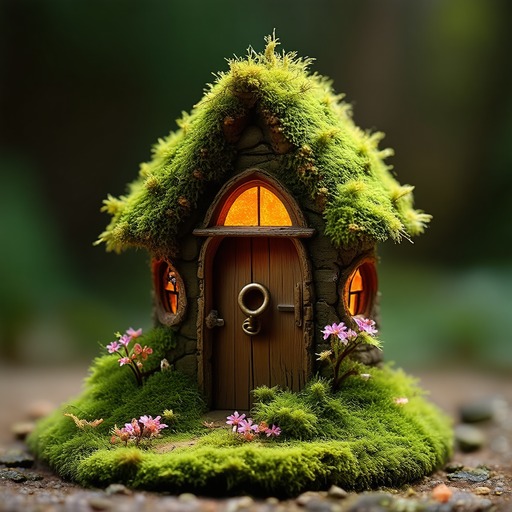}} & \raisebox{-\height}{\rimg[width=0.145\textwidth]{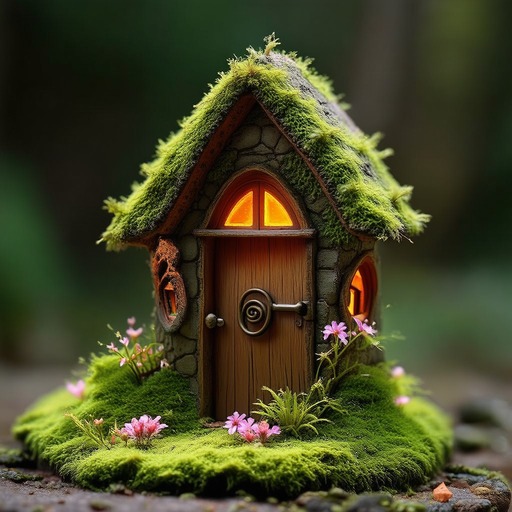}} & \vspace{0pt}\scriptsize\qplabel{PROMPT} {\color{qmute}\itshape A whimsical fairy cottage emerges from a patch of vibrant green moss, adorned with glowing windows, a wooden door with a spiral knocker, and [\dots]}\newline\qdlabel{DESCRIPTION} Two wooden door: positioned centrally, featuring a small porch with warm brown tones and vibrant red accents. Top right of the frame contains two black doors placed on either side; smaller potted plant in front of pink, greenery including one hand-drawn knosties to show blue curtains around the house. Central foreground: located below the structure, displaying white studded by colorful [\dots] \\[3pt]
  \end{tabular}\end{adjustbox}\end{qualband}
  \caption{\textbf{Trained on FLUX-Reason-6M, with decoded descriptions}: SD3.5 Medium, and the SFT control and \method both trained on the $114$k FLUX-Reason-6M images (\cref{tab:supp_open_form}), on held-out FLUX-Reason-6M prompts (same seed).}
  \label{fig:supp_open_desc}
\end{figure*}

\begin{figure*}[!p]
\centering
\begin{tcbraster}[raster columns=2,raster equal height,raster column skip=5pt,raster left skip=0pt,raster right skip=0pt]
\begin{tcolorbox}[judgebox=sdours]
{\small\bfseries\color{sdcol}Rewrite (Qwen3.5-35B-A3B, text only)}\\[3pt]
\qchip{sdband}{Instructions}\\[2pt]
{\footnotesize\itshape You rewrite an image description into a ``subject description'' in a fixed house style. A subject
description states only the composition of the image: which objects, people and animals are present, how many of each, what
each looks like (shape, colours, materials, clothing, markings, any text written on it), where each is in the frame (left,
right, centre, top, bottom, foreground, midground, background, on top of, in front of, behind, next to), and how they relate
to each other (holding, standing on, leaning on, wearing, next to).\\
Rules: Organise it by entity: one block per object, person, animal or group, introduced by a short label and a colon (for
example ``Young woman:'', ``Two dogs:'', ``Wooden table:'', ``Background:''); separate blocks with semicolons or sentences.
State counts explicitly and name colours precisely. Leave out lighting, shadows, atmosphere, mood, emotion, camera, lens,
framing, depth of field, blur, focus, photographic or artistic style, image quality, and any interpretation of what the scene
means. Use only facts stated in the input. Never add an object, count, colour or position the input does not state; if the
input gives no position for something, give none. Plain English, present tense, no preamble, no markdown, no bullet points.
About 50 to 150 words; fewer for a simple scene. Output only the subject description.}\\[4pt]
\qchip{sdband}{Input}\\[2pt]
{\footnotesize Three example descriptions of other images in the house style, the sample's prompt and its GCoT
(temperature $0.3$, no thinking).}
\end{tcolorbox}
\begin{tcolorbox}[judgebox=sdours]
{\small\bfseries\color{sdcol}Fidelity check (Claude Opus 5.5)}\\[3pt]
\qchip{sdband}{Instructions}\\[2pt]
{\footnotesize\itshape The file contains an ORIGINAL image description and a REWRITE of it. The rewrite was meant to restate
the original in a fixed format (one block per object or person, counts, colours, positions, relations), using ONLY facts the
original states, and to leave out lighting, mood and camera terms.\\
Go through the REWRITE statement by statement (an object, an attribute such as colour, material, count, clothing or written
text, a position or a relation). Label each one:
supported: the original states it or it follows directly from what the original states;
added: a concrete detail the original does not state (a new object, attribute, count, colour, position or relation); a more
specific word than the original uses also counts as added;
contradicted: the original states something incompatible.}\\[4pt]
\qchip{sdband}{Output}\\[2pt]
{\footnotesize\ttfamily n\_statements, n\_supported, added: [\dots], contradicted: [\dots]}\\[3pt]
{\scriptsize\color{qmute}One sample per judge, in a fresh session with no other context. The judge also lists facts of the original that the rewrite leaves out; we do not use them here.}
\end{tcolorbox}
\end{tcbraster}
\caption{\textbf{Prompts of the conversion and of its check.} Left: the instructions with which Qwen3.5-35B-A3B rewrites a
FLUX-Reason-6M GCoT into our corpus' description form, from the text alone. Right: the instructions with which Claude Opus 5.5 checks
each rewrite against its GCoT (\cref{tab:supp_open_fid}), leaving out file handling.}
\label{fig:supp_open_prompts}
\end{figure*}

{\small\bibliographystyle{ieeenat_fullname}\bibliography{main}}

\end{document}